\documentclass{article}

\makeatletter
\def\input@path{{Formatting_Instructions_For_NeurIPS_2026/}}
\makeatother
\PassOptionsToPackage{numbers,compress}{natbib}
\usepackage[main, final]{neurips_2026}
\usepackage[utf8]{inputenc}
\usepackage[T1]{fontenc}
\usepackage{subcaption} 
\usepackage{graphicx}
\usepackage{microtype}

\usepackage{xcolor}
\definecolor{citecolor}{HTML}{0071BC}
\usepackage[pagebackref=true,colorlinks,citecolor=citecolor]{hyperref}
\usepackage{booktabs} %

\usepackage{placeins}

\usepackage{url}
\usepackage{algorithm}
\usepackage{algorithmic}

\usepackage{titlesec}
\titlespacing\section{0pt}{3pt plus 1pt minus 1pt}{2pt plus 1pt minus 1pt}
\titlespacing\subsection{0pt}{2pt plus 1pt minus 1pt}{1pt plus 1pt minus 1pt}

\renewcommand{\paragraph}[1]{\vspace{.1em}\noindent\textbf{#1}}

\usepackage[british, english, american]{babel}

\newcommand{\R}{\mathbb{R}}

\usepackage{amsmath}
\usepackage{amssymb}
\usepackage{mathtools}
\usepackage{amsthm}
\usepackage{wrapfig}
\usepackage{listings}
\usepackage{xcolor}
\usepackage{colortbl}
\usepackage{tabulary}
\usepackage{etoolbox}
\usepackage{multirow}
\usepackage{enumitem}
\usepackage[british, american]{babel}

\newcommand{\tablestyle}[2]{\setlength{\tabcolsep}{#1}\renewcommand{\arraystretch}{#2}\centering
\small
}
\newcolumntype{x}[1]{>{\centering\arraybackslash}p{#1pt}}
\newcolumntype{y}[1]{>{\raggedright\arraybackslash}p{#1pt}}
\newcolumntype{z}[1]{>{\raggedleft\arraybackslash}p{#1pt}}

\newcommand\tablefontsize{%
  \footnotesize
}

\newcommand{\vv}[1]{\mathbf{#1}}
\newcommand{\x}{\vv{x}}
\newcommand{\y}{\vv{y}}
\newcommand{\V}{\vv{V}}
\newcommand{\yp}{{\textcolor{blue}{\vv{y}^{+}}}}
\newcommand{\yn}{{\textcolor{orange}{\vv{y}^{-}}}}
\newcommand{\Vp}[1]{{\textcolor{blue}{\vv{V}^{+}_{\textcolor{black}{#1}}}}}
\newcommand{\Vn}[1]{{\textcolor{orange}{\vv{V}^{-}_{\textcolor{black}{#1}}}}}

\newcommand{\e}{{\boldsymbol{\epsilon}}}
\newcommand{\otheta}{{\hat{\theta}}}
\newcommand{\mypara}[1]{\noindent\textbf{#1}\quad}

\makeatletter
\def\onedot{.}

\def\eg{\emph{e.g}\onedot} 
\def\ie{\emph{i.e}\onedot} 
 
 \def\vs{\emph{vs}\onedot}

\makeatother

\definecolor{codeblue}{rgb}{0.25,0.5,0.5}
\definecolor{codekw}{rgb}{0.85, 0.18, 0.50}
\definecolor{codesign}{RGB}{0, 0, 255}
\definecolor{codefunc}{rgb}{0.85, 0.18, 0.50}

\lstdefinelanguage{PythonFuncColor}{
  language=Python,
  keywordstyle=\color{blue}\bfseries,
  commentstyle=\color{codeblue},
  stringstyle=\color{orange},
  showstringspaces=false,
  basicstyle=\ttfamily\small,
  literate=
    {model}{{\color{codefunc}model}}{1}
    {compute_V}{{\color{codefunc}compute\_V}}{1}
    {randn}{{\color{codefunc}randn}}{1}
    {stopgrad}{{\color{codefunc}stopgrad}}{1}
    {mean}{{\color{codefunc}mean}}{1}
    {cat}{{\color{codefunc}cat}}{1}
    {eye}{{\color{codefunc}eye}}{1}
    {affinity}{{\color{codefunc}affinity}}{1}
    {softmax}{{\color{codefunc}softmax}}{1}
    {mse_loss}{{\color{codefunc}mse\_loss}}{1}
    {sqrt}{{\color{codefunc}sqrt}}{1}
    {split}{{\color{codefunc}split}}{1}
    {sum}{{\color{codefunc}sum}}{1}
    {concat}{{concat} }{1}
}

\usepackage[capitalize,noabbrev]{cleveref}

\theoremstyle{plain}
\newtheorem{theorem}{Theorem}[section]
\newtheorem{proposition}[theorem]{Proposition}

\theoremstyle{definition}

\usepackage[export]{adjustbox}
\usepackage[font=small]{caption}
\theoremstyle{remark}

\usepackage[disable,textsize=tiny]{todonotes}

\title{Generative Modeling via Drifting}

\author{%
  Mingyang Deng$^{1}$ \quad He Li$^{1}$ \quad Tianhong Li$^{1}$ \quad Yilun Du$^{2}$ \quad Kaiming He$^{1}$ \\
  $^{1}$MIT \qquad $^{2}$Harvard University
}

\begin{document}

\maketitle

\begin{abstract}
\vspace{-.5em}
Generative modeling can be formulated as learning a mapping $f$ such that its pushforward distribution matches the data distribution.
The pushforward behavior can be carried out iteratively at inference time, \eg, in diffusion/flow-based models.
In this paper, we propose \textit{Drifting Models}, a new generative modeling paradigm that evolves the pushforward distribution during training and naturally admits one-step inference.
We introduce a drifting field that governs the sample movement and achieves 
equilibrium when the distributions match. 
This leads to a training objective that allows the neural network optimizer to evolve the distribution.
In experiments, our one-step generator achieves state-of-the-art one-step results on ImageNet 256$\times$256, with FID 1.54 in latent space and 1.61 in pixel space.
We hope that our work opens up new opportunities for high-quality one-step generation.
\vspace{-.5em}
\end{abstract}

\noindent{\centering
  \includegraphics[width=0.82\textwidth]{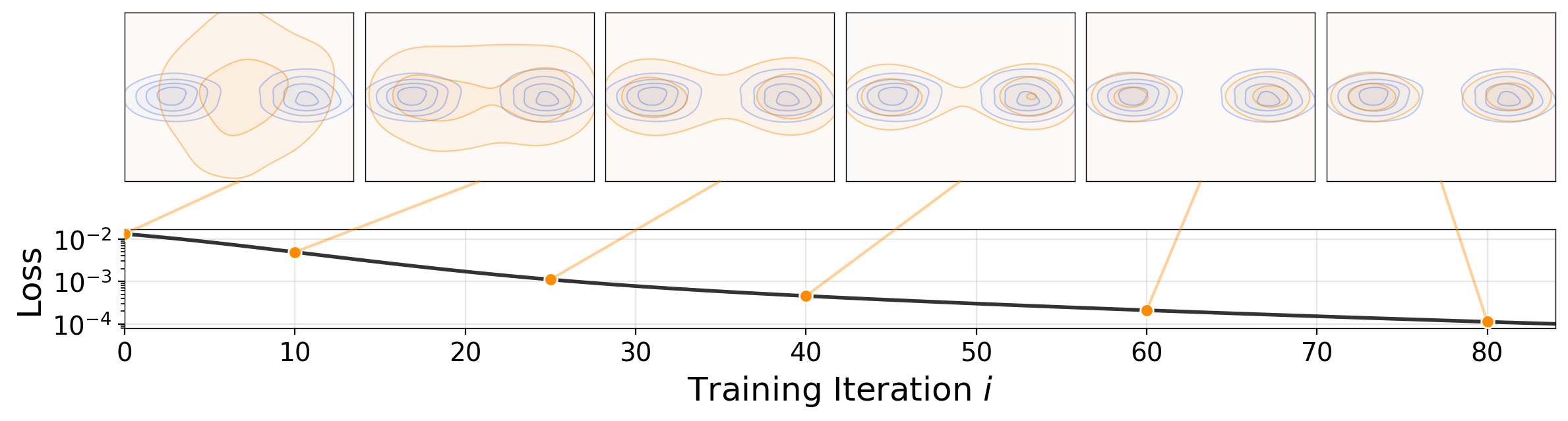}
  \vspace{-0.25em}
  \captionof{figure}
  {\textbf{Drifting Model.}
  A network $f$ performs a pushforward operation: $q={f}_\# p_{\text{prior}}$, mapping a prior distribution $p_{\text{prior}}$ (\eg, Gaussian, not shown here) to a pushforward distribution $q$ ({\color{orange}orange}).
  The goal of training is to approximate the data distribution $p_{\text{data}}$ ({\color{blue}blue}).
  As training iterates, we obtain a sequence of models $\{f_i\}$, which corresponds to a sequence of distributions $\{q_i\}$.
  Our Drifting Model focuses on the evolution of $\{q_i\}$ at \textit{training-time}. We introduce a drifting field that approaches zero when $q$ matches $p_{\text{data}}$, and guides the evolution. This drifting field provides a loss function (y-axis, in log-scale) for training.
  }
  \label{fig:teaser}
  \vspace{-5pt}
\par}

\section{Introduction}

Generative models are commonly regarded as more challenging than discriminative models.
While discriminative modeling typically focuses on mapping individual samples to their corresponding labels, generative modeling concerns mapping from one distribution to another.
This can be expressed as learning a mapping $f$ such that the \mbox{\textit{pushforward}} of a prior distribution $p_\text{prior}$ matches the data distribution, namely, $f_\# p_\text{prior} \approx p_{\text{data}}$.
Conceptually, generative modeling learns a \textit{functional} (here, $f_\#$) that maps from one function (here, a distribution) to another.

The ``pushforward'' behavior can be realized \textit{iteratively} at \textit{inference} time, \eg, in prevailing paradigms such as Diffusion~\cite{sohl2015deep} and Flow Matching~\cite{lipman2022flow}. When generating, these models map noisier samples to slightly cleaner ones, progressively evolving the sample distribution toward the data distribution.
This modeling philosophy can be viewed as decomposing a complex pushforward map (\ie, $f_\#$) into a chain of more feasible transformations, applied at inference time.

In this paper, we propose \emph{Drifting Models}, a new generative modeling paradigm that learns a pushforward map by evolving it during \textit{training} time, thereby removing the need for an iterative inference procedure. The mapping $f$ is represented by a single-pass, non-iterative network. As the training process is inherently iterative in deep learning optimization, it can be naturally viewed as evolving the pushforward
distribution, $f_\# p_\text{prior}$, through the update of $f$. See Fig.~\ref{fig:teaser}.

To drive the evolution of the training-time pushforward, we introduce a
\emph{drifting field} $\V$ that governs the sample movement.
This field depends on the generated distribution and the data distribution.
By definition, this field becomes zero when the two distributions match, thereby reaching an equilibrium in which the samples no longer drift.

Building on this formulation, we propose a simple training objective that minimizes the \textit{drift} of the generated samples. This objective induces sample movements and thereby evolves the underlying pushforward distribution through iterative optimization (\eg, SGD). We further introduce the designs of the drifting field, the neural network model, and the training algorithm.

Drifting Models naturally perform \textit{single-step} (\mbox{``1-NFE''}) generation and achieve strong empirical performance.
On ImageNet 256$\times$256, we obtain a 1-NFE
FID of \textbf{1.54} under the standard latent-space generation protocol, achieving a new state-of-the-art among single-step methods. This result remains competitive even when compared with \textit{multi-step} diffusion-/flow-based models.
Further, under the more challenging \textit{pixel}-space generation protocol (\ie, without latents), we reach a 1-NFE FID of \textbf{1.61}, substantially outperforming previous pixel-space methods. 
These results suggest that Drifting Models offer a promising framework for high-quality, efficient generative modeling.

\vspace{-5pt}
\section{Related Work}
\label{sec:related_work}

\vspace{-.5em}
\mypara{Diffusion-/Flow-based Models.}
Diffusion models (\eg, \citep{sohl2015deep,ho2020denoising,song2020score}) and their flow-based counterparts (\eg, \citep{lipman2022flow,liu2022flow,albergo2023stochastic}) formulate noise-to-data mappings through \textit{differential equations} (SDEs or ODEs). At the core of their inference-time computation is an \textit{\mbox{iterative}} update, \eg, of the form $\x_{i+1}=\x_{i} + \Delta\x_{i}$, such as with an Euler solver. The update $\Delta\x_{i}$ depends on the neural network $f$, and as a result, generation involves multiple steps of network evaluations.

A growing body of work has focused on reducing the steps of diffusion-/flow-based models.
Distillation-based methods 
(\eg, \citep{salimans2022progressive,luo2023diff,yin2024one,zhou2024score}) distill a pretrained multi-step model into a single-step one.
Another line of research aims to train one-step diffusion/flow models from scratch (\eg, \citep{song2023consistency,frans2024one,boffi2025flow,geng2025mean}). To achieve this, these methods utilize the SDE/ODE dynamics in training by approximating the induced trajectories.
In contrast, our work presents a conceptually different framework and does not rely on SDE/ODE formulations in diffusion/flow models (more discussions in Appendix~\ref{app:related_connections}).

\mypara{Generative Adversarial Networks (GANs).}
GANs \cite{goodfellow2014generative} are a classical family of models that train a generator by discriminating generated samples from real data.
Like GANs, our method involves a single-pass network $f$ that maps noise to data, whose ``goodness'' is evaluated by a loss function; however, unlike GANs, our method does not rely on adversarial optimization.
Implicit Maximum Likelihood Estimation (IMLE) \cite{li2018implicit} trains a generator without a discriminator, by pulling the nearest generated sample toward each data sample; it has no repulsion between generated samples.

\mypara{Variational Autoencoders (VAEs).}
VAEs~\cite{kingma2013auto} optimize the ELBO, consisting of reconstruction and KL terms.
Classical VAEs are one-step generators with Gaussian priors, while modern uses often treat VAEs as tokenizers with learned priors from diffusion or autoregressive models~\cite{rombach2022high,esser2021taming}.

\mypara{Normalizing Flows (NFs).}
NFs~\cite{rezende2015variational,dinh2016density,zhai2024normalizing} learn data-to-noise mappings by maximizing sample log-likelihood.
They require invertible architectures with tractable Jacobians. Conceptually, NFs are one-step generators at inference, with computation performed by the \mbox{inverse} of the network.

\mypara{Moment Matching.} 
Moment-matching methods \cite{dziugaite2015training,li2015generative} seek to minimize the Maximum Mean Discrepancy (MMD) between the generated and data distributions.
Moment Matching has recently been extended to one-/few-step diffusion \cite{zhou2025inductive}.
Related to MMD, our approach also leverages the concepts of kernel functions and positive/negative samples. However, our approach focuses on a drifting field that explicitly governs the sample drifts at training time; MMD relates to a special case of this field. Further discussion is in Appendix~\ref{sec:kernel_generators} and~\ref{sec:theory_mmd_relation}.

\mypara{Contrastive Learning.}
Our drifting field is driven by positive samples from the data distribution and negative samples from the generated distribution.
This is conceptually related to the positive and negative samples and softmax-normalized weights in \textit{contrastive representation learning}, \eg, \cite{hadsell2006dimensionality,oord2018representation,he2020momentum,chen2020simple}.
Contrastive learning has also been extended to generative models, \eg, in energy-based models trained with contrastive divergence \cite{hinton2002training,lecun2006tutorial,xie2016theory,nijkamp2019learning,du2019implicit,du2021improved}, in GANs \cite{unterthiner2017coulomb,kang2020contragan}, or in Flow Matching \cite{stoica2025contrastive}; we discuss the relation to Coulomb GAN in Appendix~\ref{sec:coulomb_gan}. 

\mypara{Drifting Fields in Existing Generative Models.}
For one-step generative methods that minimize a loss between
two distributions $p$ and $q$, our framework implies the existence of an underlying drifting field $\V$. 
In particular, one may define $\V{\propto}{-}\frac{\partial\mathcal L }{\partial {\x}}$, which vanishes at the minimum of $\mathcal L$, attained when $p = q$.
Accordingly, we can derive the implicit drifting fields of several existing one-step generative models, including GANs~\cite{goodfellow2014generative}, DMD-style methods \cite{yin2024one,luo2023diff,zhou2024score}, and MMD-based methods~\cite{li2015generative,dziugaite2015training}, which we elaborate on in Appendix~\ref{app:related_connections}. Nevertheless, our method directly models a drifting field with the desired properties, without requiring it to be the gradient of a discrepancy loss $\mathcal L$.

\section{Drifting Models for Generation}
\label{sec:methodology}

We propose Drifting Models, which frame generative modeling as a \emph{training-time} evolution of the pushforward distribution via a drifting field.
This naturally allows one-step generation at inference.

\subsection{Pushforward at Training Time}
\label{subsec:training_time_pushforward}

Consider a neural network $f: \mathbb{R}^C \mapsto \mathbb{R}^D$ with input $\e \sim p_{\e}$ (\eg, noise in dimension $C$) and output $\x = f(\e) \in \mathbb{R}^D$. In general, the input and output dimensions need not be equal.

We denote the distribution of the network output by $q$, \ie, $\x = f(\e) \sim q$.
In probability theory, $q$ is referred to as the \textit{pushforward} distribution of $p_\e$ under $f$, denoted by:
\begin{equation}
q = f_{\#}p_\e.
\end{equation}
Here, ``$f_{\#}$'' denotes the pushforward induced by $f$.
Intuitively, it means that $f$ transforms a distribution $p_{\e}$ into another distribution $q$.
The goal of generative modeling is to find $f$ such that $f_{\#}p_{\e} \approx p_\text{data}$.

Since neural network \textit{training} is inherently iterative (\eg, SGD), the training process produces a sequence of models $\{f_i\}$, where $i$ denotes the training iteration.
This corresponds to a sequence of pushforward distributions $\{q_i\}$ during training, where
$q_i = [f_i]_{\#}p_{\e}$ for each $i$.
The training process progressively evolves $q_i$ to match $p_{\text{data}}$.

When the network $f$ is updated,
a sample at training iteration $i$ is implicitly ``drifted'' as:
$\x_{i+1} = \x_{i} + \Delta\x_{i}$, where $ \Delta\x_{i}:=f_{i+1}(\e) - f_{i}(\e)$ arises from parameter updates to $f$.
This implies that the update of $f$ determines the ``\textit{residual}'' of $\x$, which we refer to as the ``drift''.

\subsection{Drifting Field for Training}
\label{subsec:training_objective}

Next, we define a \emph{\textbf{drifting field}} to govern the training-time evolution of the samples $\x$ and, consequently, the pushforward distribution $q$.
A drifting field is a function that computes $ \Delta\x$ given $\x$. Formally, denoting this field by $\V_{p,q}(\cdot)\colon \R^d \to \R^d$, we have: 
\begin{equation}
\x_{i+1}=\x_i+\V_{p,q_i}(\x_i),
\end{equation}
Here, $\x_i=f_i(\e) \sim q_i$ and after drifting we denote $\x_{i+1}\sim q_{i+1}$.
The subscripts $p,q$ denote that this field depends on $p$ (\eg, $p=p_\text{data}$) and the current distribution $q$.

Ideally, when $p=q$, we want all $\x$ to stop drifting \ie, $\V=\textbf{0}$.
We consider the following proposition:
\begin{proposition}
Consider an \textbf{anti-symmetric} drifting field:
\begin{equation}
\label{eq:asym}
\V_{p,q}(\x) = - \V_{q,p}(\x),\quad \forall \x.
\end{equation}
Then we have: $\quad q=p \quad \Rightarrow \quad \V_{p,q}(\x) = \mathbf{0},\forall \x$.
\end{proposition}

The proof is straightforward\footnotemark.
Intuitively, anti-symmetry means that swapping $p$ and $q$ simply flips the sign of the drift.
This proposition implies that if the pushforward distribution $q$ matches the data distribution $p$, the drift is zero for any sample and the model achieves an equilibrium.

\footnotetext{$q{=}p \Rightarrow \V_{p,q}{=}\V_{q,p}{=}{-}\V_{p,q} \Rightarrow \V_{p,q}{=}\textbf{0}$}

We note that the converse implication, \ie, $\V_{p,q}=\mathbf{0}\Rightarrow q=p$, is false in general for arbitrary choices of $\V$.
For our kernelized formulation (Sec.~\ref{subsec:instantiation_field}), we give sufficient conditions under which $\V_{p,q}\approx\mathbf{0}$ implies $q\approx p$ (Appendix~\ref{sec:theory_identifiability}).

\paragraph{Training Objective.}
The property of equilibrium motivates a definition of a training objective.
Let $f_\theta$ be a network parameterized by $\theta$, and $\x = f_\theta(\e)$ for $\e \sim p_\e$. At the equilibrium where $\V=\textbf{0}$, we set up the following \textit{fixed-point} relation:
\begin{equation}
\label{eq:eq_cond}
f_{\otheta}(\e)
=
f_{\otheta}(\e)
+
\V_{p,q_{\otheta}}\!\big(f_{\otheta}(\e)\big).
\end{equation}
Here, $\otheta$ denotes the optimal parameters at the equilibrium, and $q_\otheta$ denotes the pushforward of $f_{\otheta}$.

This equation motivates a fixed-point iteration during training. At iteration $i$, we seek to satisfy:
\begin{equation}
f_{\theta_{i+1}}(\e)
\leftarrow
f_{\theta_{i}}(\e)
+
\V_{p,q_{\theta_{i}}}\!\big(f_{\theta_{i}}(\e)\big).
\end{equation}
We convert this update rule into a loss function:
\begin{equation}
\label{eq:train_loss}
\mathcal{L}
=
\mathbb{E}_{\e}
\Big[
\big\|
\underbrace{f_{\theta}(\e)}_{\text{prediction}}
-
\underbrace{
\text{\texttt{stopgrad}}\big(
f_{\theta}(\e)
+
\V_{p,q_{\theta}}\!\big(f_{\theta}(\e)\big)
\big)
}_{\text{frozen target}}
\big\|^2
\Big].
\end{equation}
Here, the stop-gradient operation provides a frozen state from the last iteration, following \cite{chen2021exploring,song2023improved}.
Intuitively, we compute a frozen target and move the network prediction toward it.

We note that the \textit{value} of our loss function $\mathcal{L}$ is equal to $\mathbb{E}_{\e}
\big[\|\V(f(\e))\|^2 \big]$, that is, the squared norm of the drifting field $\V$. 
With the stop-gradient formulation, our solver does not directly back-propagate through $\V$, because $\V$ depends on $q_\theta$ and back-propagating through a distribution is nontrivial.
Instead, our formulation minimizes this objective \textit{indirectly}: it moves $\x=f_{\theta}(\e)$ towards its drifted version, \ie, towards $\x+\Delta\x$ that is frozen at this iteration.

\subsection{Designing the Drifting Field}
\label{subsec:instantiation_field}

The field $\V_{p,q}$ depends on distributions $p$ and $q$. To obtain a computable formulation, we consider:
\begin{equation}
\label{eq:kernel_K}
\V_{p,q}(\x)=\mathbb E_{\yp \sim p}\mathbb E_{\yn \sim q} [\mathcal{K}(x,\yp,\yn)],
\end{equation}
where $\mathcal{K}(\cdot,\cdot,\cdot)$ is a kernel-like function describing interactions among three sample points. $\mathcal{K}$ can optionally depend on $p$ and $q$.
Our framework supports a broad class of functions $\mathcal{K}$, as long as $\V=\text{0}$ when $p=q$. 

\begin{figure}[t]
\vspace{-.5em}
\centering
\begin{minipage}[t]{0.48\linewidth}
\vspace{-.5em}
\vspace{0pt}%
\centering
\includegraphics[width=0.8\linewidth, trim=0 2.5cm 0 0, clip]{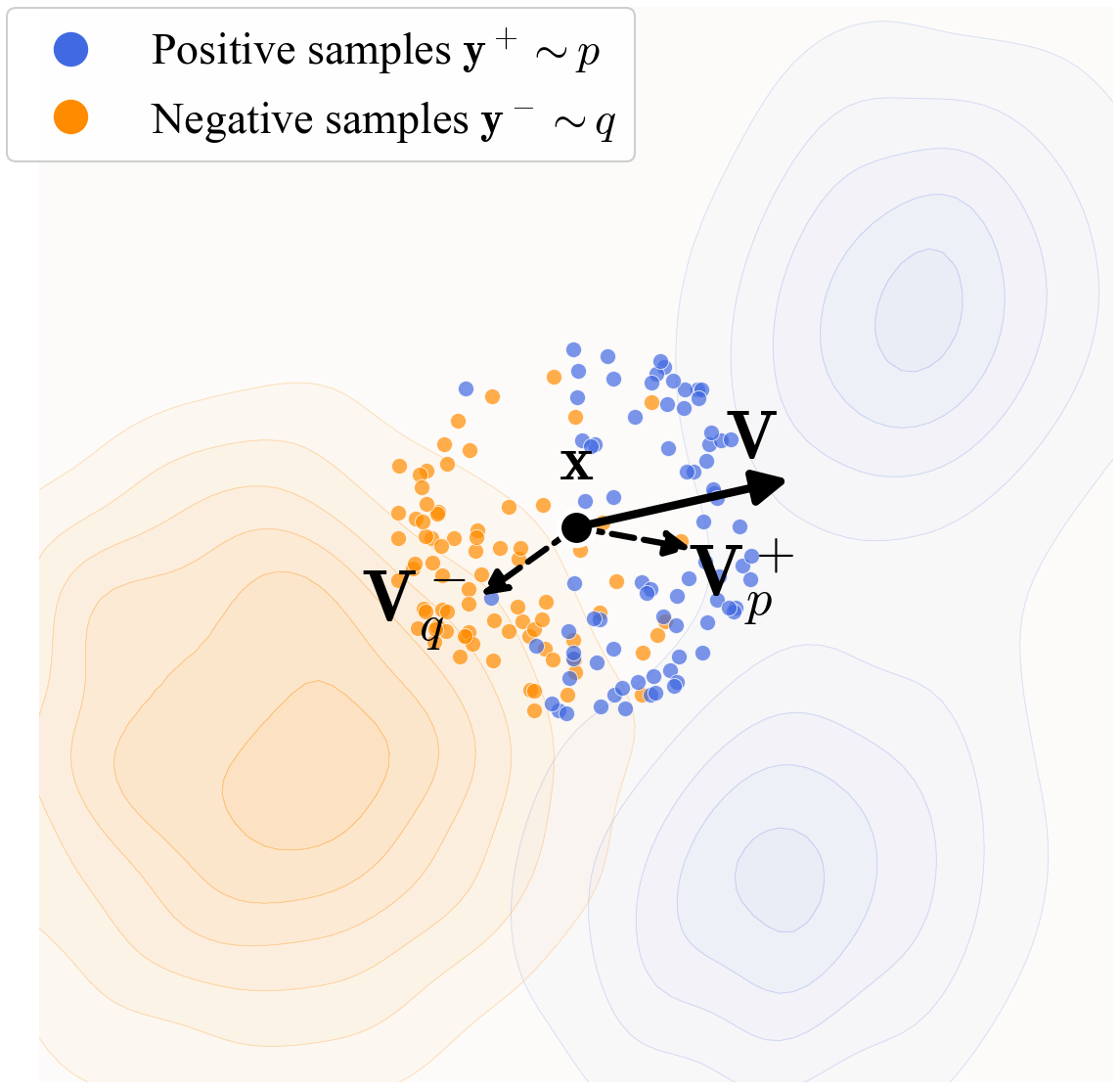}
\caption{\textbf{Illustration of drifting a sample.}
A generated sample $\mathbf{x}$ (black) drifts according to $\V=\V^+_{p}-\V^-_{q}$:
it is attracted by positive samples (blue) and repulsed by negative samples (orange).
}
\label{fig:micro_shift}
\end{minipage}%
\hfill
\begin{minipage}[t]{0.48\linewidth}
\vspace{0pt}%

\begin{algorithm}[H]
\caption{
\textbf{Training Loss.}
Note: for brevity, here the negative samples \texttt{y\_neg} are from the same batch of generated data,
though they can include other sources of negatives.
}
\label{alg:training_code}
\begin{lstlisting}
# f: generator; 
# y_pos: [N_pos, D], data samples
e = randn([N, C]) # noise
x = f(e) # [N, D], generated samples
y_neg = x # reuse x as negatives
V = compute_V(x, y_pos, y_neg)
x_drifted = stopgrad(x + V)
loss = mse_loss(x - x_drifted)
\end{lstlisting}
\end{algorithm}
\end{minipage}
\vspace{-.5em}
\end{figure}

For the instantiation in this work, we introduce a form of $\V$ driven by attraction and repulsion.
We define the following fields inspired by the \textit{mean-shift} method \cite{cheng1995mean}:
\begin{equation}
\begin{aligned}
\Vp{p}(\x)
&:= \frac{1}{Z_{p}} {\mathbb{E}_{p}\!\left[k(\x,\yp)(\yp - \x)\right]}, \qquad
\Vn{q}(\x)
&:= \frac{1}{Z_{q}} 
{\mathbb{E}_{q}\!\left[k(\x,\yn)(\yn - \x)\right]}.
\end{aligned}
\label{eq:meanshift}
\end{equation}
Here, $Z_{p}$ and $Z_{q}$ are normalization factors:
\begin{equation}
\label{eq:Z}
\begin{aligned}
Z_{p}(\x) := \mathbb{E}_{p} [k(\x,\yp)],\qquad 
Z_{q}(\x) := \mathbb{E}_{q} [k(\x,\yn)].
\end{aligned}
\end{equation}
Intuitively, Eq.~(\ref{eq:meanshift}) computes the weighted mean of the vector difference $\y - \x$. The weights are given by a kernel $k(\cdot, \cdot)$ normalized by (\ref{eq:Z}).
We then define $\V$ as:
\begin{equation}
\V_{p,q}(\x) := \Vp{p}(\x) - \Vn{q}(\x).
\label{eq:V_pos_neg}
\end{equation}
Intuitively, this field can be viewed as attracting by the data distribution $p$ and repulsing by the sample distribution $q$.
This is illustrated in Fig.~\ref{fig:micro_shift}.
Substituting Eq.~(\ref{eq:meanshift}) into
Eq.~(\ref{eq:V_pos_neg}), we obtain:
\begin{equation}
\V_{p,q}(\x)
= \frac{1}{Z_p Z_q} {\mathbb{E}_{p,q}\!\left[k(\x,\yp)k(\x,\yn)(\yp - \yn)\right]}.
\label{eq:VZZ}
\end{equation}
Here, the vector difference reduces to $\y^+ - \y^-$; the weight is computed from two kernels and normalized jointly.
This form is an instantiation of Eq.~(\ref{eq:kernel_K}).
It is easy to see that $\V$ is anti-symmetric: $\V_{p,q}=-\V_{q,p}$. In general, our method does not require $\V$ to be decomposed into attraction and repulsion; it only requires $\V=\text{0}$ when $p=q$.

\paragraph{Kernel.}
The kernel $k(\cdot, \cdot)$ can be a function that measures the similarity. In this paper, we adopt:
\begin{equation}
\label{eq:kernel}
k(\x, \y) = \exp\left(-\frac{1}{\tau} \|\x - \y\|\right),
\end{equation}
where $\tau$ is a temperature and $\|\cdot\|$ is $\ell_2$-distance. We view $\tilde{k}(\x, \y) \triangleq\frac{1}{Z}k(\x, \y)$ as a normalized kernel, which absorbs the normalization in Eq.~(\ref{eq:VZZ}).

In practice, we implement $\tilde{k}$ using a \textit{softmax} operation, with logits given by $-\frac{1}{\tau} \|\x - \y\|$, where the softmax is taken over $\y$.
This softmax operation is similar to that of InfoNCE \cite{oord2018representation} in contrastive learning.
In our implementation, we further apply an extra softmax normalization over the set of $\{\x\}$ within a batch, which slightly improves performance in practice. This additional normalization does not alter the antisymmetric property of the resulting $\V$.

\paragraph{Equilibrium and Matched Distributions.}
Since our training loss in Eq.~(\ref{eq:train_loss}) encourages minimizing $\|\V\|^2$, we hope that $\V \approx \textbf{0}$ leads to $q\approx p$.
While this implication does not hold for arbitrary choices of $\V$, we empirically observe that decreasing the value of $\|\V\|^2$ correlates with improved generation quality. In Appendix~\ref{sec:theory_identifiability}, we prove identifiability in special cases and provide a broader heuristic: for our kernelized construction, zero drift imposes many bilinear constraints on $(p,q)$, which under mild non-degeneracy assumptions force approximate matching of $p$ and $q$.

\paragraph{Stochastic Training.}
In stochastic training (\eg, mini-batch optimization), we estimate $\V$ by approximating the expectations in Eq.~(\ref{eq:VZZ}) with empirical means.
For each training step, we draw $N$ samples of noise $\e \sim 
p_\e$ and compute a batch of $\x=f_\theta(\e)\sim q$.
The generated samples also serve as the negative samples in the same batch, \ie, $\y^- \sim q$.
On the other hand, we sample $N_\text{pos}$ data points $\y^+ \sim p_\text{data}$. The drifting field $\V$ is computed in this batch of positive and negative samples. 
Alg.~\ref{alg:training_code} provides the pseudocode for such a training step, where \texttt{compute\_V} is given in \cref{sec:appendix_compute_V}.

\subsection{Drifting in Feature Space}
\label{subsec:feature_space}

Thus far, we have defined the objective (\ref{eq:train_loss}) directly in the raw data space. Our formulation can be extended to any feature space.
Let $\phi$ denote a feature extractor (\eg, an image encoder) operating on real or generated samples. We rewrite the loss (\ref{eq:train_loss}) in the feature space as:
\begin{equation}
\mathbb{E}\!\left[
\left\|
\phi(\x)
-
\text{\texttt{stopgrad}}\Big(
\phi(\x)
+
\V\big(\phi(\x)\big)
\Big)
\right\|^2
\right].
\label{eq:feature_space}
\end{equation}
Here, $\x=f_\theta(\e)$ is the output (\eg, images) of the generator. $\V$ is defined in the feature space: in practice, this means that
$\phi(\y^+)$ and $\phi(\y^-)$ serve as the positive/negative samples.
It is worth noting that feature encoding is a training-time operation and is not used at inference time.

This can be further extended to multiple features, \eg, at multiple scales and locations:
\begin{equation}
\sum_j\mathbb{E}\!\left[
\left\|
\phi_j(\x)
-
\text{\texttt{stopgrad}}\Big(
\phi_j(\x)
+
\V\big(\phi_j(\x)\big)
\Big)
\right\|^2
\right].
\label{eq:feature_space_multi}
\end{equation}
Here, $\phi_j$ represents the feature vectors at the $j$-th scale and/or location from an encoder $\phi$.
With a ResNet-style image encoder \cite{he2016deep}, we compute drifting losses across multiple scales and locations, which provides richer gradient information for training.

The feature extractor plays an important role in the generation of high-dimensional data.
As our method is based on the kernel $k(\cdot, \cdot)$ for characterizing sample similarities, it is desired for semantically similar samples to stay close in the feature space. This goal is aligned with self-supervised learning (\eg, \citep{he2020momentum,chen2020simple}). 
We use pre-trained self-supervised models as the feature extractor.

\paragraph{Relation to Perceptual Loss.}
Our feature-space loss is related to perceptual loss \cite{zhang2018unreasonable} but is conceptually different. The perceptual loss minimizes: $\|\phi(\x)-\phi(\x_\text{target})\|_2^2$, that is, the regression target is $\phi(\x_\text{target})$ and requires pairing $\x$ with its target.
In contrast, our regression target in (\ref{eq:feature_space}) is
$
\phi(\x)
+
\V\big(\phi(\x)\big)
$, where the drifting is in the feature space and requires no pairing. 
In principle, our feature-space loss aims to match the pushforward distributions $\phi_\#q$ and $\phi_\#p$.

\paragraph{Relation to Latent Generation.}
Our feature-space loss is \textit{orthogonal} to the concept of latent-space generators \cite{rombach2022high}.
In our case, when using $\phi$,
the generator $f$ can still produce outputs in the pixel space or the latent space of a tokenizer.
If the generator $f$ is in the latent space and the feature extractor $\phi$ is in the pixel space, the tokenizer decoder is applied before extracting features from $\phi$.

\subsection{Classifier-Free Guidance}
\label{subsec:additive_guidance}

Classifier-free guidance (CFG)~\cite{ho2022classifier} improves generation quality by extrapolating between class-conditional and unconditional distributions.
Our method naturally supports a related form of guidance.

In our model, given a class label $c$ as the condition, the underlying target distribution $p$ now becomes $p_{\text{data}}(\cdot|c)$, from which we can draw positive samples: $\y^+ \sim p_{\text{data}}(\cdot|c)$.
To achieve guidance, we draw \textit{negative} samples either from generated samples or \textit{real} samples from different classes.
Formally, the negative sample distribution is now:
\begin{equation}
\tilde{q}(\cdot|c) \triangleq (1-\gamma)\,q_\theta(\cdot|c)   + \gamma\,p_{\text{data}}(\cdot | \varnothing).
\label{eq:cfg1}
\end{equation}
Here, $\gamma \in [0, 1)$ is a mixing rate,
and $p_{\text{data}}(\cdot | \varnothing)$ denotes the \textit{unconditional} data distribution\footnotemark{}.
The goal of learning is to find $\tilde{q}(\cdot|c)=p_\text{data}(\cdot|c)$.
Substituting it into (\ref{eq:cfg1}), we obtain:
\begin{equation}
q_\theta(\cdot|c) = \alpha\, p_{\text{data}}(\cdot|c) - (\alpha - 1)\, p_{\text{data}}(\cdot | \varnothing).
\label{eq:cfg2}
\end{equation}
where $\alpha=\frac{1}{1-\gamma} \geq 1$.
This implies that $q_\theta(\cdot|c)$ is to approximate a linear combination of conditional and unconditional data distributions. This follows the spirit of original CFG.

\footnotetext{This should be the data distribution excluding the class $c$. For simplicity, we use the unconditional data distribution.}

In practice, Eq.~(\ref{eq:cfg1}) means that we sample extra negative examples from the data in $p_{\text{data}}(\cdot | \varnothing)$, in addition to the generated data. The distribution $q_\theta(\cdot|c)$ corresponds to a class-conditional network $f_\theta(\cdot | c)$, similar to common practice \cite{ho2022classifier}.
We note that, in our method, CFG is a \textit{training-time} behavior by design: the one-step (1-NFE) property is preserved at inference time.

\section{Implementation for Image Generation}
\label{sec:impl}

We describe our implementation for image generation on ImageNet \cite{deng2009imagenet} at resolution 256$\times$256. Full implementation details are provided in Appendix \ref{sec:appendix_impl}.

\mypara{Tokenizer and Architecture.} By default, we perform generation in latent space \cite{rombach2022high}. We adopt the standard SD-VAE tokenizer, which produces a 32$\times$32$\times$4 latent space.
Our generator ($f_\theta$) has a DiT-like \cite{peebles2023scalable} architecture.
Its input is 32$\times$32$\times$4-dim Gaussian noise $\e$, and its output is the generated latent $\x$ of the same dimension.
We use a patch size of 2, \ie, like DiT/2.
Our model uses adaLN-zero \cite{peebles2023scalable} for processing class-conditioning or other extra conditioning.

\mypara{CFG conditioning.} We follow \cite{geng2025improved} and adopt CFG-conditioning. At training time, a CFG scale $\alpha$ (Eq.~(\ref{eq:cfg2})) is randomly sampled. Negative samples are prepared based on $\alpha$ (Eq.~(\ref{eq:cfg1})), and the network is conditioned on this value. At inference time, $\alpha$ can be freely specified. Details are in \ref{sec:appendix_cfg}.

\mypara{Batching.} The pseudo-code in Alg.~\ref{alg:training_code} describes a batch of $N=N_\text{neg}$ generated samples. In practice, when class labels are involved, we sample a batch of $N_\text{c}$ class labels. For each label, we perform Alg.~\ref{alg:training_code} \textit{independently}. Accordingly, the \textit{effective} batch size is $B=N_\text{c}{\times}N$, which consists of $N_\text{c}{\times}N$ negatives and $N_\text{c}{\times}N_\text{pos}$ positives.
We define a ``training epoch'' based on the number of generated samples $\x$.
In particular, each iteration generates $B$ samples, and one epoch corresponds to $N_\text{data}/B$ iterations for a dataset of size $N_\text{data}$.

\mypara{Feature Extractor.} Our model is trained with drifting loss in a feature space (Sec.~\ref{subsec:feature_space}). 
The feature extractor $\phi$ is an image encoder.
We mainly consider a ResNet-style \cite{he2016deep} encoder, pre-trained by self-supervised learning, \eg, MoCo \cite{he2020momentum} and SimCLR \cite{chen2020simple}.
When these pre-trained models operate in pixel space, we apply the VAE decoder to map our generator’s latent-space output back to pixel space for feature extraction. Gradients are backpropagated through the feature encoder and VAE decoder.
We also study an MAE \cite{he2022masked} pre-trained in latent space (detailed in \ref{sec:appendix_latent_mae}).
For all ResNet-style models, features are extracted from multiple stages (\ie, multi-scale feature maps). The drifting loss in (\ref{eq:feature_space}) is computed per scale and then combined. We discuss details in \ref{sec:appendix_feature_extraction}.

\mypara{Pixel-space Generation.} 
While our experiments primarily focus on latent-space generation, our models support pixel-space generation. In this case, $\e$ and $\x$ are both 256$\times$256$\times$3. We use a patch size of 16 (\ie, DiT/16). The feature extractor $\phi$ is directly on the pixel space.

\section{Experiments}

\subsection{Toy Experiments}
\begin{figure}[t]
    \centering
    \hspace{-2em}
    \begin{minipage}[t]{0.43\linewidth}
        \vspace{0pt}
        \centering
        \includegraphics[height=12em]{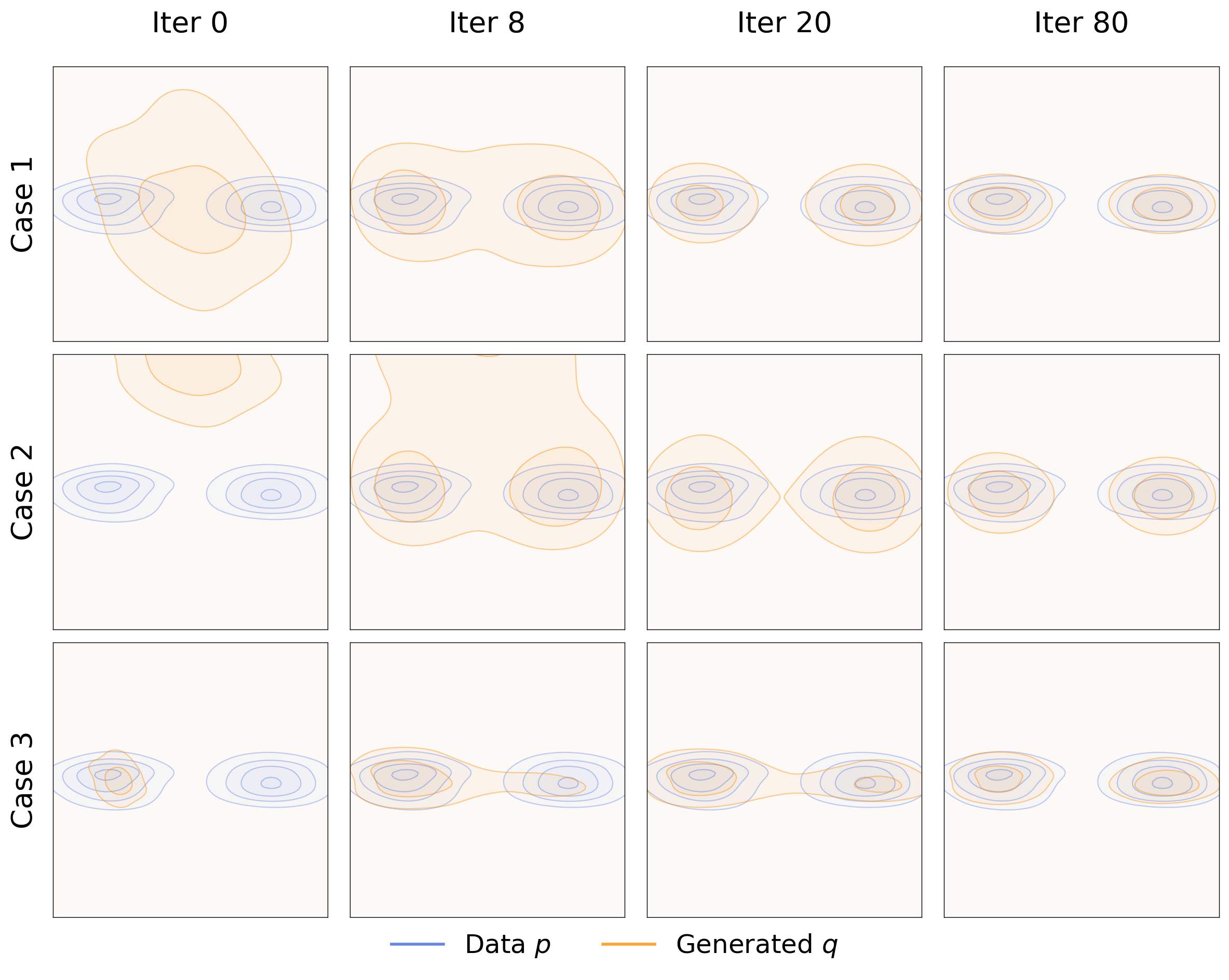}
        \vspace{.15em}
        \caption{\textbf{Evolution of the generated distribution.}
        The generated distribution $q$ ({\color{orange}orange}) evolves toward a bimodal target $p$ ({\color{blue}blue}) across three initializations.
        }
        \label{fig:drifting_evolution_macro}
    \end{minipage}\hfill
    \begin{minipage}[t]{0.56\linewidth}
        \vspace{0pt}
        \centering
        \includegraphics[width=\linewidth]{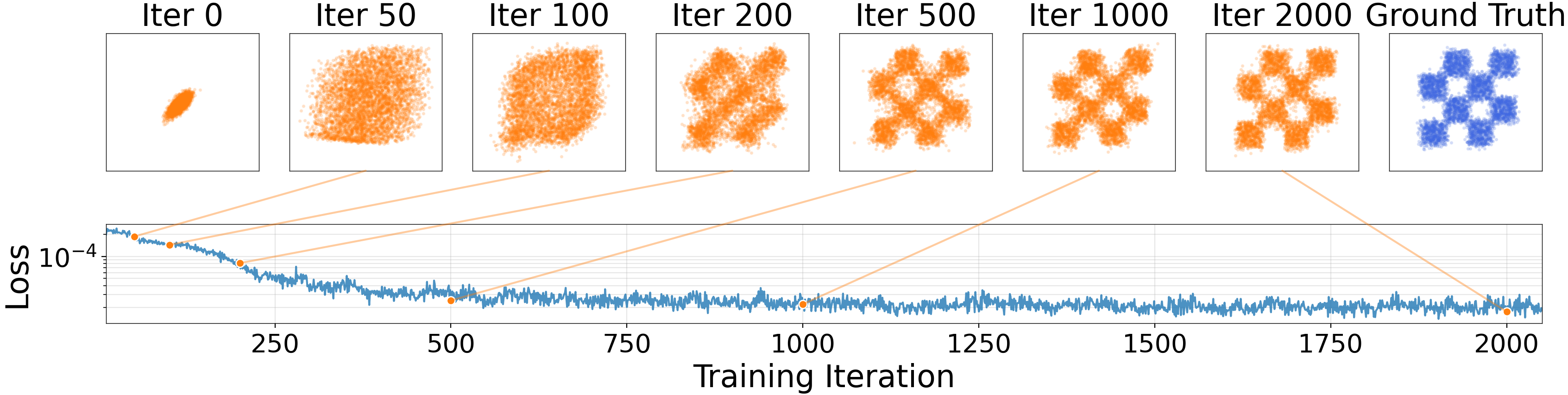}\\[2pt]
        \includegraphics[width=\linewidth]{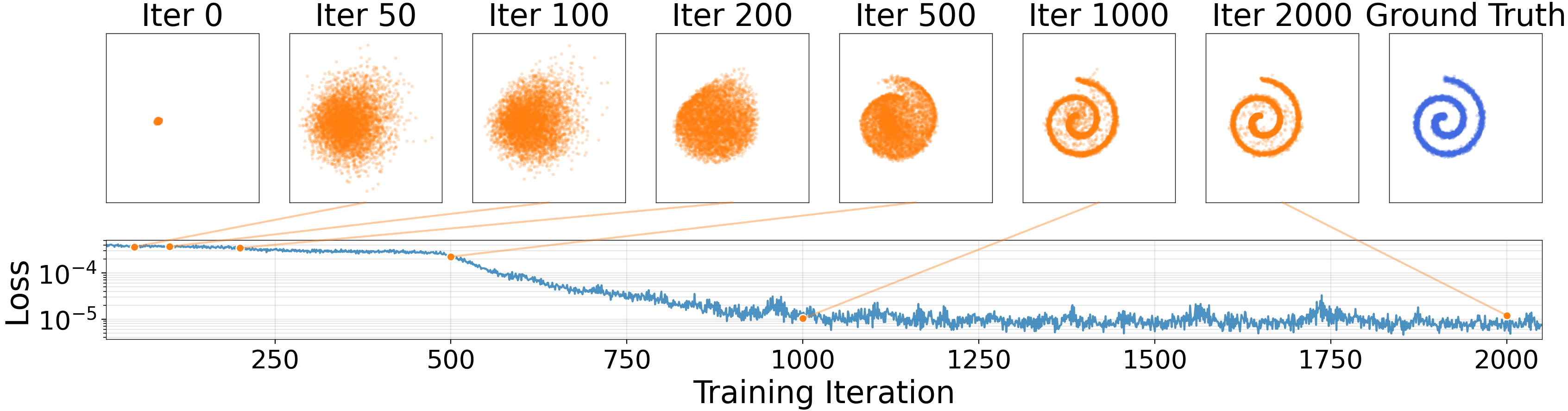}
        \vspace{-.75em}
        \caption{\textbf{Evolution of samples.}
        We show generated points sampled at different training iterations, along with their loss values.
        The loss (whose value equals $\|\V\|^2$) decreases as the distribution converges to the target (y-axis log-scale).
        }
        \label{fig:toy_distributions}
    \end{minipage}
    \vspace{.5em}
\end{figure}

\mypara{Evolution of the generated distribution.}
\cref{fig:drifting_evolution_macro} visualizes a 2D toy case, where $q$ evolves toward a bimodal distribution $p$ at training time, under three initializations.
In this toy example, our method approximates the target distribution without exhibiting mode collapse.
This holds even when $q$ is initialized in a collapsed single-mode state (bottom).
This provides intuition into why our method is robust to mode collapse:
if $q$ collapses onto one mode, other modes of $p$ will attract the samples, allowing them to continue moving and pushing $q$ to continue evolving.

\mypara{Evolution of the samples.}
\cref{fig:toy_distributions} shows the training process on two 2D cases.
A small MLP generator is trained.
The loss, whose value equals $\|\V\|^2$, decreases as the generated distribution converges to the target.
This is in line with our motivation that reducing the drift and pushing towards the equilibrium will approximately yield $p = q$.

\subsection{ImageNet Experiments}
\label{subsec:imagenet_exp}

We evaluate our models on ImageNet 256$\times$256. Ablation studies use a B/2 model on the SD-VAE latent space, trained for 100 epochs.
The drifting loss is in a feature space computed by a latent-MAE encoder.
We report FID \cite{fid} on 50K generated images.
We analyze the results as follows.

\mypara{Equilibrium.}
Our method relies on the equilibrium of the drifting field, \ie, $\V{=}\mathbf{0}$ when $p{=}q$, which is ensured by anti-symmetry; see Eq.~(\ref{eq:asym}). In Table~\ref{tab:ablation_antisymmetry}, we conduct a \textit{destructive} study that intentionally breaks this equilibrium, by breaking the anti-symmetry.
The anti-symmetric case (our ablation default) works well, while other cases fail catastrophically.

Intuitively, for a sample $\x$,
we want attraction from $p$ to be canceled by repulsion from $q$ when $p$ and $q$ match. This equilibrium is not achieved in the destructive cases.
Anti-symmetry is sufficient for this equilibrium, but not necessary: a field that is not anti-symmetric but still vanishes when $p{=}q$ performs on par with our default (Appendix~\ref{sec:appendix_equilibrium}).

\DeclareRobustCommand{\bgpos}[1]{%
  \begingroup
  \setlength{\fboxsep}{1pt}%
  \raisebox{0pt}[\height][0pt]{\colorbox{blue!15}{#1}}%
  \endgroup
}
\DeclareRobustCommand{\bgneg}[1]{%
  \begingroup
  \setlength{\fboxsep}{1pt}%
  \raisebox{0pt}[\height][0pt]{\colorbox{orange!15}{#1}}%
  \endgroup
}
\begin{table}[t]
\vspace{-1.5em}
\hspace{-2em}
\begin{minipage}[t]{0.43\textwidth}
\vspace{0pt}
    \centering
    \tablestyle{4pt}{1.02}
    \tablefontsize
    \begin{tabular}{y{72}x{50}z{25}}
    \toprule
    case & drifting field $\V$ & FID \\
    \midrule
    \textbf{anti-symmetry} & $\V^+ - \V^-$ & \textbf{8.46} \\
    \midrule
    1.5$\times$ attraction & $1.5\V^+ - \V^-$ & 41.05 \\
    1.5$\times$ repulsion & $\V^+ - 1.5\V^-$ & 46.28 \\
    2.0$\times$ attraction & $2\V^+ - \V^-$ & 86.16 \\
    2.0$\times$ repulsion & $\V^+ - 2\V^-$ & 112.84 \\
    attraction-only & $\V^+$ & 177.14 \\
    \bottomrule
    \end{tabular}
    \vspace{.5em}
    \caption{
    \textbf{Importance of equilibrium:}
    breaking the anti-symmetry, and with it the equilibrium, leads to failure. Here, the anti-symmetric case is defined in Eq.~(\ref{eq:V_pos_neg}) and Eq.~(\ref{eq:VZZ});
    other destructive cases are defined in similar ways.
    (Setting: B/2 model, 100 epochs)
    }
    \label{tab:ablation_antisymmetry}
\end{minipage}\hfill
\begin{minipage}[t]{0.56\textwidth}
    \vspace{.5em}
    \centering
    \begin{minipage}[t]{0.48\linewidth}
    \vspace{0pt}
        \centering
        \tablestyle{2.5pt}{1.02}
        \tablefontsize
        \begin{tabular}{c >{\columncolor{blue!15}}c c | c | r}
        \toprule
        $N_\text{c}$ & {$N_\text{pos}$} & $N_\text{neg}$ & $B$ & FID  \\
        \midrule
        64 &  1 & 64 & 4096 & 20.43 \\
        64 &  16 & 64 & 4096 & 10.39 \\
        64 &  32 & 64 & 4096 & 8.97 \\
        64 &  \textbf{64} & 64 & 4096 & \textbf{8.46} \\
        \bottomrule
        \end{tabular}
    \end{minipage}
    \hfill
    \begin{minipage}[t]{0.48\linewidth}
    \vspace{0pt}
        \centering
        \tablestyle{2.5pt}{1.02}
        \tablefontsize
        \begin{tabular}{c c >{\columncolor{orange!15}}c | c | r}
        \toprule
        $N_\text{c}$ & $N_\text{pos}$ & $N_\text{neg}$ & $B$ & FID  \\
        \midrule
        512 & 8 & 8 & 4096 & 11.82 \\
        256 & 16 & 16 & 4096 & 10.16 \\
        128 & 32 & 32 & 4096 & 9.32 \\
        64 & 64 & \textbf{64} & 4096 & \textbf{8.46} \\
        \bottomrule
        \end{tabular}
    \end{minipage}
    \vspace{.5em}
    \caption{
    \textbf{Allocation of positive and negative samples.}
    In both sub-tables, we control the total compute by fixing the epochs (100) and the batch size $B=N_\text{c}{\times}N_\text{neg}$ (4096). Here, $N_\text{c}$ is for class labels. Under the same budget, increasing \bgpos{positive} samples (\textbf{left}) and \bgneg{negative} samples (\textbf{right}) improves generation quality. (Setting: B/2 model, 100 epochs)
    }
    \label{tab:ablation_scaling}
\end{minipage}
\vspace{-2.em}
\end{table}

\mypara{Allocation of Positive and Negative Samples.}
Our method samples positive and negative examples to estimate $\V$ (see Alg.~\ref{alg:training_code}).
Table~\ref{tab:ablation_scaling} studies \bgpos{$N_\text{pos}$} and \bgneg{$N_\text{neg}$} under fixed epochs and batch size $B$. It shows that larger $N_\text{pos}$ and $N_\text{neg}$ is beneficial. Larger sample sizes are expected to improve the accuracy of the estimated $\V$ and hence the generation quality.
This observation aligns with results in contrastive learning \cite{oord2018representation,he2020momentum,chen2020simple}, where larger sample sets improve representation learning.

\mypara{Feature Space for Drifting.}
Our model computes the drifting loss in a feature space (Sec.~\ref{subsec:feature_space}). Table~\ref{tab:ablation_drift_space} compares the feature encoders.
Using the public pre-trained encoders from SimCLR \cite{chen2020simple} and MoCo v2 \cite{chen2020improved}, our method obtains decent results.

\begin{wraptable}{l}{0.56\linewidth}
\vspace{0pt}
\centering
\tablestyle{3pt}{1.0}
\tablefontsize
\begin{tabular}{l | c c c c | r}
\toprule
\multicolumn{1}{c}{} & \multicolumn{4}{c}{feature encoder ($\phi$)} & \\
\cmidrule(lr){2-5}
\multicolumn{1}{c}{SSL method} & arch & block & width & \multicolumn{1}{c}{SSL ep.} & FID \\
\midrule
SimCLR & ResNet & bottleneck & 256 & 800 & 11.05 \\
MoCo-v2 & ResNet & bottleneck & 256 & 800 & 8.41 \\
\midrule
\rowcolor{black!7} latent-MAE & ResNet & basic & 256 & 192 & 8.46 \\
latent-MAE & ResNet & basic & 384 & 192 & 7.26 \\
latent-MAE & ResNet & basic & 512 & 192 & 6.49 \\
latent-MAE & ResNet & basic & 640 & 192 & 6.30 \\
\midrule
latent-MAE & ResNet & basic & 640 & 1280 & 4.28 \\
latent-MAE + cls ft & ResNet & basic & 640 & 1280 & \textbf{3.36} \\
\bottomrule
\end{tabular}
\vspace{.5em}
\caption{\textbf{Feature space for drifting.}
Standard SimCLR and MoCo encoders are competitive; our customized latent-MAE performs best and improves with wider models and longer pre-training.
(Setting: B/2 generator, 100 epochs; shaded row: ablation default.)
}
\label{tab:ablation_drift_space}
\end{wraptable}

These standard encoders operate in the pixel domain, which requires running the VAE decoder at training.
To circumvent this, we pre-train a ResNet-style model with the MAE objective \cite{he2022masked}, directly on the latent space.
The feature space produced by this ``latent-MAE'' performs strongly (Table~\ref{tab:ablation_drift_space}). Increasing the MAE encoder width and the number of pre-training epochs both improve generation quality; fine-tuning it with a classifier (`cls ft') boosts further to 3.36 FID.

The comparison in Table~\ref{tab:ablation_drift_space} shows that the quality of the feature encoder plays an important role.
We hypothesize that this is because our method depends on a kernel $k(\cdot, \cdot)$ (see Eq.~(\ref{eq:kernel})) to measure sample similarity. Samples that are closer in feature space generally yield stronger drift, providing richer training signals. This goal is aligned with the motivation of self-supervised learning.
A strong feature encoder reduces the occurrence of a nearly ``flat'' kernel (\ie, $k(\cdot, \cdot)$ vanishes because all samples are far away).

On the other hand, we report that we were unable to make our method work on ImageNet without a feature encoder. In this case, the kernel may fail to effectively describe similarity, even in the presence of a latent VAE. We leave further study of this limitation for future work (see also Appendix~\ref{sec:appendix_encoder_discussion}).

\mypara{Contribution of Each Design.}
In Table~\ref{tab:ladder} (Appendix~\ref{sec:theory_mmd_relation}), we start from a classical kernel-based generator (GMMN \cite{li2015generative}), which does not produce reasonable results in our setting (214.6 FID), and add the designs of this paper one at a time.
Computing the field in our multi-scale feature space (Sec.~\ref{subsec:feature_space}) gives 18.61 FID; drift normalization, guidance, and our normalized drifting field further improve it to 8.46.
These intermediate results are ablations of our method, rather than existing baselines: to our knowledge, training a generator from scratch with a kernel-based objective in the feature space of a frozen encoder first appears in this paper (Appendix~\ref{sec:kernel_generators}).

\begin{wraptable}{r}{0.35\linewidth}
\vspace{0pt}
    \tablestyle{3pt}{1.0}
    \tablefontsize
    \begin{tabular}{l | c c | c}
    \toprule
    case    & arch & epoch  & FID \\
    \midrule
    baseline (Table~\ref{tab:ablation_drift_space})   & B/2  & 100  & 3.36 \\
    \midrule
    longer  & B/2  & 320  & 2.51 \\
    longer + hyper-param & B/2  & 1280 & 1.75 \\
    larger  & L/2  & 1280 & \textbf{1.54} \\
    \bottomrule
    \end{tabular}
    \vspace{.5em}
    \caption{
    \textbf{From ablation to final setting}. We train our model for more epochs, adjust hyper-parameters for this regime, and use a
larger model size.
    }
\label{tab:ablation_final}
\end{wraptable}

\mypara{System-level Comparisons.} In addition to the ablation setting, we train stronger variants and summarize them in Table~\ref{tab:ablation_final}.
We compare with previous methods in Table~\ref{tab:in256_latent_pixel} (left).

Our method achieves \textbf{1.54} FID with \textit{native} 1-NFE generation. It outperforms all previous 1-NFE methods, which are based on approximating diffusion-/flow-based trajectories.
Notably, our Base-size model competes with previous XL-size models.
Our best model (FID 1.54) uses a CFG scale of 1.0, which corresponds to ``no CFG’’ in flow-based methods.
Our CFG formulation exhibits a tradeoff between FID and IS (see \ref{section:exp_cfg}), similar to standard CFG.

\newcommand\headspace{\hspace{.2em}}
\colorlet{mygray}{black!40}
\newcommand{\cg}{\color{mygray}}
\begin{table}[!t]
    \centering
    \begin{minipage}[t]{0.49\textwidth}
        \vspace{0pt}
        \centering
        \tablestyle{2pt}{1.025}
        \tablefontsize
        \resizebox{\linewidth}{!}{%
        \begin{tabular}{@{}l c c c r r@{}}
        \toprule
        method & space & params & NFE & FID$\downarrow$ & IS$\uparrow$ \\    \midrule
        \rowcolor[gray]{0.9} \multicolumn{6}{l}{\textit{Multi-step Diffusion/Flows}} \\
        \headspace \cg DiT-XL/2 \tiny\cite{peebles2023scalable}  & \cg \tiny SD-VAE & \cg \tiny 675M+49M & \cg \tiny 250$\times$2 & \cg 2.27 & \cg 278.2 \\
        \headspace \cg SiT-XL/2 \tiny\cite{ma2024sit} & \cg \tiny SD-VAE & \cg \tiny 675M+49M & \cg \tiny 250$\times$2 & \cg 2.06 & \cg 270.3 \\
        \headspace \cg SiT-XL/2+REPA \tiny\cite{yu2024representation}& \cg \tiny SD-VAE & \cg \tiny 675M+49M & \cg \tiny 250$\times$2 & \cg 1.42 & \cg 305.7 \\
        
        \headspace \cg LightningDiT-XL/2 \tiny\cite{yao2025reconstruction}& \cg \tiny VA-VAE & \cg \tiny 675M+70M & \cg \tiny 250$\times$2 & \cg 1.35 & \cg 295.3 \\
        \headspace \cg DDT-XL/2\tiny\cite{wang2025ddt}& \cg \tiny SD-VAE & \cg \tiny 675M+49M & \cg \tiny 250$\times$2 & \cg 1.26 & \cg 310.6 
        \\
        \headspace \cg RAE+DiT$^{\text{DH}}$-XL/2 \tiny\cite{zheng2025diffusion} & \cg \tiny RAE & \cg \tiny 839M+415M & \cg \tiny 50$\times$2 & \cg \textbf{1.13} & \cg 262.6 \\
        \midrule
        \rowcolor[gray]{0.9} \multicolumn{6}{l}{\textit{Single-step Diffusion/Flows}} \\
        \headspace iCT-XL/2 \tiny\cite{song2023improved}       & \tiny SD-VAE & \tiny 675M+49M & \tiny 1 & 34.24 & -- \\
        \headspace Shortcut-XL/2  \tiny\cite{frans2024one} & \tiny SD-VAE & \tiny 675M+49M & \tiny 1 & 10.60 & 102.7 \\
        \headspace MeanFlow-XL/2 \tiny\cite{geng2025mean}  & \tiny SD-VAE & \tiny 676M+49M & \tiny 1 & 3.43 & 247.5 \\
        \headspace TiM-XL/2 \tiny\cite{wang2025transition}  & \tiny SD-VAE & \tiny 664M+49M & \tiny 1 & 3.26 & 210.3 \\
         \headspace $\alpha$-Flow-XL/2 \tiny\cite{zhang2025alphaflow}  & \tiny SD-VAE & \tiny 676M+49M & \tiny 1 & 2.58 & -- \\
         
        \headspace AdvFlow-XL/2 \tiny\cite{lin2025adversarial} & \tiny SD-VAE & \tiny 673M+49M & \tiny 1 & 2.38 & 284.2 \\
        \headspace iMeanFlow-XL/2 \tiny\cite{geng2025improved}    & \tiny SD-VAE & \tiny 610M+49M & \tiny 1 & 1.72 & 282.0 \\
        \midrule
        \rowcolor[gray]{0.9} \multicolumn{6}{l}{\textit{Drifting Models}} \\
        \headspace \textbf{Drifting Model, B/2} & \tiny SD-VAE & \tiny 133M+49M & \tiny 1 & 1.75 & 263.2 \\
        \headspace \textbf{Drifting Model, L/2} & \tiny SD-VAE & \tiny 463M+49M & \tiny 1 & \textbf{1.54} & 258.9 \\
        \bottomrule
        \end{tabular}
        }
    \end{minipage}\hfill
    \begin{minipage}[t]{0.49\textwidth}
        \vspace{0pt}
        \centering
        \tablestyle{2pt}{1.0}
        \tablefontsize
        \resizebox{\linewidth}{!}{%
        \begin{tabular}{@{}l c c c c r@{}}
        \toprule
        method & space & params & NFE & FID$\downarrow$ & IS$\uparrow$ \\
        \midrule
        \rowcolor[gray]{0.9} \multicolumn{6}{l}{\textit{Multi-step Diffusion/Flows}} \\
        \headspace \cg SiD, UViT/2 \tiny\cite{hoogeboom2023simple} & \cg pix & \cg 2.5B & \cg 1000$\times$2 & \cg 2.44 & \cg 256.3  \\
        \headspace \cg VDM++, UViT/2 \tiny\cite{kingma2023understanding} & \cg pix & \cg 2.5B & \cg 256$\times$2 & \cg 2.12 & \cg 267.7 \\
        \headspace \cg SiD2, UViT/2 \tiny\cite{hoogeboom2024simpler} & \cg pix & \cg -- & \cg 512$\times$2 & \cg 1.73 & \cg -- \\
        \headspace \cg SiD2, UViT/1 \tiny\cite{hoogeboom2024simpler} & \cg pix & \cg -- & \cg 512$\times$2 & \cg \textbf{1.38} & \cg -- \\
        \headspace \cg JiT-G/16 \tiny\cite{li2025back} & \cg pix & \cg 2B & \cg 100$\times$2 & \cg 1.82 & \cg 292.6 \\
        \headspace \cg PixelDiT/16 \tiny\cite{yu2025pixeldit} & \cg pix & \cg 797M & \cg 200$\times$2 & \cg 1.61 & \cg 292.7 \\
        \midrule
        \rowcolor[gray]{0.9} \multicolumn{6}{l}{\textit{Single-step Diffusion/Flows}} \\
        \headspace EPG-L/16 \tiny\cite{lei2025there} & pix & 540M & 1 & 8.82 & -- \\
        \headspace pMF-H/32 \tiny\cite{lu2026one} & pix & 959M & 1 & 2.48 & 284.9 \\
        \midrule
        \rowcolor[gray]{0.9} \multicolumn{6}{l}{\textit{GANs}} \\
        \headspace BigGAN \tiny\cite{brock2018large} & pix & 112M & 1 & 6.95 & 152.8 \\
        \headspace GigaGAN \tiny\cite{kang2023scaling} & pix & 569M & 1 & 3.45 & 225.5 \\
        \headspace StyleGAN-XL \tiny\cite{sauer2022stylegan} & pix & 166M & 1 & 2.30 & 265.1 \\
        \midrule
        \rowcolor[gray]{0.9} \multicolumn{6}{l}{\textit{Drifting Models}} \\
        \headspace \textbf{Drifting Model, B/16} & pix & 134M & 1 & 1.76 & 299.7 \\
        \headspace \textbf{Drifting Model, L/16} & pix & 464M & 1 & \textbf{1.61} & 307.5 \\
        \bottomrule
        \end{tabular}
        }
    \end{minipage}
    \vspace{.75em}
    \caption{\textbf{System-level comparison on ImageNet 256$\times$256.}
    FID on 50K images, with CFG when applicable.
    \textbf{Left}: latent-space methods (params: generator $+$ decoder).
    \textbf{Right}: pixel-space methods (params: generator only).
    All generators are trained from scratch (\ie, not distilled).}
    \label{tab:in256_latent_pixel}
    \vspace{-.5em}
\end{table}

We provide uncurated qualitative results in Appendix~\ref{sec:appendix_qualitative_results}, Fig.~\ref{fig:latent_uncurated_samples_1}-\ref{fig:latent_uncurated_samples_4}, with CFG 1.0.
Moreover, Fig.~\ref{fig:comparison_imf_1}-\ref{fig:comparison_imf_5} show a side-by-side comparison with improved MeanFlow (iMF) \cite{geng2025improved}, a recent state-of-the-art one-step method. Our best model uses a classification-finetuned latent-MAE encoder. We also provide ablations without classification fine-tuning and an additional DINO-FD evaluation in Appendix~\ref{sec:appendix_no_cls_ft}, and Precision/Recall in Appendix~\ref{sec:appendix_pr}.

\mypara{Pixel-space Generation.} Our method can naturally work \textit{without} the latent VAE, \ie, the generator $f$ directly produces 256$\times$256$\times$3 images. The feature encoder is applied on the generated images for drifting loss. We use a configuration similar to the latent variant; details are in Appendix \ref{sec:appendix_impl}.

Table~\ref{tab:in256_latent_pixel} (right) compares different pixel-space generators.
Our \textit{one-step}, \textit{pixel-space} method achieves \textbf{1.61} FID, which outperforms or competes with previous multi-step methods.
Comparing with other one-step, pixel-space methods (GANs), our method achieves 1.61 FID using only 87G FLOPs; by comparison, StyleGAN-XL produces 2.30 FID using 1574G FLOPs.
More ablations are in \ref{section:exp_pixel_space}.

\subsection{Robotics Control}
Beyond image generation, we further evaluate our method on robotics control. Our experiment protocols follow \textit{Diffusion Policy} \citep{chi2025diffusion}. At the core of Diffusion Policy is a multi-step, diffusion-based generator; we replace it with our one-step Drifting Model.
We directly compute drifting loss on the \textit{raw} representations for control, using no feature space. Results are in Table~\ref{tab:robotics_combined}.

\begin{table}[t]
\centering
\tablestyle{3.5pt}{1.0}
\tablefontsize
\begin{minipage}[t]{0.49\linewidth}
\vspace{0pt}
\centering
\begin{tabular}{llcc}
\toprule
& & Diffusion Policy & \textbf{Drifting Policy} \\
Task & Setting & NFE: 100 & \textbf{NFE: 1} \\
\midrule
\multicolumn{4}{l}{\textit{\textbf{Single-Stage Tasks} (State \& Visual Observation)}} \\
\midrule
\multirow{2}{*}{Lift} & State & {0.98} & \textbf{1.00} \\
 & Visual & {\textbf{1.00}} & \textbf{1.00} \\
\addlinespace[0.3em]
\multirow{2}{*}{Can} & State & {0.96} & \textbf{0.98} \\
 & Visual & {0.97} & \textbf{0.99} \\
\addlinespace[0.3em]
\multirow{2}{*}{ToolHang} & State & {0.30} & \textbf{0.38} \\
 & Visual & {\textbf{0.73}} & 0.67 \\
\addlinespace[0.3em]
\multirow{2}{*}{PushT} & State & {\textbf{0.91}} & 0.86 \\
 & Visual & {0.84} & \textbf{0.86} \\
\bottomrule
\end{tabular}
\end{minipage}\hfill
\begin{minipage}[t]{0.49\linewidth}
\vspace{0pt}
\centering
\begin{tabular}{llcc}
\toprule
& & Diffusion Policy & \textbf{Drifting Policy} \\
Task & Setting & NFE: 100 & \textbf{NFE: 1} \\
\midrule
\multicolumn{4}{l}{\textit{\textbf{Multi-Stage Tasks} (State Observation)}} \\
\midrule
\multirow{2}{*}{BlockPush} & Phase 1 & {0.36} & \textbf{0.56} \\
 & Phase 2 & {0.11} & \textbf{0.16} \\
\addlinespace[0.3em]
\multirow{4}{*}{Kitchen} & Phase 1 & {\textbf{1.00}} & \textbf{1.00} \\
 & Phase 2 & {\textbf{1.00}} & \textbf{1.00} \\
 & Phase 3 & {\textbf{1.00}} & 0.99 \\
 & Phase 4 & {\textbf{0.99}} & 0.96 \\
\bottomrule
\end{tabular}
\end{minipage}
\vspace{0.5em}
\caption{\textbf{Robotics Control: Comparison with Diffusion Policy.}
The evaluation protocol follows Diffusion Policy \citep{chi2025diffusion}.
This table involves four single-stage tasks and two multi-stage tasks.
Success rates are reported as the average over the last 10 checkpoints.
}
\label{tab:robotics_combined}
\end{table}

Our 1-NFE model is broadly competitive with, and often matches or exceeds, Diffusion Policy with 100 NFE. This comparison suggests that Drifting Models can serve as a promising generative model across domains.

\section{Discussion and Conclusion}
\label{sec:conclusion}

We present \emph{Drifting Models}, a $\V$-centric framework for one-step generative modeling.
At the core of our model is the idea of modeling the evolution of pushforward distributions \textit{during training}, governed by an explicit drifting field $\V$.
This allows us to focus on the update rule, \ie, $\x_{i+1} = \x_{i} + \Delta \x_i$, during the iterative training process.
This is in contrast with diffusion-/flow-based models, which perform the iterative update at \textit{inference} time. Our method naturally performs one-step inference.

Given that our methodology is substantially different, many open questions remain.
For example, although we show that $q=p \Rightarrow \V=\mathbf{0}$, the converse implication does not generally hold in theory.
While our designed $\V$ performs well empirically and we provide some identifiability arguments in Appendix~\ref{sec:theory_identifiability}, a fully general statement of $\V_{p,q}\equiv\mathbf{0}\Rightarrow p=q$ remains open.

From a practical standpoint, although our paper presents an effective instantiation of drifting modeling, many of our design decisions may remain sub-optimal. For example, the design of the drifting field and its kernels, the feature encoder, and the generator architecture remain open for future exploration.
Our method also has limitations: it currently relies on a feature encoder for high-dimensional data, and our image-generation experiments are confined to class-conditional ImageNet (see Appendix~\ref{sec:appendix_encoder_discussion}).

From a broader perspective, our work reframes iterative neural network training as a mechanism for distribution evolution, in contrast to the differential equations underlying diffusion-/flow-based models.
We hope this perspective inspires future realizations of this mechanism.
 
\begin{ack}
We greatly thank Google TPU Research Cloud (TRC) for granting us access to TPUs. We thank Michael Albergo, Ziqian Zhong, Yilun Xu, Zhengyang Geng, Hanhong Zhao, Jiangqi Dai, Alex Fan, and Shaurya Agrawal for helpful discussions.
Mingyang Deng is partially supported by funding from MIT-IBM Watson AI Lab.
Competing interests: Kaiming He is also with Google DeepMind (part-time), and Yilun Du is also with NVIDIA (part-time). Mingyang Deng has interned at Pika, Meta, and Google DeepMind.
\end{ack}

\bibliography{references}

\bibliographystyle{abbrvnat}

\clearpage
\newpage
\appendix
\section{Additional Implementation Details}
\label{sec:appendix_impl}

\cref{tab:hypers_all} summarizes the configurations and hyper-parameters for ablation studies and system-level comparisons. 
We provide detailed experimental configurations for reproducibility. All ablation studies share a common default setup, while system-level comparisons use scaled-up configurations. More implementation details are described as follows.

\begin{table*}[t]
    \centering
    \caption{\textbf{Configurations for ImageNet 256$\times$256}.
    }
    \vspace{.5em}
    \label{tab:hypers_all}
    \tablestyle{1pt}{1.03}
    \scriptsize
    \begin{tabular}{@{}y{82}|x{54}|x{54}x{54}|x{54}x{54}@{}}
    \toprule
    & \shortstack{\textbf{ablation}\\\textbf{default}}
    & \shortstack{\textbf{B/2, latent}\\Table~\ref{tab:in256_latent_pixel} (left)}
    & \shortstack{\textbf{L/2, latent}\\Table~\ref{tab:in256_latent_pixel} (left)}
    & \shortstack{\textbf{B/16, pixel}\\Table~\ref{tab:in256_latent_pixel} (right)}
    & \shortstack{\textbf{L/16, pixel}\\Table~\ref{tab:in256_latent_pixel} (right)} \\
    \midrule
    \rowcolor[gray]{0.9} \multicolumn{6}{l}
    {\textit{\textbf{Generator Architecture}}} \\
    arch & DiT-B/2 & DiT-B/2 & DiT-L/2 & DiT-B/16 & DiT-L/16 \\
    input size & 32$\times$32$\times$4 & 32$\times$32$\times$4 & 32$\times$32$\times$4 & 256$\times$256$\times$3 & 256$\times$256$\times$3 \\
    patch size & 2$\times$2 & 2$\times$2 & 2$\times$2 & 16$\times$16 & 16$\times$16 \\
    hidden dim & 768 & 768 & 1024 & 768 & 1024 \\
    depth & 12 & 12 & 24 & 12 & 24 \\
    register tokens & 16 & 16 & 16 & 16 & 16 \\
    style embedding tokens & 32 & 32 & 32 & 32 & 32 \\
    \midrule
    \rowcolor[gray]{0.9} \multicolumn{6}{l}{\textit{\textbf{Feature Encoder for Drifting Loss}}} \\
    arch & ResNet & ResNet & ResNet & ResNet + ConvNeXt-V2 & ResNet + ConvNeXt-V2 \\
    SSL pre-train method & latent-MAE & latent-MAE & latent-MAE & pixel-MAE & pixel-MAE  \\
    ResNet: input size & 32$\times$32$\times$4 & 32$\times$32$\times$4 &
    32$\times$32$\times$4 &
    256$\times$256$\times$3 &
    256$\times$256$\times$3 \\
    ResNet: conv$_\text{1}$ stride & 1 & 1 & 1 & 8 & 8 \\
    ResNet: base width & 256 & 640 & 640 & 640 & 640 \\
    ResNet: block type & \multicolumn{5}{c}{basic} \\
    ResNet: blocks / stage & \multicolumn{5}{c}{[3, 4, 6, 3]} \\
    ResNet: size / stage & \multicolumn{5}{c}{[32$^2$, 16$^2$, 8$^2$, 4$^2$]} \\
    MAE: masking ratio & \multicolumn{5}{c}{50\%} \\
    MAE: pre-train epochs & 192 & 1280 & 1280 & 1280 & 1280 \\
    classification finetune & No & 3k steps & 3k steps & 3k steps & 3k steps \\
    \midrule
    \rowcolor[gray]{0.9} \multicolumn{6}{l}{\textit{\textbf{Generator Optimizer}}} \\
    optimizer & \multicolumn{5}{c}{AdamW ($\beta_1$ = 0.9, $\beta_2$ = 0.95)} \\
    learning rate & 2e-4 & 4e-4 & 4e-4 & 2e-4 & 4e-4 \\
    weight decay & 0.01 & 0.0 & 0.01 & 0.01 & 0.01 \\
    warmup steps & 5k & 10k & 10k & 10k & 10k \\
    gradient clip & 2.0 & 2.0 & 2.0 & 2.0 & 2.0 \\
    training steps & 30k & 200k & 200k & 100k & 100k \\
    training epochs & 100 & 1280 & 1280 & 640 & 640 \\
    EMA decay & 0.999 & \multicolumn{4}{c}{\{0.999, 0.9995, 0.9998, 0.9999\}}\\
    \midrule
    \rowcolor[gray]{0.9} \multicolumn{6}{l}{\textit{\textbf{Drifting Loss Computation}}} \\
    class labels $N_\text{c}$ & 64 & 128 & 128 & 128 & 128 \\
    positive samples $N_\text{pos}$ & 64 & 128 & 64 & 128 & 128 \\
    generated samples $N_\text{neg}$ & 64 & 64 & 64 & 64 & 64 \\
    effective batch $B$ ($N_\text{c}{\times}N_\text{neg}$) & 4096 & 8192 & 8192 & 8192 & 8192 \\
    temperatures $\tau$ & \multicolumn{5}{c}{\shortstack{\{0.02, 0.05, 0.2\}: one loss per $\tau$;\\sum all loss terms}} \\
    \midrule
    \rowcolor[gray]{0.9} \multicolumn{6}{l}{\textit{\textbf{CFG Configuration}}} \\
    train: CFG $\alpha$ range & $[1, 4]$ & $[1, 4]$ & $[1, 4]$ & $[1, 4]$ & $[1, 4]$ \\
    train: CFG $\alpha$ sampler
    & $p_{-3}$
    & $p_{-5}$
    & $\tfrac12(\delta_1{+}p_{-3})$
    & $p_{-5}$
    & $p_{-5}$ \\
    \multicolumn{6}{@{}l@{}}{\scriptsize \emph{Notation:} $p_{-k}(\alpha)\propto\alpha^{-k}$ on $[1,4]$; $\delta_1$ always samples $\alpha{=}1$.} \\
    train: uncond samples $N_\text{uncond}$ & 16 & 32 & 32 & 32 & 32 \\
    inference: CFG $\alpha$ search & \multicolumn{5}{c}{[1.0, 3.5]} \\
    \bottomrule
    \end{tabular}
\end{table*}
 
\subsection{Pseudo-code for Computing Drifting Field $\V$}
\label{sec:appendix_compute_V}

Alg.~\ref{alg:drift_code} provides the pseudo-code for computing $\V$. The computation is based on taking empirical means in Eq.~(\ref{eq:VZZ}) and (\ref{eq:kernel}), which are implemented as softmax over $\y$-sample axis. In practice, we further normalize over the $\x$-sample axis, also implemented as softmax on the same logit matrix. 
We ablate its influence in \ref{sec:appendix_kernel_normalizatio}.

\renewcommand{\yn}{\y^{-}}
\renewcommand{\yp}{\y^{+}}
\newcommand{\B}{\mathcal{B}}

It is worth noting that this implementation preserves the desired property of $\V$.
In principle, this implementation can be viewed as a Monte Carlo estimation of a drifting field:
\begin{equation}
\V_{p,q}(\x)=\mathbb E_{\B,p,q}[\tilde K_\B(\x, \yp)\tilde K_\B(\x, \yn)(\yp-\yn)],    
\end{equation}
where $\B$ consists of other samples in the batch and $\tilde K_\B$ denote normalizing the distance based on statistics within $\B$. This $\V$ also satisfies $\V_{p,p}(\x)=\mathbf 0$, since when $p=q$, the term  $\tilde K_\B(\yp, x)\tilde K_\B(\yn,x)(\yp-\yn)$ cancels out with the term $\tilde K_\B(\yn, x)\tilde K_\B(\yp,x)(\yn-\yp)$. 

\begin{algorithm}[t]
\caption{Computing the drifting field $\V$.}
\label{alg:drift_code}
\begin{lstlisting}
def compute_V(x, y_pos, y_neg, T):
  # x: [N, D]
  # y_pos: [N_pos, D]
  # y_neg: [N_neg, D]
  # T: temperature

  # compute pairwise distance
  dist_pos = cdist(x, y_pos)  # [N, N_pos]
  dist_neg = cdist(x, y_neg)  # [N, N_neg]
  
  # ignore self (if y_neg is x)
  dist_neg += eye(N) * 1e6

  # compute logits
  logit_pos = -dist_pos / T
  logit_neg = -dist_neg / T

  # concat for normalization
  logit = cat([logit_pos, logit_neg], dim=1)
  # normalize along both dimensions
  A_row = logit.softmax(dim=-1)
  A_col = logit.softmax(dim=-2)
  A = sqrt(A_row * A_col) 

  # back to [N, N_pos] and [N, N_neg]
  A_pos, A_neg = split(A, [N_pos,], dim=1)

  # compute the weights
  W_pos = A_pos  # [N, N_pos]
  W_neg = A_neg  # [N, N_neg]
  W_pos *= A_neg.sum(dim=1,keepdim=True)
  W_neg *= A_pos.sum(dim=1,keepdim=True)

  drift_pos = W_pos @ y_pos # [N_x, D]
  drift_neg = W_neg @ y_neg # [N_x, D]

  V = drift_pos - drift_neg
  return V 
\end{lstlisting}
\end{algorithm} 
\subsection{Generator Architecture}
\label{sec:appendix_generator}

\paragraph{Input and output.}
The input to the generator consists of random noise along with conditioning:
\[
f_\theta: (\e, c, \alpha) \mapsto \x
\]
where $\e$ denotes random variables, $c$ is a class label, and $\alpha$ is the CFG strength.
$\e$ may consist of both continuous random variables (\eg, Gaussian noise) and discrete ones (\eg, uniformly distributed integers; see random style embeddings).
For latent-space models, the output $\mathbf{x} \in \mathbb{R}^{32\times 32\times 4}$ is in the SD-VAE latent space.
For pixel-space models, the output $\mathbf{x} \in \mathbb{R}^{256\times 256\times 3}$ is directly an image.

\paragraph{Transformer.}
We adopt a DiT-style Transformer~\citep{peebles2023scalable}. Following~\citep{yao2025reconstruction}, we use SwiGLU~\citep{shazeer2020glu}, RoPE~\citep{su2024roformer}, RMSNorm~\citep{zhang2019root}, and QK-Norm~\citep{henry2020query}.
The input Gaussian noise is patchified into 256$=$16$\times$16 tokens (patch size 2$\times$2 for latent, 16$\times$16 for pixel).
Conditioning $(c, \alpha)$ is processed by adaLN, as well as by in-context conditioning tokens.
The output tokens are unpatchified back to the target shape.

\paragraph{In-context tokens.}
Following~\cite{li2025back}, we prepend 16 learnable tokens to the sequence for in-context conditioning \cite{peebles2023scalable}.
These tokens are formed by summing the projected conditioning vector with positional embeddings.

\paragraph{Random style embeddings.}
Our framework allows arbitrary noise distributions beyond Gaussians.
Inspired by StyleGAN~\cite{sauer2022stylegan}, we introduce an additional 32 ``style tokens'': each of which is a random index into a codebook of 64 learnable embeddings.
These are summed and added to the conditioning vector.
This does not change the sequence length and introduces negligible overhead in terms of parameters and FLOPs. This table reports the effect of style embeddings on our ablation default:

\begin{center}
\tablestyle{6pt}{1.02}
\tablefontsize
\begin{tabular}{c|cc}
\toprule
 & w/o style & w/ style \\
\midrule
FID & 8.86 & \textbf{8.46} \\
\bottomrule
\end{tabular}
\end{center}

In contrast to diffusion-/flow-based methods, our method can naturally handle different types of noise or random variables.
With random style embeddings, the input random variables consist of two parts: (1) Gaussian noise, and (2) discrete indices for style embeddings. Our model $f$ produces the pushforward distribution of their joint distribution. 

\subsection{Implementation of ResNet-style MAE}
\label{sec:appendix_latent_mae}

In addition to standard self-supervised learning models (MoCo \cite{he2020momentum}, SimCLR\cite{chen2020simple}), we develop a customized ResNet-style MAE model as the feature encoder for drifting loss.

\paragraph{Overview.}
Unlike standard MAE \cite{he2022masked}, which is based on ViT \cite{Dosovitskiy2021}, our MAE trains a convolutional ResNet that provides multi-scale features.
For latent-space models, the input and output have dimension 32$\times$32$\times$4;
for pixel-space models, the input and output have dimension 256$\times$256$\times$3.

Our MAE consists of a ResNet-style encoder paired with a deconvolutional decoder in a U-Net-style \cite{ronneberger2015u} encoder-decoder architecture.
We only use the ResNet-style encoder for feature extraction when computing the drifting loss.

\paragraph{MAE Encoder.}
The encoder follows a classical ResNet \cite{he2016deep} design.
It maps an input to multi-scale feature maps (4 scales in ResNet):
\[
\text{Encoder}: \mathbf{x} \mapsto \{\mathbf{f}_1, \mathbf{f}_2, \mathbf{f}_3, \mathbf{f}_4\}
\]
Here, a feature map $\mathbf{f}_i$ has dimension $H_i{\times}W_i{\times}C_i$, with $H_i{\times}W_i \in \{32^2, 16^2, 8^2, 4^2\}$ and $C_i \in \{C, 2C, 4C, 8C\}$ for a base width $C$.

The architecture follows standard ResNet~\citep{he2016deep} design, with GroupNorm (GN)~\citep{wu2018group} used in place of BatchNorm (BN)~\citep{ioffe2015batch}.
All residual blocks are ``basic'' blocks (\ie, each consisting of two $3{\times}3$ convolutions).
Following the standard ResNet-34 \citep{he2016deep}: 
the encoder has a 3${\times}$3 convolution (without downsampling) and 4 stages with $[3, 4, 6, 3]$ blocks; downsampling (stride 2) happens at the first block of stages 2 to 4.

For latent-space (\ie, latent-MAE), the input of this ResNet is 32$\times$32$\times$4; for pixel-space, the 256$\times$256$\times$3 input is first patchified (by a 8$\times$8 patch) into 32$\times$32$\times$192. The ResNet operates on the input with $H{\times}W$ $=$ 32$\times$32.

\paragraph{MAE Decoder.}
The decoder returns to the input shape via deconvolutions and skip connections:
\[
\text{Decoder}: \{\mathbf{f}_4, \mathbf{f}_3, \mathbf{f}_2, \mathbf{f}_1\} \mapsto \hat{\mathbf{x}}.
\]

It starts with a $3{\times}3$ convolutional block on $\mathbf{f}_4$, followed by 4 upsampling blocks.
Each upsampling block performs: bilinear 2${\times}$2 upsampling $\to$ concatenating with encoder's skip connection $\to$ GN $\to$ two $3{\times}3$ convolutions with GN and ReLU.
A final 1${\times}$1 convolution produces the output channels. For the pixel-space, the decoder unpatchifies back to the original resolution after the last layer.

\paragraph{Masking.}
The MAE is trained to reconstruct randomly masked inputs.
Unlike the ViT-based MAE~\citep{he2022masked}, which removes the masked tokens from the sequence, we simply zero out masked patches.
For the input of a shape $H{\times}W$ $=$ 32$\times$32 (in either the latent- or pixel-based case), we mask 2${\times}$2 patches by zeroing.
Each patch is independently masked with 50\% probability.

\paragraph{MAE training.}
We minimize the $\ell_2$ reconstruction loss on the masked regions.
We use AdamW \cite{Loshchilov2019} with learning rate $4{\times}10^{-3}$ and a batch size of 8192. EMA with decay 0.9995 is used.
Following \cite{he2022masked}, we apply random resized crop augmentation to the input (for the latent setting, images are augmented before being passed through the VAE encoder).

\paragraph{Classification fine-tuning.}
For our best feature encoder (last row of Table~\ref{tab:ablation_drift_space}), we fine-tune the MAE model with a linear classifier head.
The loss is $\lambda \mathcal{L}_{\text{cls}} + (1-\lambda)\mathcal{L}_{\text{recon}}$. We fine-tune all parameters in this MAE for 3k iterations, where $\lambda$ follows a linear warmup schedule, increasing from $0$ to $0.1$ over the first 1k iterations and remaining constant at $0.1$ for the rest of the training.

\subsection{Other Pretrained Feature Encoders}
\label{sec:appendix_other_encoders}

In addition to our customized MAE, we also evaluate other feature encoders for computing the drifting loss.

\paragraph{MoCo and SimCLR.} We evaluate publicly available self-supervised encoders trained on ImageNet in pixel space: MoCo \cite{he2020momentum,chen2020improved} SimCLR \cite{chen2020simple}. We use the ResNet-50 variant. 
For latent-space generation, we apply the VAE decoder to map generator outputs from latent space (32${\times}$32${\times}$4) to pixel space (256${\times}$256${\times}$3) before feature extraction.
Gradients are backpropagated through both the feature extractor and the VAE decoder.

\paragraph{MAE with ConvNeXt-V2.} In our pixel-space generator, we also investigate ConvNeXt-V2 \citep{woo2023convnext} as the feature encoder. We note that ConvNeXt-V2 is a self-supervised pre-trained model using the MAE objective, followed by classification fine-tuning. Like ResNet, ConvNeXt-V2 is a multi-stage architecture.

\subsection{Multi-scale Features for Drifting Loss}
\label{sec:appendix_feature_extraction}

Given an image, the feature encoder produces feature maps at multiple scales, with multiple spatial locations per scale.
We compute one drifting loss per feature (\eg, per scale and/or per location).
Specifically, we compute the kernel, the drift, and the resulting loss independently for each feature. The resulting losses are summed.

For each stage in a ResNet, we extract features from the output of every 2 residual blocks, together with the final output.
This yields a set of feature maps, each of shape $H_i{\times}W_i{\times}C_i$. For each feature map, we produce:
\begin{enumerate}[itemsep=2pt, topsep=2pt, parsep=4pt, partopsep=4pt,label=(\alph*)]
    \item $H_i{\times}W_i$ vectors, one per location (each $C_i$-dim);
    \item 1 global mean and 1 global std (each $C_i$-dim);
    \item $\frac{H_i}{2}{\times}\frac{W_i}{2}$ vectors of means and $\frac{H_i}{2}{\times}\frac{W_i}{2}$ vectors of stds (each $C_i$-dim), computed over 2${\times}$2 patches;
    \item $\frac{H_i}{4}{\times}\frac{W_i}{4}$ vectors of means and $\frac{H_i}{4}{\times}\frac{W_i}{4}$ vectors of stds (each $C_i$-dim), computed over 4${\times}$4 patches.
\end{enumerate}
In addition, for the encoder's input ($H_0{\times}W_0{\times}C_0$), we compute the mean of squared values ($x^2$) per channel and obtain a $C_0$-dim vector.

All resulting vectors here are $C_i$-dim.
We compute one drifting loss for each of these $C_i$-dim vectors. All these losses, in addition to the vanilla drifting loss without $\phi$, are summed.
This table compares the effect of these designs on our ablation default:
\begin{center}
\tablestyle{6pt}{1.02}
\tablefontsize
\begin{tabular}{c|ccc}
\toprule
 & (a,b) & (a-c) & (a-d) \\
\midrule
FID & 9.58 & 9.10 & \textbf{8.46} \\
\bottomrule
\end{tabular}
\end{center}
This shows that our method benefits from richer feature sets.
We note that once the feature encoder is run, the computational cost of our drifting loss is negligible: computing multi-scale, multi-location losses incurs little overhead compared to computing a single loss.

\subsection{Feature and Drift Normalization}
\label{sec:appendix_feature_and_loss_normalization}

To balance the multiple loss terms from multiple features, we perform normalization for each feature $\phi_j$, where, $\phi_j$ denotes a feature at a specific spatial location within a given scale (see \ref{sec:appendix_feature_extraction}). 
Intuitively, we want to perform normalization such that the kernel $k(\cdot, \cdot)$ and the drift $\V$ are insensitive to the absolute magnitude of features.
This allows our model to robustly support different feature encoders (see Table~\ref{tab:ablation_drift_space}) as well as a rich set of features from one encoder.

\paragraph{Feature Normalization.} Consider a feature $\phi_j \in \mathbb{R}^{C_j}$. We define a normalization scale $S_j \in \mathbb{R}^{1}$ and the normalized feature is denoted by:
\begin{equation}
\tilde{\phi}_j := {\phi}_j / S_j.
\end{equation}
When using $\tilde{\phi}_j$, the $\ell_2$ distance computed in Eq.~(\ref{eq:kernel}) is:
\begin{equation}
dist_j(\x, \y) = \|\tilde{\phi}_j(\x) - \tilde{\phi}_j(\y)\|,
\end{equation}
where $\x$ denotes a generated sample and $\y$ denotes a positive/negative sample, and $\tilde{\phi}_j(\cdot)$ means extracting their feature at $j$. We want the \textit{average} distance to be $\sqrt{C_j}$:
\begin{equation}
\mathrm{E}_\x \mathrm{E}_\y [ dist_j(\x, \y)] \approx \sqrt{C_j}.
\end{equation}
To achieve this, we set the normalization scale $S_j$ as:
\begin{equation}
S_j = \frac{1}{\sqrt{C_j}} \mathrm{E}_\x \mathrm{E}_\y [ \|{\phi}_j(\x) - {\phi}_j(\y)\| ]
\end{equation}
In practice, we use all $\x$ and $\y$ samples in a batch to compute the empirical mean in place of the expectation. We reuse the \texttt{cdist} computation in Alg.~\ref{alg:drift_code} for computing the pairwise distances. We apply stop-gradient to $S_j$, because this scalar is conceptually computed from samples from the previous batch.

With the normalized feature, the kernel in Eq.~(\ref{eq:kernel}) is set as:
\begin{equation}
k(\x, \y) = \exp\left(-\frac{1}{\tilde{\tau_j}} \|\tilde{\phi}_j(\x) - \tilde{\phi}_j(\y)\|\right),
\end{equation}
where $\tilde{\tau_j}:=\tau{\cdot}\sqrt{C_j}$. By doing so, the value of temperature $\tau$ does not depend on the feature magnitude or feature dimensionality. We set $\tau \in$ \{0.02, 0.05, 0.2\} (discussed next).

\paragraph{Drift Normalization.} When using the feature $\phi_j$, the resulting drift is in the same feature space as $\phi_j$, denoted as $\V_j$. We perform a drift normalization on $\V_j$, for each feature $\phi_j$. Formally, we define a normalization scale $\lambda_j \in \mathbb{R}^1$ and denote:
\begin{equation}
\tilde{\V}_j:=\V_j / \lambda_j.
\end{equation}
Again, we want the normalized drift to be insensitive to the feature magnitude:
\begin{equation}
\mathbb{E} \left[ \frac{1}{C_j}\| \tilde{\V}_j \|^2 \right] \approx 1.
\end{equation}
To achieve this, we set $\lambda_j$ as:
\begin{equation}
\label{eq:lambda}
\lambda_j = \sqrt{\mathbb{E} \left[ \frac{1}{C_j} \| \V_j \|^2 \right]}.
\end{equation}
In practice, the expectation is replaced with the empirical mean computed over the entire batch.

With the normalized feature and normalized drift, the drifting loss of the feature $\phi_j$ is:
\begin{equation}
\label{eq:mse_loss_j}
\mathcal{L}_j = \text{MSE}(\tilde\phi_j({\x}) -  \texttt{sg}(\tilde\phi_j({\x}) + \tilde{\V}_j)),
\end{equation}
where MSE denotes mean squared error.
The overall loss is the sum across all features:
$\mathcal{L} = \sum_j {\mathcal{L}_j}$.

\paragraph{Multiple temperatures.}
Using normalized feature distances, the value of temperature $\tau$ determines \textit{what is considered ``nearby''}.
To improve robustness across different features and across different pretrained models we study, we adopt multiple temperatures.

Formally, for each $\tau$ value, we compute the normalized drift as described above, denoted by $\tilde{\V}_{j,\tau}$. Then we compute an aggregated field: $\tilde{\V}_j \leftarrow \sum_\tau \tilde{\V}_{j,\tau}$, and use it for the loss in \cref{eq:mse_loss_j}.

This table shows the effect of multiple temperatures on our ablation default:
\begin{center}
    \tablestyle{4pt}{1.05}
    \tablefontsize
    \begin{tabular}{l | x{32} x{32} x{32} | c}
    \toprule
    $\tau$ & 0.02 & 0.05 & 0.2 & \{0.02, 0.05, 0.2\} \\
    \midrule
    FID & 10.62 & \textbf{8.67} & 8.96 & \textbf{8.46} \\
    \bottomrule
    \end{tabular}
\end{center}
Using multiple temperatures can achieve slightly better results than using a single optimal temperature. We fix $\tau \in $ \{0.02, 0.05, 0.2\} and do not require tuning this hyperparameter across different configurations.

\paragraph{Normalization across spatial locations.}
For a feature map of resolution $H_i{\times}W_i$, there are $H_i{\times}W_i$ per-location features.
Separately computing the normalization for each location would be slow and unnecessary.
We assume that features at different locations within the same feature map share the same normalization scale. Accordingly, we concatenate all $H_i{\times}W_i$ locations and compute the normalization scale over all of them.
The feature normalization and drift normalization are both performed in this way.
\subsection{Classifier-Free Guidance (CFG)}
\label{sec:appendix_cfg}

To support CFG, at training time, we sample a CFG strength $\alpha$ for each class label and generate samples from an $\alpha$-conditioned generator:
\[
\x=f_\theta(\e,c,\alpha), \qquad \x\sim q_\theta(\cdot|c,\alpha).
\]
The main text suppresses this explicit $\alpha$ argument for readability; here we write it out to match the implementation.
The scalar $\alpha$ is embedded by the generator's conditioning pathway.

When computing the drifting field for generated samples from $q_\theta(\cdot|c,\alpha)$, we use the other $N_\text{neg}{-}1$ samples from $q_\theta(\cdot|c,\alpha)$ as negatives, together with $N_\text{unc}$ additional \mbox{unconditional} samples (real images from random classes).
The unconditional samples are weighted by a factor $w(\alpha)$ when computing the kernel.
The effective negative distribution is therefore also $\alpha$-conditioned:
\[
\tilde{q}(\cdot|c,\alpha) \triangleq \frac{(N_\text{neg}{-}1) \cdot q_\theta(\cdot|c,\alpha) + N_\text{unc}  w(\alpha) \cdot p_{\text{data}}(\cdot|\varnothing)}{(N_\text{neg}{-}1) + N_\text{unc} w(\alpha)}.
\]
Comparing this expression with Eqs.~(\ref{eq:cfg1})--(\ref{eq:cfg2}), we have:
\[
\gamma = \frac{N_\text{unc} w(\alpha)}{(N_\text{neg}{-}1) + N_\text{unc} w(\alpha)}
\]
and
\[
\alpha = \frac{1}{1-\gamma} = \frac{(N_\text{neg}{-}1) + N_\text{unc} w(\alpha)}{N_\text{neg}{-}1}.
\]
Given a CFG strength $\alpha$, we compute $w(\alpha)$ accordingly and use it to weight the unconditional samples in the kernel.
The same weighting $w(\alpha)$ is also applied when computing the global distance normalization.

We train our model with CFG-conditioning \cite{geng2025improved}. At each iteration, we randomly sample $\alpha$ from a predefined distribution (Table~\ref{tab:hypers_all}) and compute the corresponding $w(\alpha)$.

At inference time, we specify a value of $\alpha$. The inference-time computation remains one-step (1-NFE).

\subsection{Sample Queue}
\label{sec:appendix_sample_queue}

Our method requires access to randomly sampled \textit{real} (positive/unconditional) data. This can be implemented using a specialized data loader. 
Instead, we adopt a \textit{sample queue} of cached data, similar to the queue used in MoCo \cite{he2020momentum}.
This implementation samples data in a statistically similar way to a specialized data loader.
For completeness, we describe our implementation as follows, while noting that a data loader would be a more principled solution.  

For each class label, we keep a queue of size 128; for unconditional samples (used in CFG), we maintain a separate global queue of size 1000.
At each training step, we push the latest 64 new real (positive/unconditional) samples, alongside their labels, into the corresponding queues; the earliest ones are dequeued.
When sampling, positive samples are drawn from the queue of the corresponding class, and unconditional samples are drawn from the global queue.
We sample without replacement.

\subsection{Training Loop}
\label{sec:appendix_training_loop}

In summary, in the training loop, each step proceeds as:
\begin{enumerate}[itemsep=2pt, topsep=2pt, parsep=2pt, partopsep=2pt]
    \item Sample a batch ($N_c$) of class labels.
    \item For each label $c$, sample a CFG scale $\alpha$.
    \item Sample a batch ($N_\text{neg}$) of noise $\e$. Feed $(\e, c, \alpha)$ to the generator $f$ to produce generated samples $\x\sim q_\theta(\cdot|c,\alpha)$;
    \item Sample positive samples (same class, $N_\text{pos}$) and unconditional samples (for CFG, $N_\text{unc}$);
    \item Extract features on all generated, positive, and unconditional samples
    \item Compute the drifting loss using the features.
    \item Run backpropagation and parameter update.
\end{enumerate}

\section{Additional Experimental Results}

\begin{table}[t]
    \centering
    \caption{\textbf{Ablations on pixel-space generation.} 
    We study generation directly in pixel space (without VAE). Applying the same MAE recipe as in latent space yields higher FID, indicating that pixel-space generation is more challenging. Combining MAE with ConvNeXt-V2 helps close this gap. Latent-space results shown for reference. The results below follow the ablation setting (B/16 model for pixel-space, 100 epochs).
    }
    \label{tab:ablation_latent_vs_pixel}
    \tablestyle{6pt}{1.05}
    \tablefontsize
    \begin{tabular}{l r r}
    \toprule
    & \multicolumn{2}{c}{FID (100-epoch)} \\
    \cmidrule{2-3}
    feature encoder $\phi$ & latent (B/2) & pixel (B/16) \\
    \midrule
    MAE (width 256, epoch 192) & \cg 8.46 & 32.11 \\
    MAE (width 640, epoch 1280) + cls ft. & \cg 3.36 & 9.35 \\
    ~+ MAE w/ ConvNeXt-V2 & \cg - & \textbf{3.70} \\
    \bottomrule
    \end{tabular}
\end{table}
\begin{table}[t]
    \centering
    \caption{\textbf{Pixel-space generation: from ablation to final setting.} Beyond the ablation setting, we compare the settings that lead to the results in Table~\ref{tab:in256_latent_pixel} (right).
    }
\label{tab:ablation_final_pixel}
    \tablestyle{6pt}{1.05}
    \tablefontsize
    \begin{tabular}{l |l r| c}
    \toprule
    case & arch & ep & FID \\
    \midrule
    (a) baseline (from Table~\ref{tab:ablation_latent_vs_pixel}) & B/16 & 100 & 3.70 \\
    \midrule
    (b) longer + hyper-param. & B/16 & 320 & 2.19 \\
    (c) longer & B/16 & 640 & 1.76 \\
    (d) larger model & L/16  & 640 & \textbf{1.61} \\
    \bottomrule
    \end{tabular}
\end{table}
\begin{table}[t]
    \centering
    \caption{\textbf{Ablation on kernel normalization.}
    Softmax normalization over both the $\x$ and $\y$ axes performs better. On the other hand, even using no normalization performs decently, showing the robustness of our method.
    (Setting: B/2 model, 100 epochs)
    }
    \label{tab:kernel_normalization}
    \tablestyle{6pt}{1.0}
    \tablefontsize
    \begin{tabular}{l r}
    \toprule
    kernel normalization & FID \\
    \midrule
    softmax over $\x$ and $\y$ (default) & \textbf{8.46} \\
    softmax over $\y$ & 8.92 \\
    no normalization & 10.54 \\
    \bottomrule
    \end{tabular}
\end{table}

\subsection{Ablations on Pixel-Space Generation}
\label{section:exp_pixel_space}

We provide more ablations on pixel-space generation in Table~\ref{tab:ablation_latent_vs_pixel} and \ref{tab:ablation_final_pixel}.
Table~\ref{tab:ablation_latent_vs_pixel} compares the effect of the feature encoder on the pixel-space generator. 
It shows that the choice of feature encoder plays a more significant role in pixel-space generation quality.
A weaker MAE encoder yields an FID of 32.11, whereas a stronger MAE encoder improves performance to an FID of 9.35.
We further add another feature encoder, ConvNeXt-V2 \cite{woo2023convnext}, which is also pre-trained with the MAE objective. This further improves the result to an FID of 3.70.

Table~\ref{tab:ablation_final_pixel} reports the results of training longer and using a larger model. Due to limited time, we train pixel-space models for 640 epochs (\vs~the latent counterpart's 1280); we expect that longer training would yield further improvements.
We achieve an FID of 1.61 for pixel-space generation. This is our result in the main paper (Table~\ref{tab:in256_latent_pixel}).

\begin{figure}[t]
    \centering
    \includegraphics[width=0.65\linewidth]{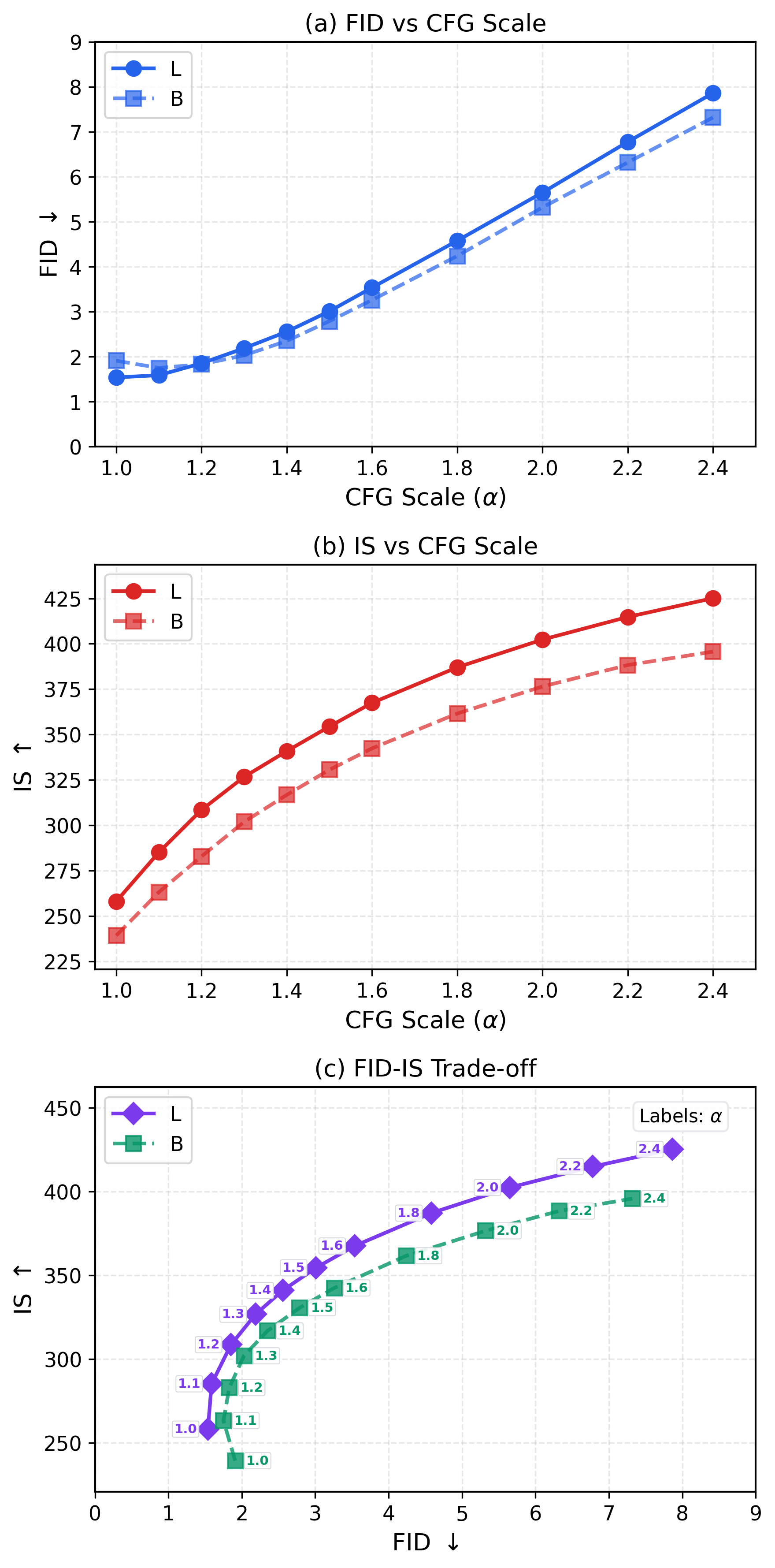}
    \caption{\textbf{Effect of CFG scale $\alpha$.} 
    \textbf{(a)}: FID vs.~$\alpha$.
    \textbf{(b)}: IS vs.~$\alpha$.
    \textbf{(c)}: IS vs.~FID.
    We show the L/2 (solid) and B/2 (dashed) models.
    Consistent with common observations in diffusion-/flow-based models, the CFG scale effectively trades off distributional coverage (as reflected by FID) against per-image quality (measured by IS).
    Notably, with the L/2 model, the optimal FID is achieved at $\alpha{=}1.0$, which is often regarded as ``w/o CFG'' in diffusion-/flow-based models. For B/2, the optimal FID is achieved at $\alpha{=}1.1$.
    }
    \label{fig:cfg_tradeoff}
\end{figure}

\subsection{Without Classification Fine-tuning, and DINO-FD Evaluation}
\label{sec:appendix_no_cls_ft}

Our main ImageNet results use a latent-MAE encoder further fine-tuned with ImageNet classification supervision (`cls ft' in Table~\ref{tab:ablation_drift_space}).
Using pretrained perceptual or representation networks as training signals is common in image generation, including networks with classification supervision (\eg, LPIPS-style perceptual losses and ConvNeXt features).
At the same time, because standard FID is itself computed in an ImageNet-trained Inception feature space, class-aligned training features may interact with the evaluation protocol~\cite{kynkaanniemi2022role}.
To separate the effect of our drifting objective from this encoder choice, we report controlled results with the same encoder \emph{before} classification fine-tuning, whose pretraining never saw class labels.
We also report DINO-FD~\cite{stein2023exposing}, a complementary FID-style evaluation based on self-supervised DINO features.

\begin{table}[!ht]
\centering
\begin{minipage}[t]{0.46\textwidth}
\vspace{0pt}
\centering
\caption{\textbf{No classification fine-tuning, L/2 model.} FID of the L/2 latent-space model trained with the same MAE-640 encoder, with and without classification fine-tuning, at different training lengths. The gap narrows as training continues; the no-cls-ft variant remains competitive with prior 1-step methods.}
\label{tab:appendix_no_cls_ft}
\tablestyle{6pt}{1.05}
\tablefontsize
\begin{tabular}{c c c}
\toprule
epochs & w/ cls ft. & w/o cls ft. \\
\midrule
320 & 1.96 & 2.31 \\
640 & 1.68 & 2.02 \\
1280 & \textbf{1.54} & \textbf{1.72} \\
\bottomrule
\end{tabular}
\end{minipage}\hfill
\begin{minipage}[t]{0.50\textwidth}
\vspace{0pt}
\centering
\caption{\textbf{DINO-FD comparison among 1-NFE methods.} We evaluate DINO-FD~\cite{stein2023exposing}, a FID-style metric computed in a self-supervised DINO feature space, for our L/2 latent-space model. Lower is better; numbers for prior 1-NFE methods are taken from \citep{stein2023exposing} and \citep{lee2025decoupled}.}
\label{tab:appendix_dino_fd}
\tablestyle{6pt}{1.05}
\tablefontsize
\begin{tabular}{l c}
\toprule
method (1-NFE) & DINO-FD$\downarrow$ \\
\midrule
StyleGAN-XL & 214.9 \\
DMF L/2 \cite{lee2025decoupled} & 199.7 \\
DMF XL/2 \cite{lee2025decoupled} & 122.3 \\
\midrule
Drifting Model L/2 (CFG 1.0) & 143.4 \\
\textbf{Drifting Model L/2} (CFG 2.0) & \textbf{96.5} \\
\bottomrule
\end{tabular}
\end{minipage}
\end{table}

Table~\ref{tab:appendix_no_cls_ft} reports FID of our L/2 latent-space model trained with the same MAE-640 encoder either with or without classification fine-tuning. Without classification fine-tuning, the model still reaches an FID of 1.72 at 1280 epochs, and the gap to the cls-ft variant shrinks with longer training (from 0.35 at 320 epochs to 0.18 at 1280 epochs). This indicates that the strong performance of our method does not crucially rely on class-aligned features, although such features do provide additional gains.

Table~\ref{tab:appendix_dino_fd} reports DINO-FD, which uses self-supervised DINO features instead of ImageNet classification features. Our L/2 model achieves a DINO-FD of 96.5 with CFG 2.0 and 143.4 with CFG 1.0, which is comparable to or better than the 1-NFE methods that report this metric \citep{stein2023exposing,lee2025decoupled}. Note that DINO-FD and standard FID prefer different CFG scales: standard FID is best at CFG~1.0 (Table~\ref{tab:in256_latent_pixel}), while DINO-FD favors a larger CFG. We report both metrics to make this evaluation dependence explicit.

\subsection{Precision and Recall}
\label{sec:appendix_pr}

To evaluate mode coverage beyond FID, we report improved Precision/Recall~\cite{kynkaanniemi2019improved} computed on 50K generated samples for our four released models in Table~\ref{tab:appendix_pr}; baseline values are taken from published results~\cite{dhariwal2021diffusion,sauer2022stylegan,peebles2023scalable,ma2024sit}.
Our models achieve recall competitive with the multi-step DiT-XL/2 and SiT-XL/2, and higher than that of the one-step GANs.
Together with the 2D examples in Figs.~\ref{fig:drifting_evolution_macro} and \ref{fig:toy_distributions} and the memorization analysis in Table~\ref{tab:appendix_clip_nn}, this supports mode coverage at ImageNet scale.

\begin{table}[!ht]
\centering
\caption{\textbf{Precision and Recall on ImageNet 256$\times$256} (50K samples)~\cite{kynkaanniemi2019improved} for our four released models, compared with published values of one-step GANs and multi-step diffusion/flow models. BigGAN-deep values are from~\cite{dhariwal2021diffusion}.}
\label{tab:appendix_pr}
\tablestyle{6pt}{1.05}
\tablefontsize
\begin{tabular}{l c c c}
\toprule
method & NFE & Precision$\uparrow$ & Recall$\uparrow$ \\
\midrule
BigGAN-deep~\cite{brock2018large} & 1 & 0.87 & 0.28 \\
StyleGAN-XL~\cite{sauer2022stylegan} & 1 & 0.78 & 0.53 \\
DiT-XL/2~\cite{peebles2023scalable} & 250$\times$2 & 0.83 & 0.57 \\
SiT-XL/2~\cite{ma2024sit} & 250$\times$2 & 0.83 & 0.59 \\
\midrule
Drifting Model, latent B/2 & 1 & 0.80 & 0.62 \\
Drifting Model, latent L/2 & 1 & 0.79 & 0.62 \\
Drifting Model, pixel B/16 & 1 & 0.82 & 0.56 \\
Drifting Model, pixel L/16 & 1 & 0.81 & 0.58 \\
\bottomrule
\end{tabular}
\end{table}

\subsection{Ablation on Kernel Normalization}
\label{sec:appendix_kernel_normalizatio}

In Eq.~(\ref{eq:VZZ}), our drifting field is weighted by normalized kernels, which can be written as:
\begin{equation}
\V(\x) = \mathbb E_{p,q} [\tilde{k}(\x, \yp) \tilde{k}(\x, \yn) (\yp - \yn)],
\end{equation}
where $\tilde{k}(\cdot, \cdot)=\frac{1}{Z}k(\cdot, \cdot)$ denotes the normalized kernel.
In principle, this normalization is approximated by a softmax operation over the axis of $\y$ samples. Our implementation (Alg.~\ref{alg:drift_code}) further applies softmax over the axis of $\x$ samples. We compare these designs, along with another variant without normalization ($Z=1$).

Table~\ref{tab:kernel_normalization} compares the three designs. Using the $\y$-only softmax performs well (8.92 FID), whereas using both $\x$ and $\y$ softmax improves the result (8.46 FID). On the other hand, even without normalization, performance remains decent, demonstrating the robustness of our method.

We note that all three variants satisfy the equilibrium condition $\V_{p,q}(\x) = \mathbf{0}$ when $p=q$. This explains why all variants perform reasonably well and why even the destructive setting (no normalization) avoids catastrophic failure.

\subsection{Ablation on Kernel Family}
\label{sec:appendix_kernel_family}

\begin{table}[!ht]
\centering
\caption{\textbf{Ablation on kernel family.}
We compare the Gaussian kernel, which admits the score-difference interpretation in Sec.~\ref{sec:theory_implicit_drifting}, with the Laplacian kernel used in our default ImageNet instantiation.
(Setting: B/2 model, 100 epochs.)
}
\label{tab:kernel_family}
\tablestyle{6pt}{1.0}
\tablefontsize
\begin{tabular}{l r}
\toprule
kernel & FID \\
\midrule
Laplacian (default) & \textbf{8.46} \\
Gaussian & 9.19 \\
\bottomrule
\end{tabular}
\end{table}

Table~\ref{tab:kernel_family} shows that both kernels work reasonably well, but the Laplacian kernel used in practice performs slightly better than the Gaussian one (8.46 vs.\ 9.19 FID). This is useful context for the discussion in Sec.~\ref{sec:theory_implicit_drifting}: the Gaussian kernel gives a clean score-difference interpretation, whereas our best practical kernel does not itself reduce to that form.

\subsection{Equilibrium without Anti-symmetry}
\label{sec:appendix_equilibrium}

The destructive cases in Table~\ref{tab:ablation_antisymmetry} break the anti-symmetry and, at the same time, make $\V$ non-zero when $p=q$.
To separate the two effects, we consider the field $g_p(\x)(\V^+{-}\V^-)$, where $g_p(\x)>0$ is the sum of the kernel values between $\x$ and the positive samples.
This field is not anti-symmetric, but it still vanishes when $p=q$.
Under our ablation setting, it achieves 8.48 FID, on par with our default (8.46).
This is consistent with Sec.~\ref{subsec:instantiation_field}: our method only requires $\V=\mathbf{0}$ when $p=q$, for which anti-symmetry is a sufficient condition.

\subsection{Ablation on CFG}
\label{section:exp_cfg}

In \cref{fig:cfg_tradeoff}, we investigate the CFG scale $\alpha$ used at inference time.
It shows that the CFG formulation developed for our models exhibits behavior similar to that observed in diffusion-/flow-based models.
Increasing the CFG scale leads to higher IS values, whereas beyond the FID sweet spot, further increases in IS come at the cost of worse FID.

Notably, with our best model (L/2), the optimal FID is achieved at $\alpha{=}1.0$, which is often regarded as ``w/o CFG'' in diffusion-/flow-based models (even though their ``w/o CFG'' setting can reduce NFE by half). 
While our method need not run an unconditional model at inference time (in contrast to standard CFG), training is influenced by the use of unconditional real samples as negatives.

\subsection{Disentangling CFG from the Drifting Field}
\label{sec:appendix_no_cfg}

We perform an ablation on the effect of CFG, by completely removing CFG from training (i.e., $N_\text{unc}=0$ unconditional negatives and $\alpha\equiv 1$), and re-train under our ablation setting (B/2, 100 epochs) to isolate the contribution of the drifting field $\V$ alone.

\begin{table}[!ht]
\centering
\caption{\textbf{No-CFG training ablation.} We disable training-time CFG ($N_\text{unc}=0$, $\alpha\equiv 1$) and compare against our default. The drifting objective remains effective on its own; CFG provides additional gains, more so when features are weaker.
(Setting: B/2 generator, 100 epochs.)
}
\label{tab:appendix_no_cfg}
\tablestyle{6pt}{1.05}
\tablefontsize
\begin{tabular}{l c c}
\toprule
feature encoder $\phi$ & w/ CFG (default) & w/o CFG \\
\midrule
latent-MAE width 256 (ablation baseline) & \textbf{8.46} & 15.99 \\
latent-MAE width 640 + cls ft. & \textbf{3.36} & 4.04 \\
\bottomrule
\end{tabular}
\end{table}

Table~\ref{tab:appendix_no_cfg} shows that the drifting objective remains effective without CFG: with the strongest feature encoder, removing CFG only degrades FID from 3.36 to 4.04. The gap is larger with weaker features (8.46 vs.\ 15.99), suggesting that CFG provides relatively more help when the base drift signal is weaker.

\subsection{Training Compute Compared with iMF}
\label{sec:appendix_compute_imf}

We compare the training compute of our L/2 latent-space model with that of improved MeanFlow (iMF-XL/2)~\cite{geng2025improved}.
We count FLOPs with the cost analysis of the XLA compiler.
For each training sample, our model takes 920 GFLOPs: a forward and backward pass of the generator, and of the frozen encoder on the generated sample.
iMF-XL/2 takes 1{,}315 GFLOPs per sample, which we measure on a Transformer of the same depth and width, including the Jacobian--vector product in its objective.
Therefore, the 800 training epochs of iMF-XL/2 cost as much as ${\approx}$1{,}143 of our 1{,}280 epochs.
At that point (step 180K), our FID is 1.55, compared with 1.72 reported for iMF-XL/2.
This count does not include the encoder forward passes on real samples, whose features could be cached.
If they are included, the 800 epochs of iMF-XL/2 cost as much as ${\approx}$860 of our epochs (step ${\approx}$135K); our FID is already 1.62 at step 130K.
These numbers also do not include the one-time pre-training of our MAE encoder, which is shared by our latent-space models.
It takes ${\approx}$22 hours on the same hardware, compared with ${\approx}$37 hours for training the L/2 generator (Appendix~\ref{sec:appendix_compute}).

\begin{figure}[t]
    \centering
    \begin{minipage}{0.49\linewidth}
        \centering
        \makebox[0.28\linewidth]{\hspace{1pt}\scriptsize \textbf{\textcolor{orange}{Generated}}}%
        \makebox[0.72\linewidth]{\scriptsize \textbf{\textcolor{blue}{Retrieved}}}
    \end{minipage}
    \hfill
    \begin{minipage}{0.49\linewidth}
        \centering
        \makebox[0.28\linewidth]{\hspace{5pt}\scriptsize \textbf{\textcolor{orange}{Generated}}}%
        \makebox[0.72\linewidth]{\hspace{5pt}\scriptsize \textbf{\textcolor{blue}{Retrieved}}}
    \end{minipage}
    \vspace{1pt}

    \includegraphics[width=0.49\linewidth]{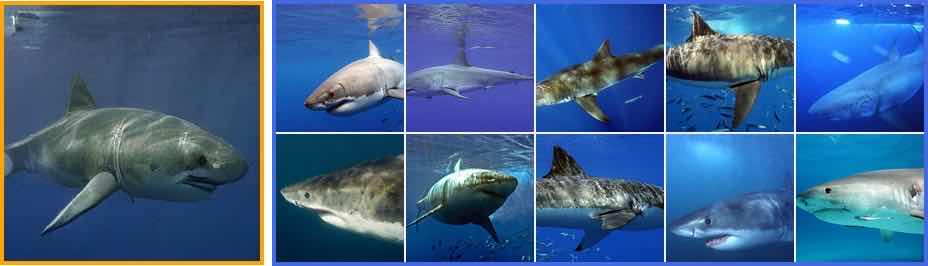}\hfill
    \includegraphics[width=0.49\linewidth]{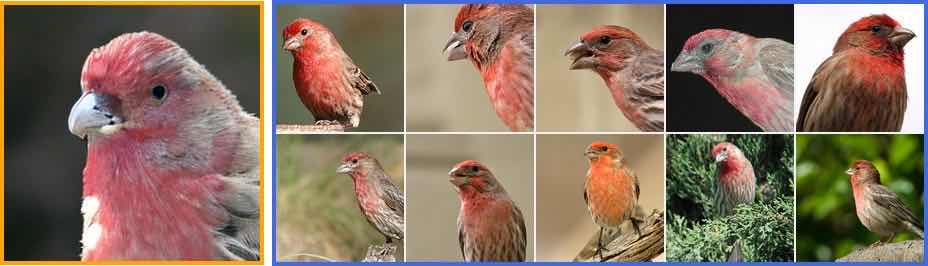}\\[1pt]
    \includegraphics[width=0.49\linewidth]{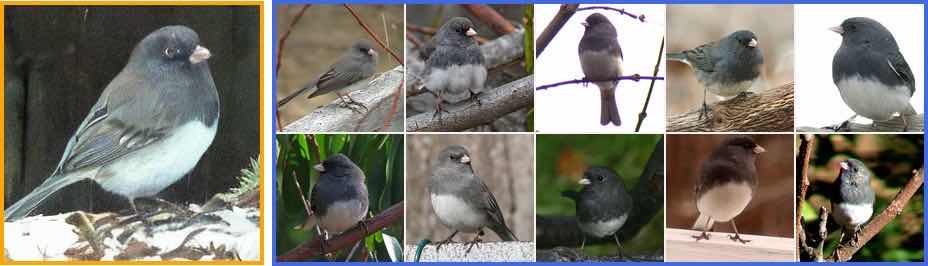}\hfill
    \includegraphics[width=0.49\linewidth]{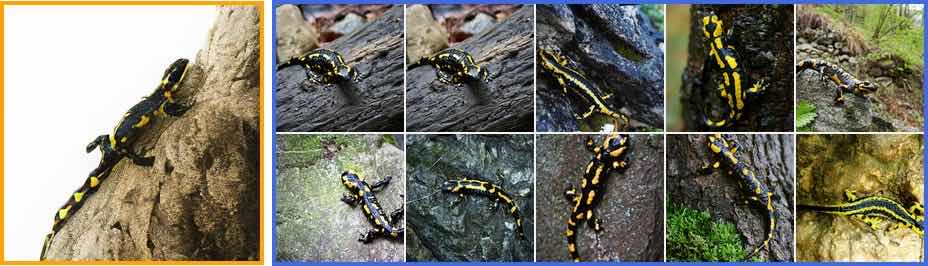}\\[1pt]
    \includegraphics[width=0.49\linewidth]{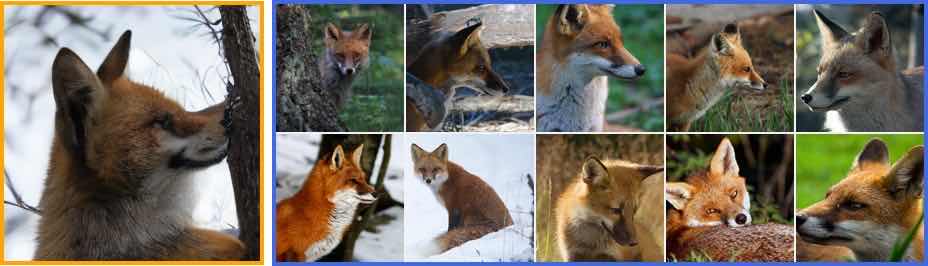}\hfill
    \includegraphics[width=0.49\linewidth]{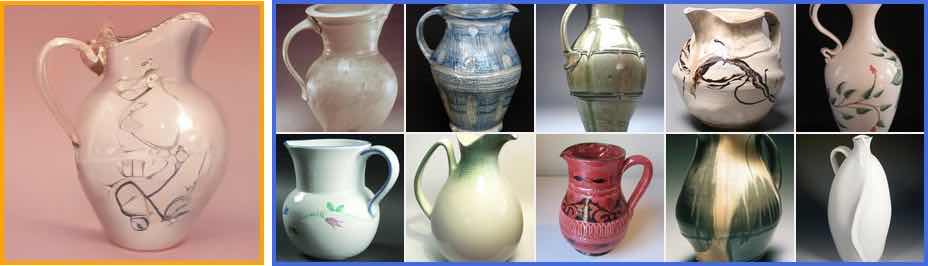}\\[1pt]
    \includegraphics[width=0.49\linewidth]{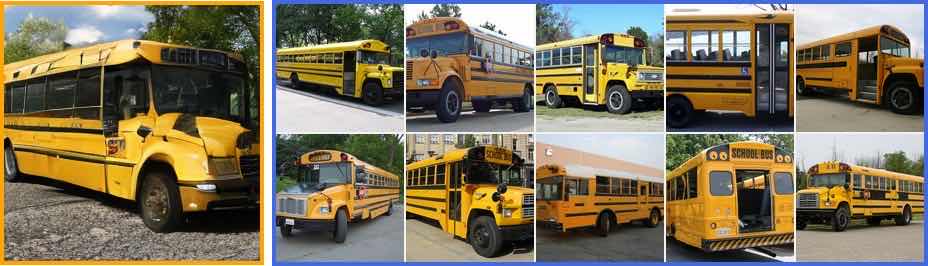}\hfill
    \includegraphics[width=0.49\linewidth]{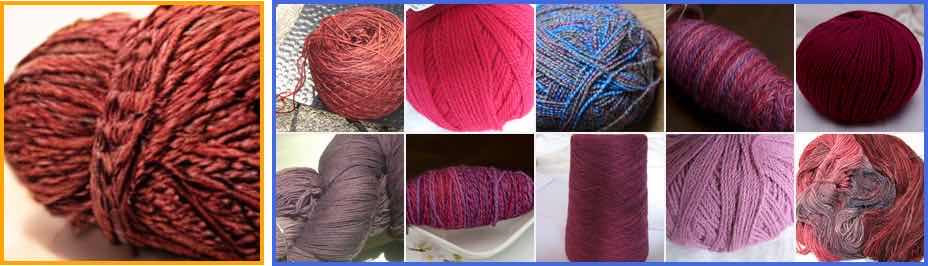}\\[1pt]
    \includegraphics[width=0.49\linewidth]{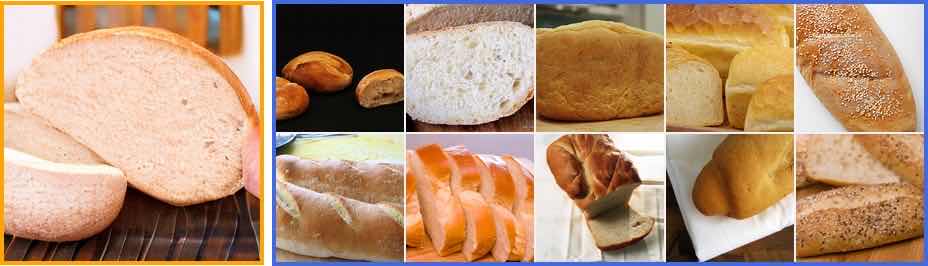}\hfill
    \includegraphics[width=0.49\linewidth]{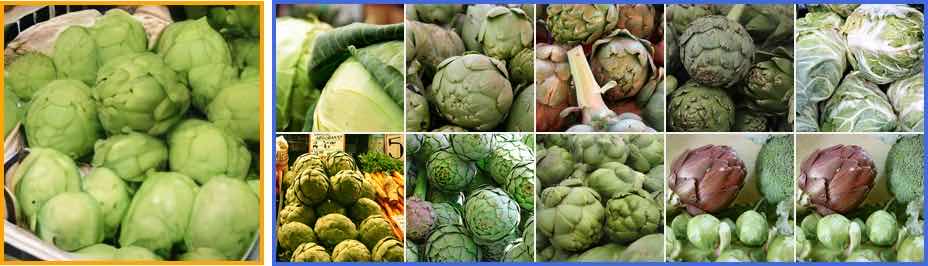}\\[1pt]
    \includegraphics[width=0.49\linewidth]{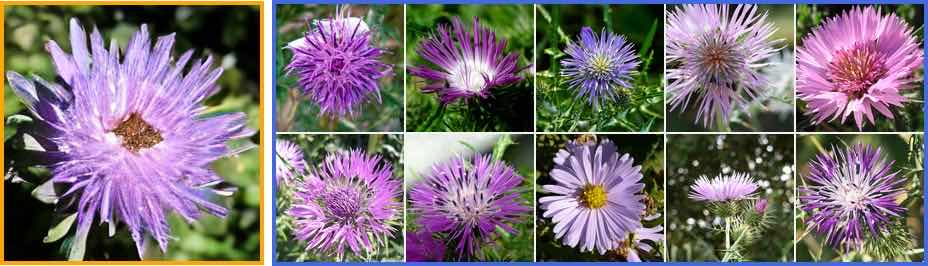}\hfill
    \includegraphics[width=0.49\linewidth]{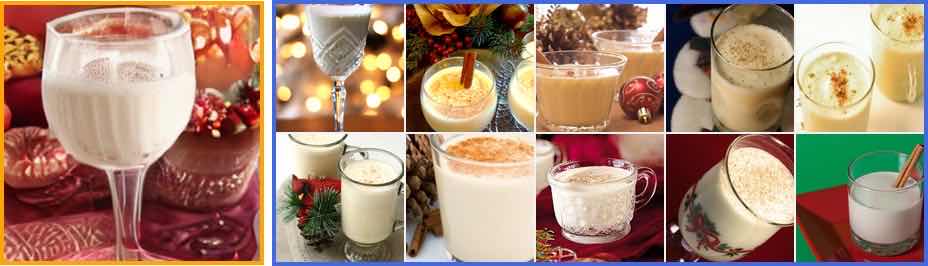}
    \caption{\textbf{Nearest neighbor analysis.} Each panel shows a \textcolor{orange}{{generated sample}} together with its top-10 nearest \textcolor{blue}{real images}. The nearest neighbors are retrieved from the ImageNet training set based on the cosine similarity using a CLIP encoder \cite{radford2021learning}. 
    Our method generates novel images that are visually distinct from their nearest neighbors.}
    \label{fig:nearest_neighbors}
\end{figure}

\subsection{Nearest Neighbor Analysis}

In \cref{fig:nearest_neighbors}, we show generated images together with their nearest real images. The nearest neighbors are retrieved from the ImageNet training set using CLIP features. 
These visualizations suggest that our method generates novel images that are visually distinct from their nearest neighbors, rather than merely memorizing training samples.

\paragraph{Quantitative analysis.}
We complement the qualitative analysis with a quantitative one. For each query sample, we compute its nearest-neighbor CLIP cosine similarity against the full ImageNet training set (1.28M images) and report the average over the query set.

\begin{table}[!ht]
\centering
\caption{\textbf{Quantitative memorization analysis via CLIP nearest-neighbor similarity.} Average nearest-neighbor CLIP cosine similarity between each query set and the ImageNet training set (1.28M). The held-out validation set provides a non-memorizing reference (since validation images are not in the training set). Our generated samples are \emph{less} similar to the training set than the held-out validation set, and on par with diffusion baselines, indicating that the model is not memorizing training images.}
\label{tab:appendix_clip_nn}
\tablestyle{6pt}{1.05}
\tablefontsize
\begin{tabular}{l c}
\toprule
query set & avg.\ nearest CLIP sim.\ to train set $\downarrow$ \\
\midrule
SiT XL/2~\cite{ma2024sit} samples (CFG 1.0) & 0.845 \\
Drifting Model L/2 samples (CFG 1.0) & 0.850 \\
SiT XL/2 samples (CFG 1.5) & 0.870 \\
\midrule
ImageNet validation set (held-out reference) & 0.890 \\
\bottomrule
\end{tabular}
\end{table}

Our generated samples have an average nearest-neighbor CLIP similarity of 0.850 to the training set, which is \emph{lower} than the held-out validation set (0.890) and similar to SiT XL/2 at the same CFG. Since the validation set is by construction not memorizing training images, this places our generator firmly in the non-memorizing regime.

\subsection{Qualitative Results}
\label{sec:appendix_qualitative_results}

\begin{figure*}[p]
    \centering
    \begin{minipage}[b]{0.48\textwidth}
        \includegraphics[width=\linewidth]{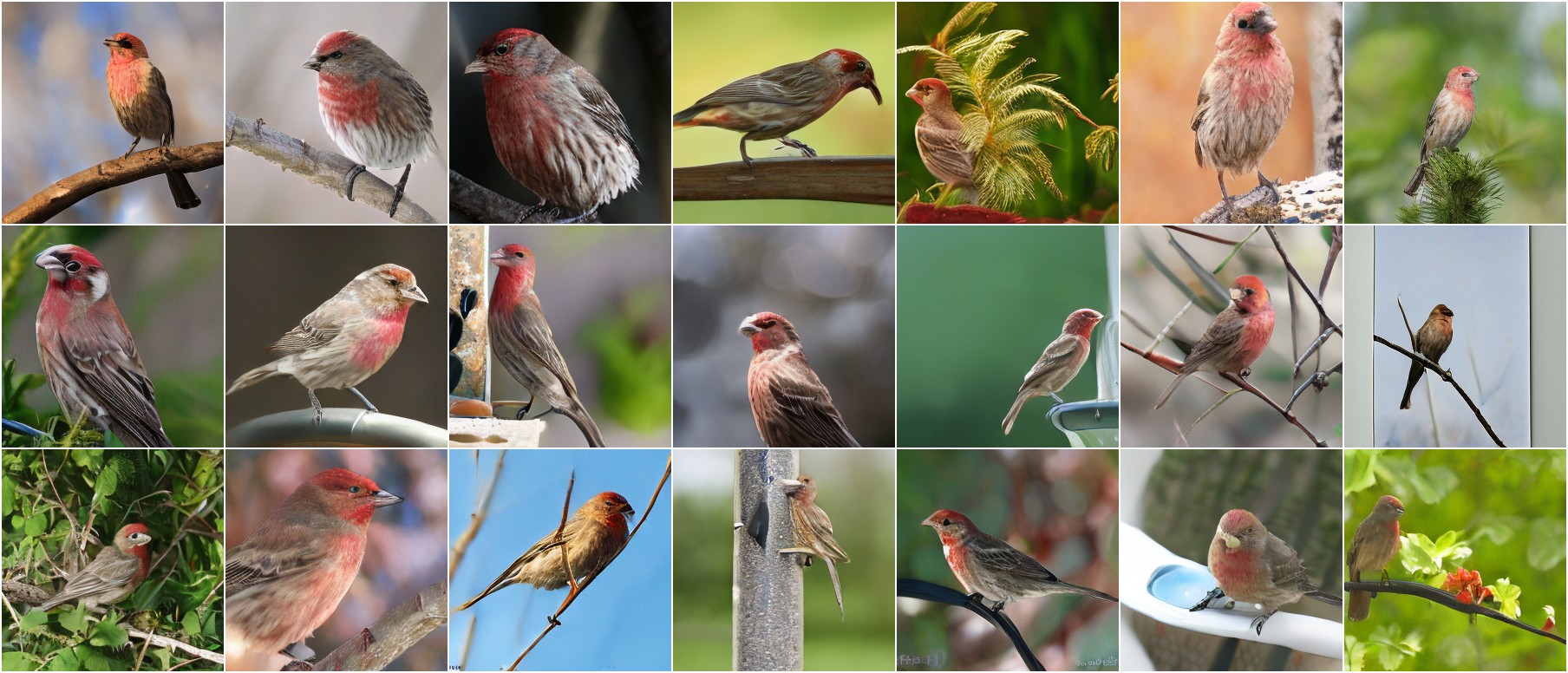}
        \centering\small Class 012: house finch, linnet, Carpodacus mexicanus
    \end{minipage}\hfill
    \begin{minipage}[b]{0.48\textwidth}
        \includegraphics[width=\linewidth]{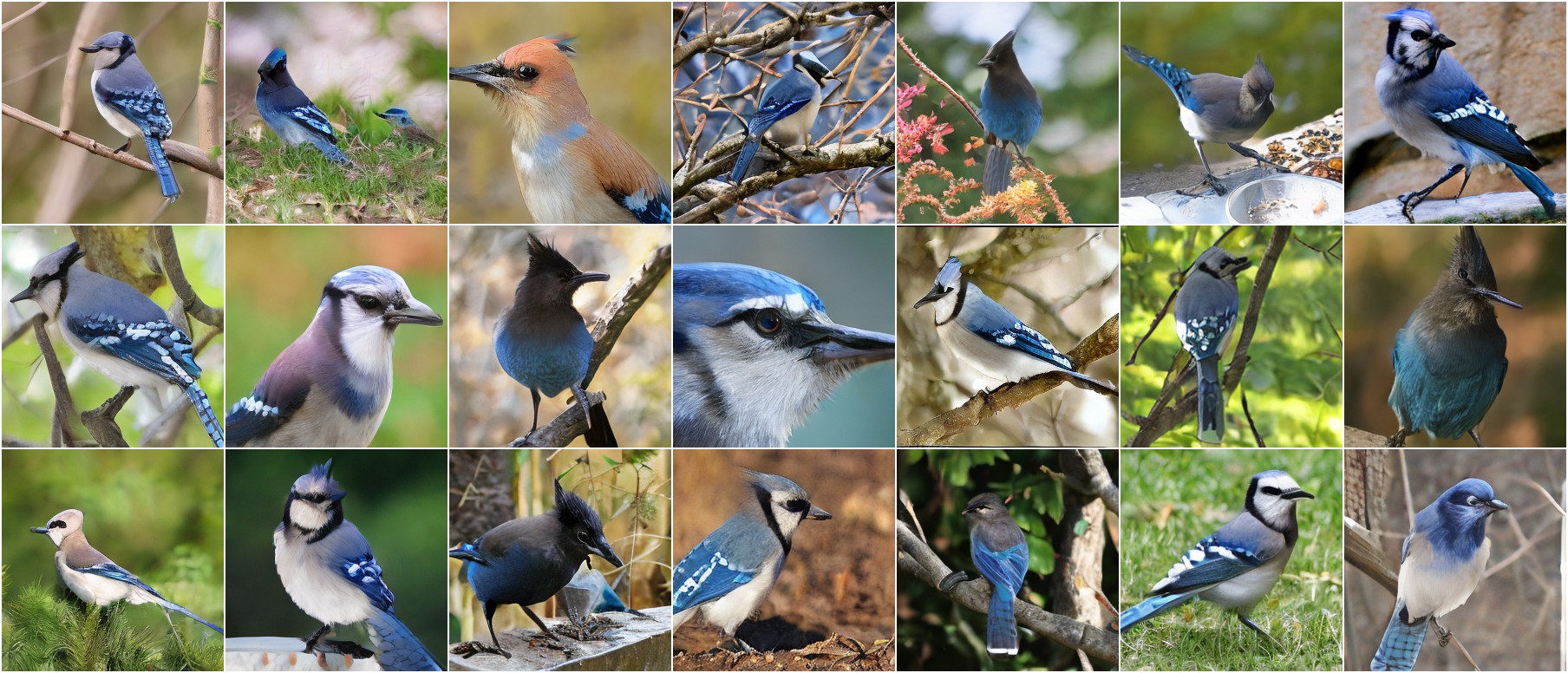}
        \centering\small Class 017: jay
    \end{minipage}\\[0.4em]
    \begin{minipage}[b]{0.48\textwidth}
        \includegraphics[width=\linewidth]{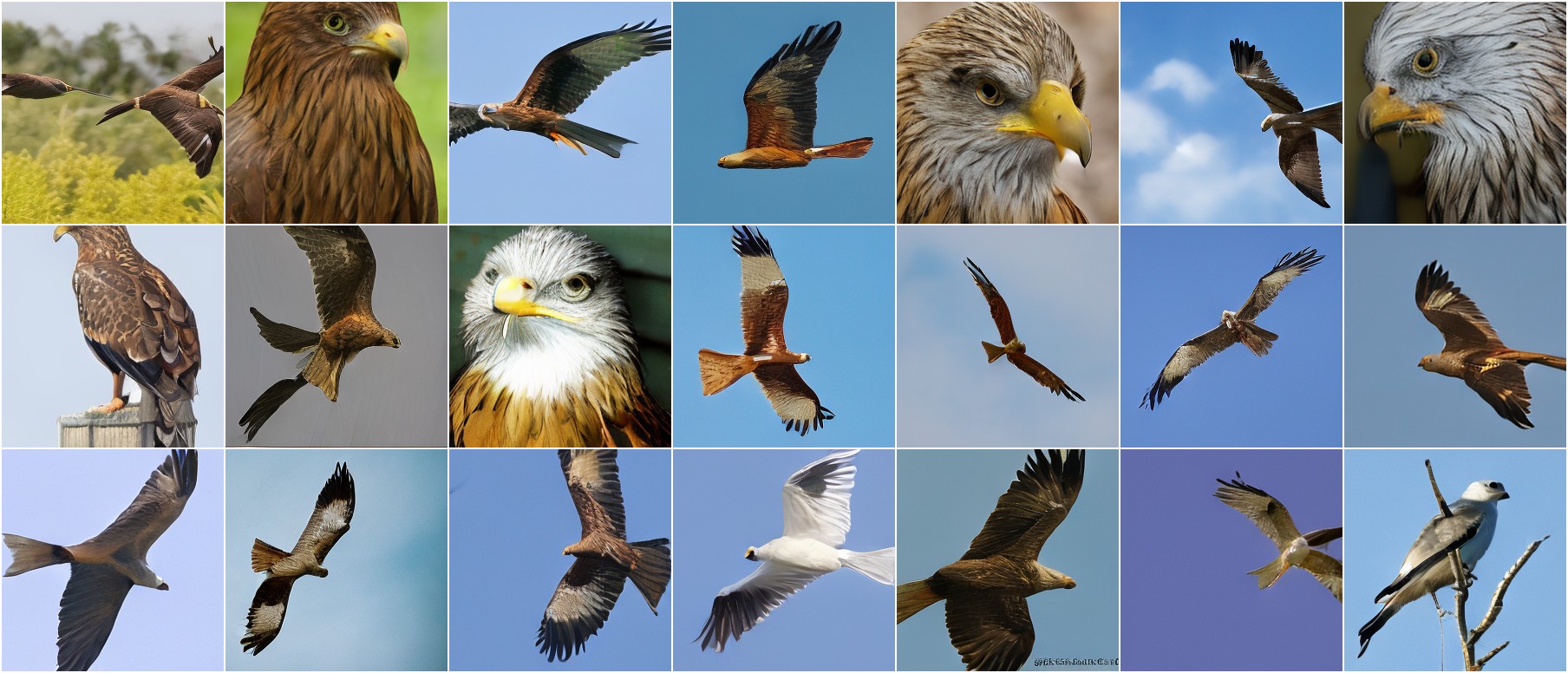}
        \centering\small Class 021: kite
    \end{minipage}\hfill
    \begin{minipage}[b]{0.48\textwidth}
        \includegraphics[width=\linewidth]{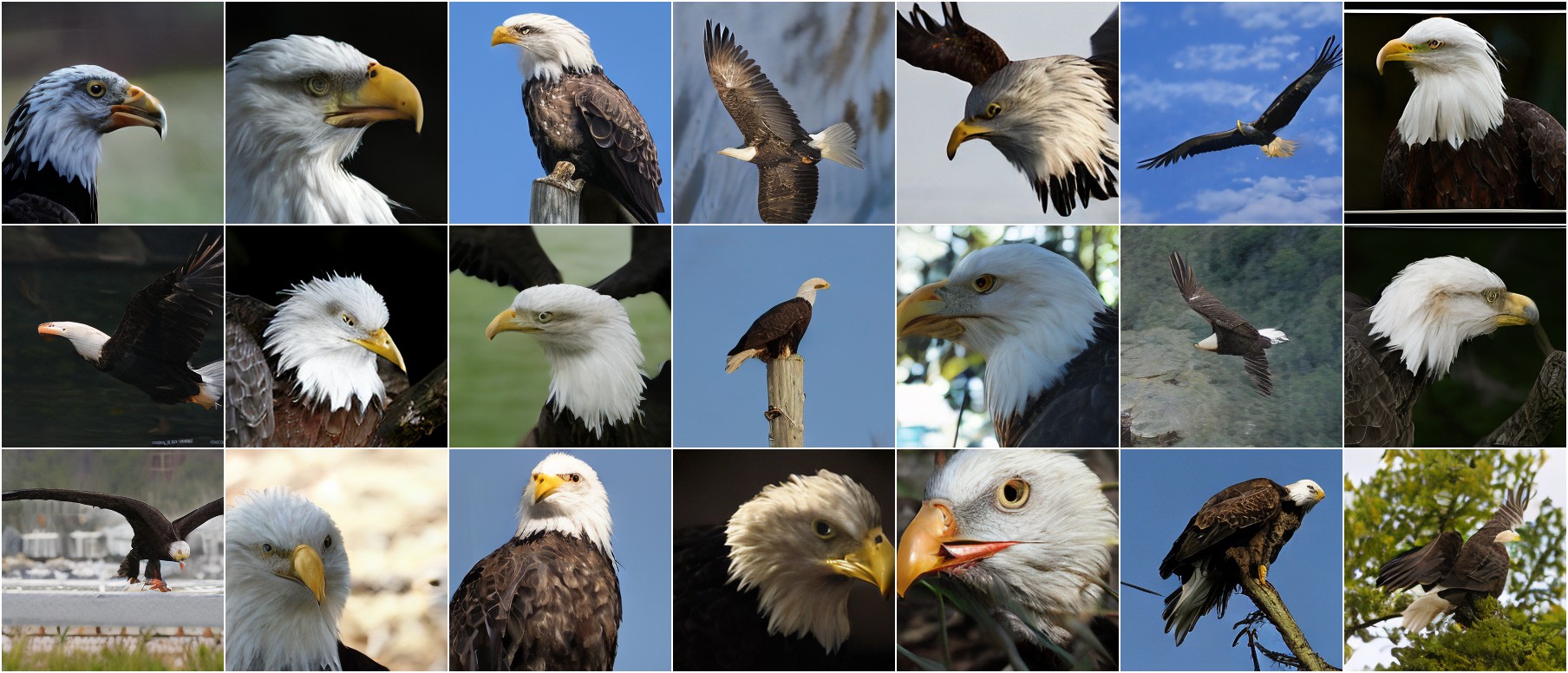}
        \centering\small Class 022: bald eagle, American eagle, Haliaeetus leucocephalus
    \end{minipage}\\[0.4em]
    \begin{minipage}[b]{0.48\textwidth}
        \includegraphics[width=\linewidth]{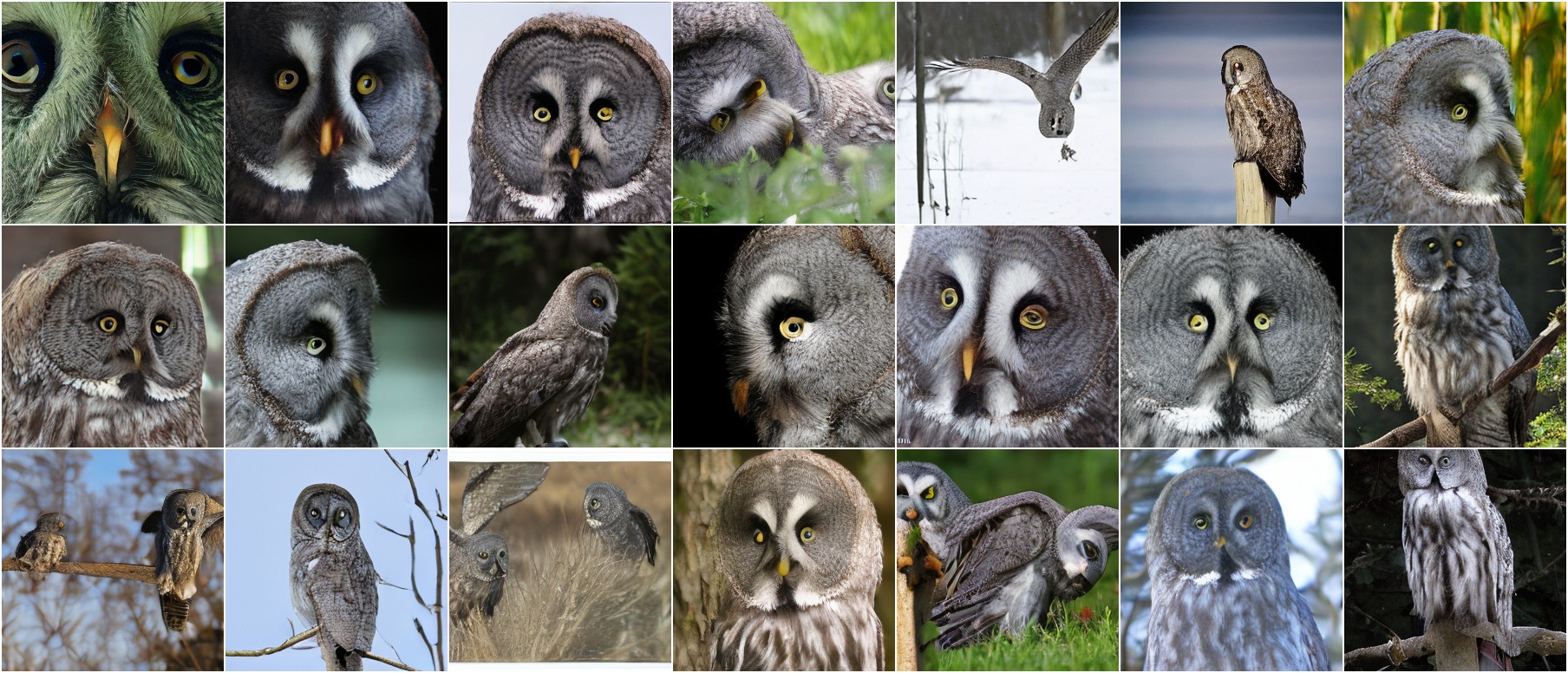}
        \centering\small Class 024: great grey owl, great gray owl, Strix nebulosa
    \end{minipage}\hfill
    \begin{minipage}[b]{0.48\textwidth}
        \includegraphics[width=\linewidth]{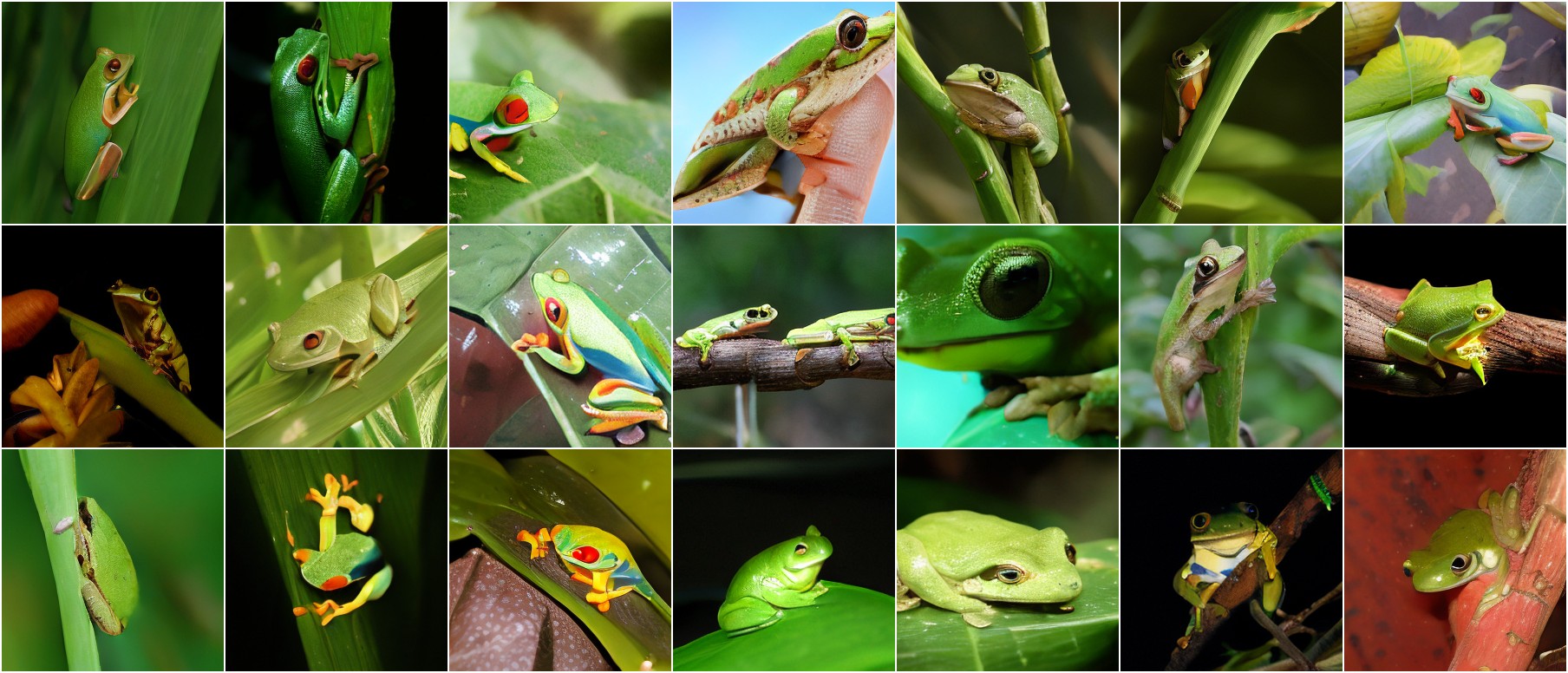}
        \centering\small Class 031: tree frog, tree-frog
    \end{minipage}\\[0.4em]
    \begin{minipage}[b]{0.48\textwidth}
        \includegraphics[width=\linewidth]{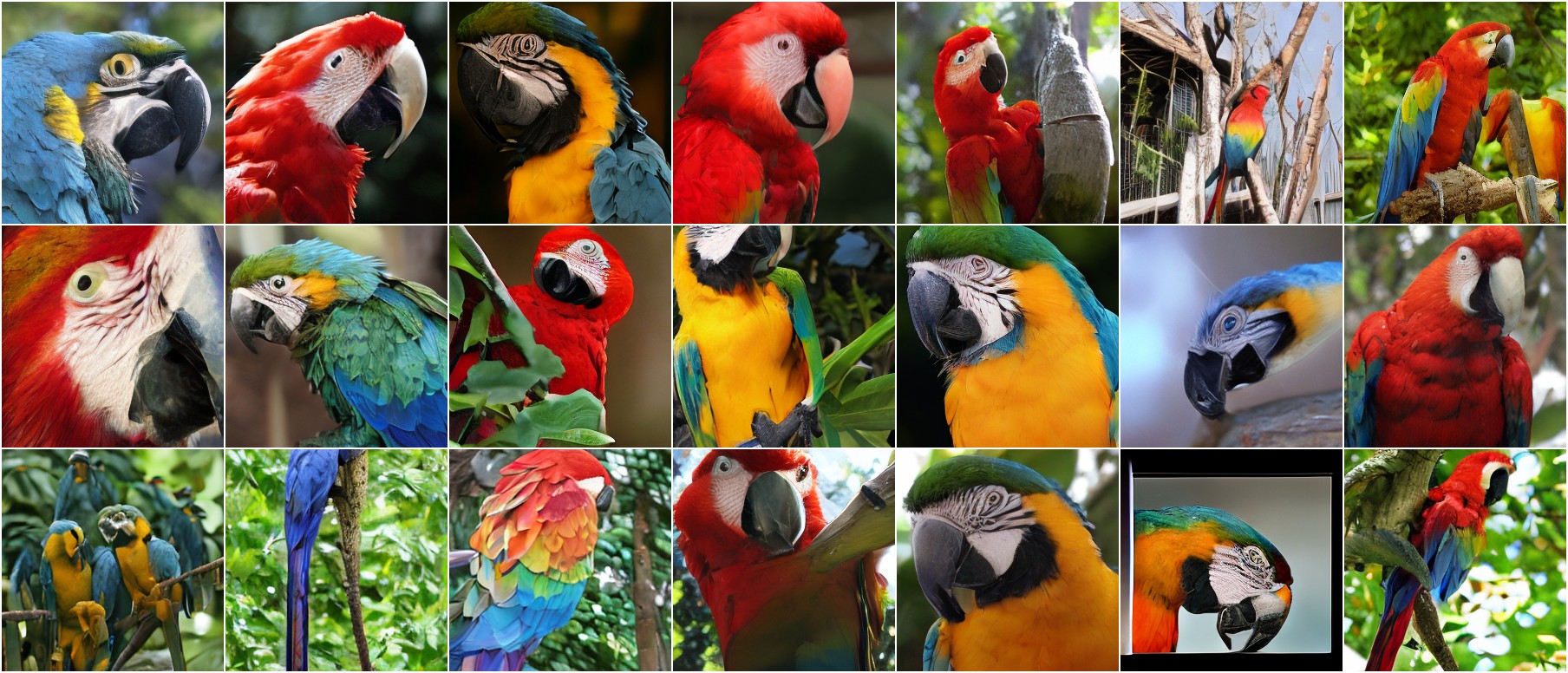}
        \centering\small Class 088: macaw
    \end{minipage}\hfill
    \begin{minipage}[b]{0.48\textwidth}
        \includegraphics[width=\linewidth]{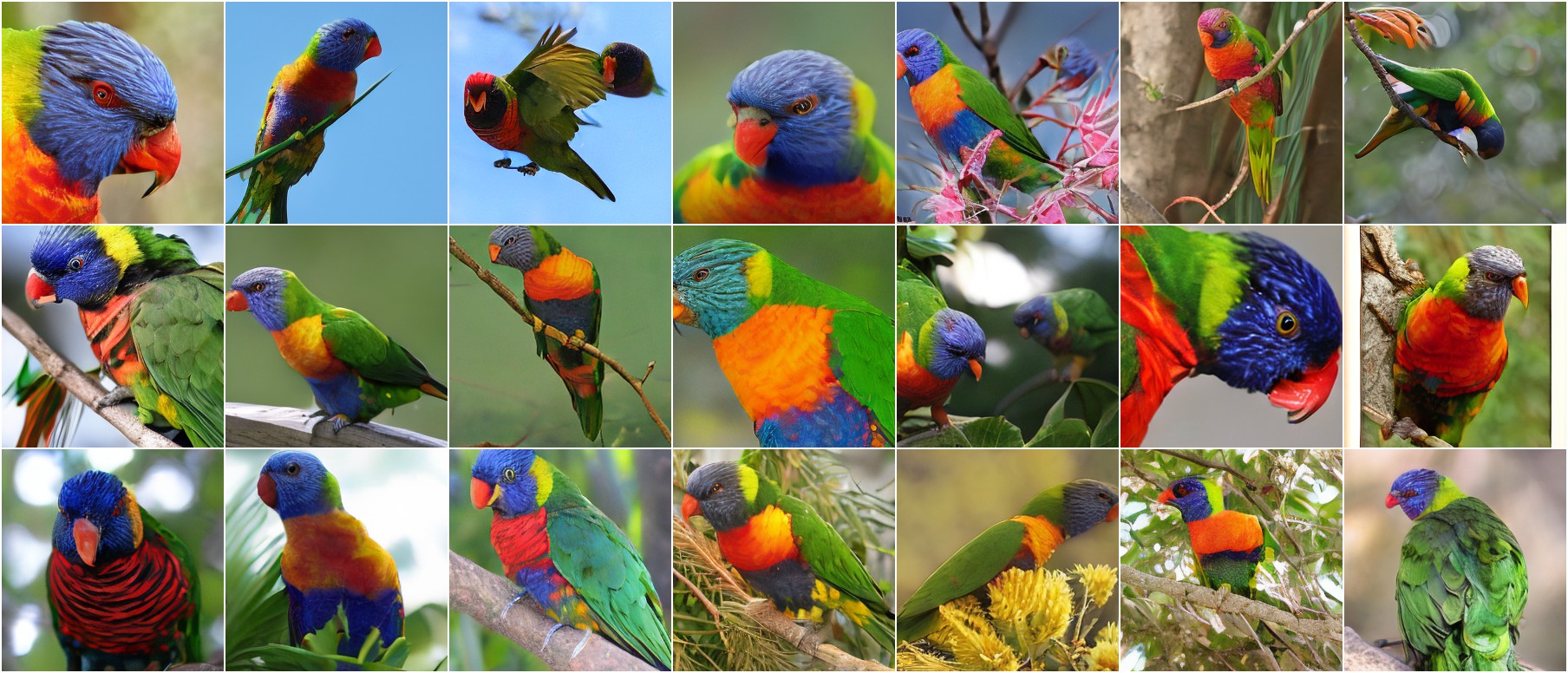}
        \centering\small Class 090: lorikeet
    \end{minipage}\\[0.4em]
    \begin{minipage}[b]{0.48\textwidth}
        \includegraphics[width=\linewidth]{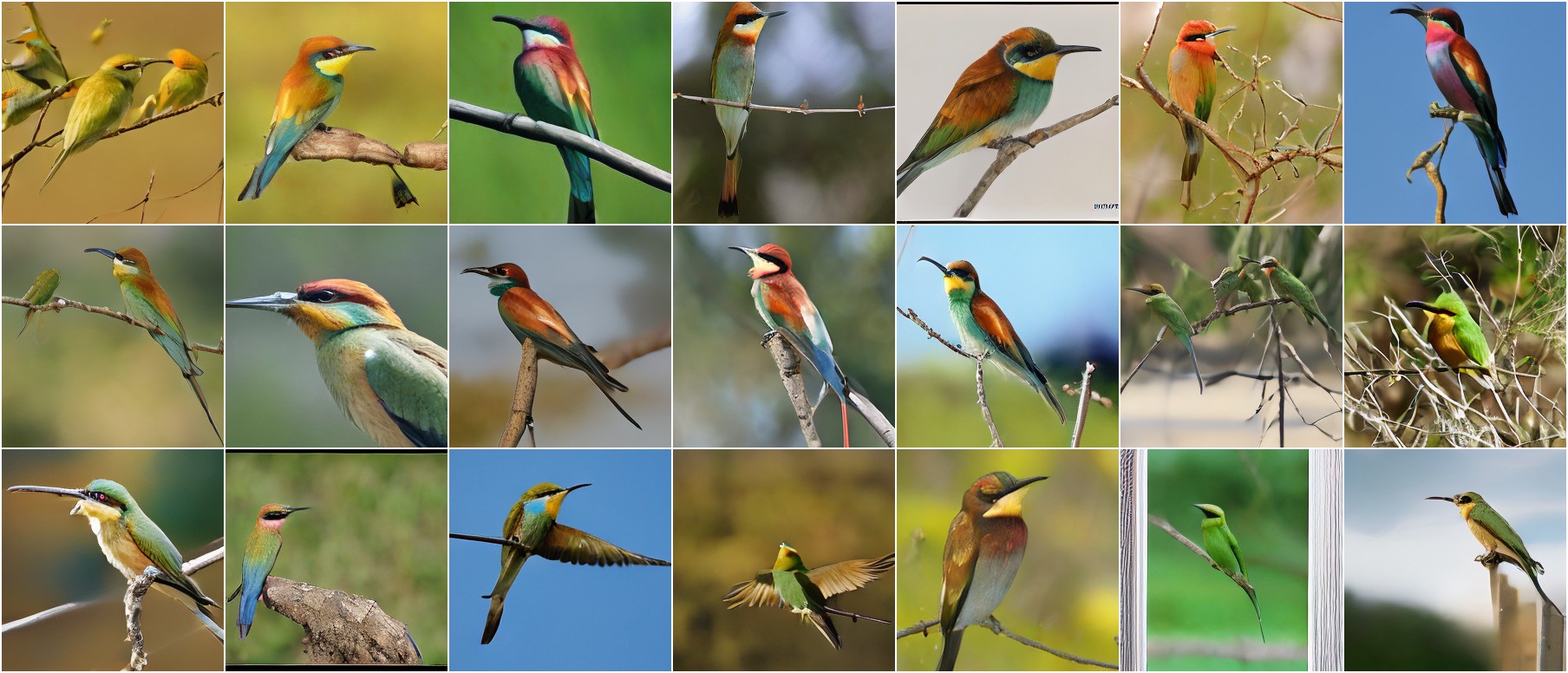}
        \centering\small Class 092: bee eater
    \end{minipage}\hfill
    \begin{minipage}[b]{0.48\textwidth}
        \includegraphics[width=\linewidth]{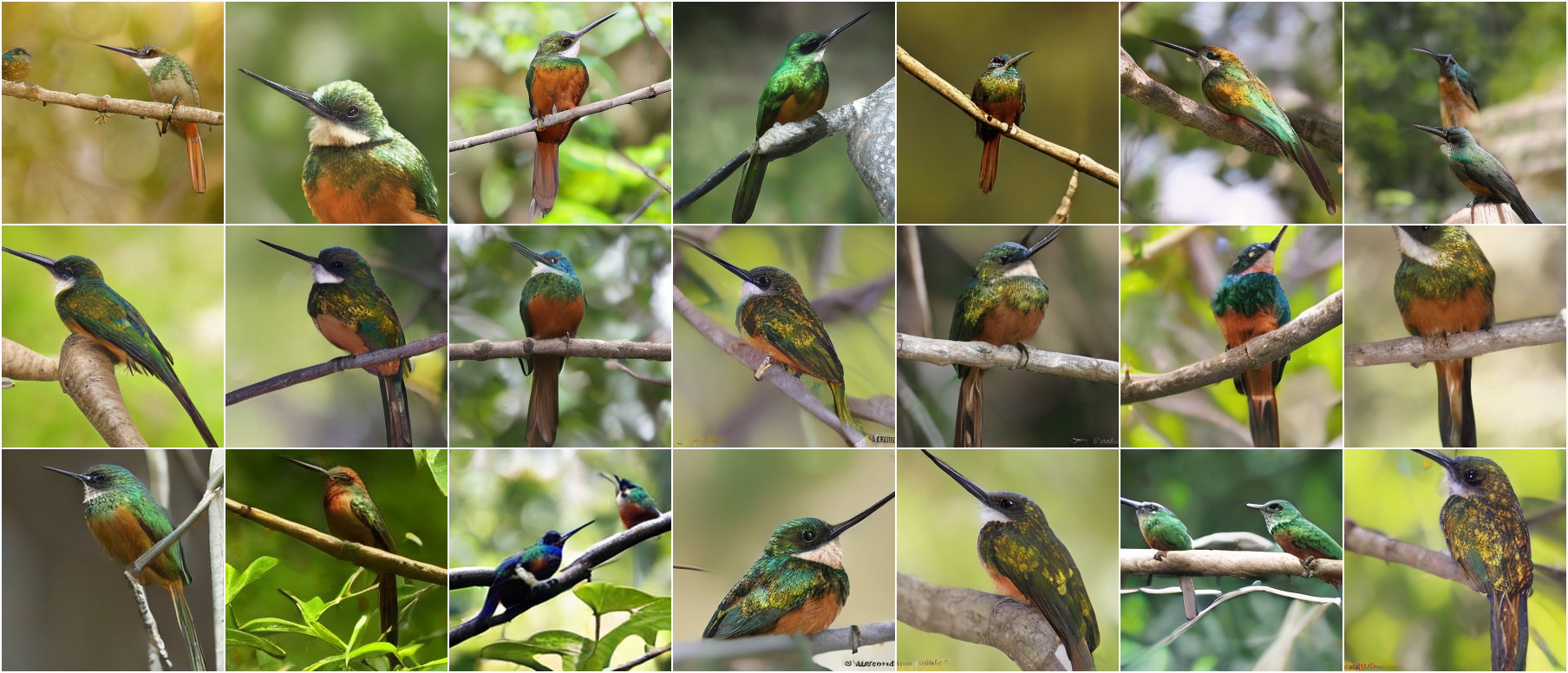}
        \centering\small Class 095: jacamar
    \end{minipage}
    \caption{\textbf{Uncurated samples from our latent-L/2 model with CFG = 1.0 (page 1/4).} FID = 1.54, IS = 258.9.}
    \label{fig:latent_uncurated_samples_1}
\end{figure*}

\begin{figure*}[p]
    \centering
    \begin{minipage}[b]{0.48\textwidth}
        \includegraphics[width=\linewidth]{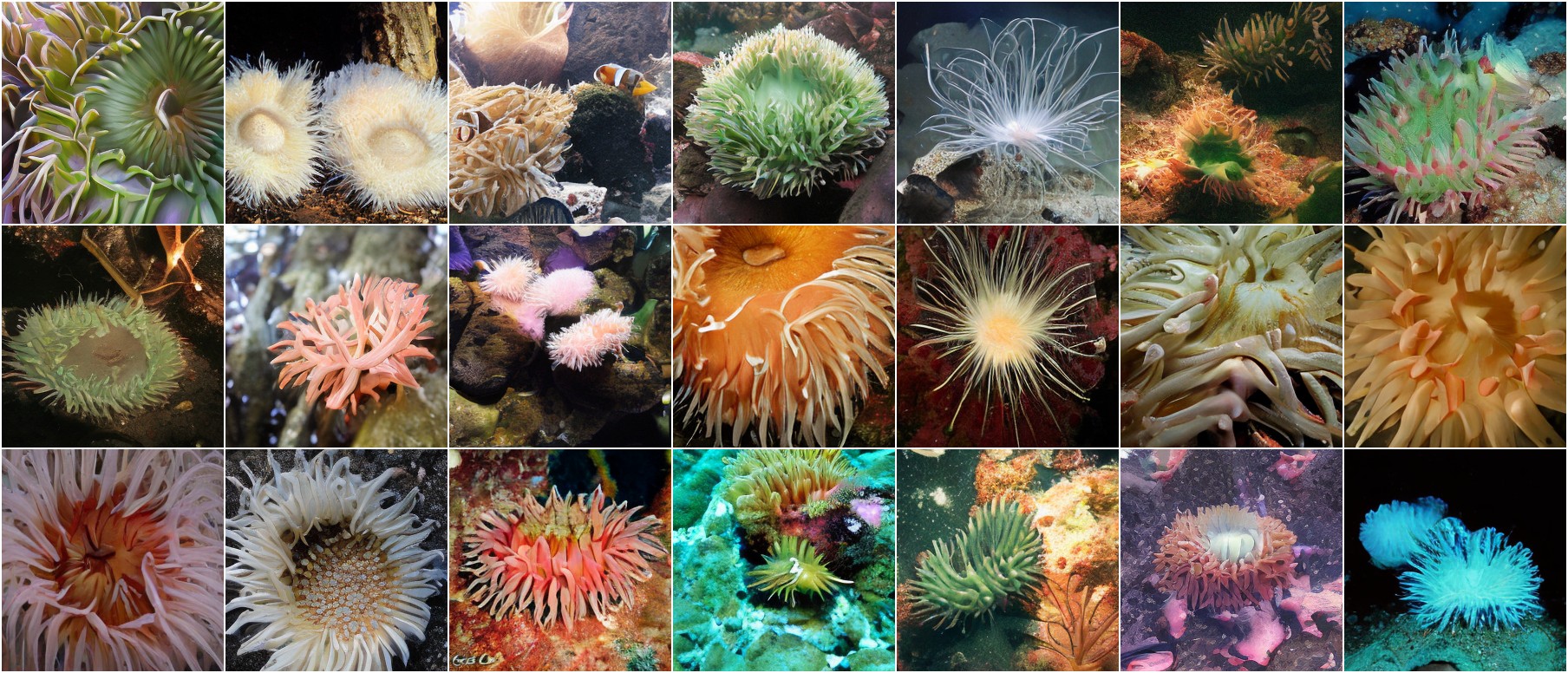}
        \centering\small Class 108: sea anemone, anemone
    \end{minipage}\hfill
    \begin{minipage}[b]{0.48\textwidth}
        \includegraphics[width=\linewidth]{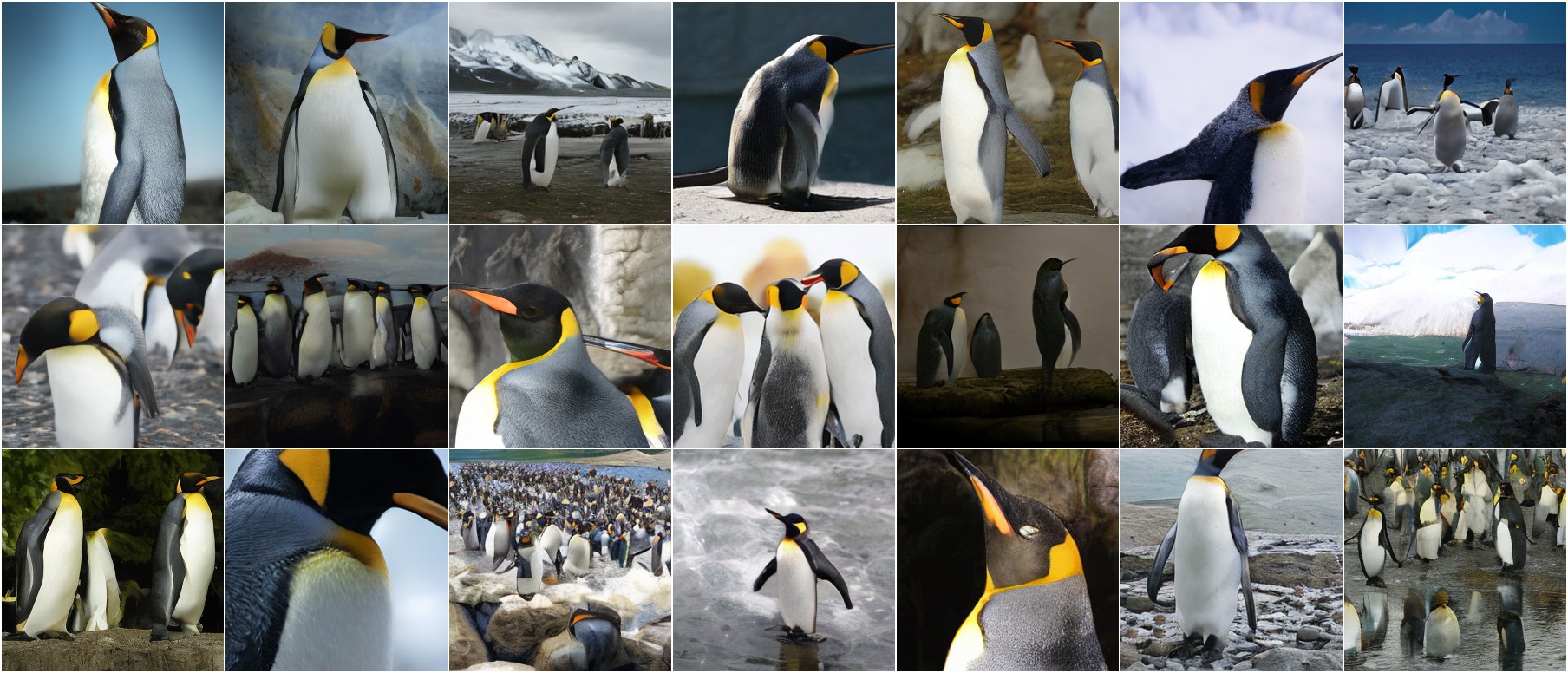}
        \centering\small Class 145: king penguin, Aptenodytes patagonica
    \end{minipage}\\[0.4em]
    \begin{minipage}[b]{0.48\textwidth}
        \includegraphics[width=\linewidth]{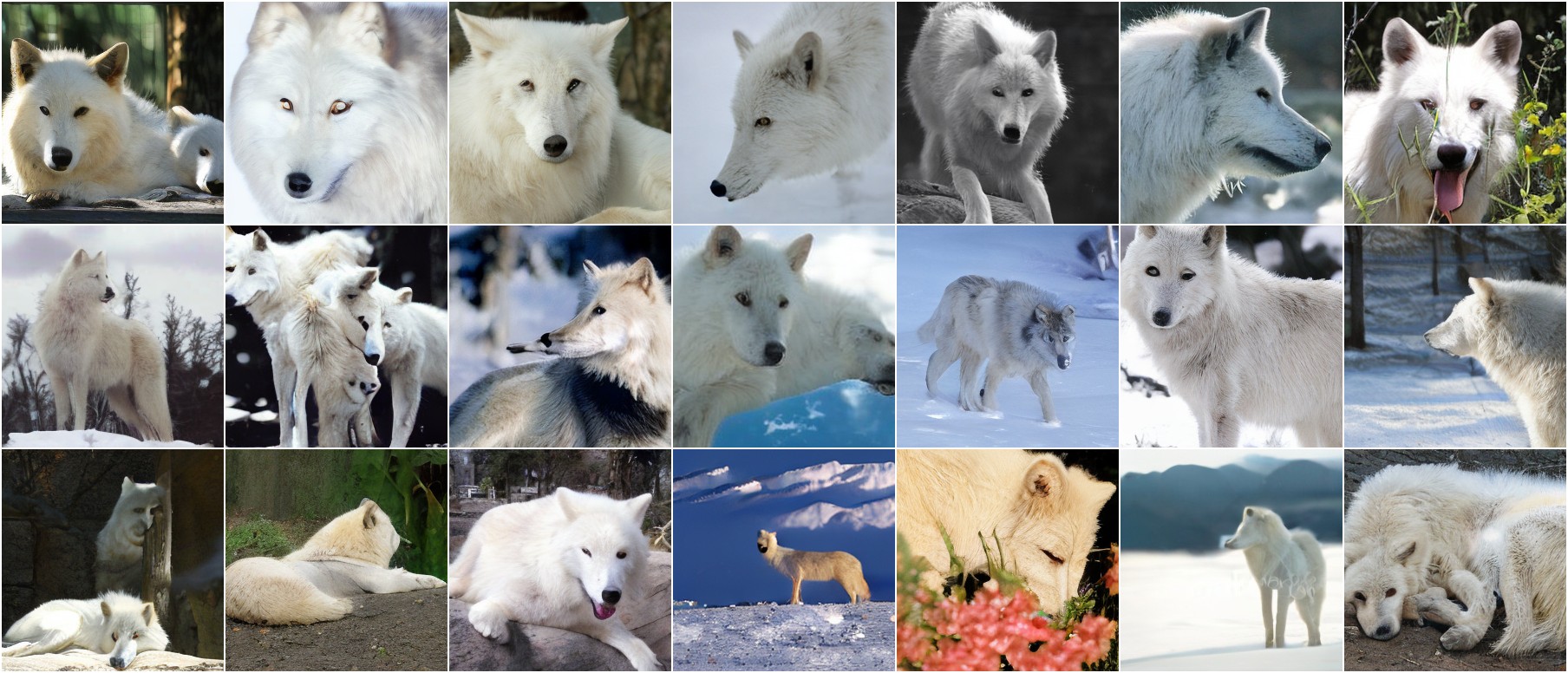}
        \centering\small Class 270: white wolf, Arctic wolf, Canis lupus tundrarum
    \end{minipage}\hfill
    \begin{minipage}[b]{0.48\textwidth}
        \includegraphics[width=\linewidth]{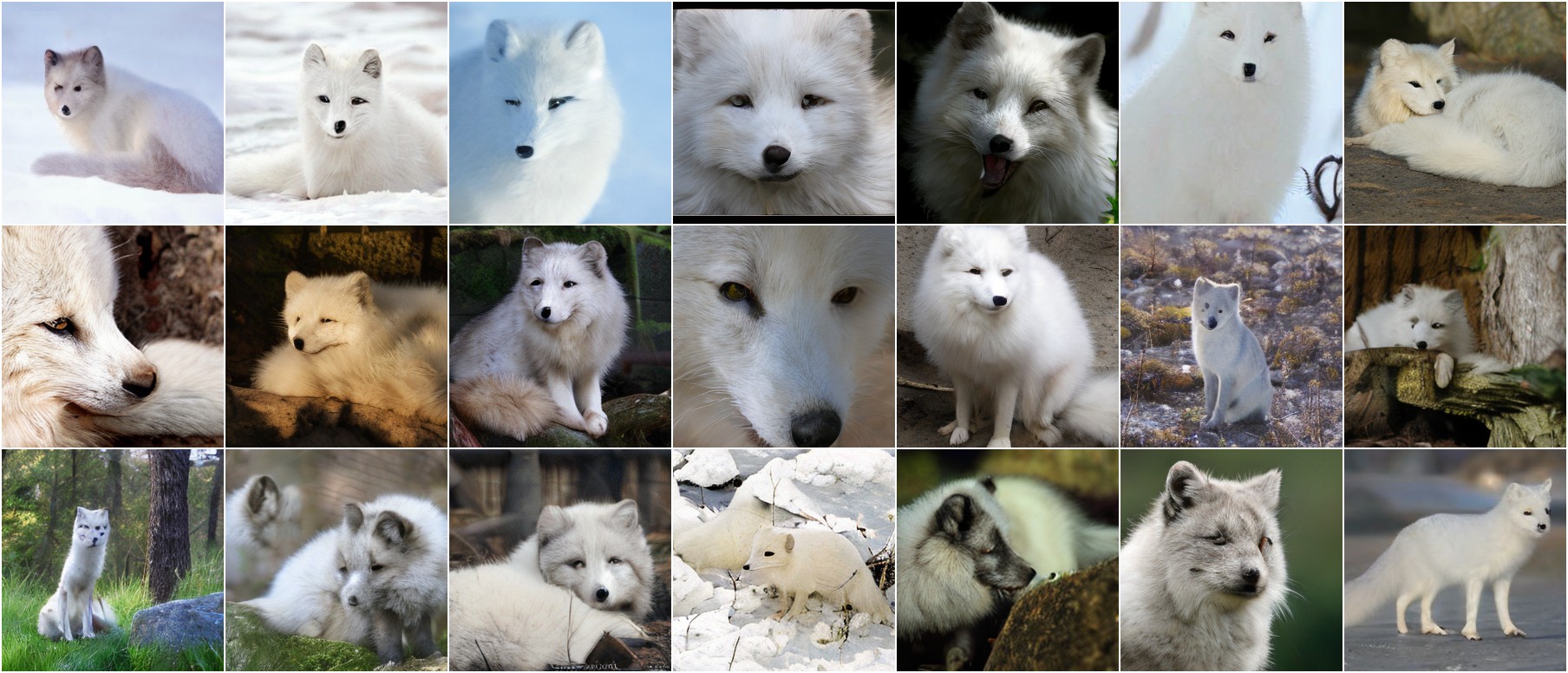}
        \centering\small Class 279: Arctic fox, white fox, Alopex lagopus
    \end{minipage}\\[0.4em]
    \begin{minipage}[b]{0.48\textwidth}
        \includegraphics[width=\linewidth]{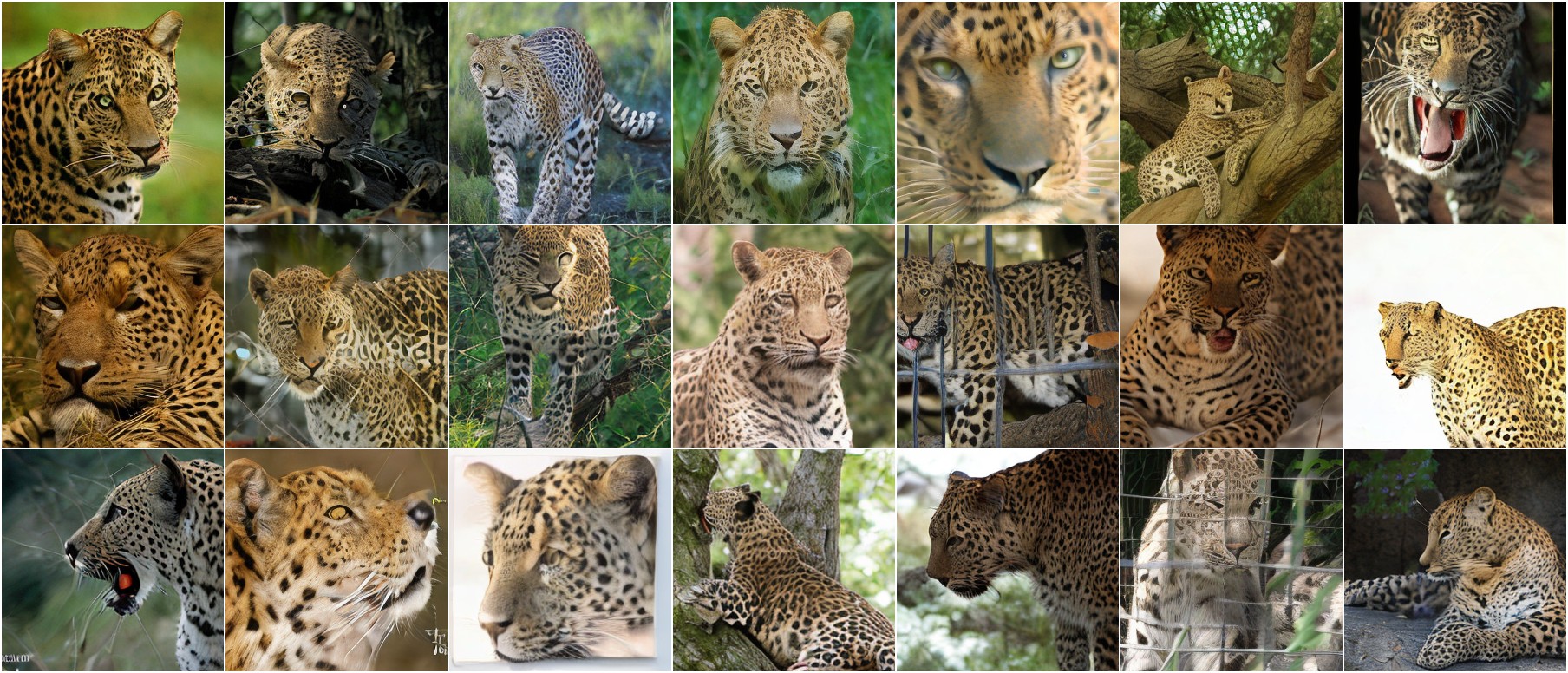}
        \centering\small Class 288: leopard, Panthera pardus
    \end{minipage}\hfill
    \begin{minipage}[b]{0.48\textwidth}
        \includegraphics[width=\linewidth]{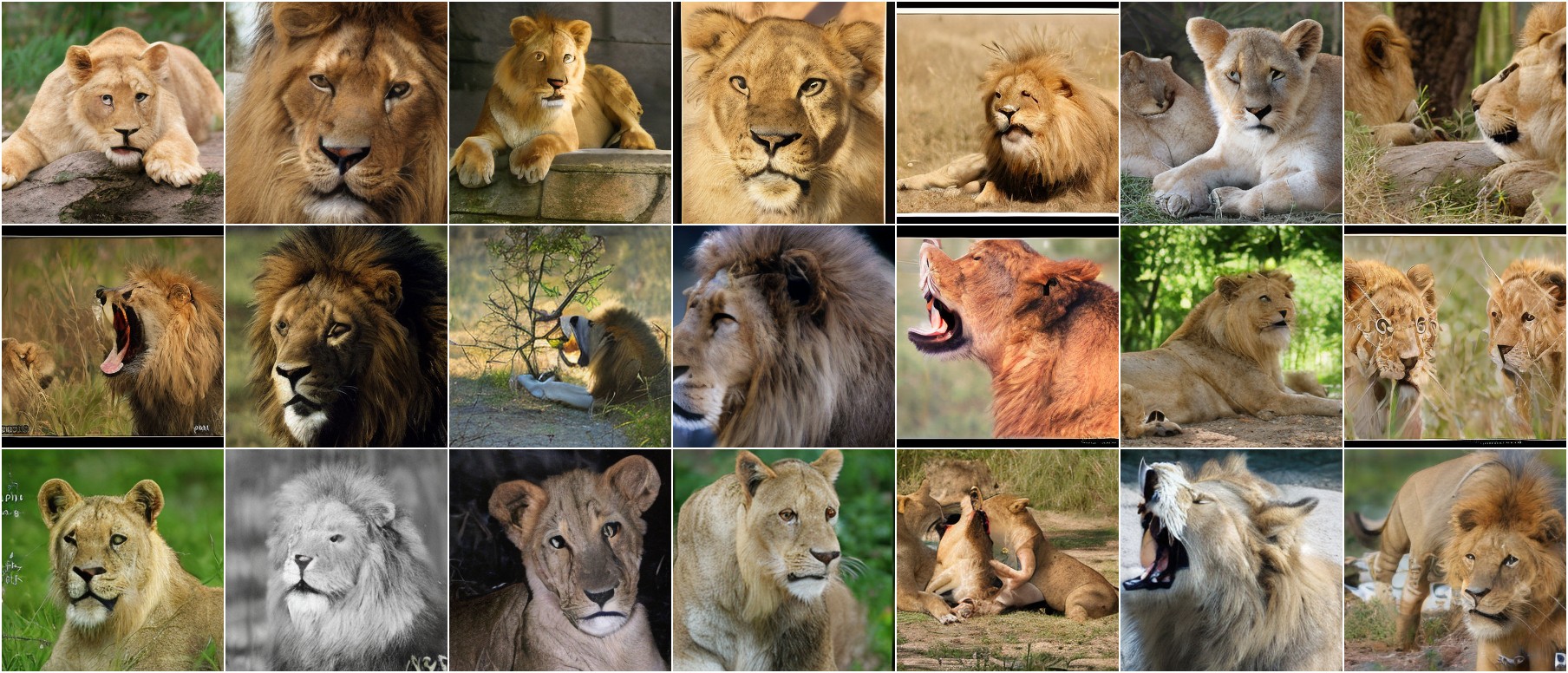}
        \centering\small Class 291: lion, king of beasts, Panthera leo
    \end{minipage}\\[0.4em]
    \begin{minipage}[b]{0.48\textwidth}
        \includegraphics[width=\linewidth]{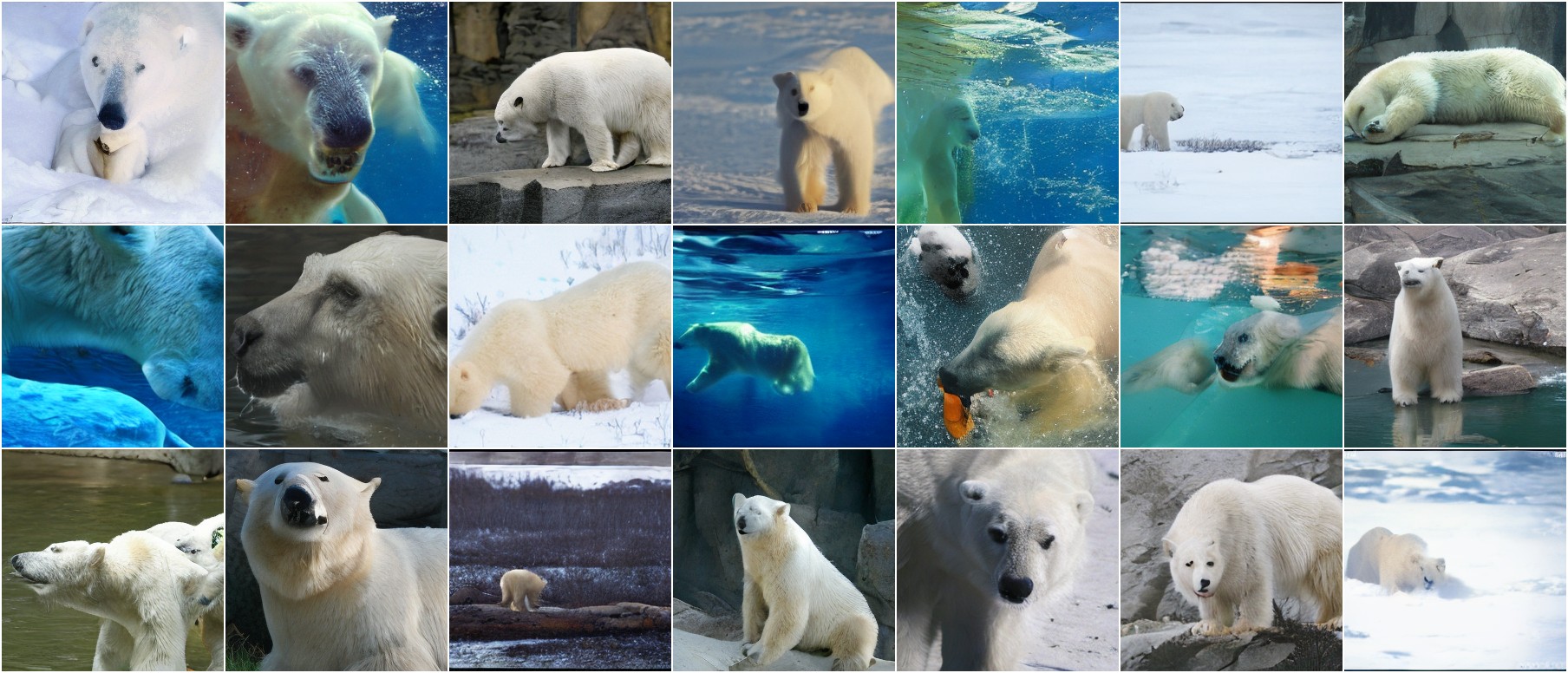}
        \centering\small Class 296: ice bear
    \end{minipage}\hfill
    \begin{minipage}[b]{0.48\textwidth}
        \includegraphics[width=\linewidth]{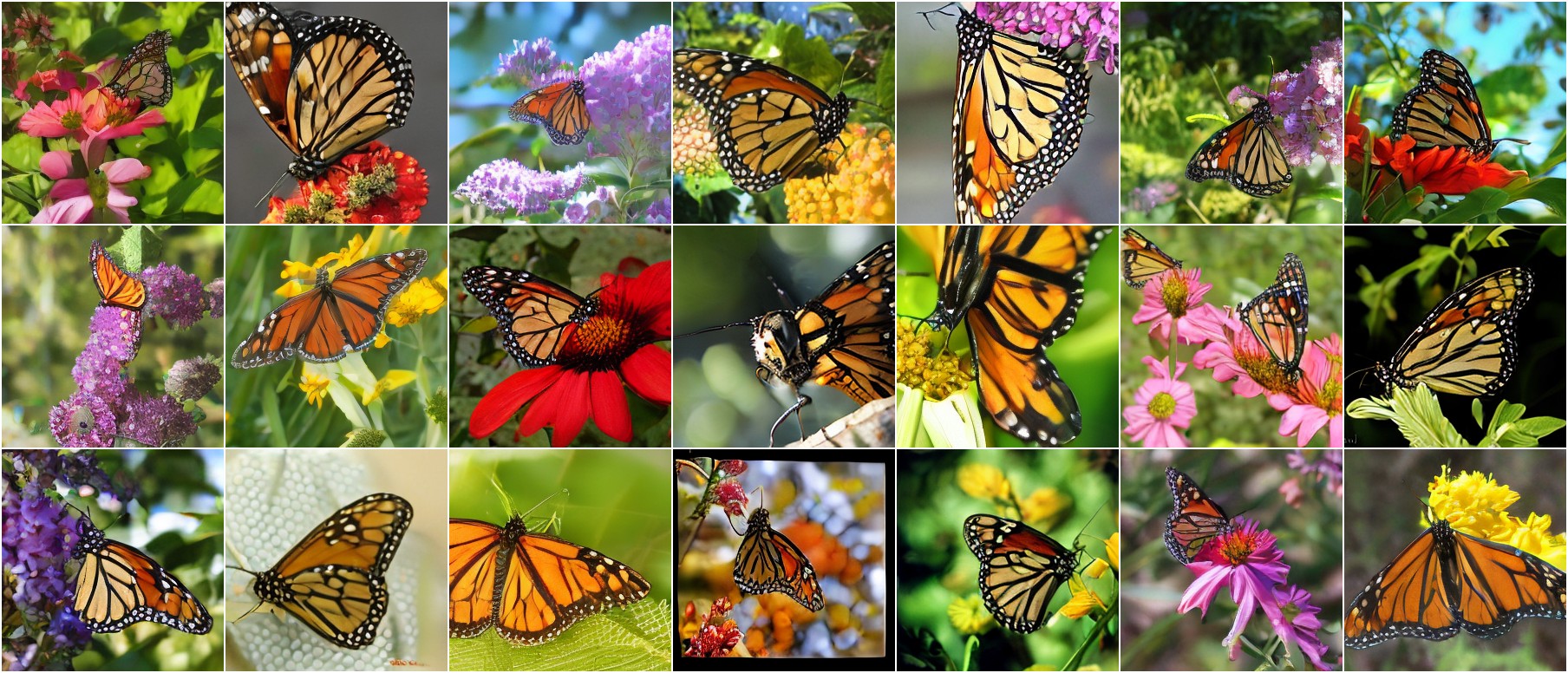}
        \centering\small Class 323: monarch, Danaus plexippus
    \end{minipage}\\[0.4em]
    \begin{minipage}[b]{0.48\textwidth}
        \includegraphics[width=\linewidth]{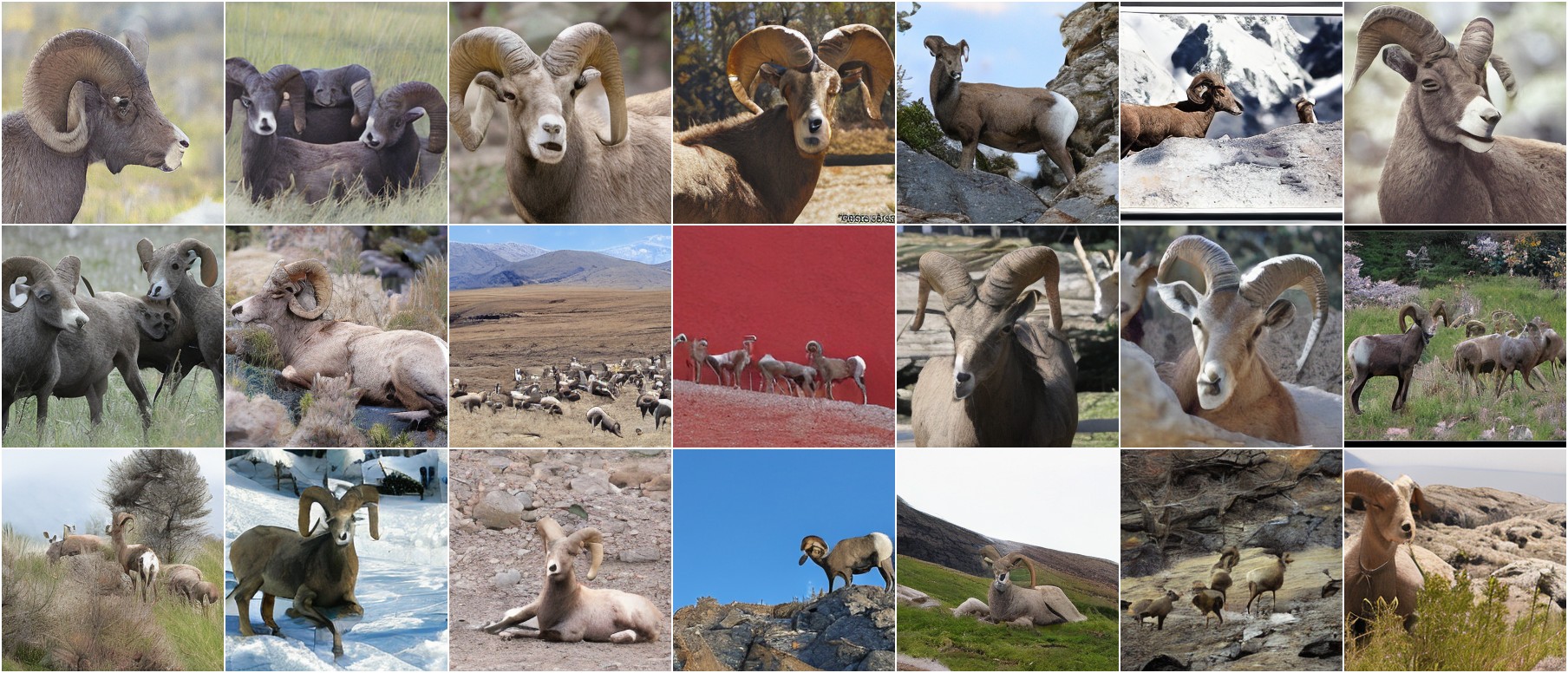}
        \centering\small Class 349: bighorn, bighorn sheep, Ovis canadensis
    \end{minipage}\hfill
    \begin{minipage}[b]{0.48\textwidth}
        \includegraphics[width=\linewidth]{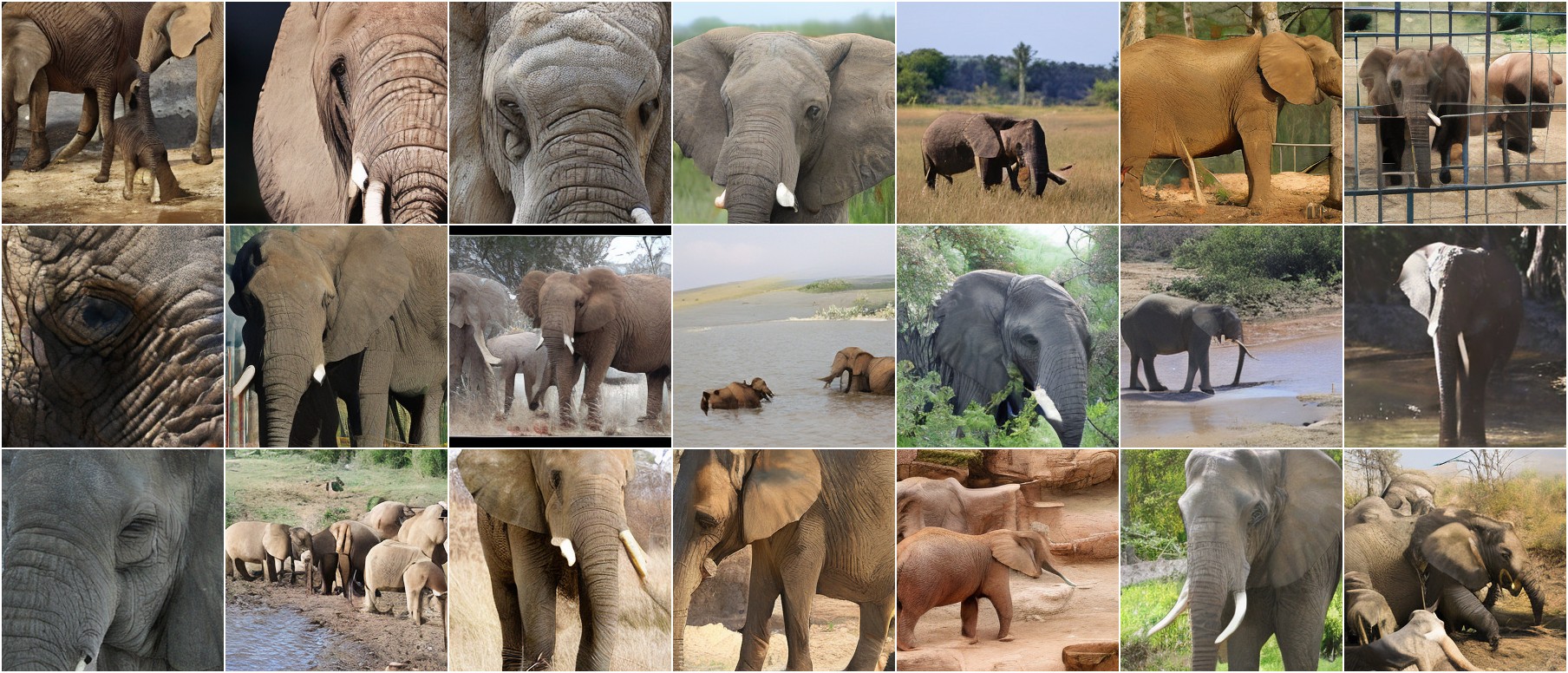}
        \centering\small Class 386: African elephant, Loxodonta africana
    \end{minipage}
    \caption{\textbf{Uncurated samples from our latent-L/2 model with CFG = 1.0 (page 2/4).} FID = 1.54, IS = 258.9.}
    \label{fig:latent_uncurated_samples_2}
\end{figure*}

\begin{figure*}[p]
    \centering
    \begin{minipage}[b]{0.48\textwidth}
        \includegraphics[width=\linewidth]{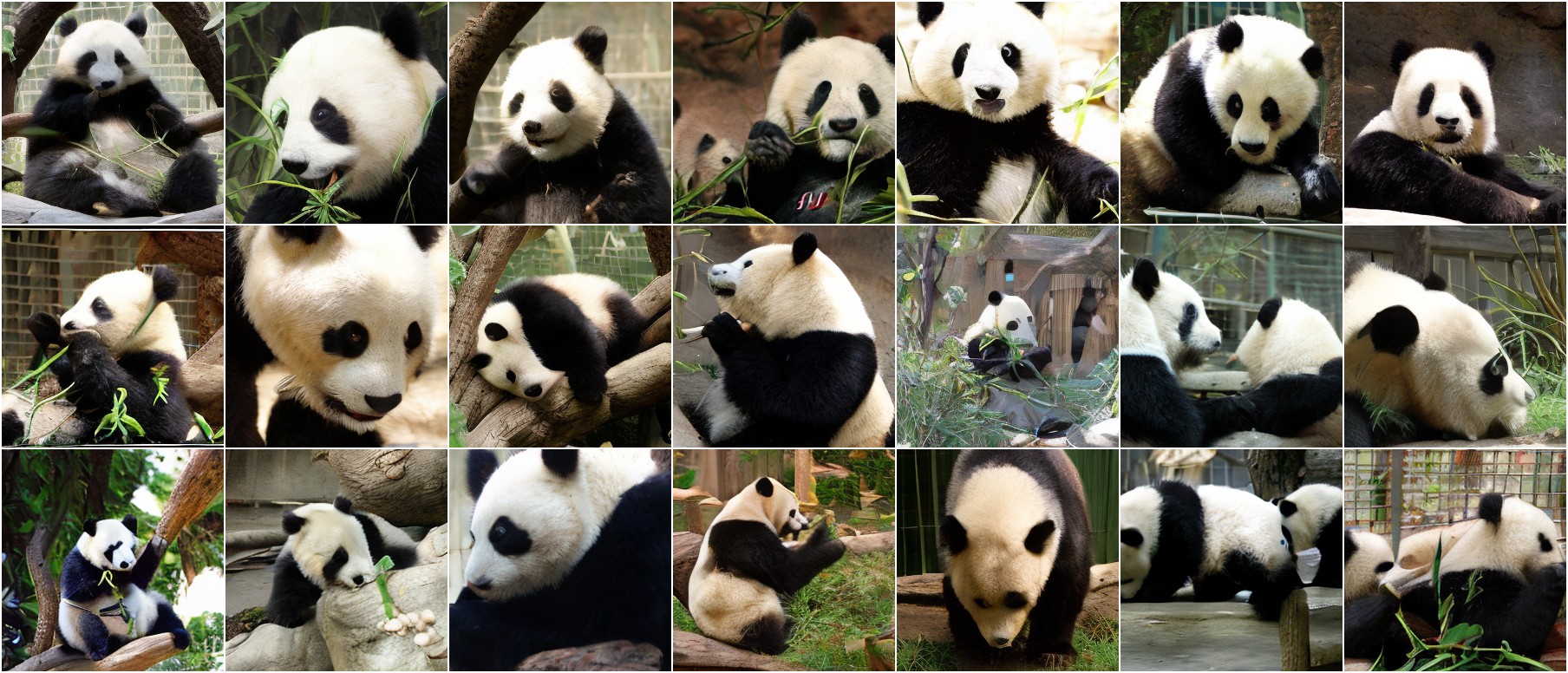}
        \centering\small Class 388: giant panda, Ailuropoda melanoleuca
    \end{minipage}\hfill
    \begin{minipage}[b]{0.48\textwidth}
        \includegraphics[width=\linewidth]{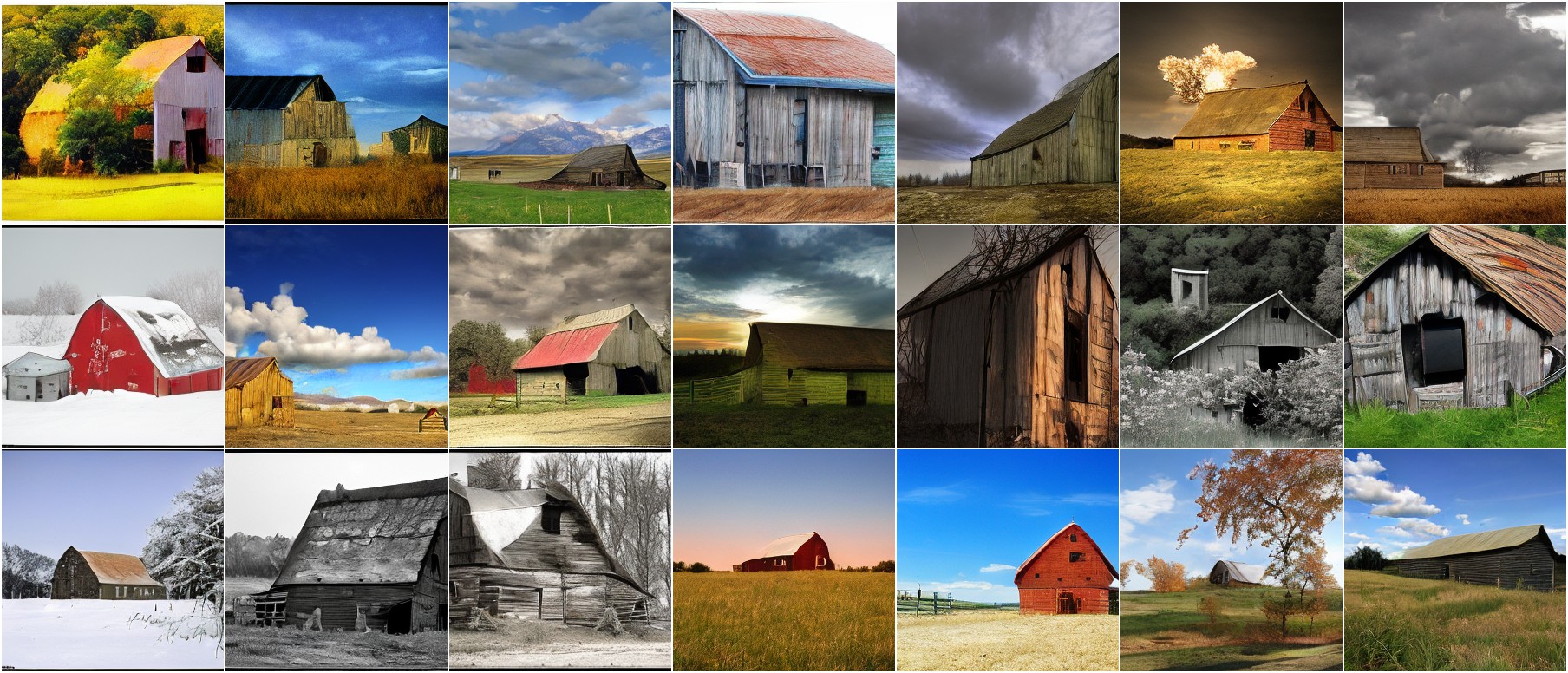}
        \centering\small Class 425: barn
    \end{minipage}\\[0.4em]
    \begin{minipage}[b]{0.48\textwidth}
        \includegraphics[width=\linewidth]{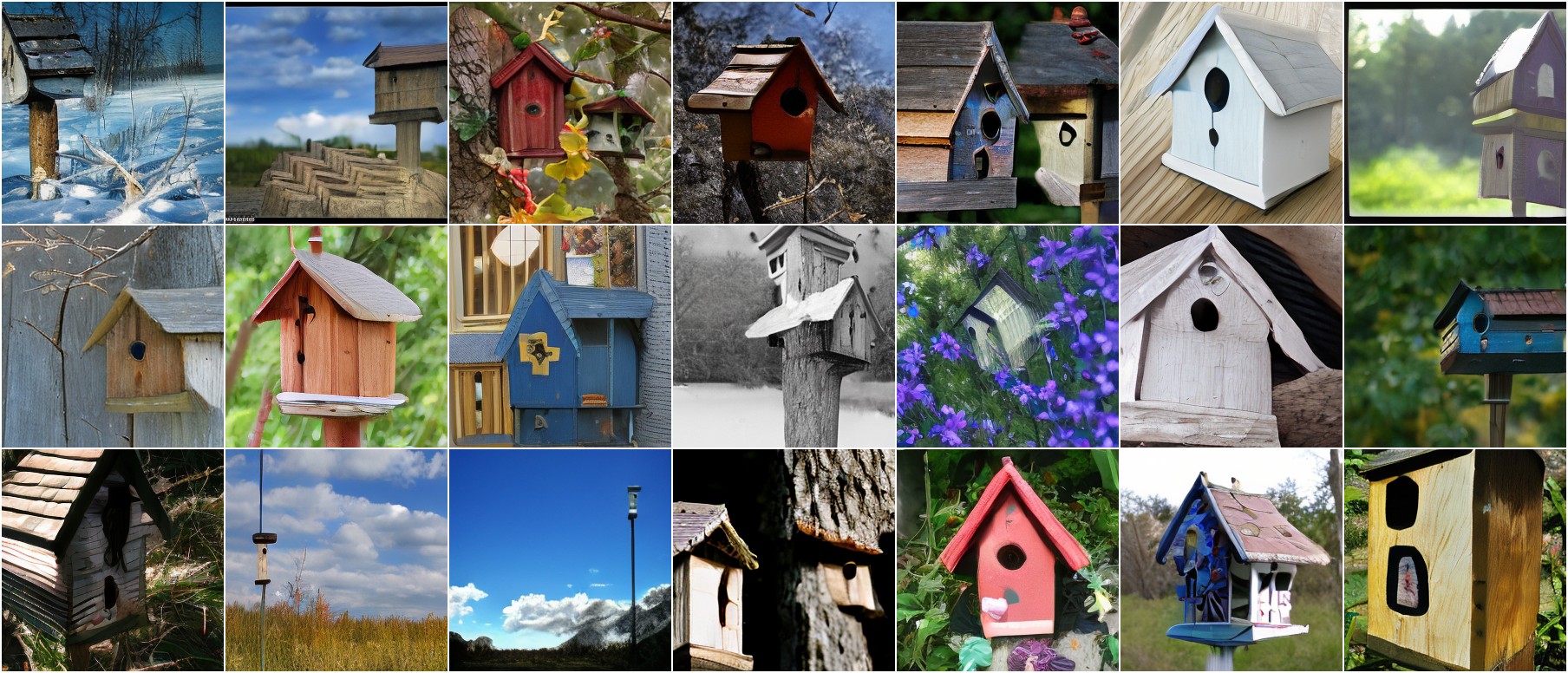}
        \centering\small Class 448: birdhouse
    \end{minipage}\hfill
    \begin{minipage}[b]{0.48\textwidth}
        \includegraphics[width=\linewidth]{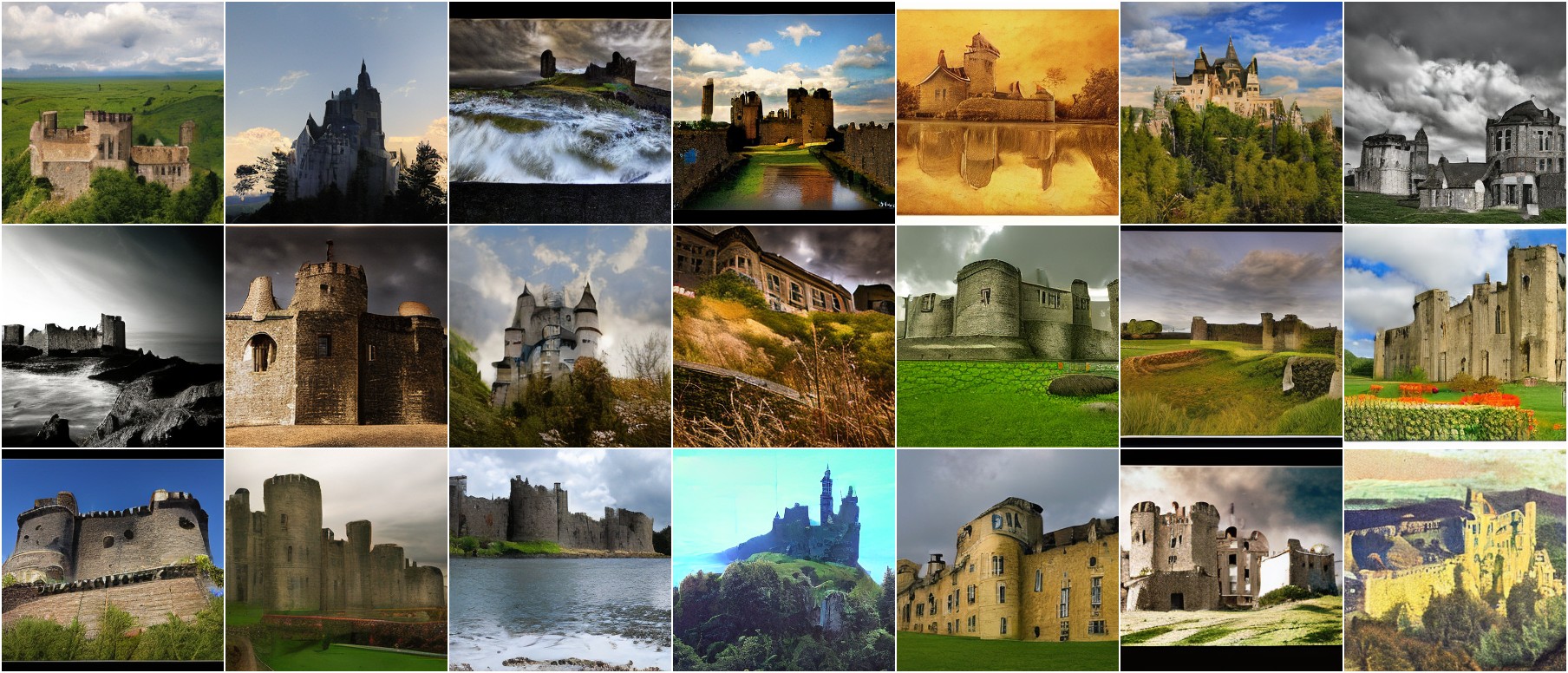}
        \centering\small Class 483: castle
    \end{minipage}\\[0.4em]
    \begin{minipage}[b]{0.48\textwidth}
        \includegraphics[width=\linewidth]{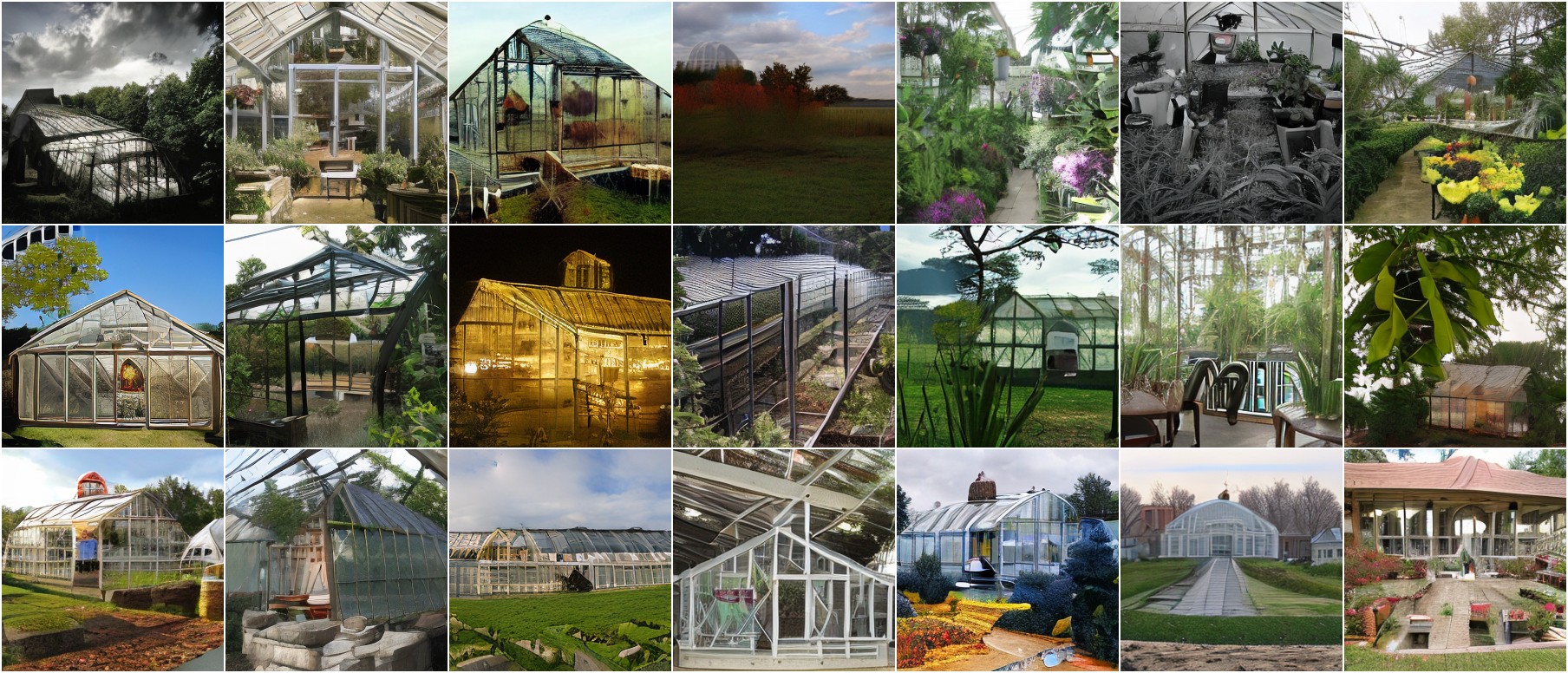}
        \centering\small Class 580: greenhouse, nursery, glasshouse
    \end{minipage}\hfill
    \begin{minipage}[b]{0.48\textwidth}
        \includegraphics[width=\linewidth]{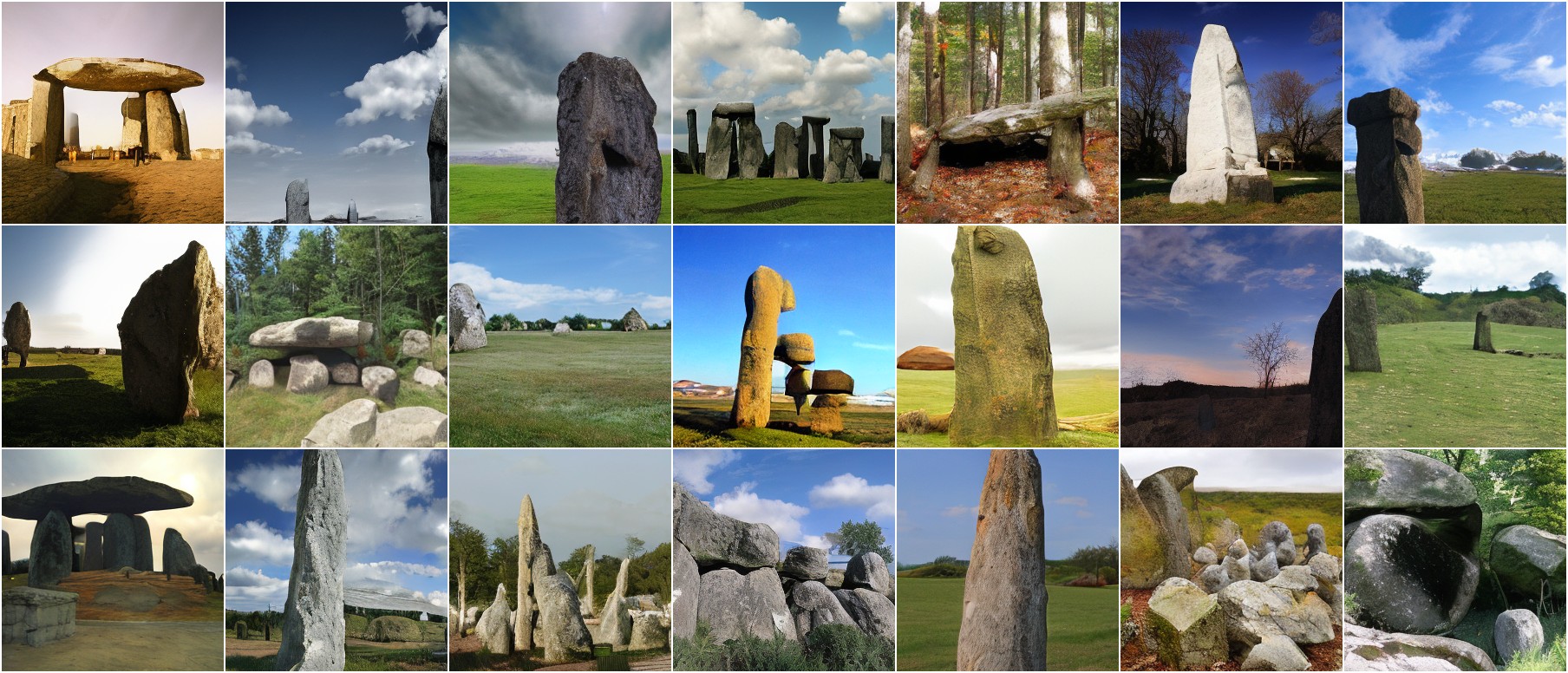}
        \centering\small Class 649: megalith, megalithic structure
    \end{minipage}\\[0.4em]
    \begin{minipage}[b]{0.48\textwidth}
        \includegraphics[width=\linewidth]{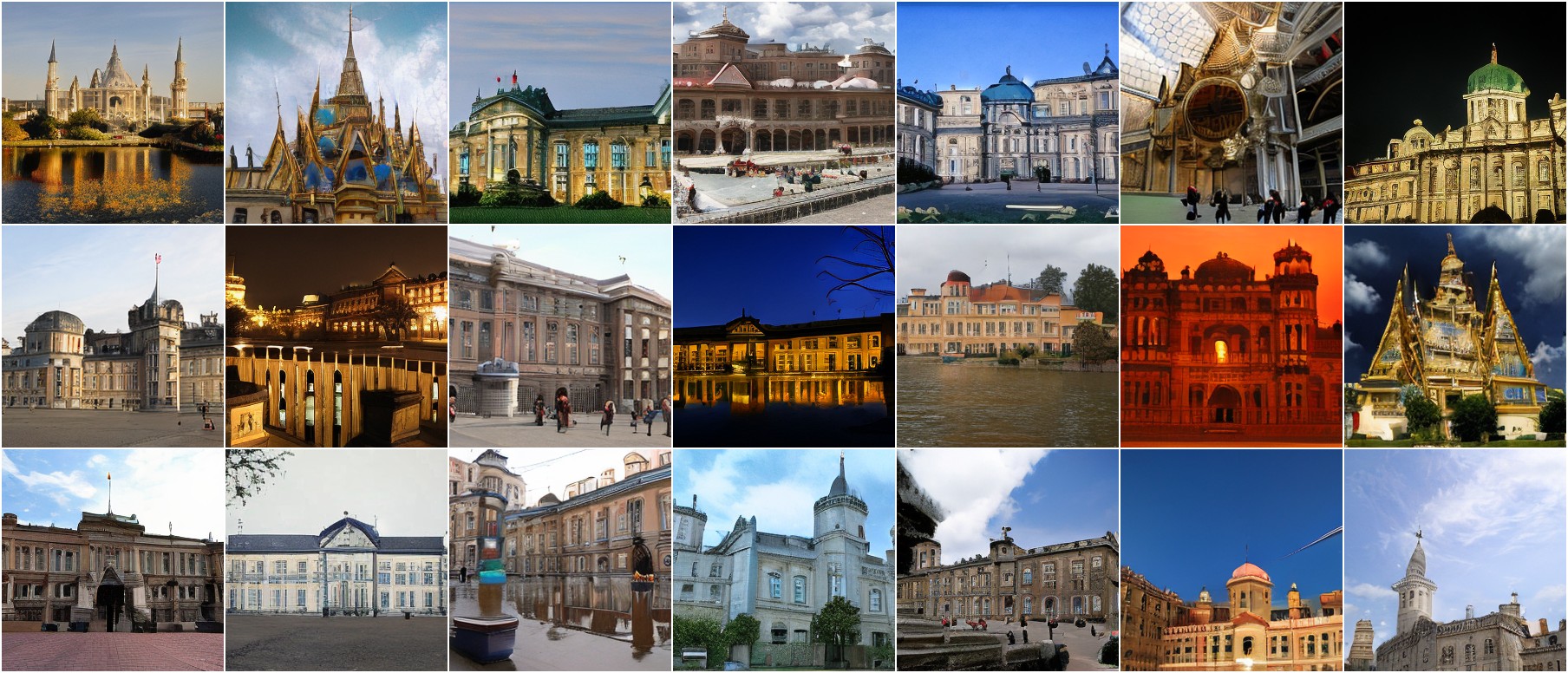}
        \centering\small Class 698: palace
    \end{minipage}\hfill
    \begin{minipage}[b]{0.48\textwidth}
        \includegraphics[width=\linewidth]{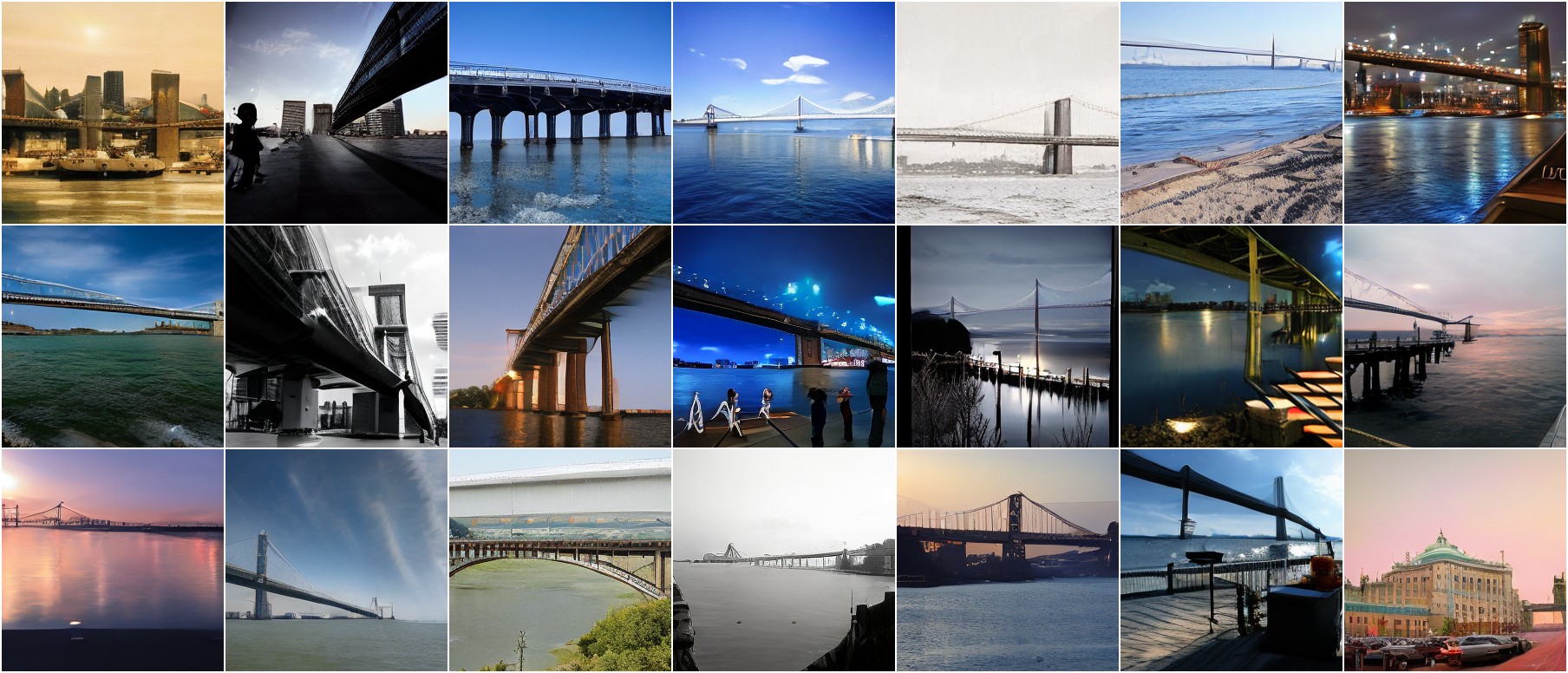}
        \centering\small Class 718: pier
    \end{minipage}\\[0.4em]
    \begin{minipage}[b]{0.48\textwidth}
        \includegraphics[width=\linewidth]{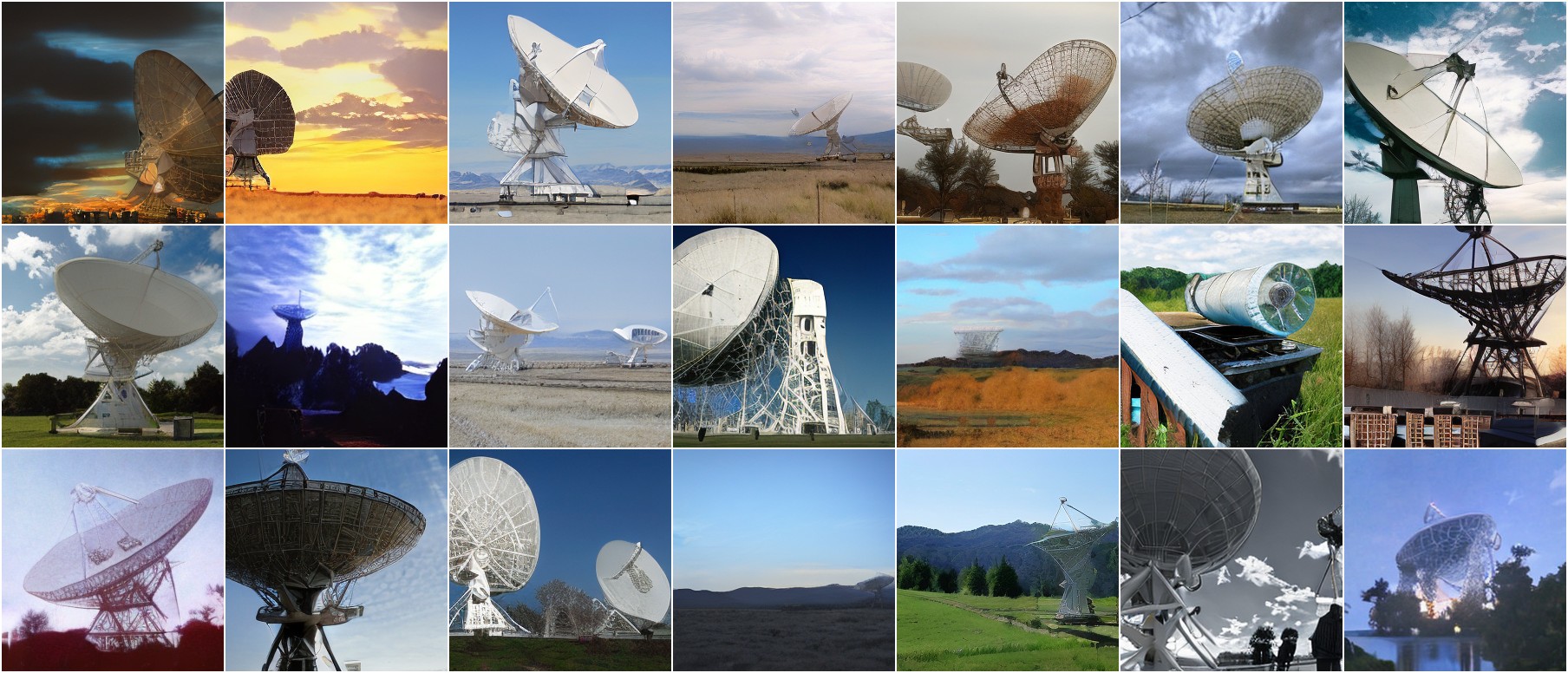}
        \centering\small Class 755: radio telescope, radio reflector
    \end{minipage}\hfill
    \begin{minipage}[b]{0.48\textwidth}
        \includegraphics[width=\linewidth]{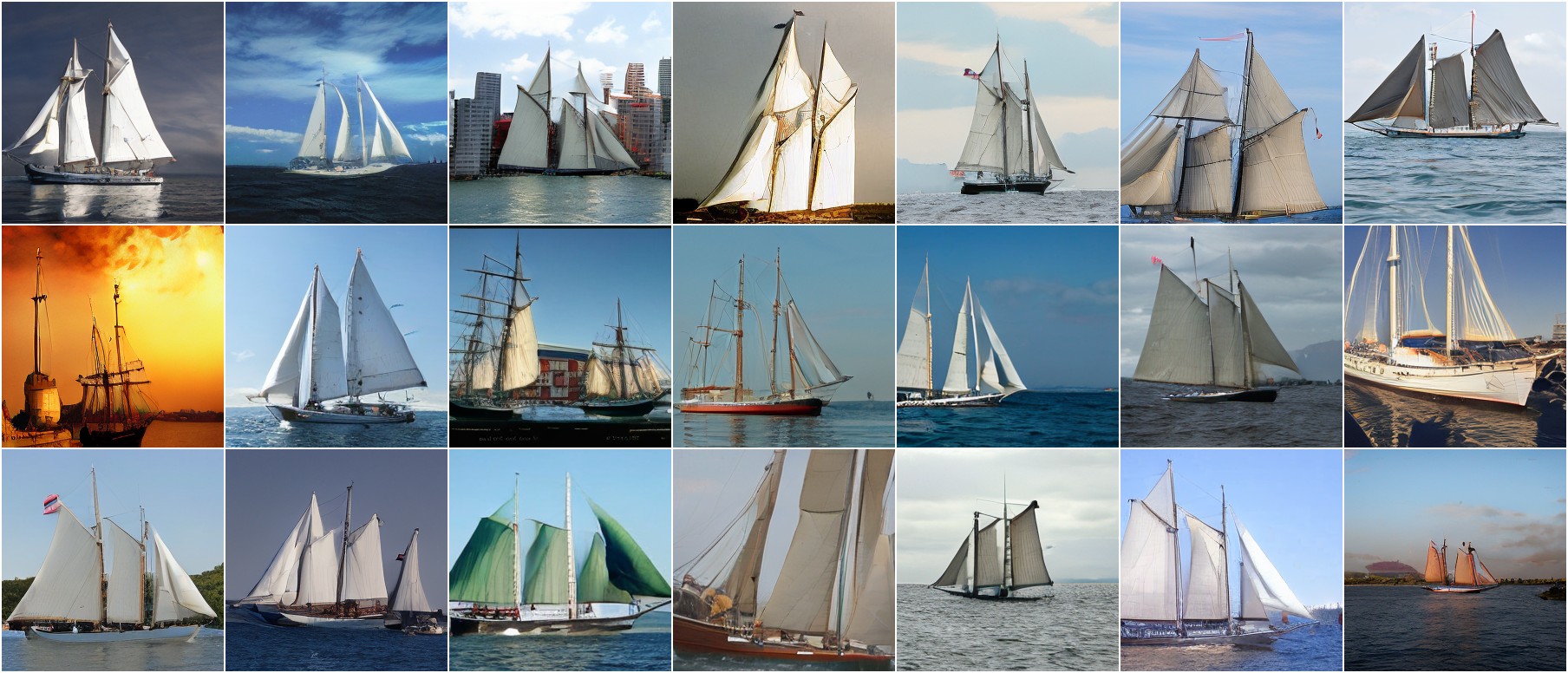}
        \centering\small Class 780: schooner
    \end{minipage}
    \caption{\textbf{Uncurated samples from our latent-L/2 model with CFG = 1.0 (page 3/4).} FID = 1.54, IS = 258.9.}
    \label{fig:latent_uncurated_samples_3}
\end{figure*}

\begin{figure*}[p]
    \centering
    \begin{minipage}[b]{0.48\textwidth}
        \includegraphics[width=\linewidth]{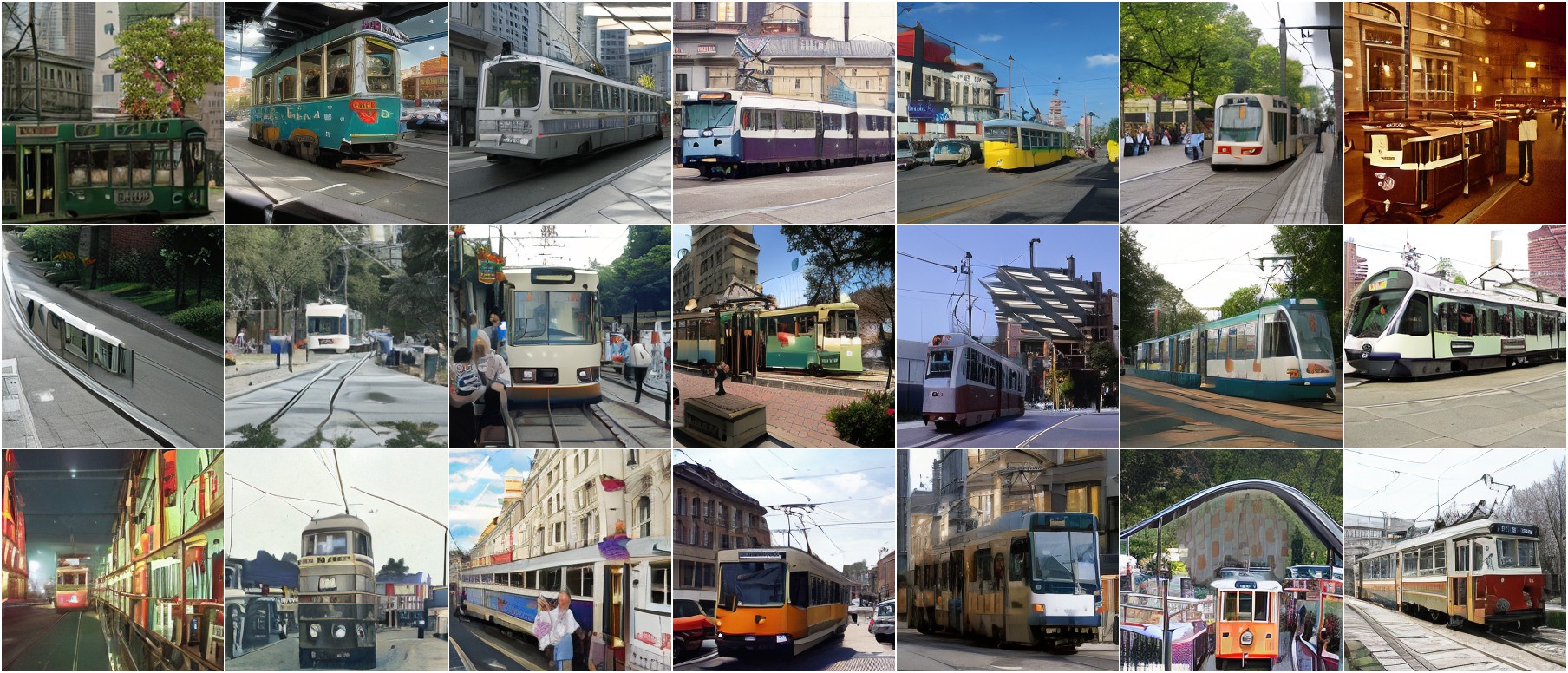}
        \centering\small Class 829: streetcar, tram, tramcar, trolley, trolley car
    \end{minipage}\hfill
    \begin{minipage}[b]{0.48\textwidth}
        \includegraphics[width=\linewidth]{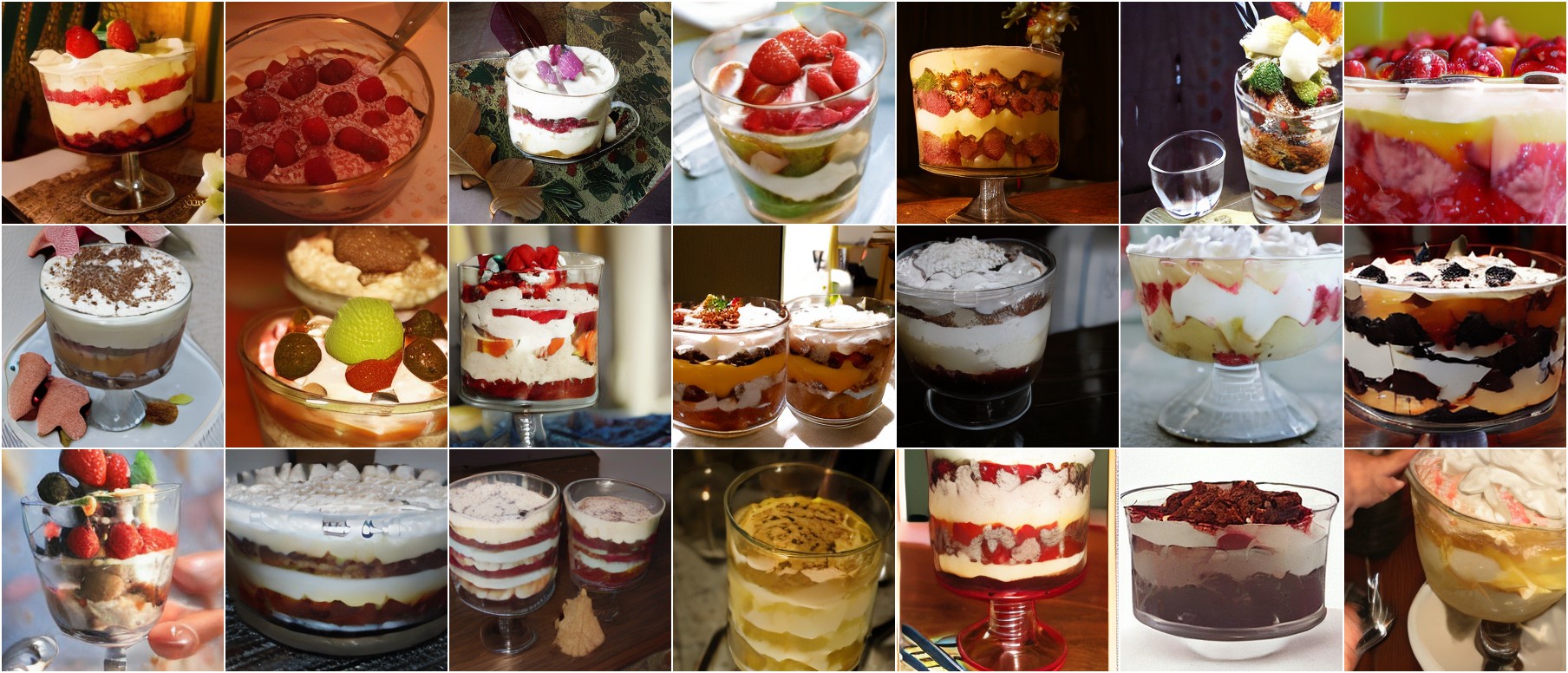}
        \centering\small Class 927: trifle
    \end{minipage}\\[0.4em]
    \begin{minipage}[b]{0.48\textwidth}
        \includegraphics[width=\linewidth]{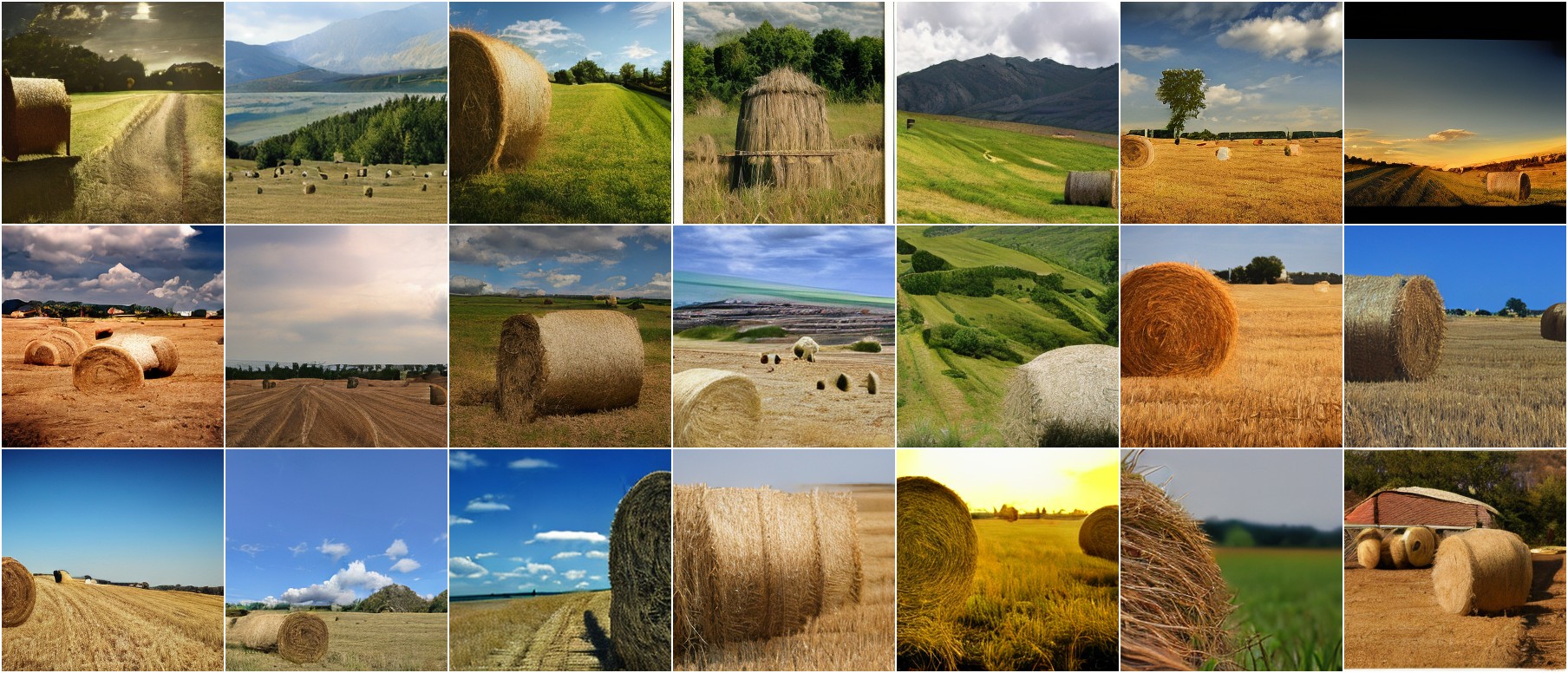}
        \centering\small Class 958: hay
    \end{minipage}\hfill
    \begin{minipage}[b]{0.48\textwidth}
        \includegraphics[width=\linewidth]{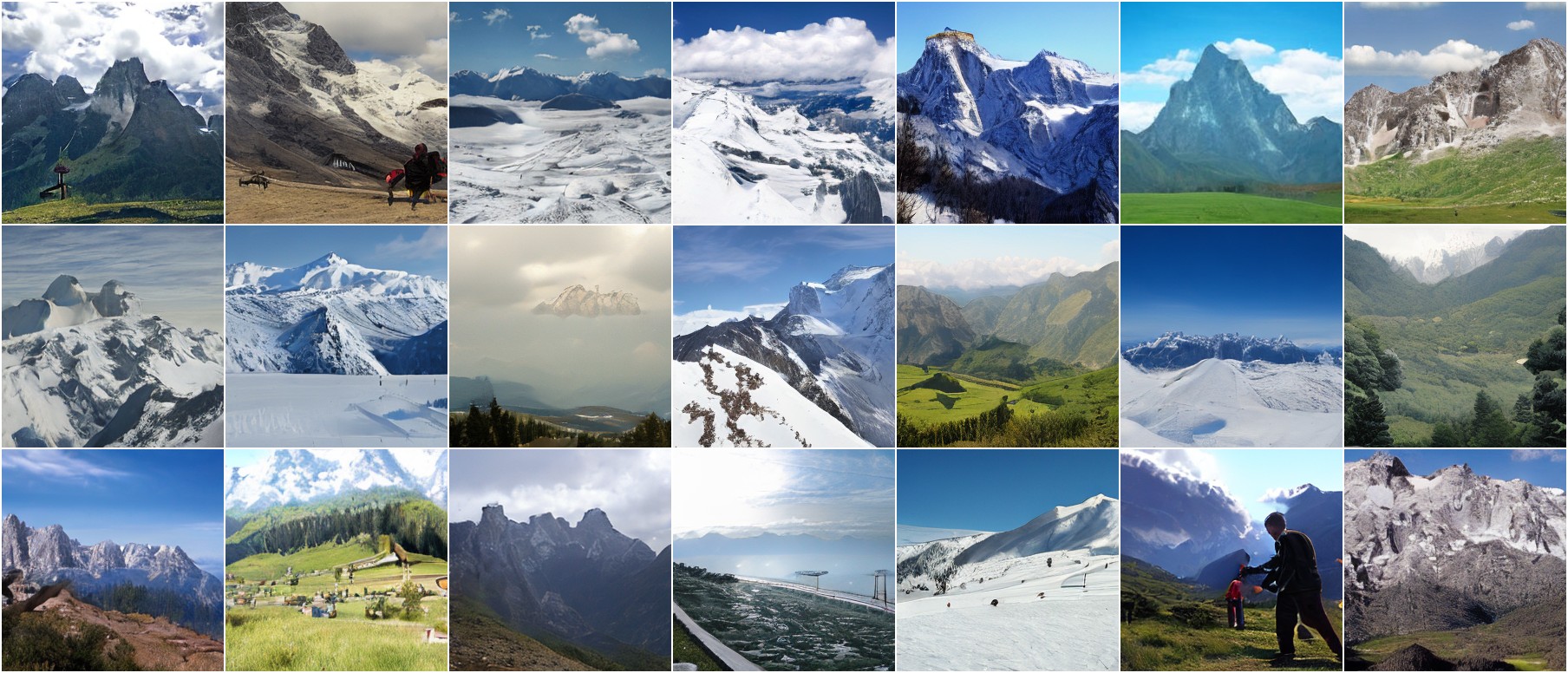}
        \centering\small Class 970: alp
    \end{minipage}\\[0.4em]
    \begin{minipage}[b]{0.48\textwidth}
        \includegraphics[width=\linewidth]{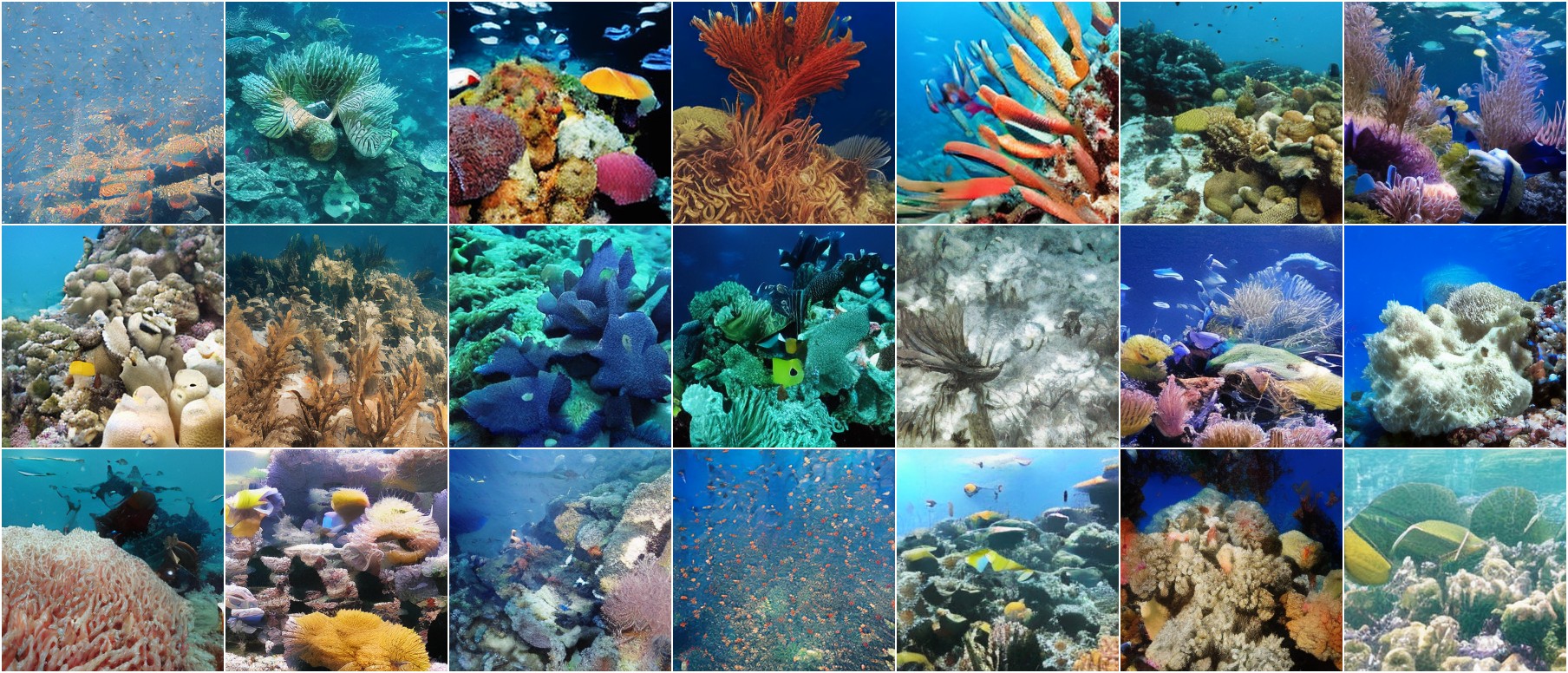}
        \centering\small Class 973: coral reef
    \end{minipage}\hfill
    \begin{minipage}[b]{0.48\textwidth}
        \includegraphics[width=\linewidth]{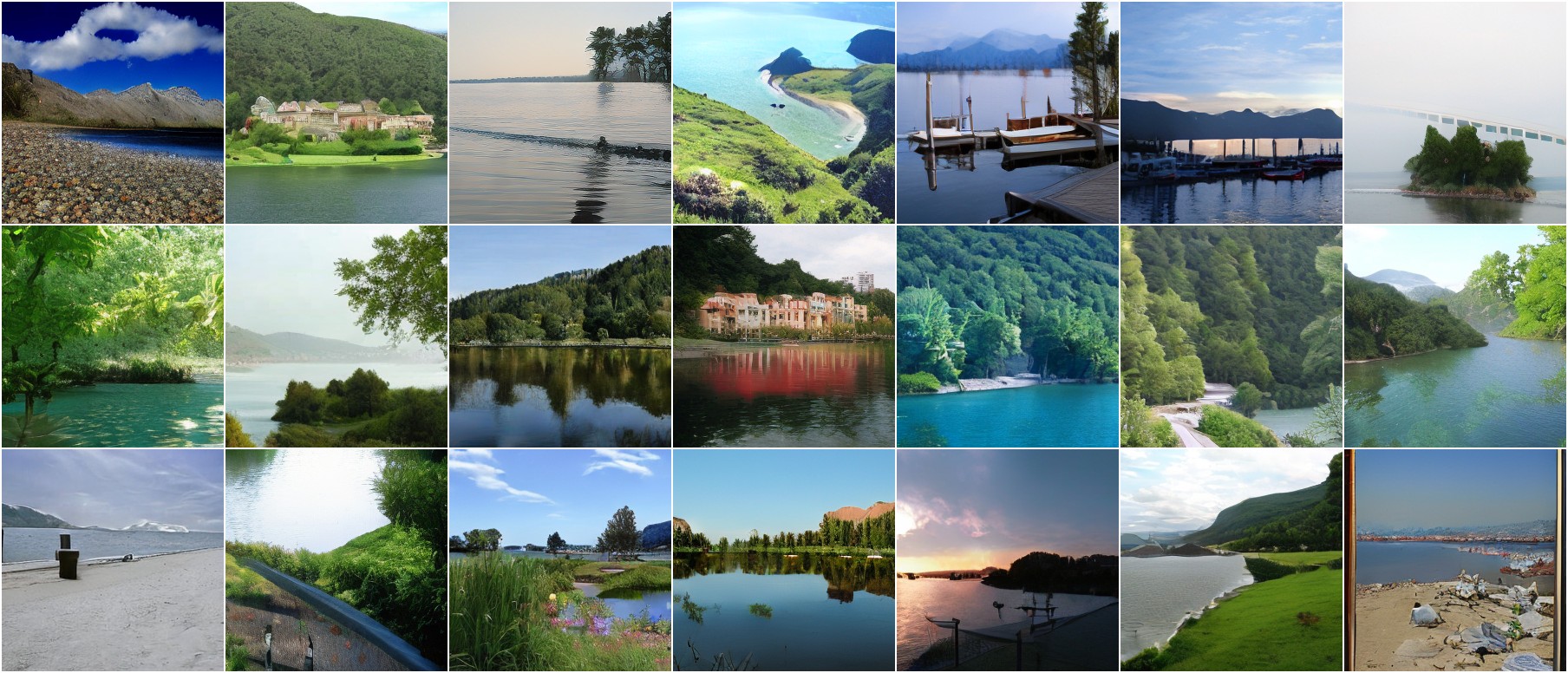}
        \centering\small Class 975: lakeside, lakeshore
    \end{minipage}\\[0.4em]
    \begin{minipage}[b]{0.48\textwidth}
        \includegraphics[width=\linewidth]{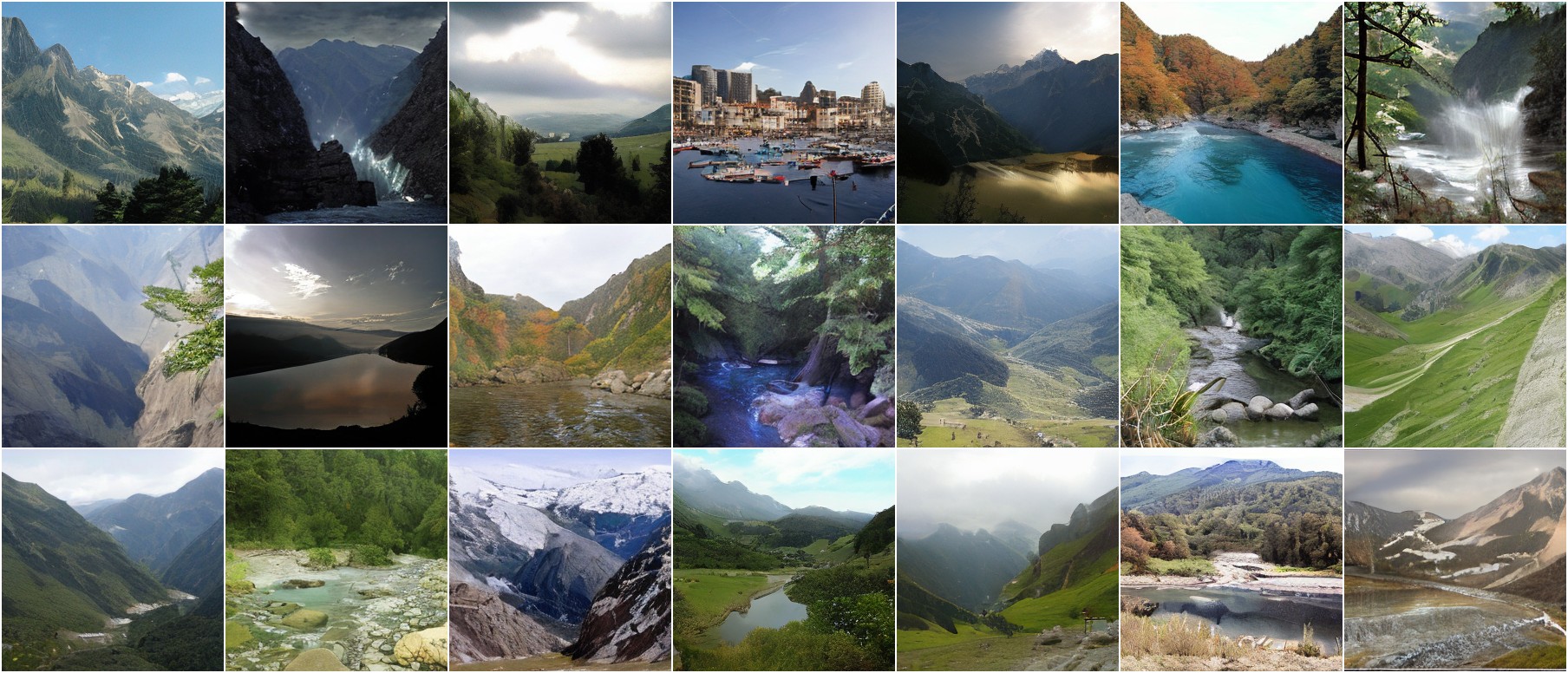}
        \centering\small Class 979: valley, vale
    \end{minipage}\hfill
    \begin{minipage}[b]{0.48\textwidth}
        \includegraphics[width=\linewidth]{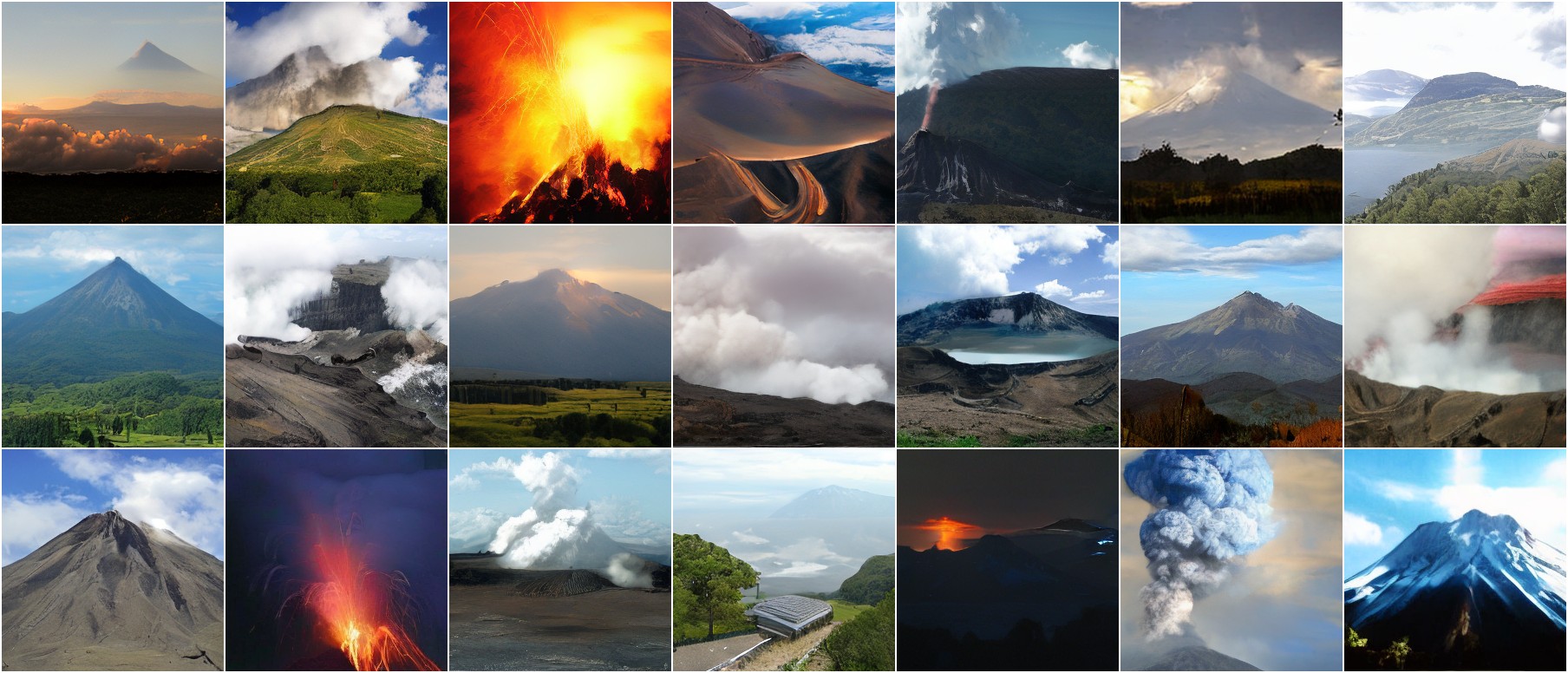}
        \centering\small Class 980: volcano
    \end{minipage}\\[0.4em]
    \begin{minipage}[b]{0.48\textwidth}
        \includegraphics[width=\linewidth]{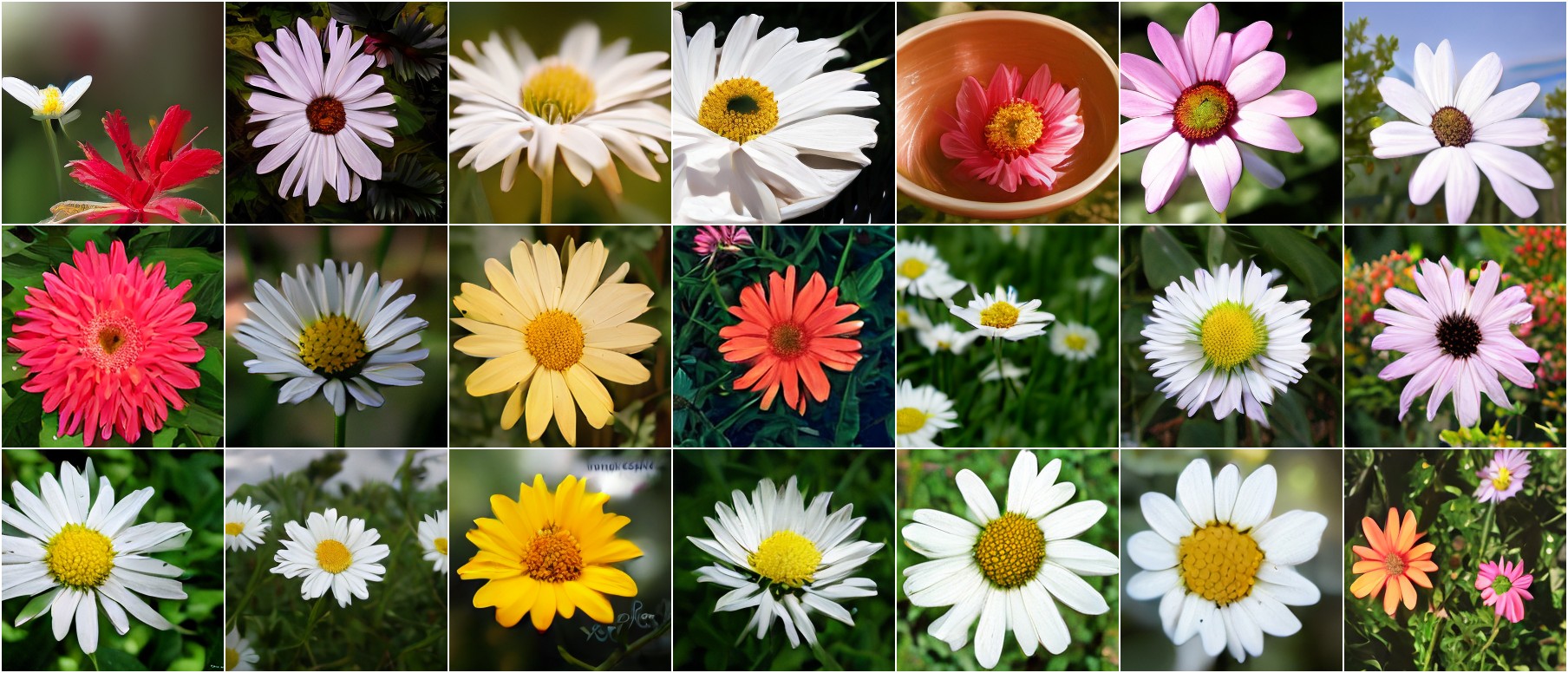}
        \centering\small Class 985: daisy
    \end{minipage}\hfill
    \begin{minipage}[b]{0.48\textwidth}
        \includegraphics[width=\linewidth]{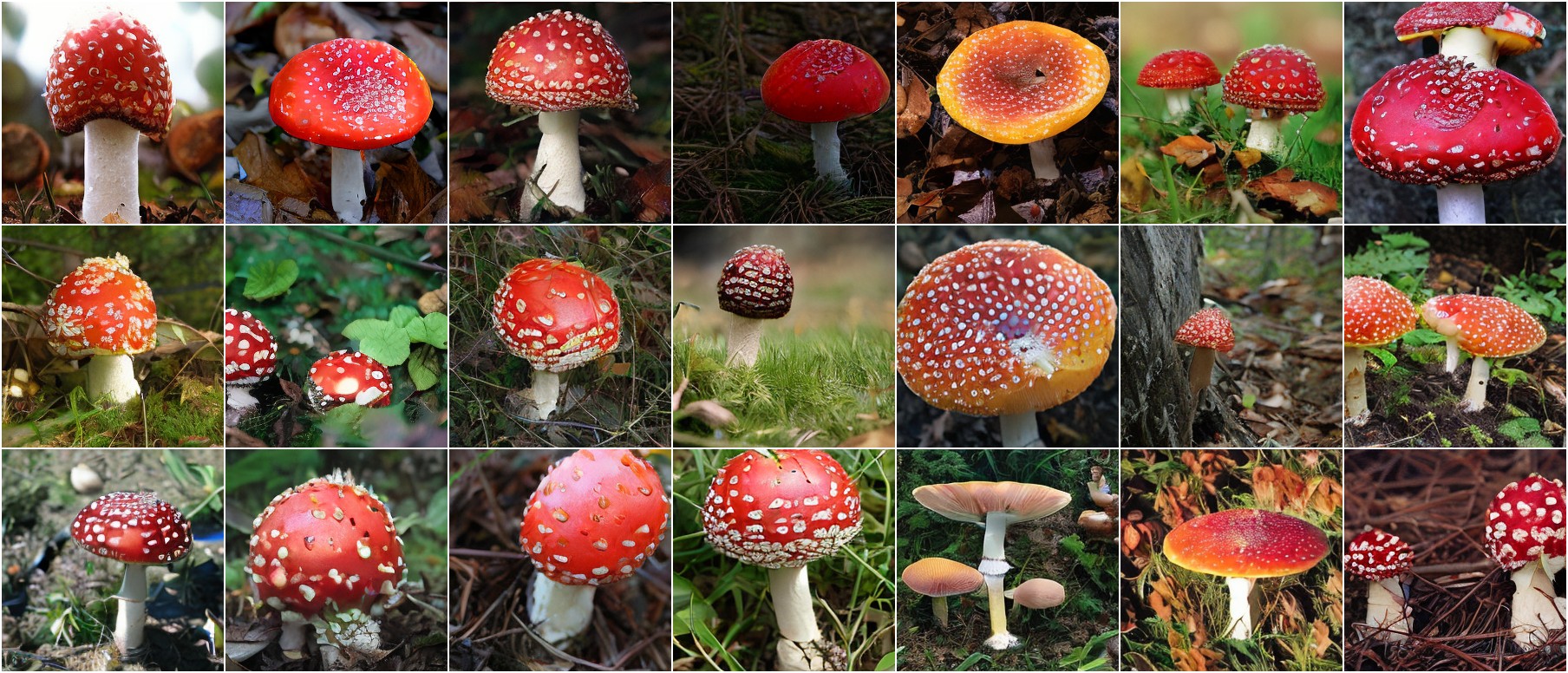}
        \centering\small Class 992: agaric
    \end{minipage}
    \caption{\textbf{Uncurated samples from our latent-L/2 model with CFG = 1.0 (page 4/4).} FID = 1.54, IS = 258.9.}
    \label{fig:latent_uncurated_samples_4}
\end{figure*}

\begin{figure*}[p]
    \centering
    \begin{minipage}{0.48\textwidth}
        \centering{ours}
    \end{minipage}\hfill
    \begin{minipage}{0.48\textwidth}
        \centering{improved MeanFlow (iMF)}
    \end{minipage}\\[0.2em]

    \begin{minipage}[b]{0.48\textwidth}
        \includegraphics[width=\linewidth]{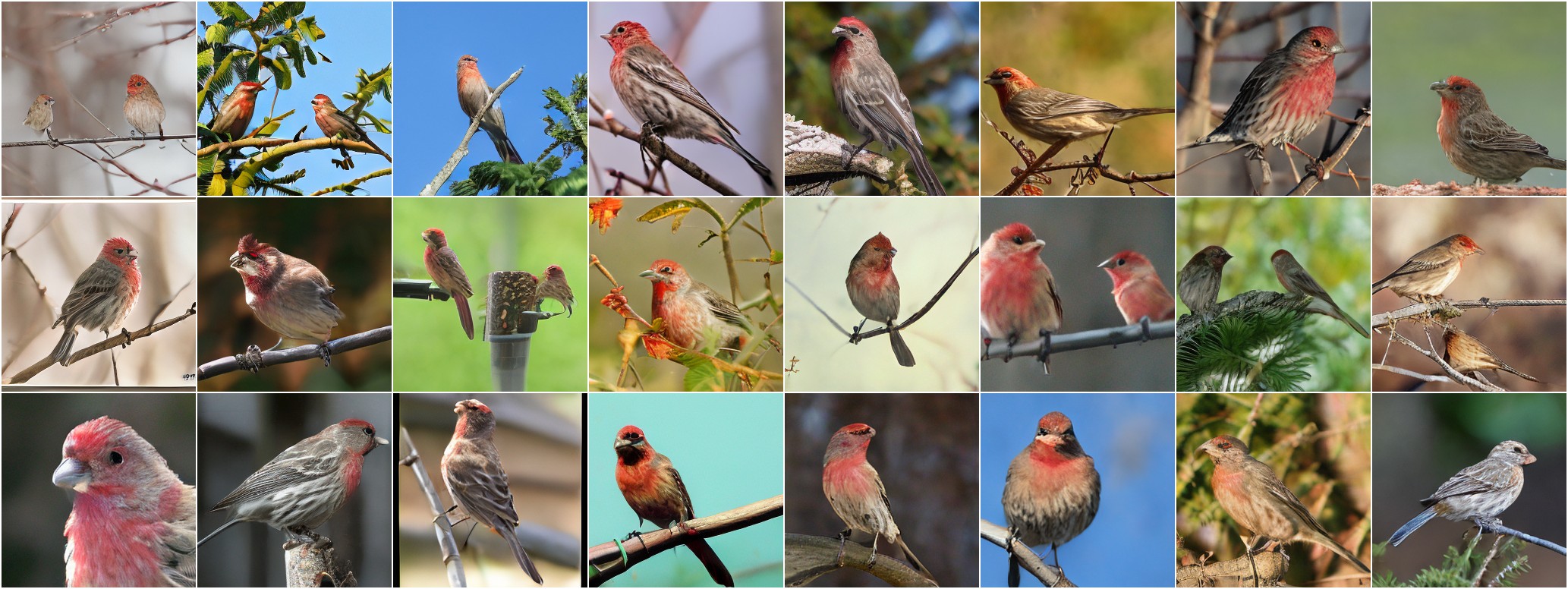}
    \end{minipage}\hfill
    \begin{minipage}[b]{0.48\textwidth}
        \includegraphics[width=\linewidth]{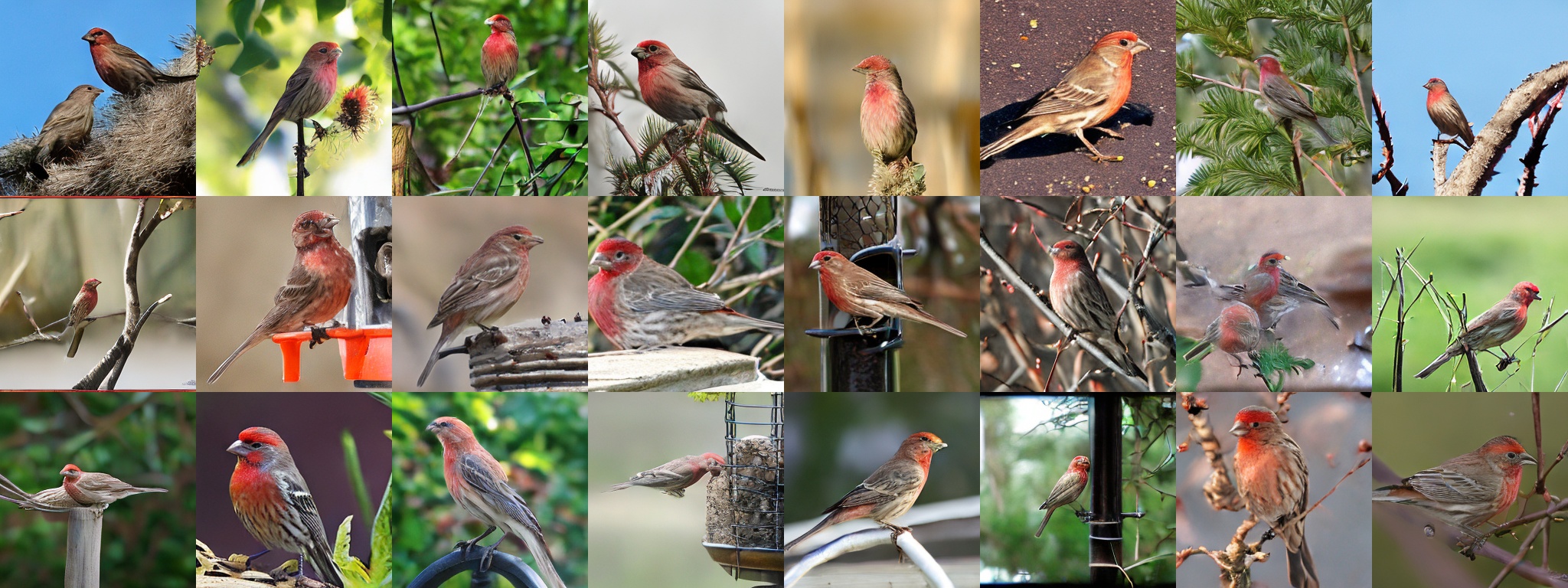}
    \end{minipage}\\[-0.2em]
    \centering\small Class 012: House finch\\[0.2em]

    \begin{minipage}[b]{0.48\textwidth}
        \includegraphics[width=\linewidth]{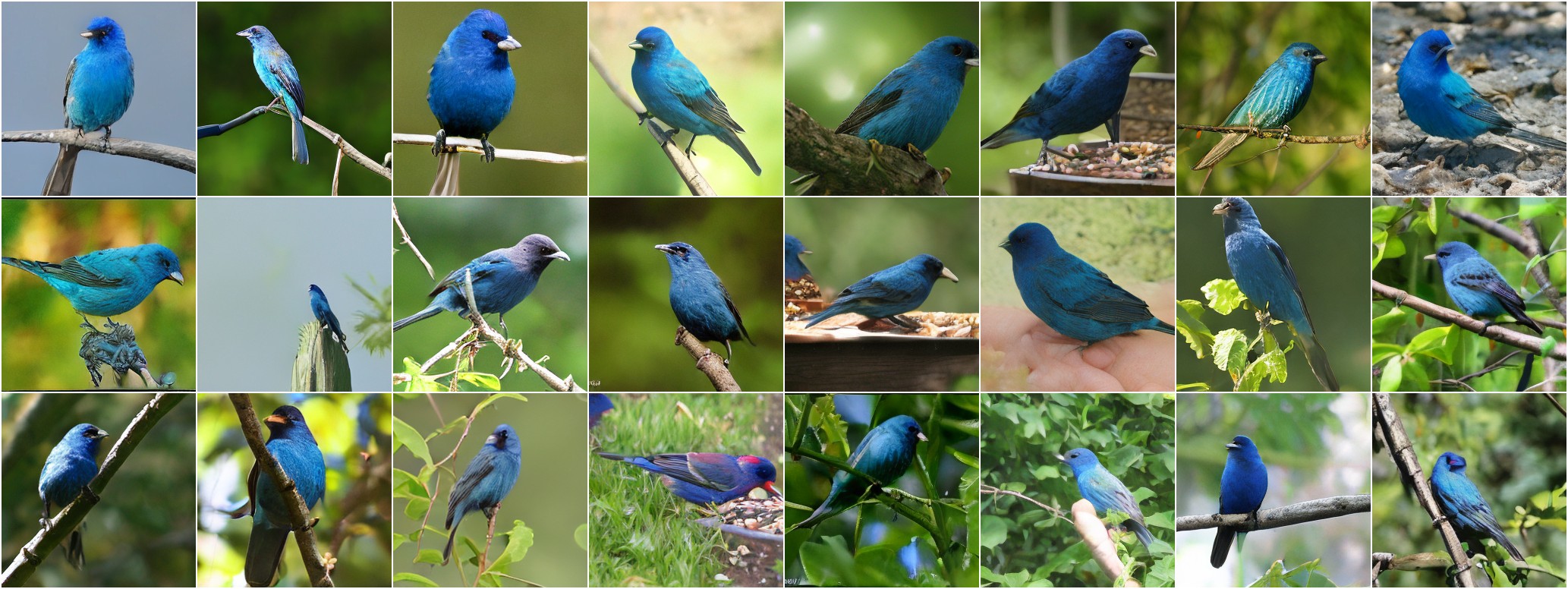}
    \end{minipage}\hfill
    \begin{minipage}[b]{0.48\textwidth}
        \includegraphics[width=\linewidth]{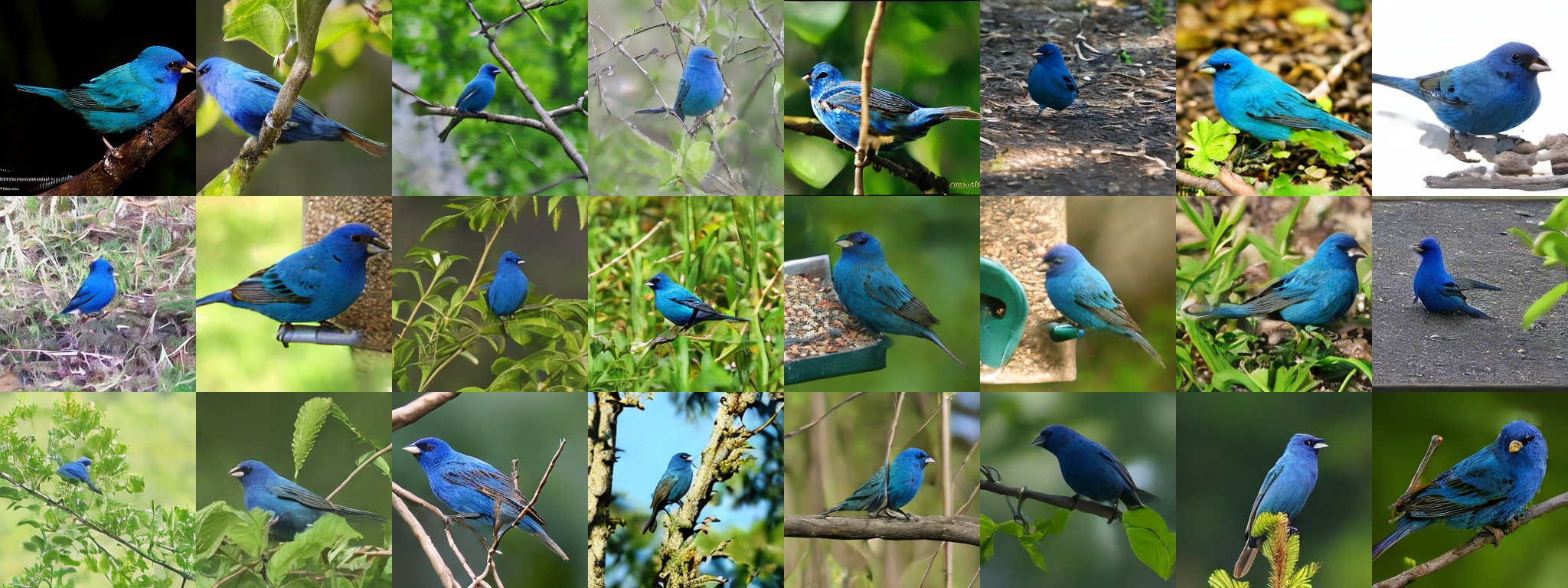}
    \end{minipage}\\[-0.2em]
    \centering\small Class 014: Indigo bunting\\[0.2em]

    \begin{minipage}[b]{0.48\textwidth}
        \includegraphics[width=\linewidth]{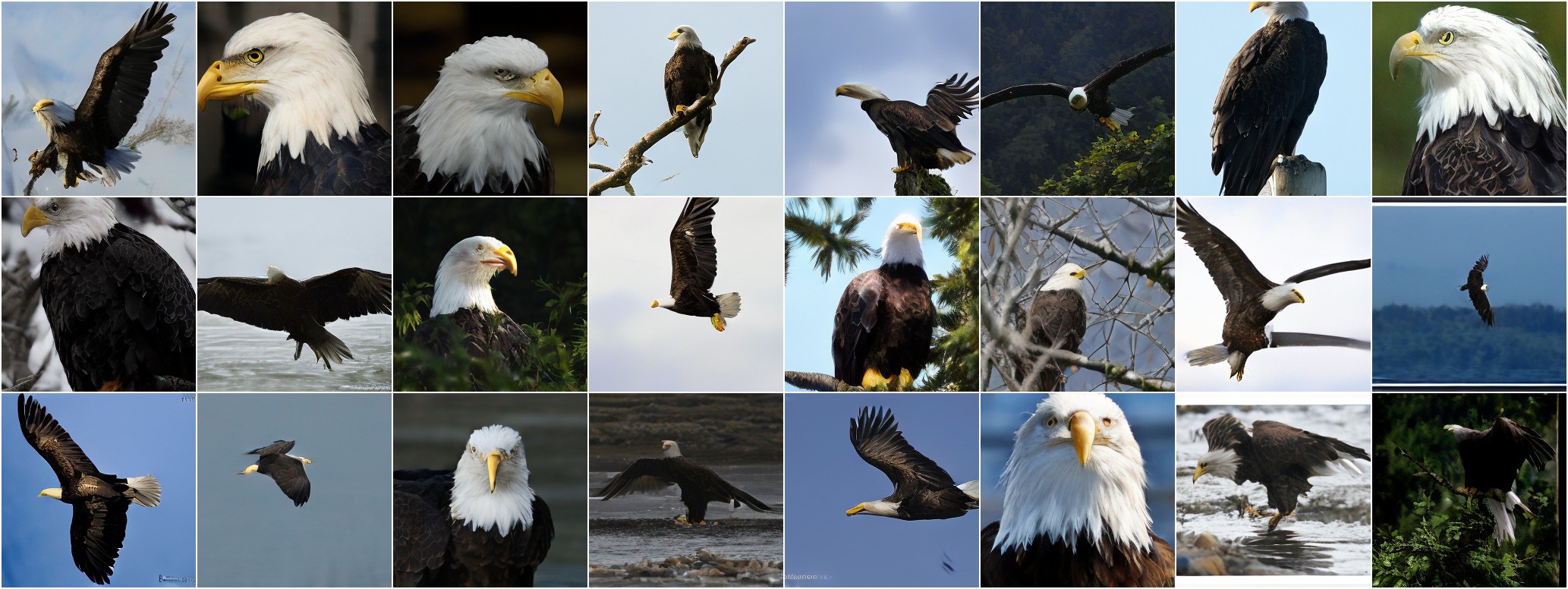}
    \end{minipage}\hfill
    \begin{minipage}[b]{0.48\textwidth}
        \includegraphics[width=\linewidth]{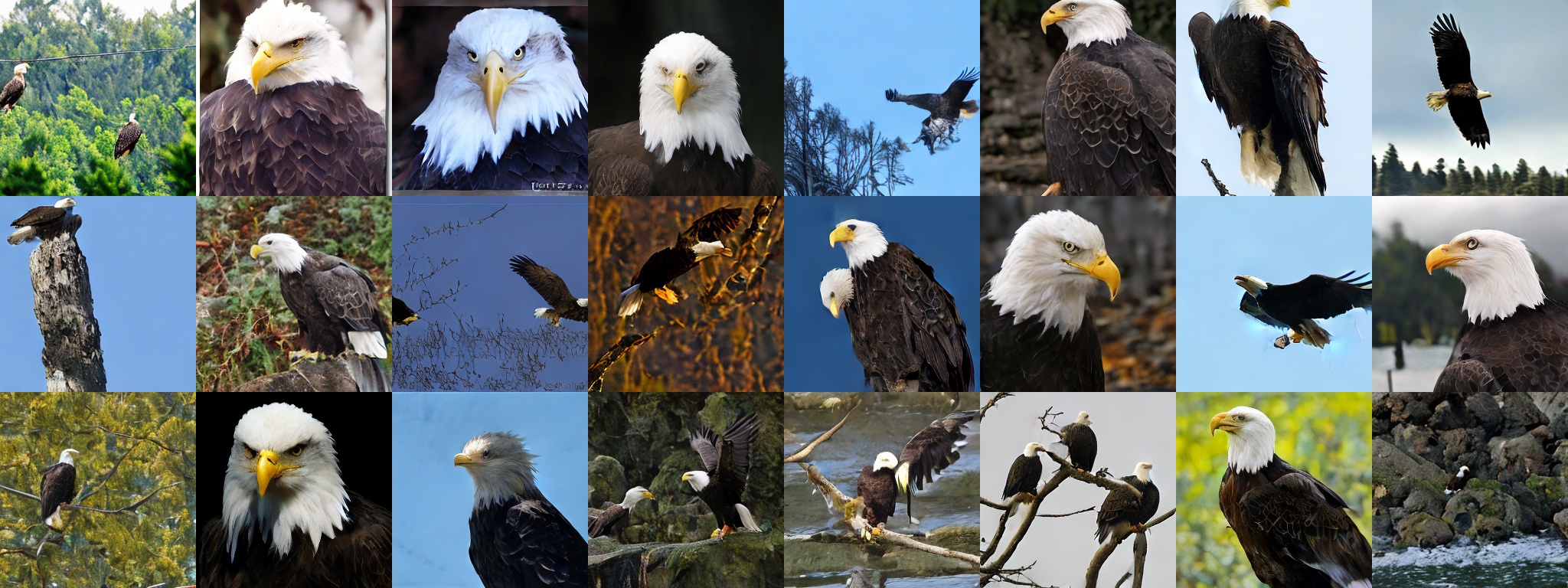}
    \end{minipage}\\[-0.2em]
    \centering\small Class 022: Bald eagle\\[0.2em]

    \begin{minipage}[b]{0.48\textwidth}
        \includegraphics[width=\linewidth]{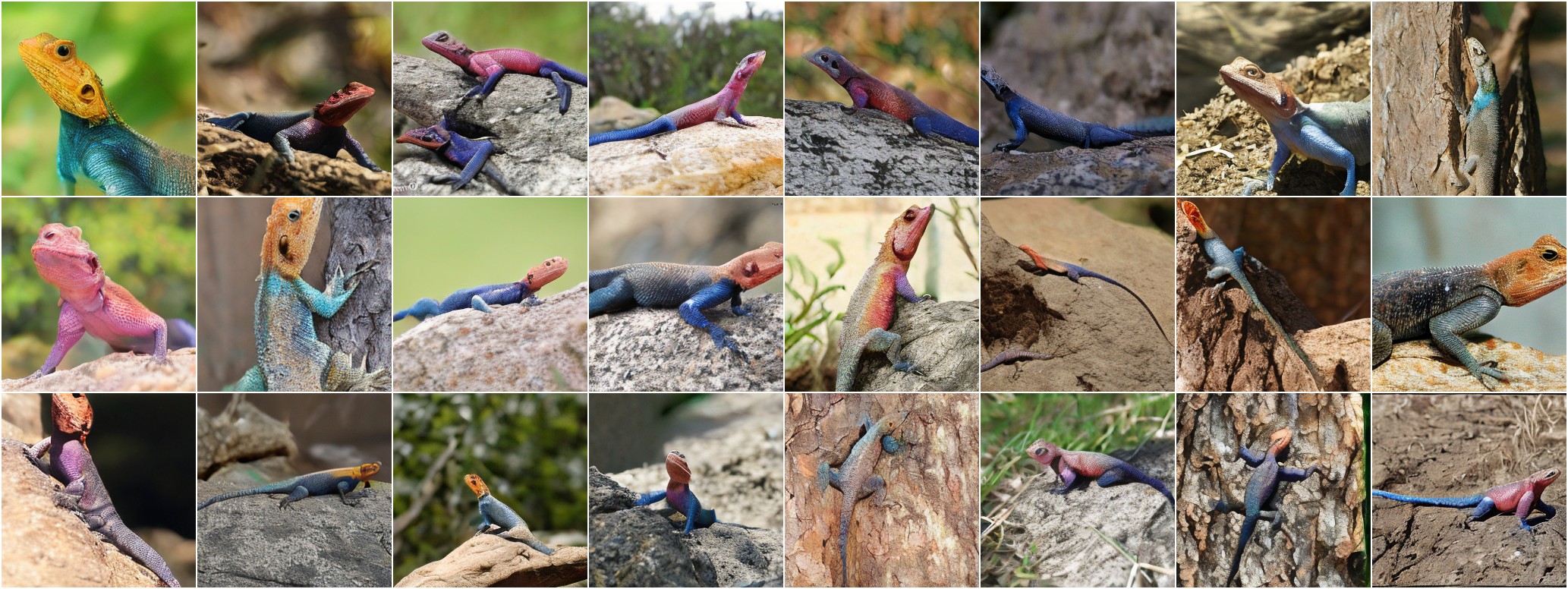}
    \end{minipage}\hfill
    \begin{minipage}[b]{0.48\textwidth}
        \includegraphics[width=\linewidth]{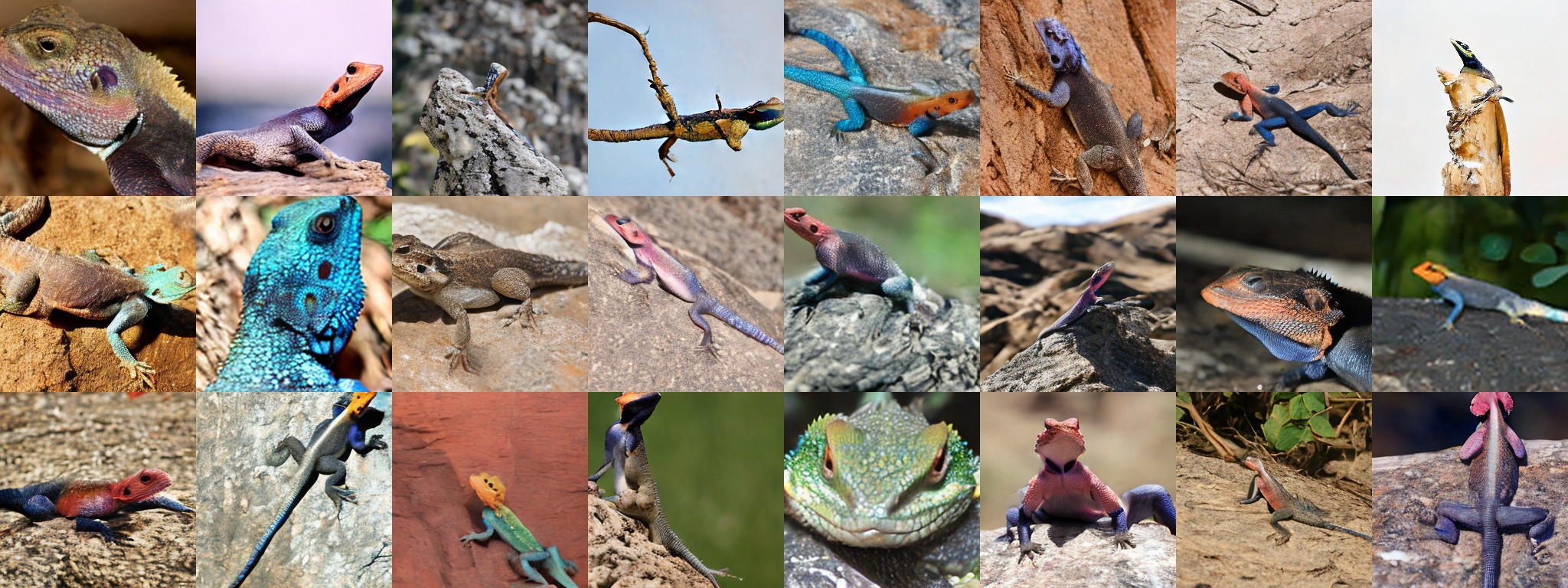}
    \end{minipage}\\[-0.2em]
    \centering\small Class 042: Agama\\[0.2em]

    \begin{minipage}[b]{0.48\textwidth}
        \includegraphics[width=\linewidth]{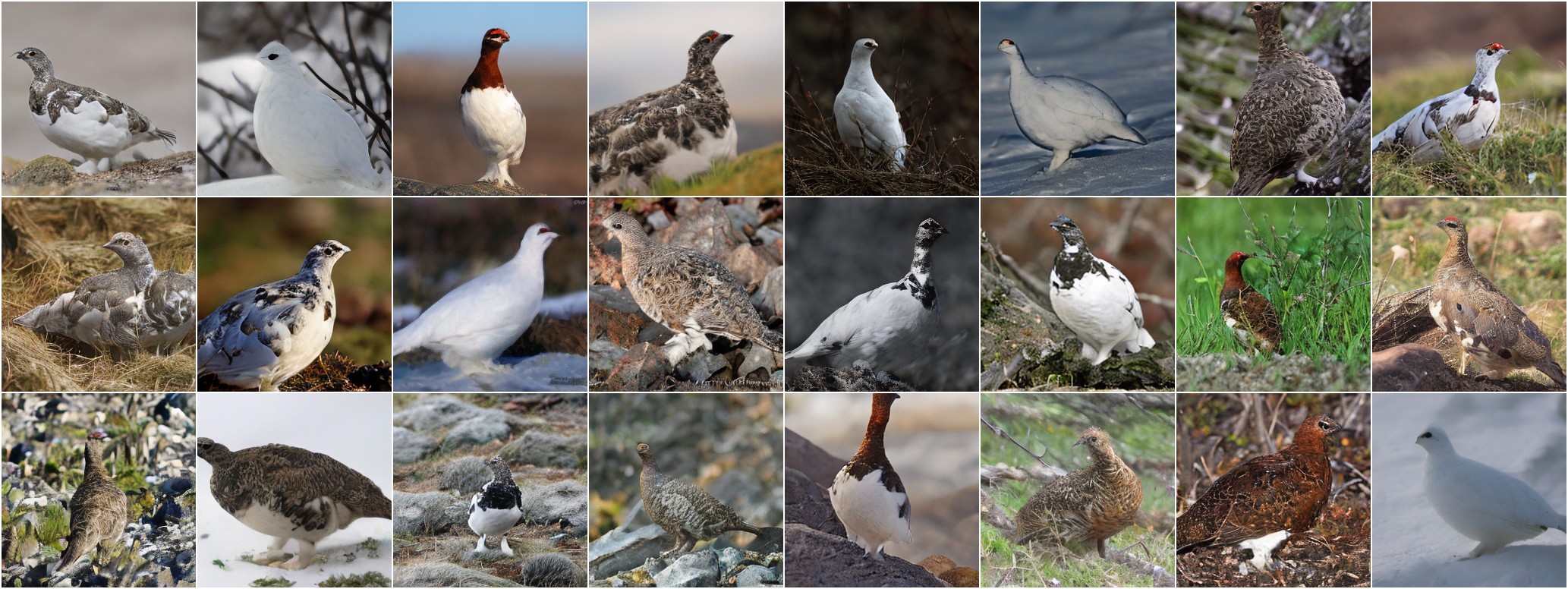}
    \end{minipage}\hfill
    \begin{minipage}[b]{0.48\textwidth}
        \includegraphics[width=\linewidth]{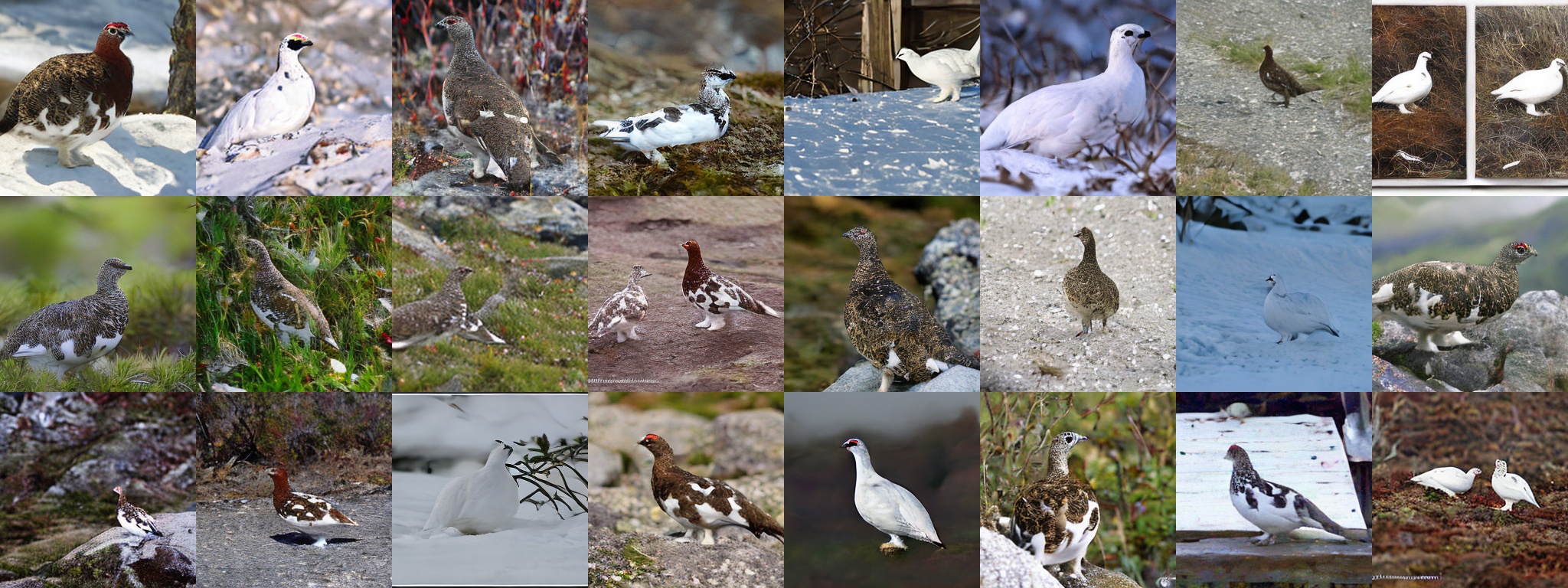}
    \end{minipage}\\[-0.2em]
    \centering\small Class 081: Ptarmigan
    
    \caption{\textbf{Side-by-side comparison with improved MeanFlow (iMF)~\cite{geng2025improved} (page 1/5).} Uncurated samples from our method (left) and iMF (right) on \textbf{all} ImageNet classes visualized in the iMF paper. Both methods generate images with a single neural function evaluation (1-NFE). The iMF visualizations use CFG $\omega{=}6.0$ and interval $[t_\text{min}, t_\text{max}]{=}[0.2, 0.8]$, achieving FID 3.92 and IS 348.2 (DiT-XL/2). For fair comparison, we set the CFG scale to match the IS of iMF visualizations, which leads to FID 3.01 and IS 354.4 (at CFG=1.5) for our method (DiT-L/2).}
    \label{fig:comparison_imf_1}
\end{figure*}

\begin{figure*}[p]
    \centering
    \begin{minipage}{0.48\textwidth}
        \centering{ours}
    \end{minipage}\hfill
    \begin{minipage}{0.48\textwidth}
        \centering{improved MeanFlow (iMF)}
    \end{minipage}\\[0.2em]

    \begin{minipage}[b]{0.48\textwidth}
        \includegraphics[width=\linewidth]{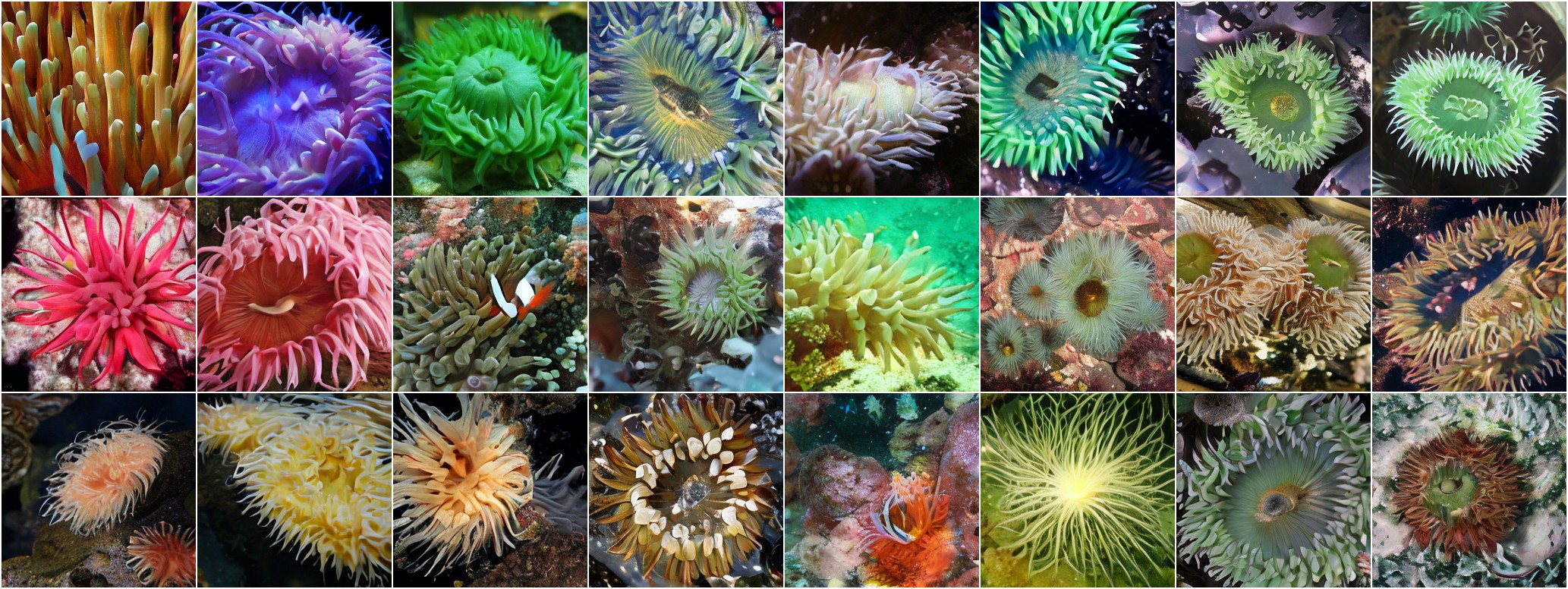}
    \end{minipage}\hfill
    \begin{minipage}[b]{0.48\textwidth}
        \includegraphics[width=\linewidth]{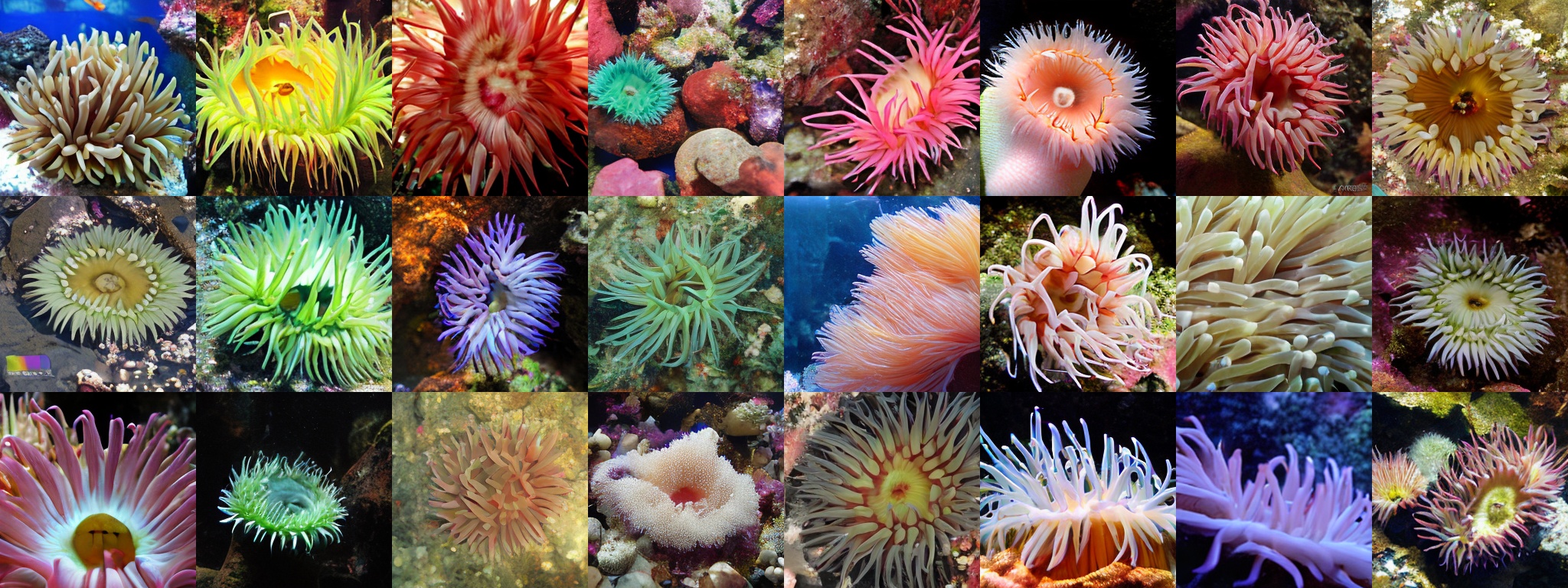}
    \end{minipage}\\[-0.2em]
    \centering\small Class 108: Sea anemone\\[0.2em]

    \begin{minipage}[b]{0.48\textwidth}
        \includegraphics[width=\linewidth]{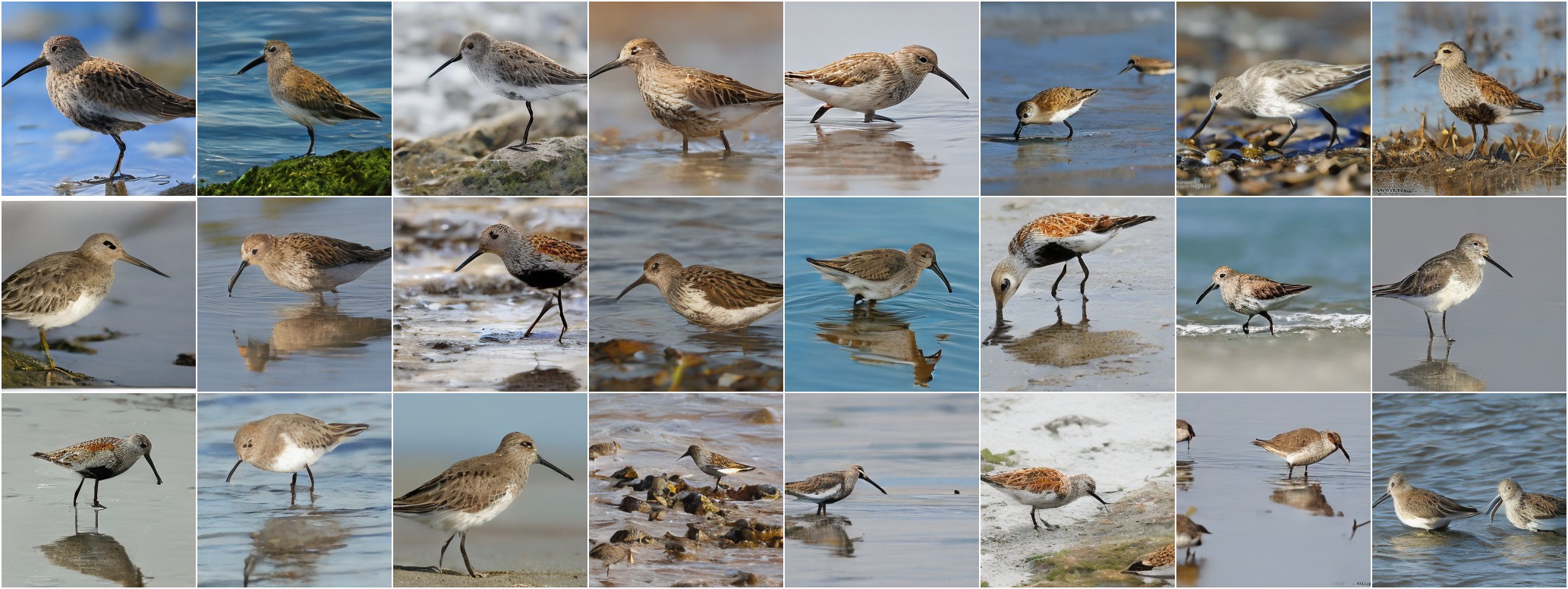}
    \end{minipage}\hfill
    \begin{minipage}[b]{0.48\textwidth}
        \includegraphics[width=\linewidth]{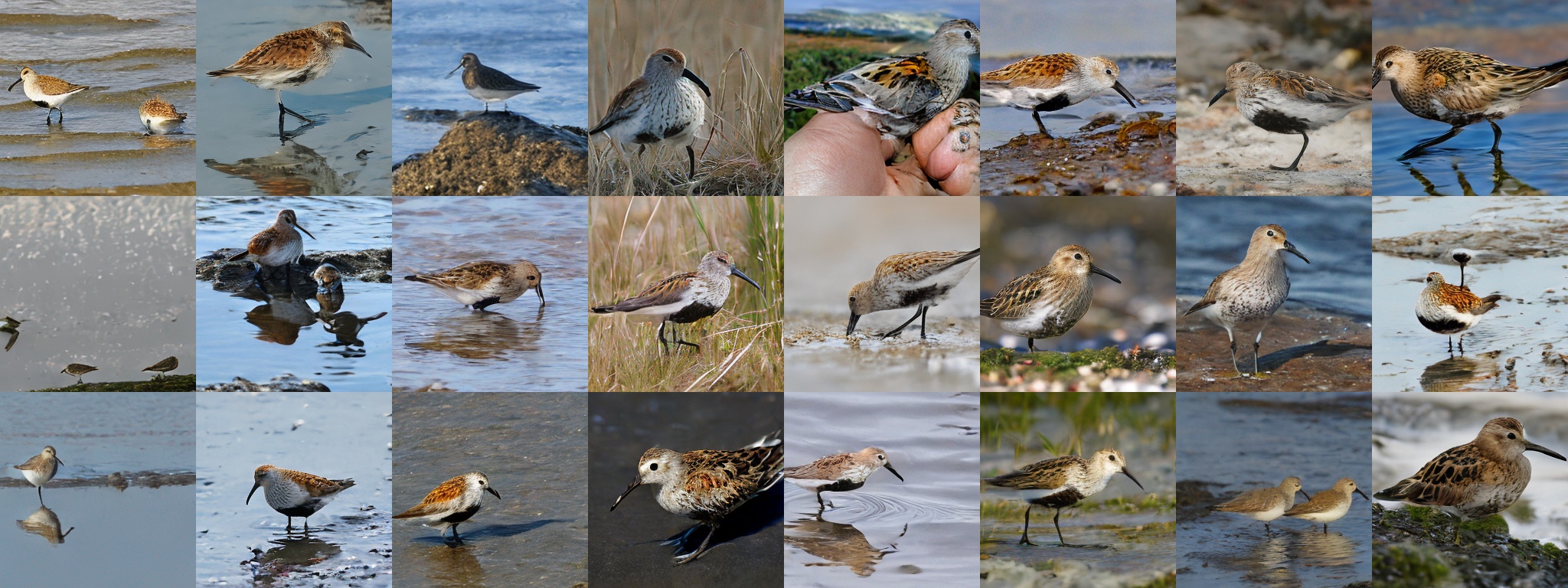}
    \end{minipage}\\[-0.2em]
    \centering\small Class 140: Red-backed sandpiper\\[0.2em]

    \begin{minipage}[b]{0.48\textwidth}
        \includegraphics[width=\linewidth]{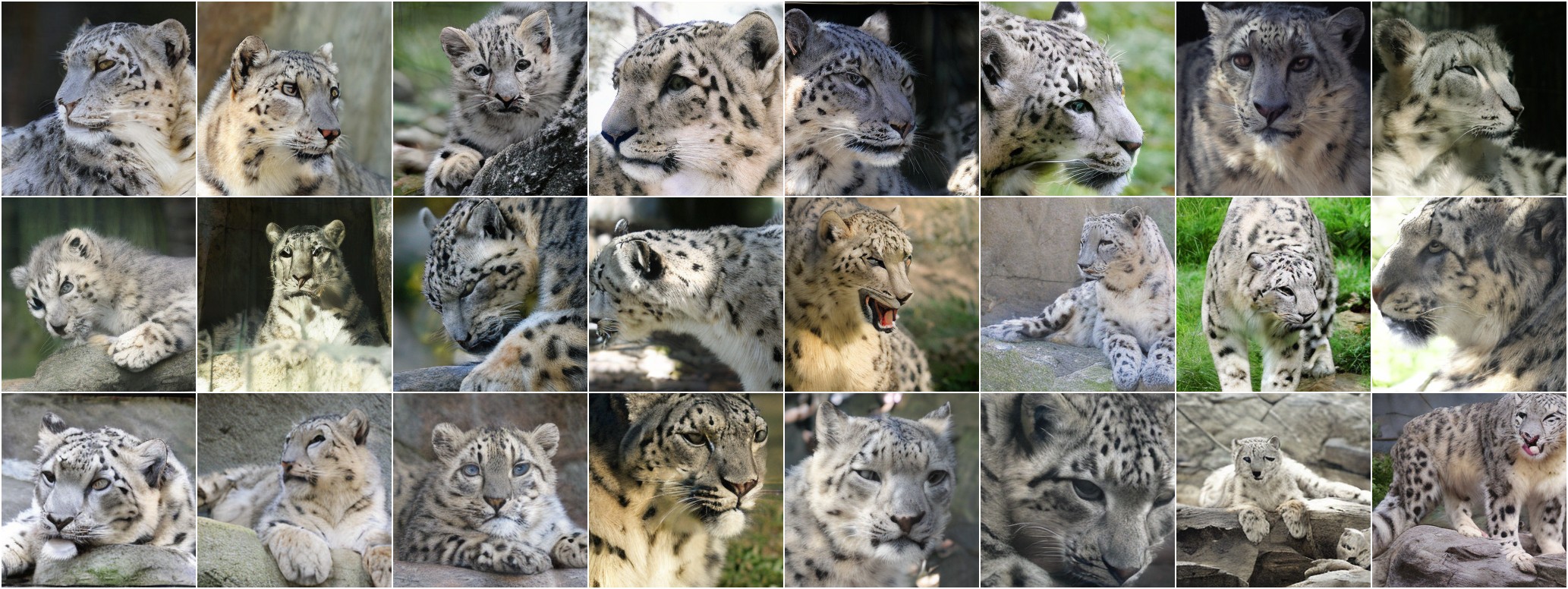}
    \end{minipage}\hfill
    \begin{minipage}[b]{0.48\textwidth}
        \includegraphics[width=\linewidth]{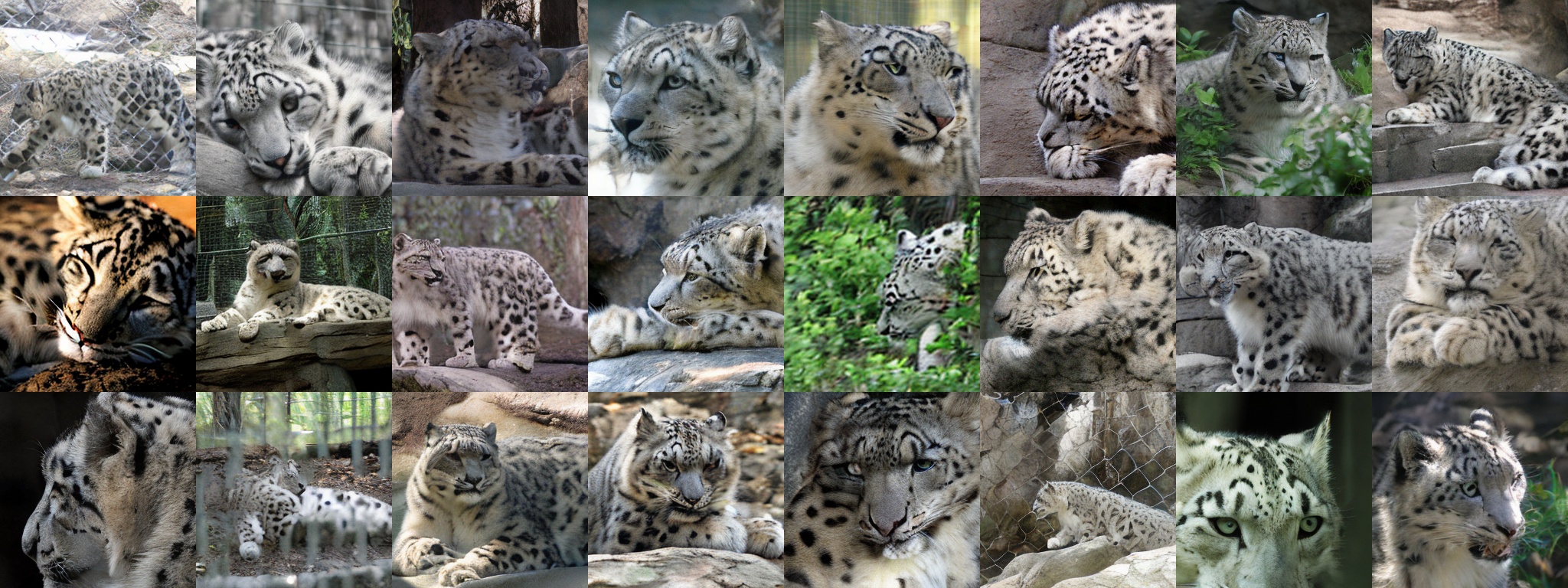}
    \end{minipage}\\[-0.2em]
    \centering\small Class 289: Snow leopard\\[0.2em]

    \begin{minipage}[b]{0.48\textwidth}
        \includegraphics[width=\linewidth]{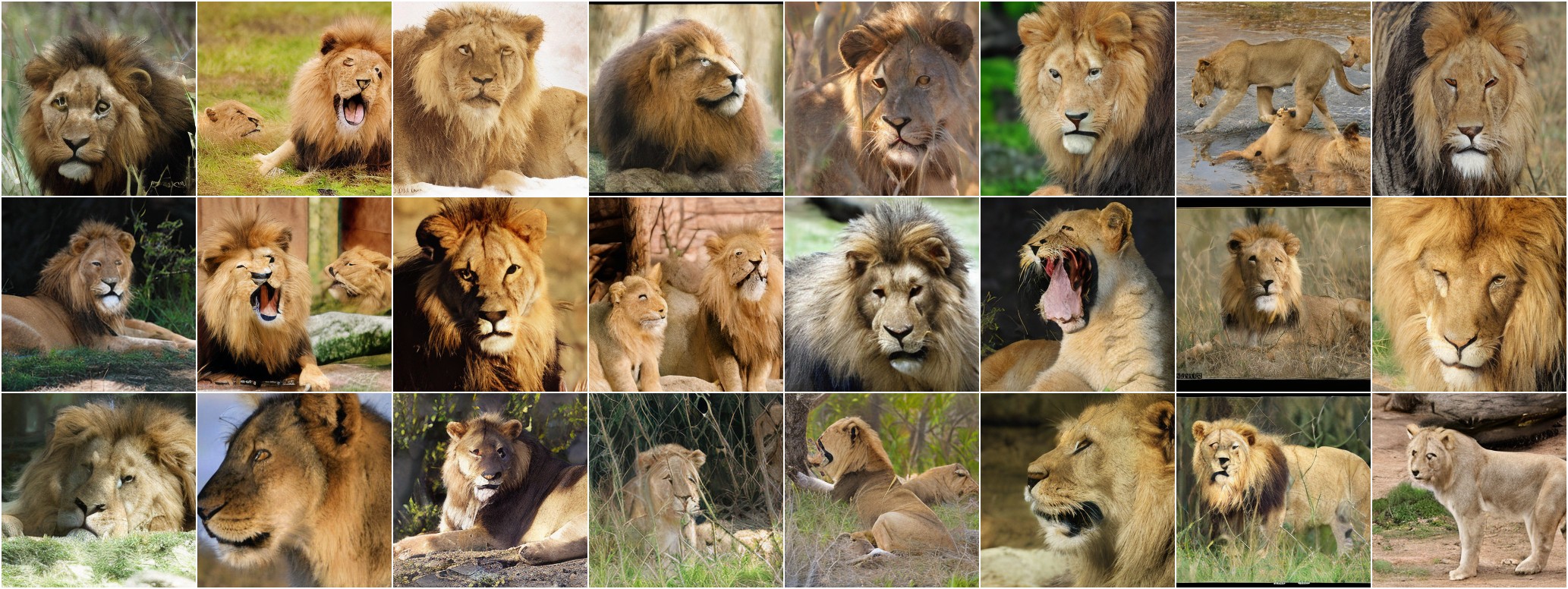}
    \end{minipage}\hfill
    \begin{minipage}[b]{0.48\textwidth}
        \includegraphics[width=\linewidth]{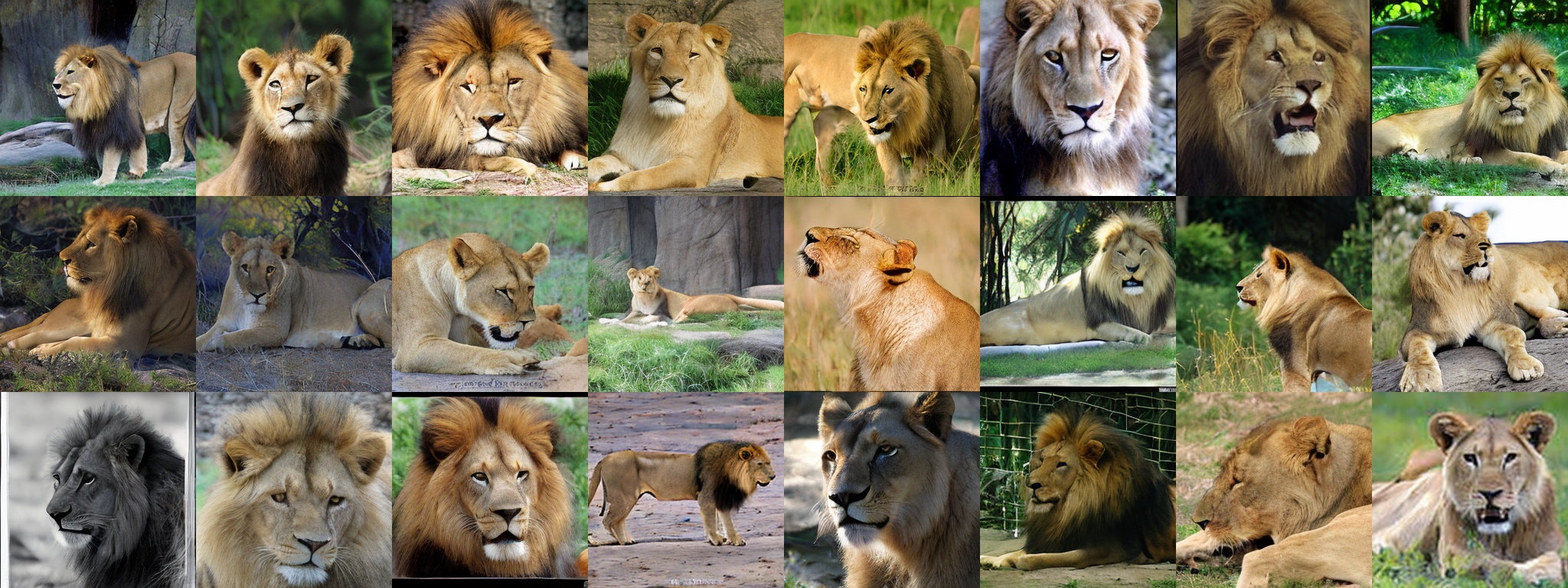}
    \end{minipage}\\[-0.2em]
    \centering\small Class 291: Lion\\[0.2em]

    \begin{minipage}[b]{0.48\textwidth}
        \includegraphics[width=\linewidth]{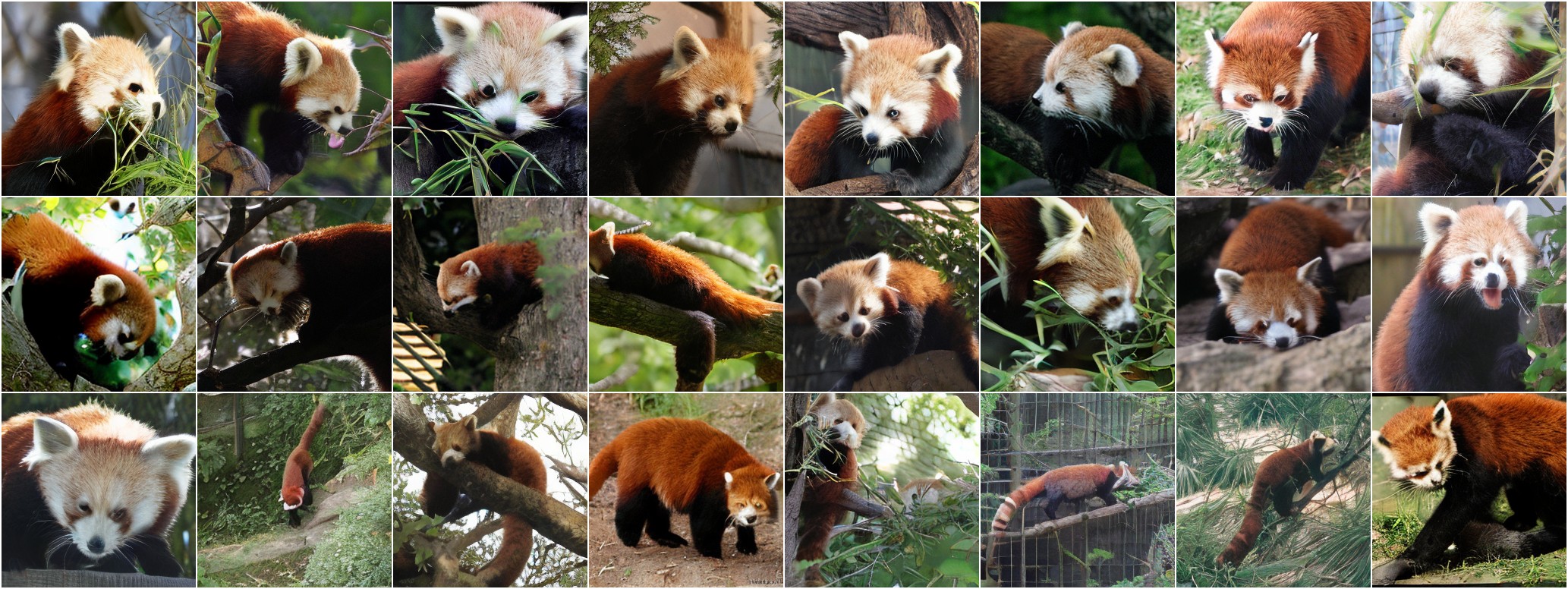}
    \end{minipage}\hfill
    \begin{minipage}[b]{0.48\textwidth}
        \includegraphics[width=\linewidth]{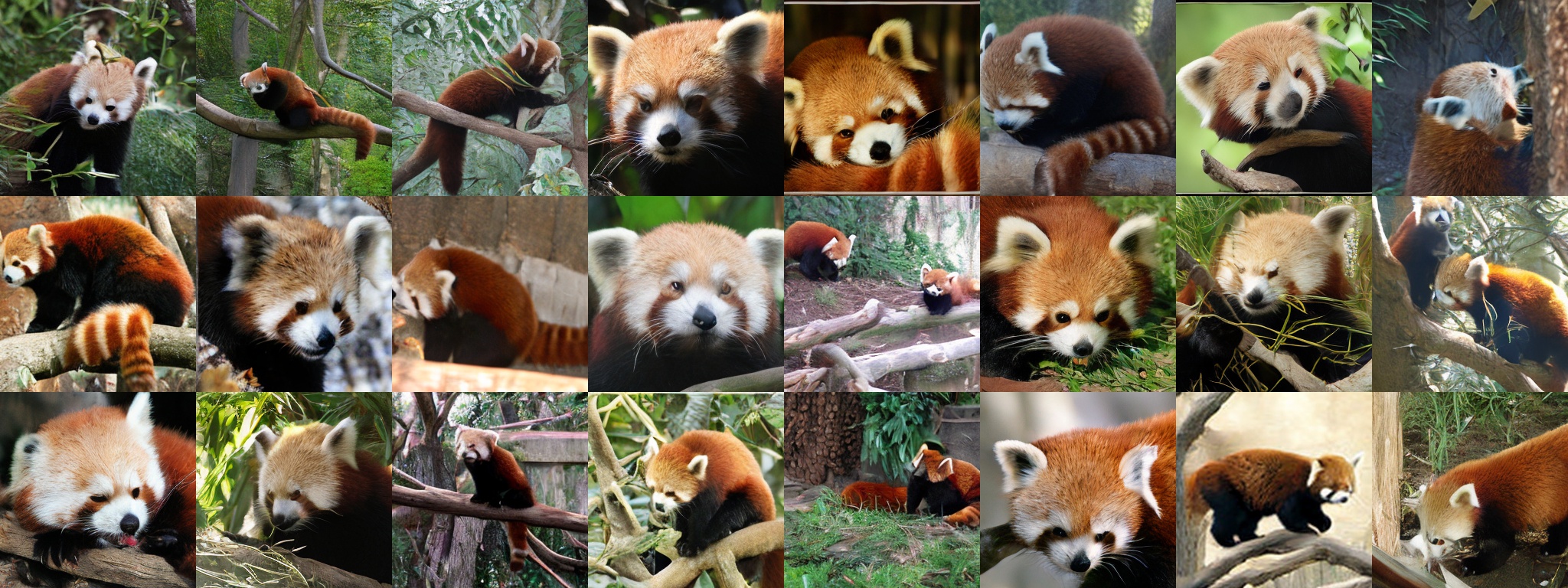}
    \end{minipage}\\[-0.2em]
    \centering\small Class 387: Lesser panda
    
    \caption{\textbf{Side-by-side comparison with improved MeanFlow (iMF)~\cite{geng2025improved} (page 2/5).} Uncurated samples from our method (left) and iMF (right) on \textbf{all} ImageNet classes visualized in the iMF paper. Both methods generate images with a single neural function evaluation (1-NFE). The iMF visualizations use CFG $\omega{=}6.0$ and interval $[t_\text{min}, t_\text{max}]{=}[0.2, 0.8]$, achieving FID 3.92 and IS 348.2 (DiT-XL/2). For fair comparison, we set the CFG scale to match the IS of iMF visualizations, which leads to FID 3.01 and IS 354.4 (at CFG=1.5) for our method (DiT-L/2).}
    \label{fig:comparison_imf_2}
\end{figure*}

\begin{figure*}[p]
    \centering
    \begin{minipage}{0.48\textwidth}
        \centering{ours}
    \end{minipage}\hfill
    \begin{minipage}{0.48\textwidth}
        \centering{improved MeanFlow (iMF)}
    \end{minipage}\\[0.2em]

    \begin{minipage}[b]{0.48\textwidth}
        \includegraphics[width=\linewidth]{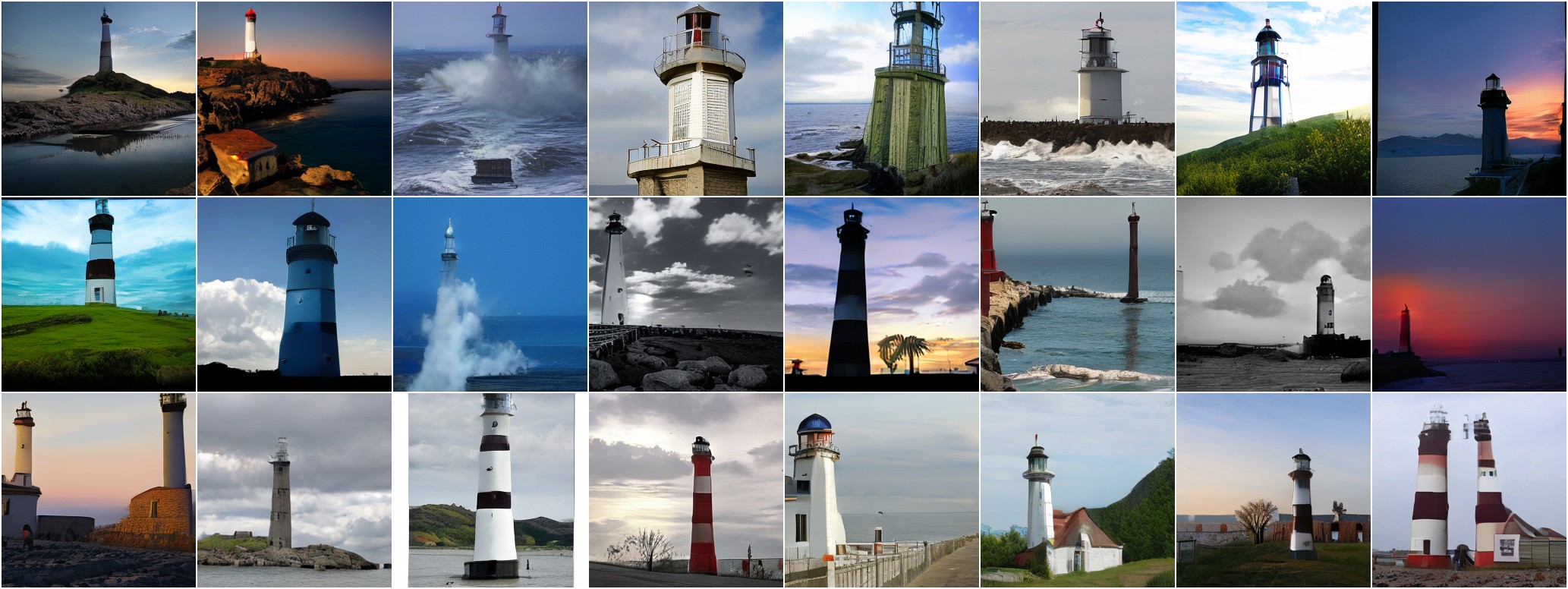}
    \end{minipage}\hfill
    \begin{minipage}[b]{0.48\textwidth}
        \includegraphics[width=\linewidth]{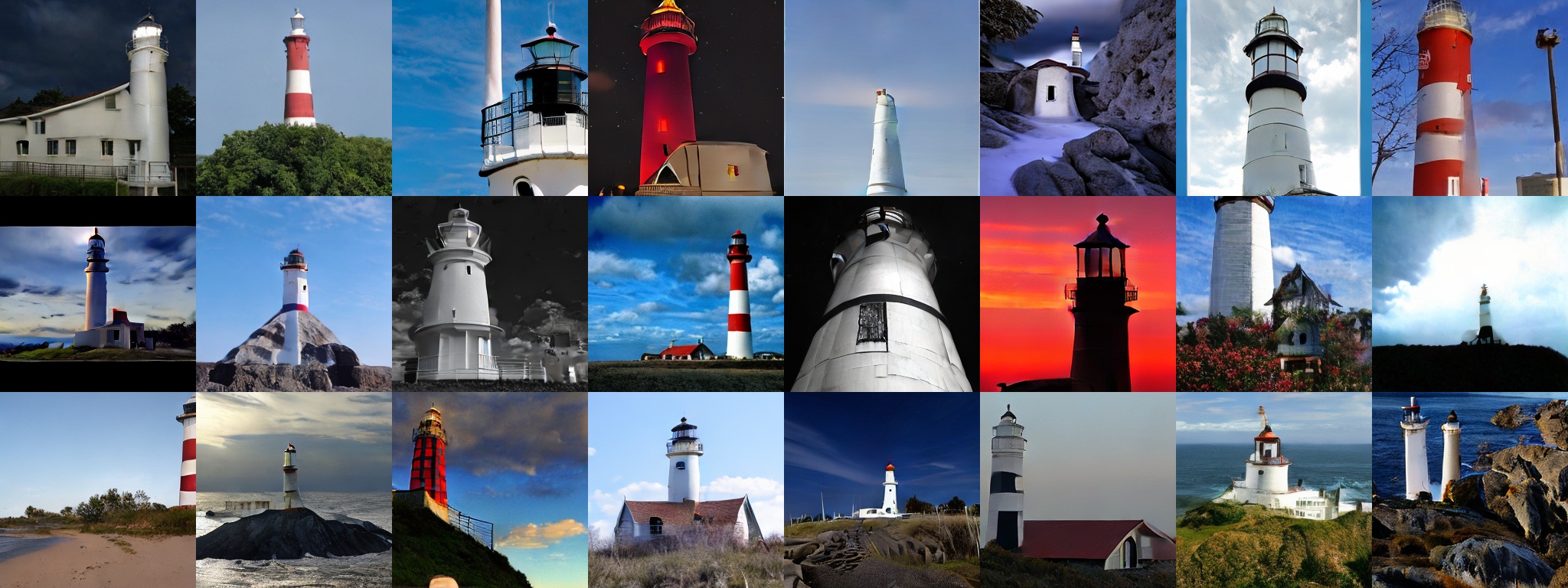}
    \end{minipage}\\[-0.2em]
    \centering\small Class 437: Beacon\\[0.2em]

    \begin{minipage}[b]{0.48\textwidth}
        \includegraphics[width=\linewidth]{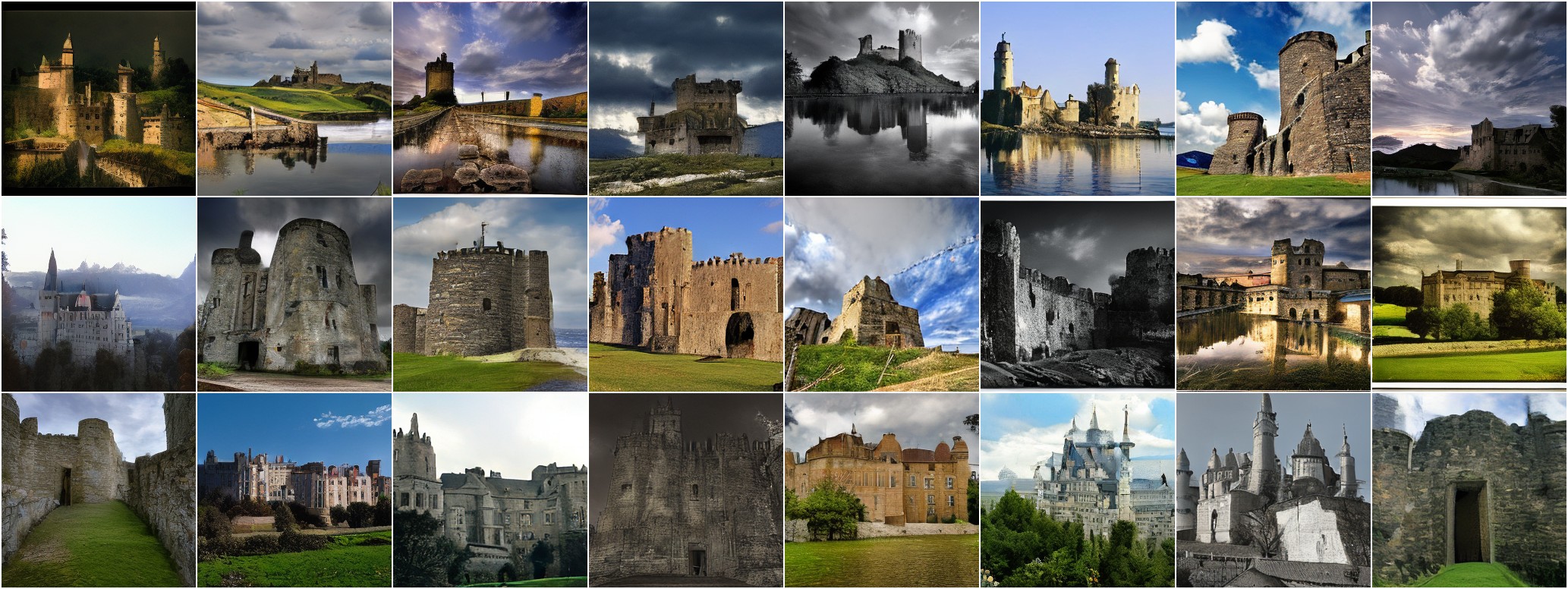}
    \end{minipage}\hfill
    \begin{minipage}[b]{0.48\textwidth}
        \includegraphics[width=\linewidth]{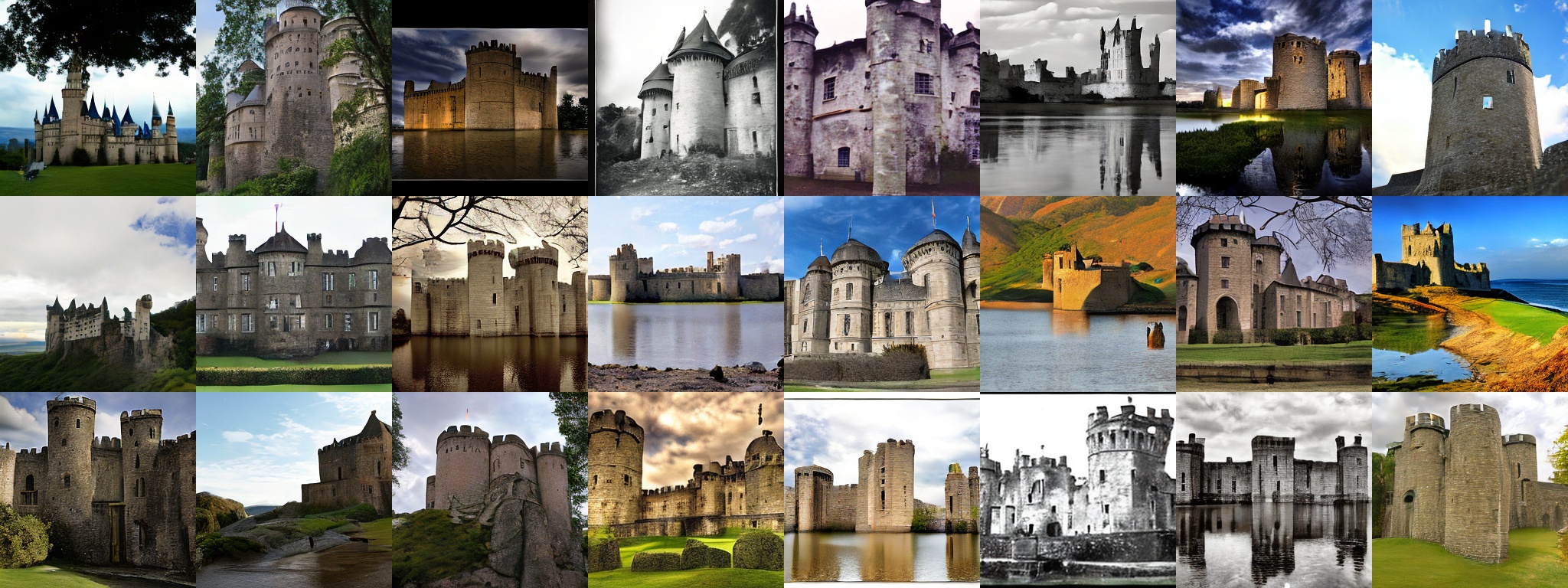}
    \end{minipage}\\[-0.2em]
    \centering\small Class 483: Castle\\[0.2em]

    \begin{minipage}[b]{0.48\textwidth}
        \includegraphics[width=\linewidth]{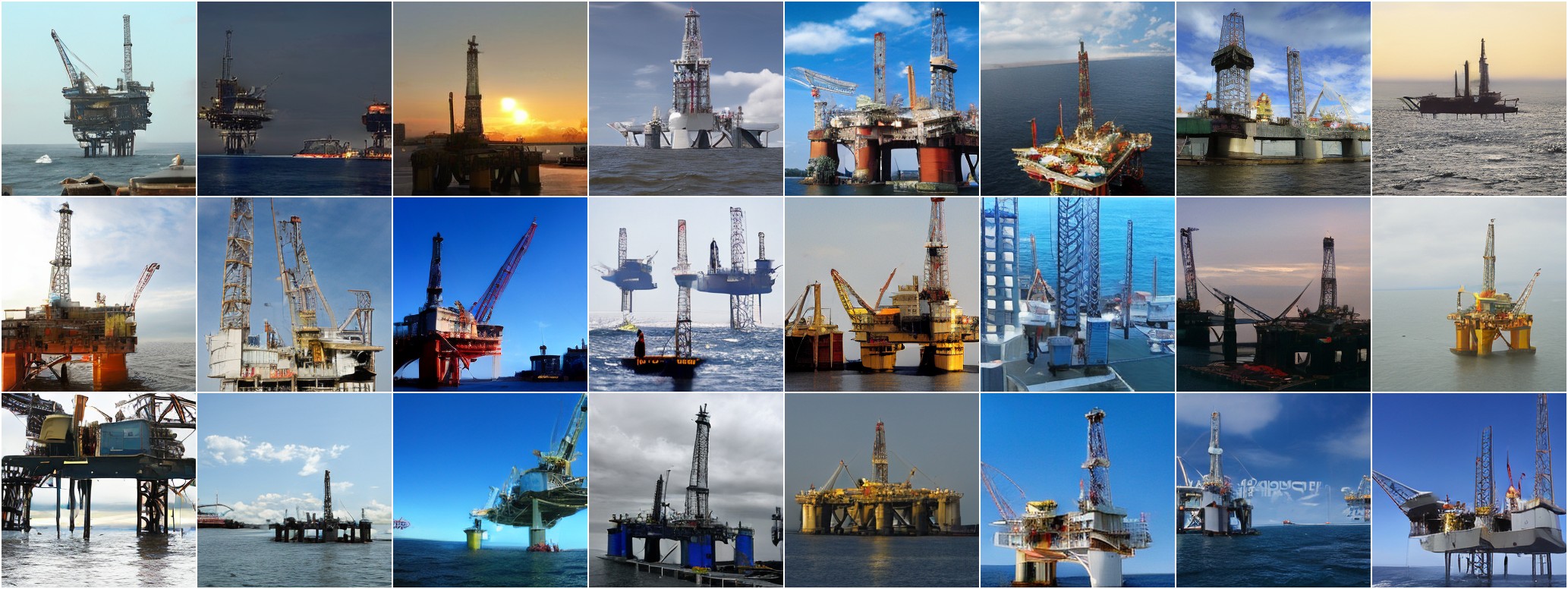}
    \end{minipage}\hfill
    \begin{minipage}[b]{0.48\textwidth}
        \includegraphics[width=\linewidth]{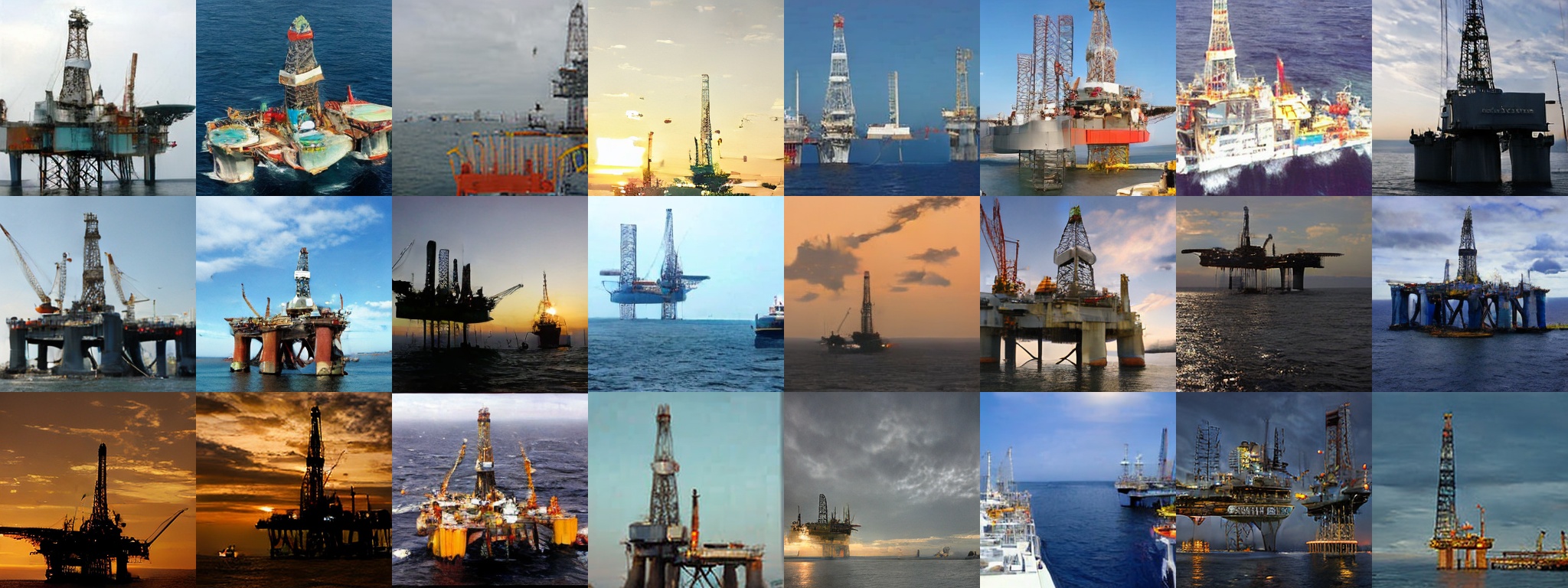}
    \end{minipage}\\[-0.2em]
    \centering\small Class 540: Drilling platform\\[0.2em]

    \begin{minipage}[b]{0.48\textwidth}
        \includegraphics[width=\linewidth]{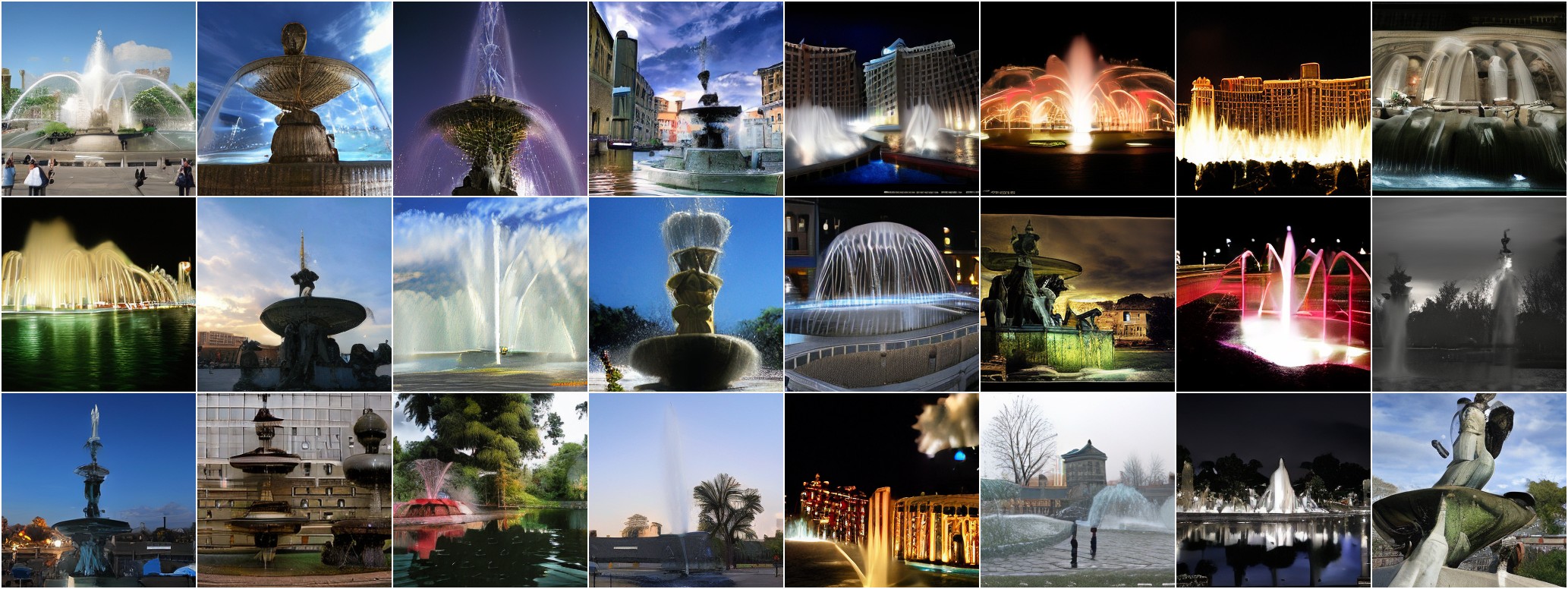}
    \end{minipage}\hfill
    \begin{minipage}[b]{0.48\textwidth}
        \includegraphics[width=\linewidth]{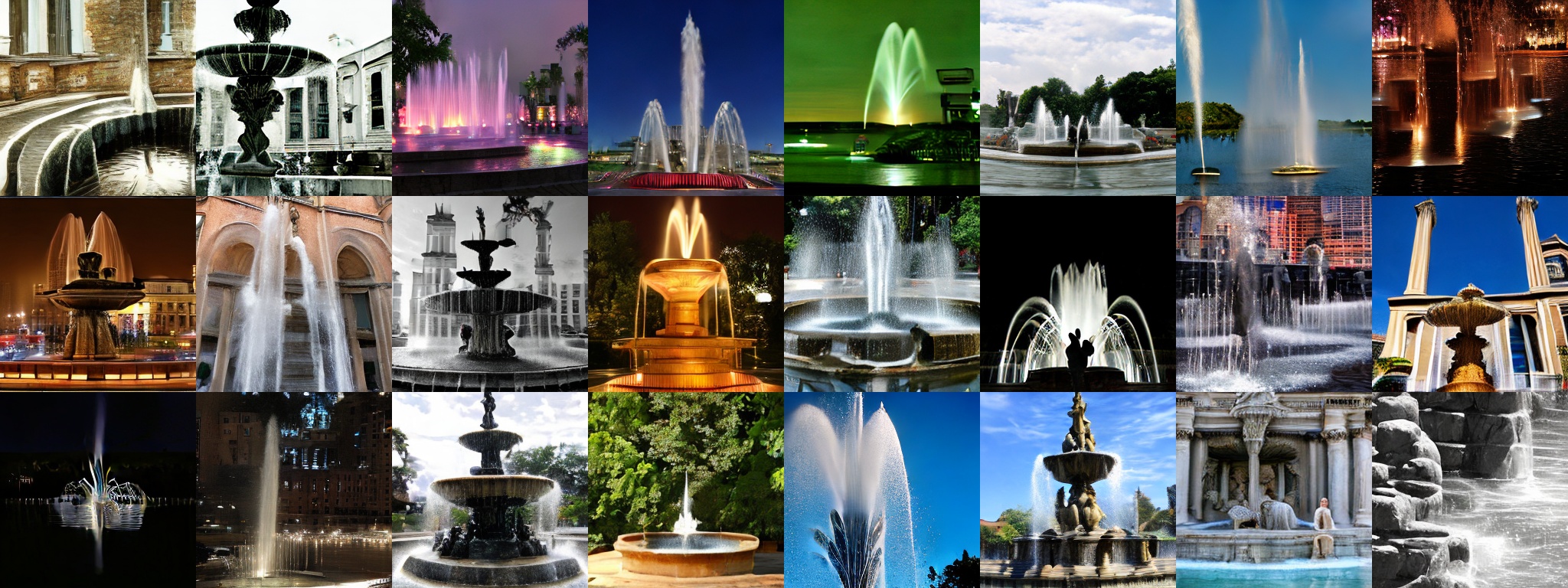}
    \end{minipage}\\[-0.2em]
    \centering\small Class 562: Fountain\\[0.2em]

    \begin{minipage}[b]{0.48\textwidth}
        \includegraphics[width=\linewidth]{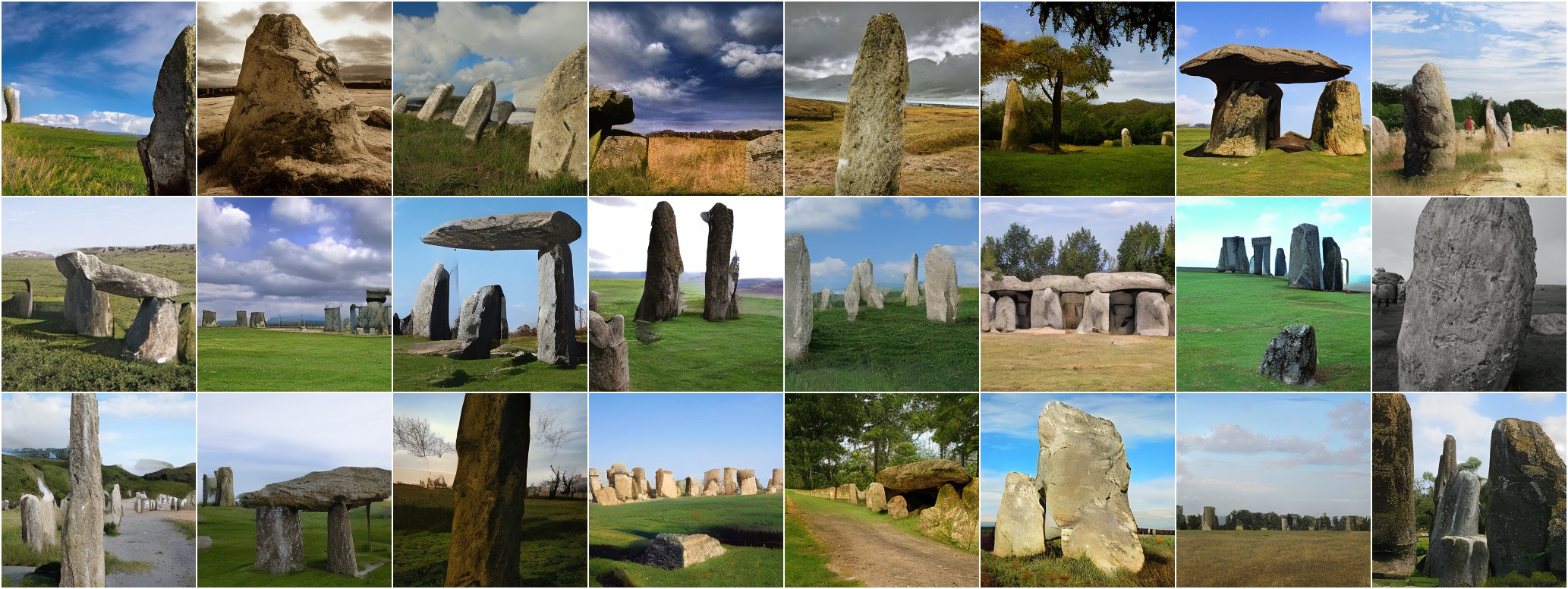}
    \end{minipage}\hfill
    \begin{minipage}[b]{0.48\textwidth}
        \includegraphics[width=\linewidth]{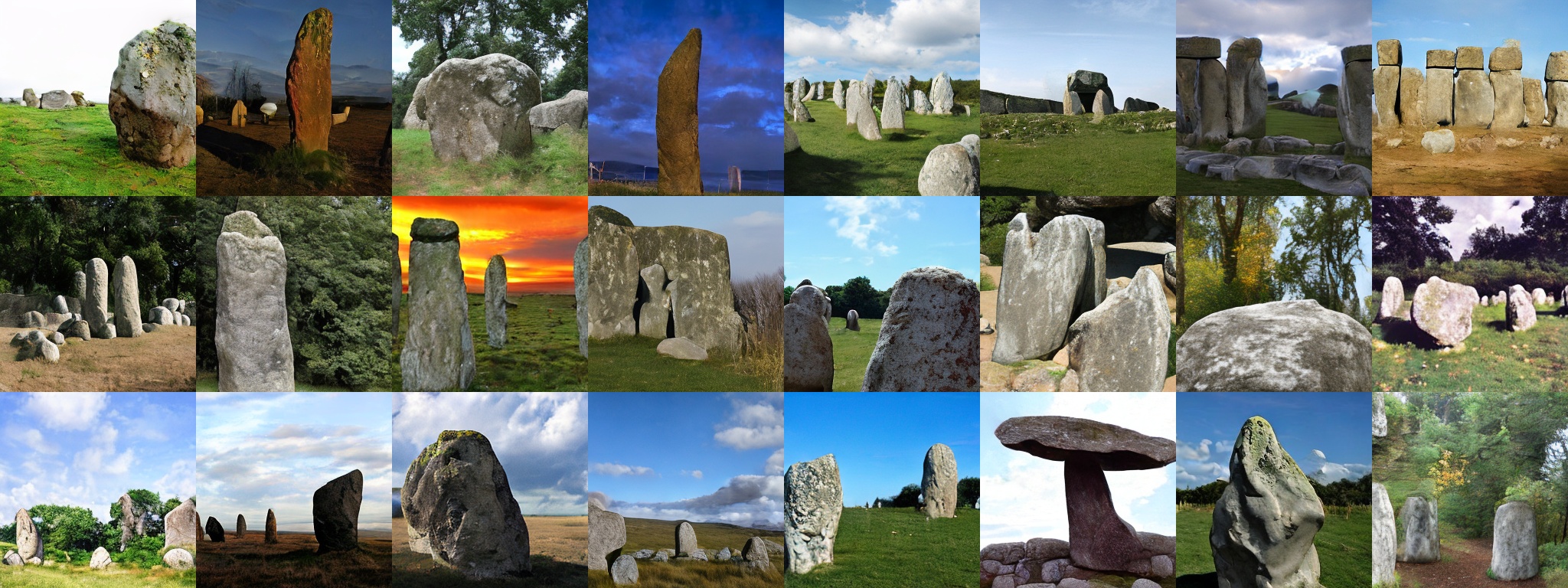}
    \end{minipage}\\[-0.2em]
    \centering\small Class 649: Megalith
    
    \caption{\textbf{Side-by-side comparison with improved MeanFlow (iMF)~\cite{geng2025improved} (page 3/5).} Uncurated samples from our method (left) and iMF (right) on \textbf{all} ImageNet classes visualized in the iMF paper. Both methods generate images with a single neural function evaluation (1-NFE). The iMF visualizations use CFG $\omega{=}6.0$ and interval $[t_\text{min}, t_\text{max}]{=}[0.2, 0.8]$, achieving FID 3.92 and IS 348.2 (DiT-XL/2). For fair comparison, we set the CFG scale to match the IS of iMF visualizations, which leads to FID 3.01 and IS 354.4 (at CFG=1.5) for our method (DiT-L/2).}
    \label{fig:comparison_imf_3}
\end{figure*}

\begin{figure*}[p]
    \centering
    \begin{minipage}{0.48\textwidth}
        \centering{ours}
    \end{minipage}\hfill
    \begin{minipage}{0.48\textwidth}
        \centering{improved MeanFlow (iMF)}
    \end{minipage}\\[0.2em]

    \begin{minipage}[b]{0.48\textwidth}
        \includegraphics[width=\linewidth]{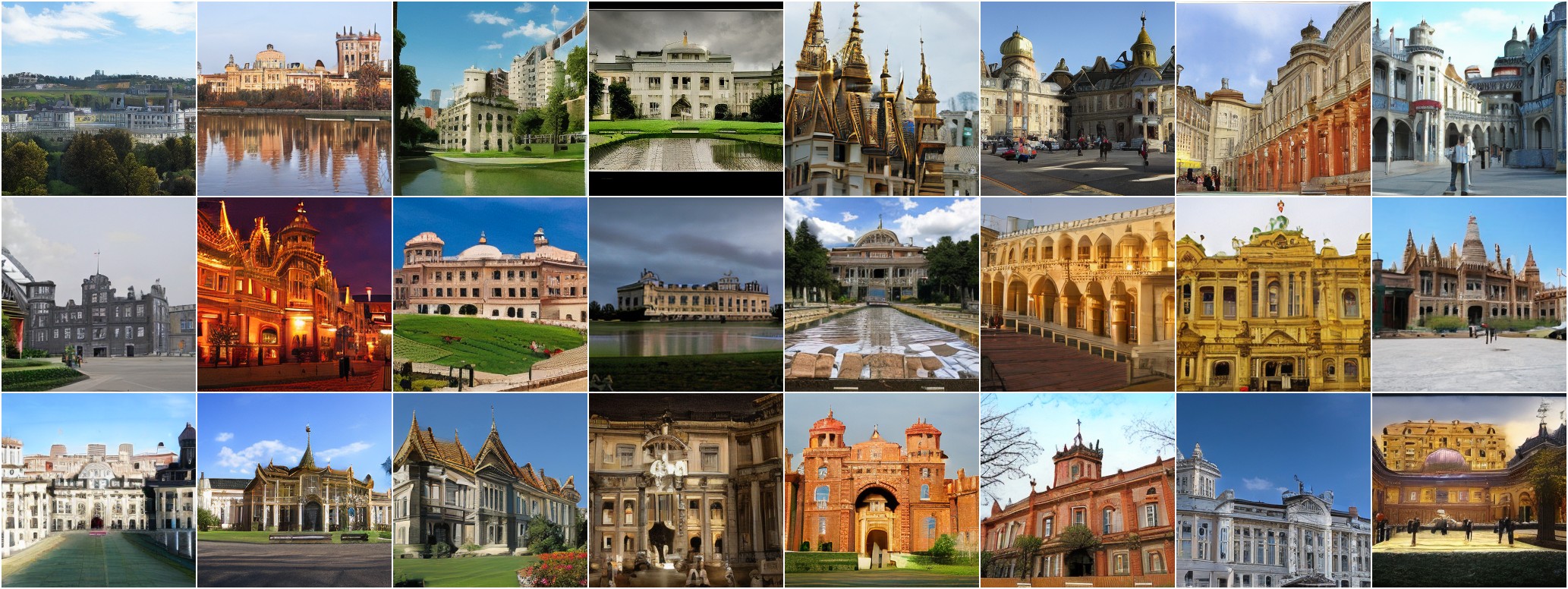}
    \end{minipage}\hfill
    \begin{minipage}[b]{0.48\textwidth}
        \includegraphics[width=\linewidth]{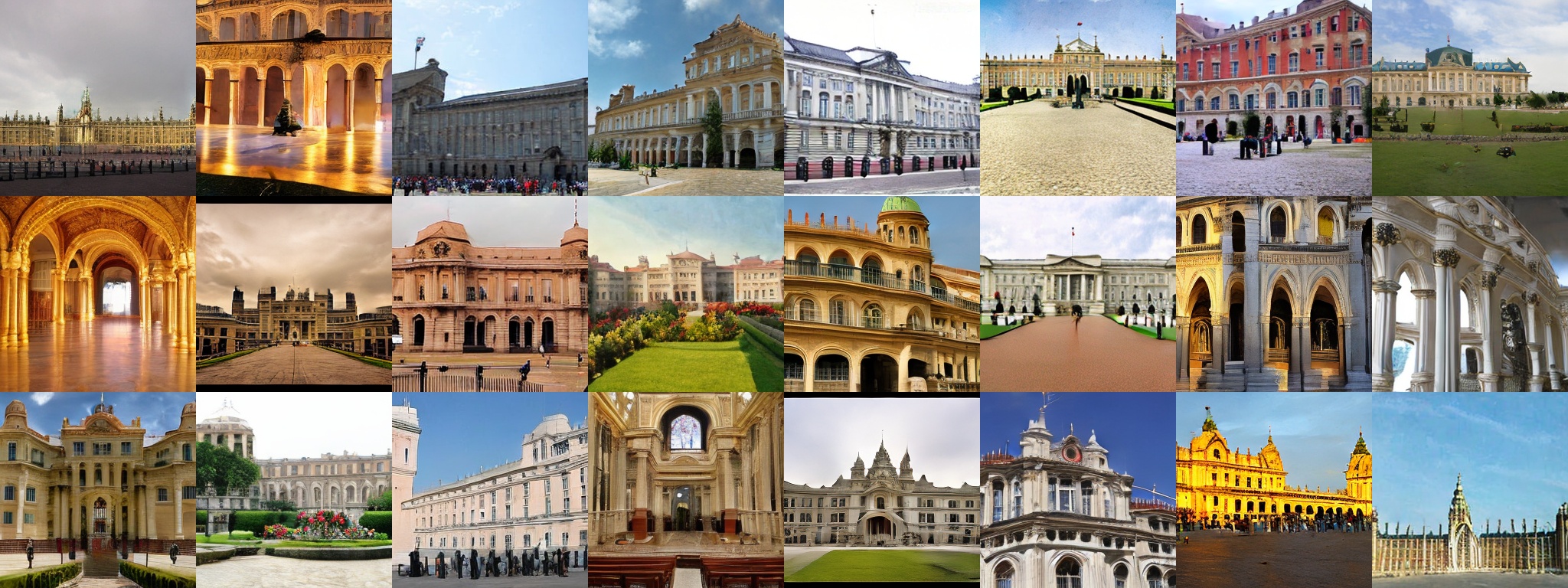}
    \end{minipage}\\[-0.2em]
    \centering\small Class 698: Palace\\[0.2em]

    \begin{minipage}[b]{0.48\textwidth}
        \includegraphics[width=\linewidth]{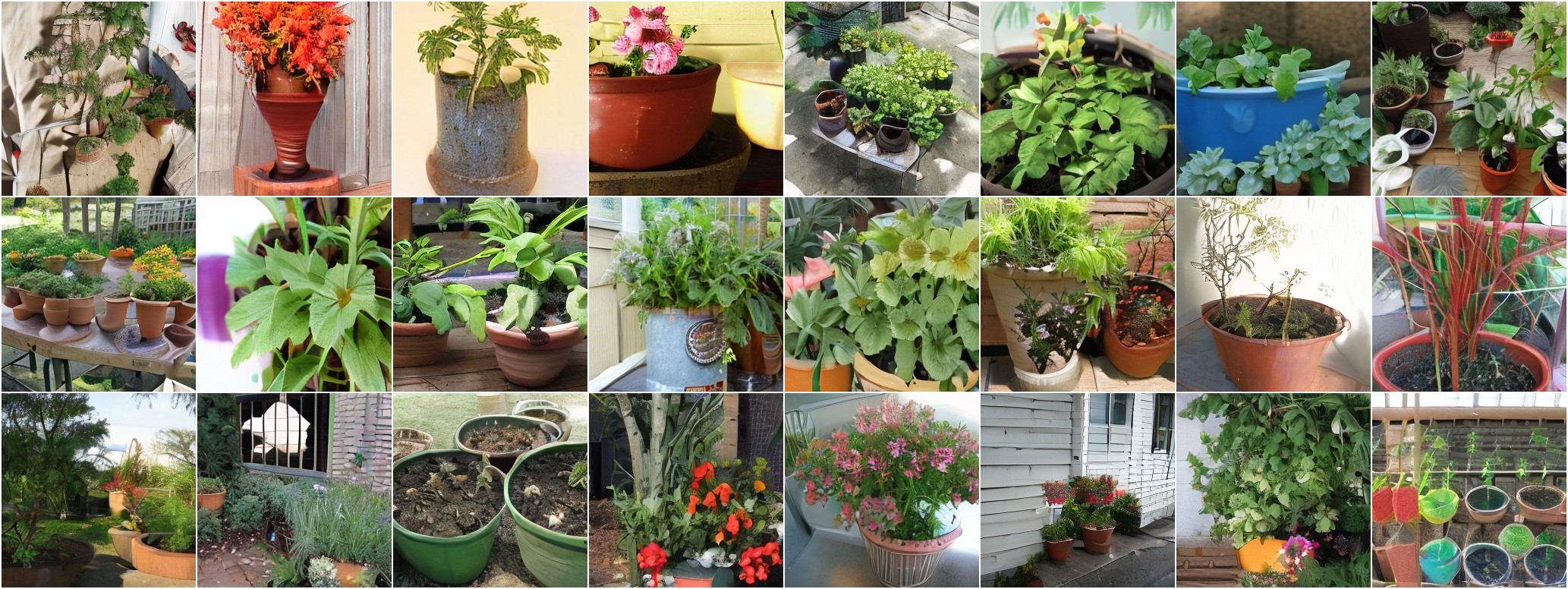}
    \end{minipage}\hfill
    \begin{minipage}[b]{0.48\textwidth}
        \includegraphics[width=\linewidth]{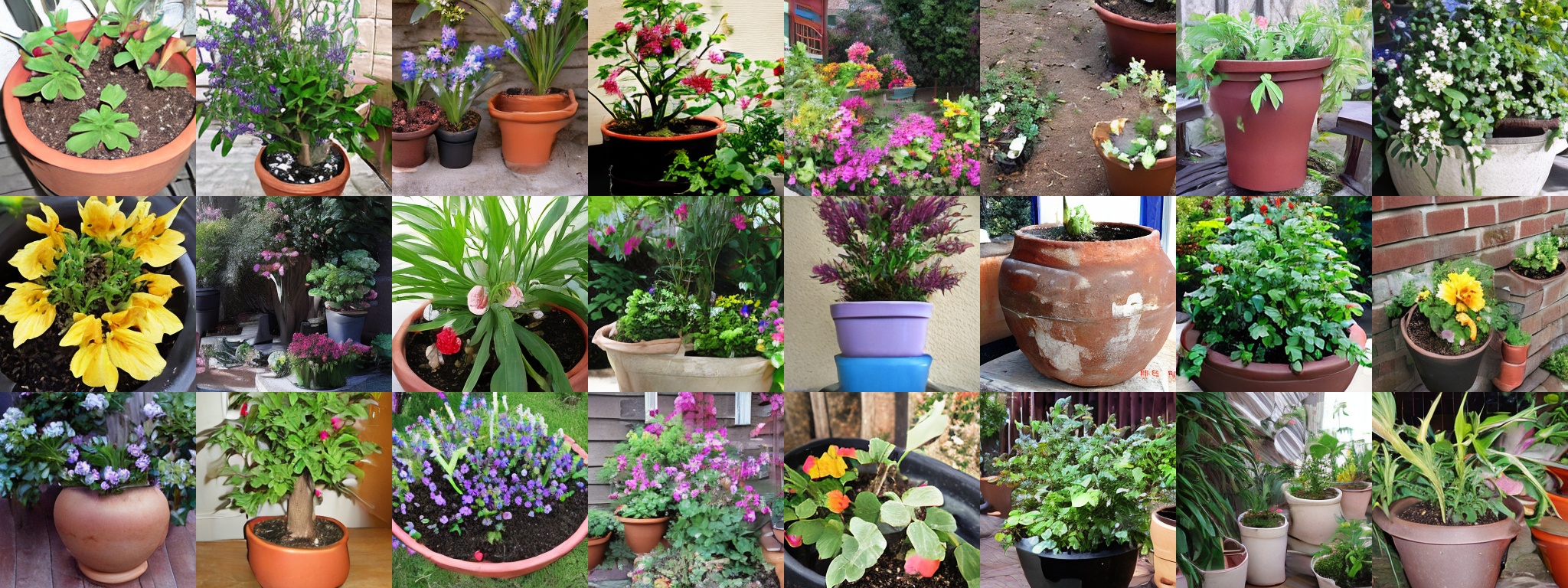}
    \end{minipage}\\[-0.2em]
    \centering\small Class 738: Pot\\[0.2em]

    \begin{minipage}[b]{0.48\textwidth}
        \includegraphics[width=\linewidth]{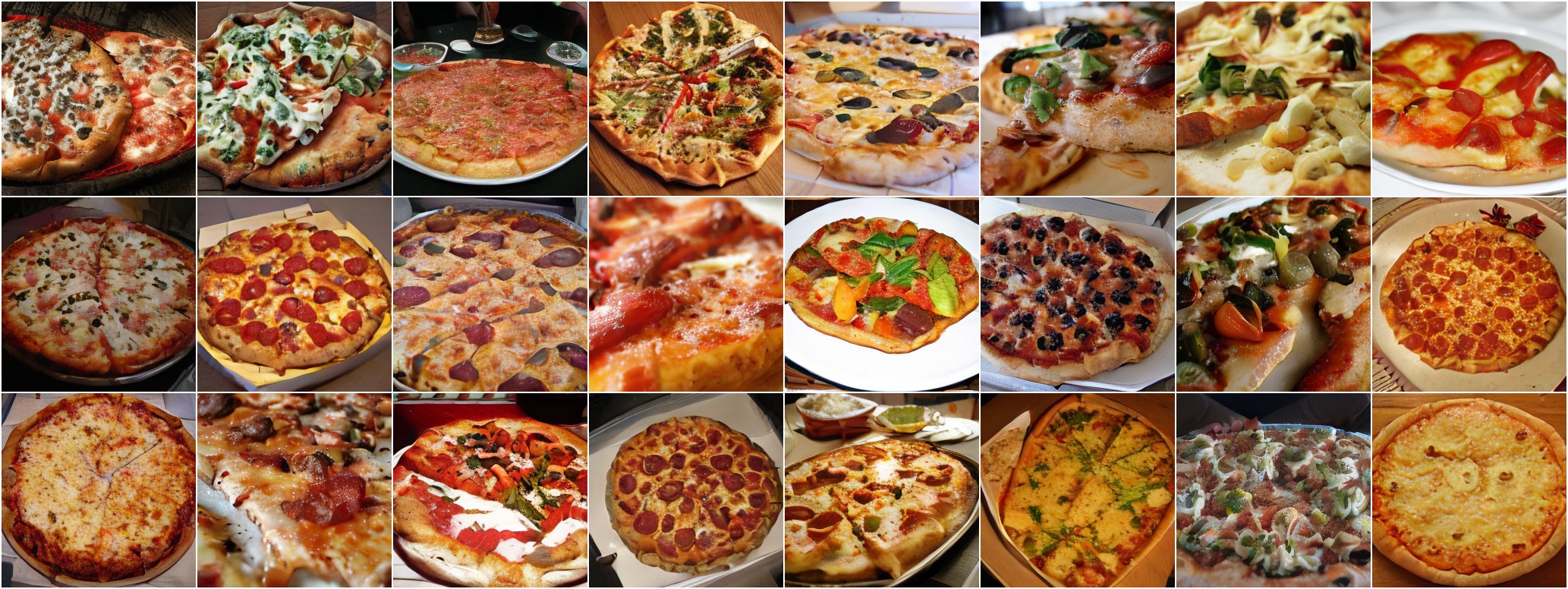}
    \end{minipage}\hfill
    \begin{minipage}[b]{0.48\textwidth}
        \includegraphics[width=\linewidth]{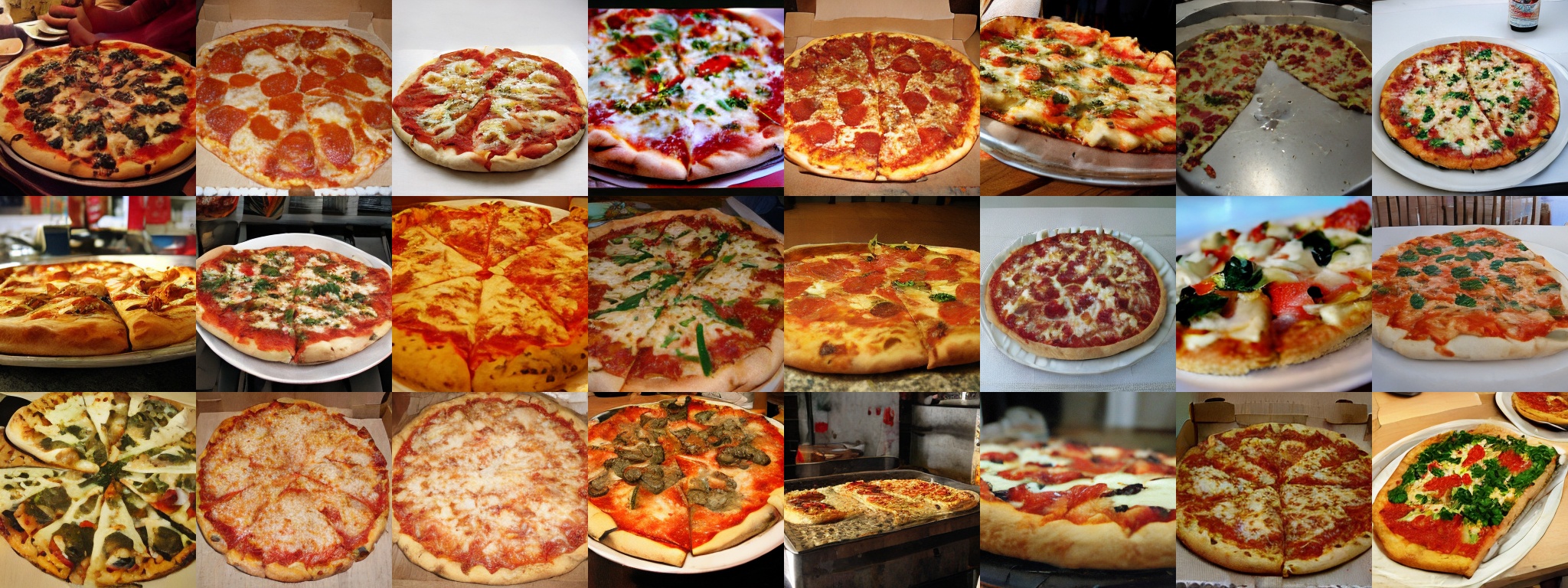}
    \end{minipage}\\[-0.2em]
    \centering\small Class 963: Pizza\\[0.2em]

    \begin{minipage}[b]{0.48\textwidth}
        \includegraphics[width=\linewidth]{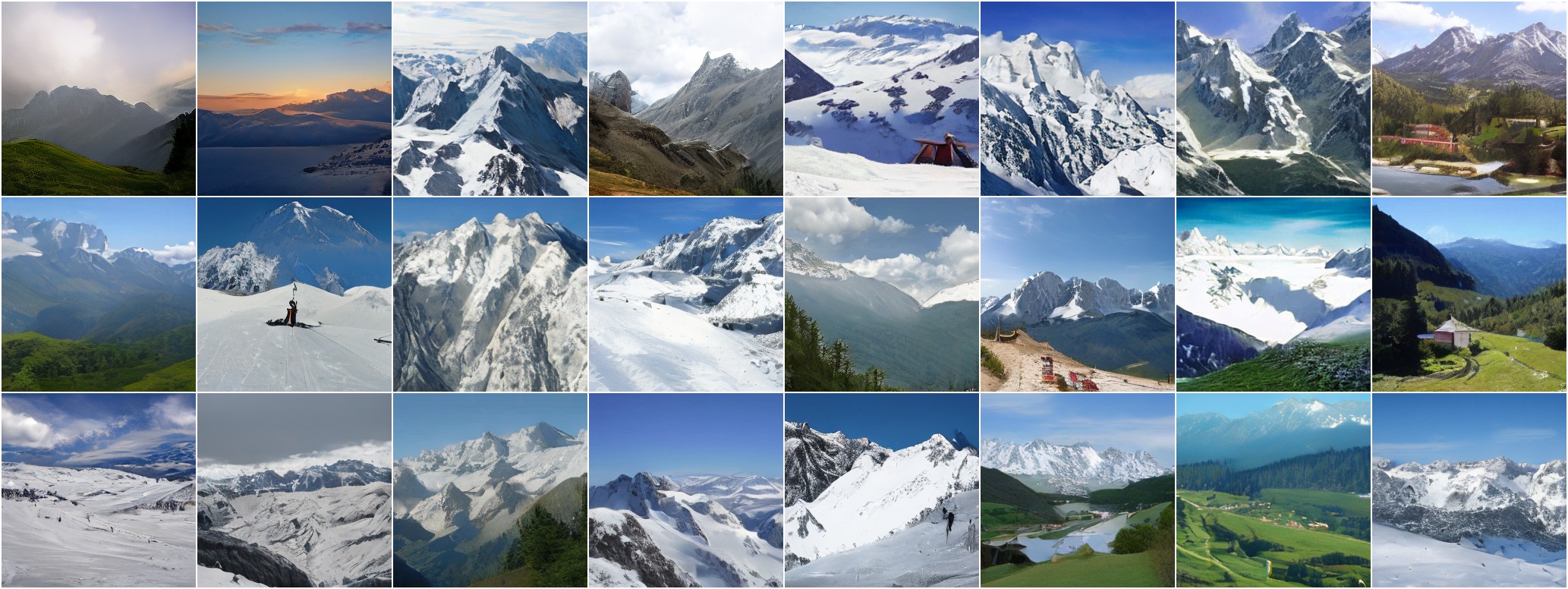}
    \end{minipage}\hfill
    \begin{minipage}[b]{0.48\textwidth}
        \includegraphics[width=\linewidth]{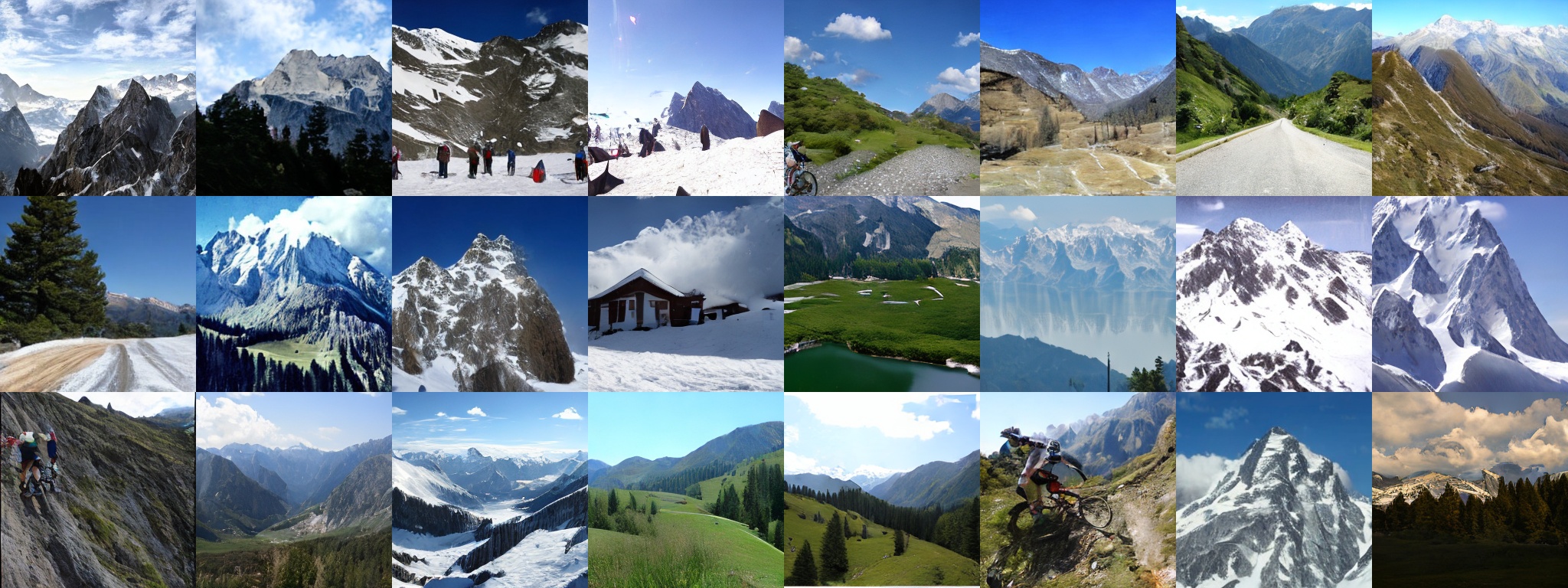}
    \end{minipage}\\[-0.2em]
    \centering\small Class 970: Alp\\[0.2em]

    \begin{minipage}[b]{0.48\textwidth}
        \includegraphics[width=\linewidth]{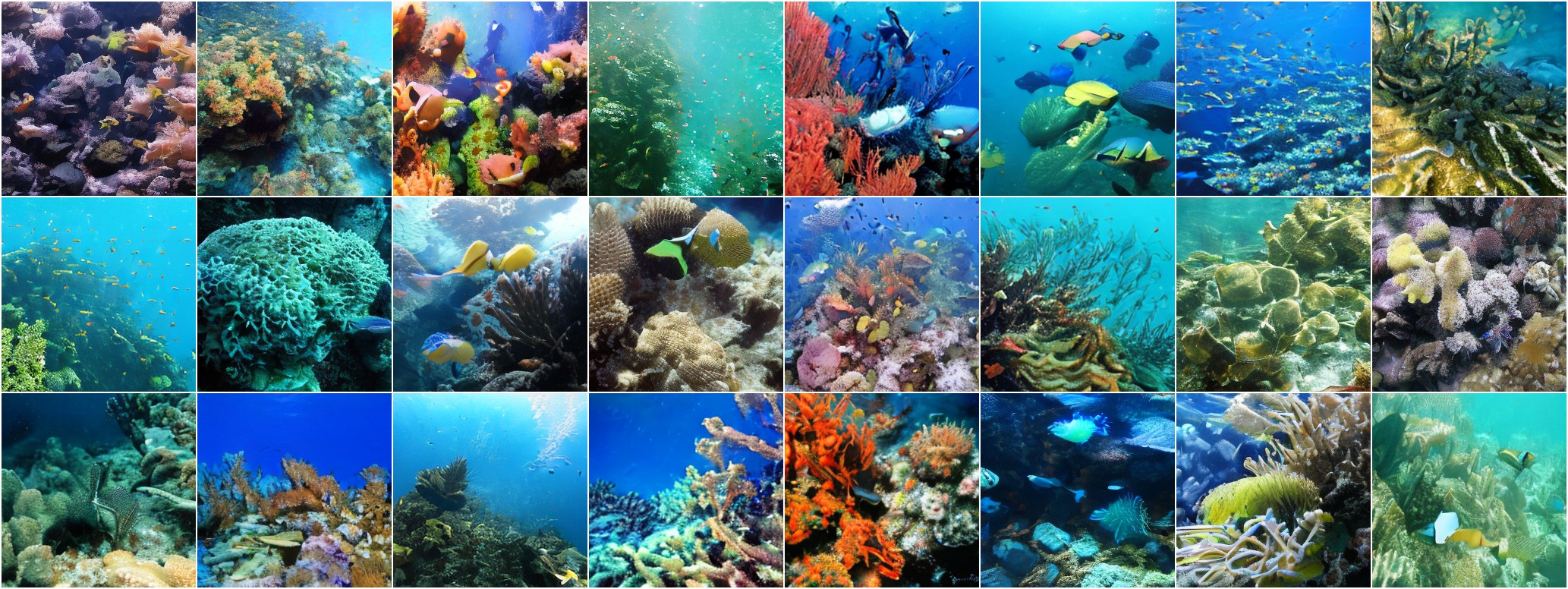}
    \end{minipage}\hfill
    \begin{minipage}[b]{0.48\textwidth}
        \includegraphics[width=\linewidth]{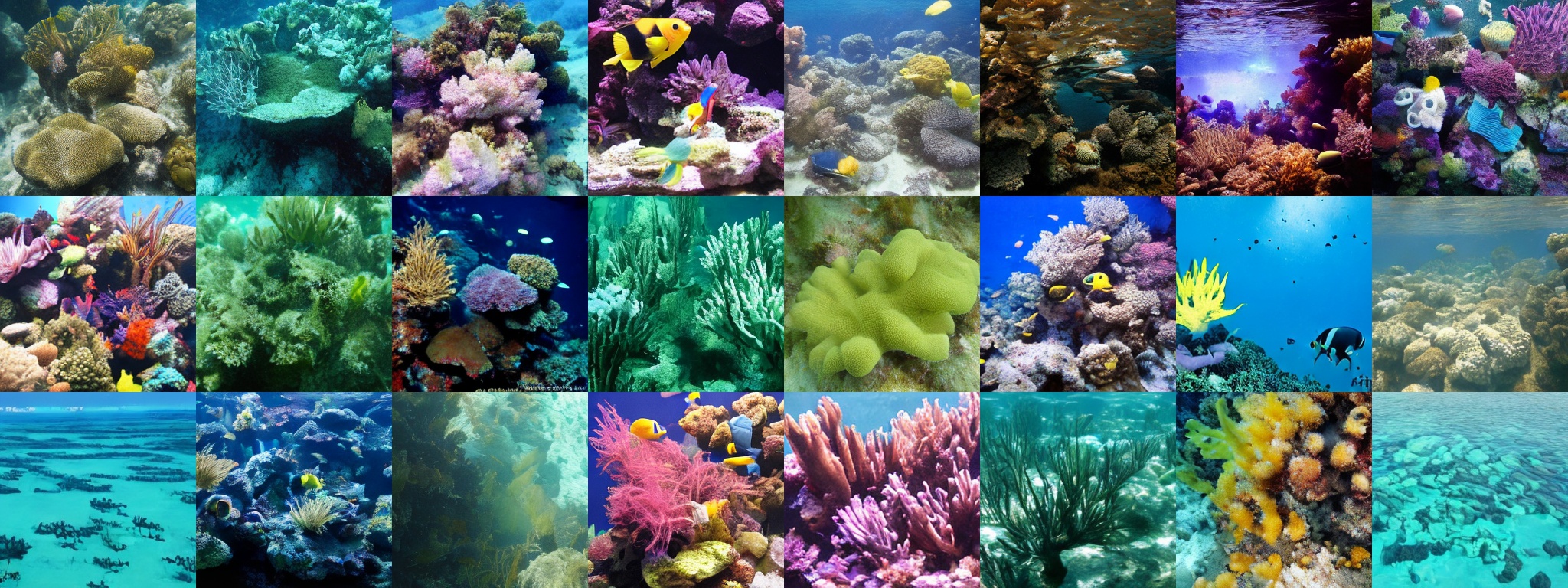}
    \end{minipage}\\[-0.2em]
    \centering\small Class 973: Coral reef
    
    \caption{\textbf{Side-by-side comparison with improved MeanFlow (iMF)~\cite{geng2025improved} (page 4/5).} Uncurated samples from our method (left) and iMF (right) on \textbf{all} ImageNet classes visualized in the iMF paper. Both methods generate images with a single neural function evaluation (1-NFE). The iMF visualizations use CFG $\omega{=}6.0$ and interval $[t_\text{min}, t_\text{max}]{=}[0.2, 0.8]$, achieving FID 3.92 and IS 348.2 (DiT-XL/2). For fair comparison, we set the CFG scale to match the IS of iMF visualizations, which leads to FID 3.01 and IS 354.4 (at CFG=1.5) for our method (DiT-L/2).}
    \label{fig:comparison_imf_4}
\end{figure*}

\begin{figure*}[p]
    \centering
    \begin{minipage}{0.48\textwidth}
        \centering{ours}
    \end{minipage}\hfill
    \begin{minipage}{0.48\textwidth}
        \centering{improved MeanFlow (iMF)}
    \end{minipage}\\[0.2em]

    \begin{minipage}[b]{0.48\textwidth}
        \includegraphics[width=\linewidth]{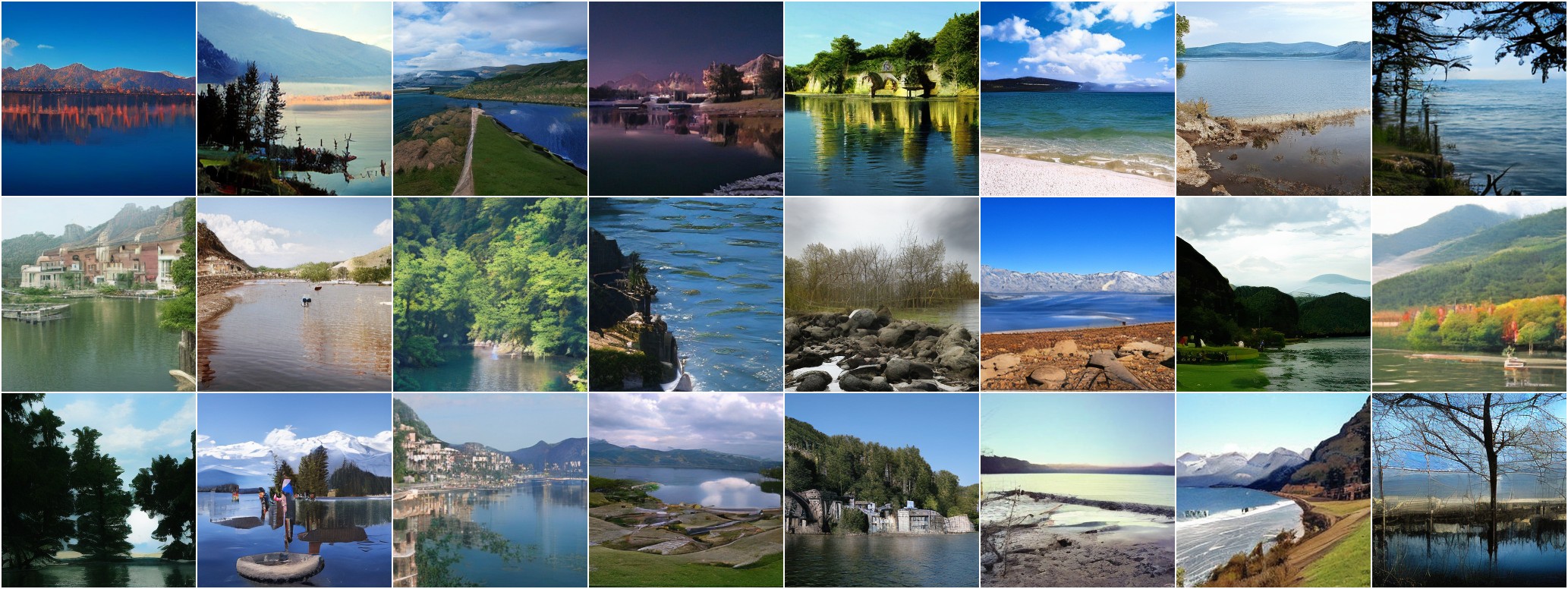}
    \end{minipage}\hfill
    \begin{minipage}[b]{0.48\textwidth}
        \includegraphics[width=\linewidth]{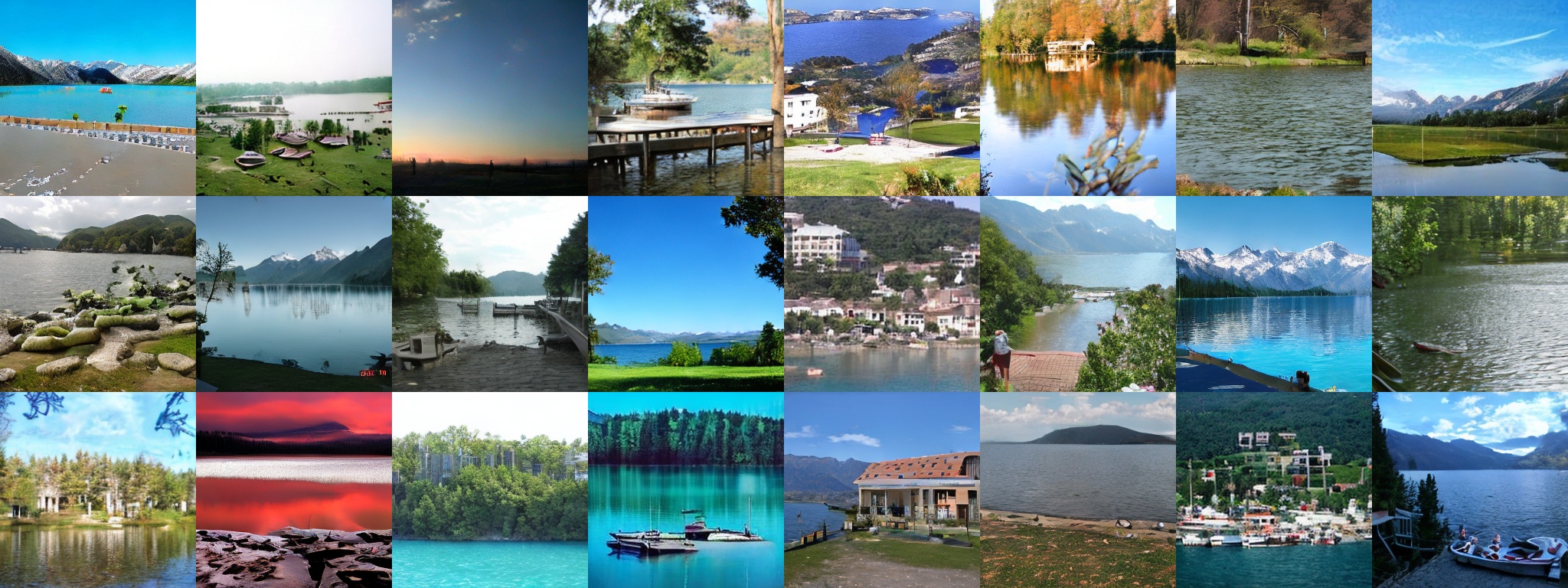}
    \end{minipage}\\[-0.2em]
    \centering\small Class 975: Lakeside\\[0.2em]

    \begin{minipage}[b]{0.48\textwidth}
        \includegraphics[width=\linewidth]{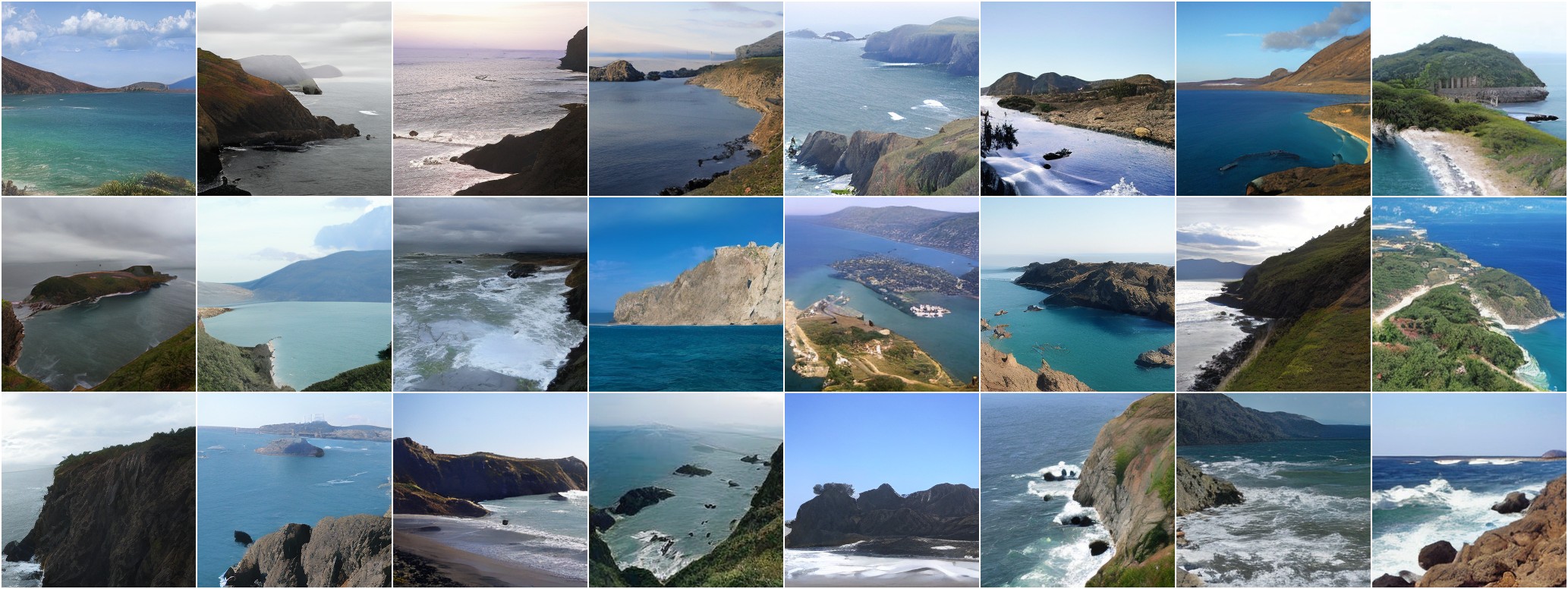}
    \end{minipage}\hfill
    \begin{minipage}[b]{0.48\textwidth}
        \includegraphics[width=\linewidth]{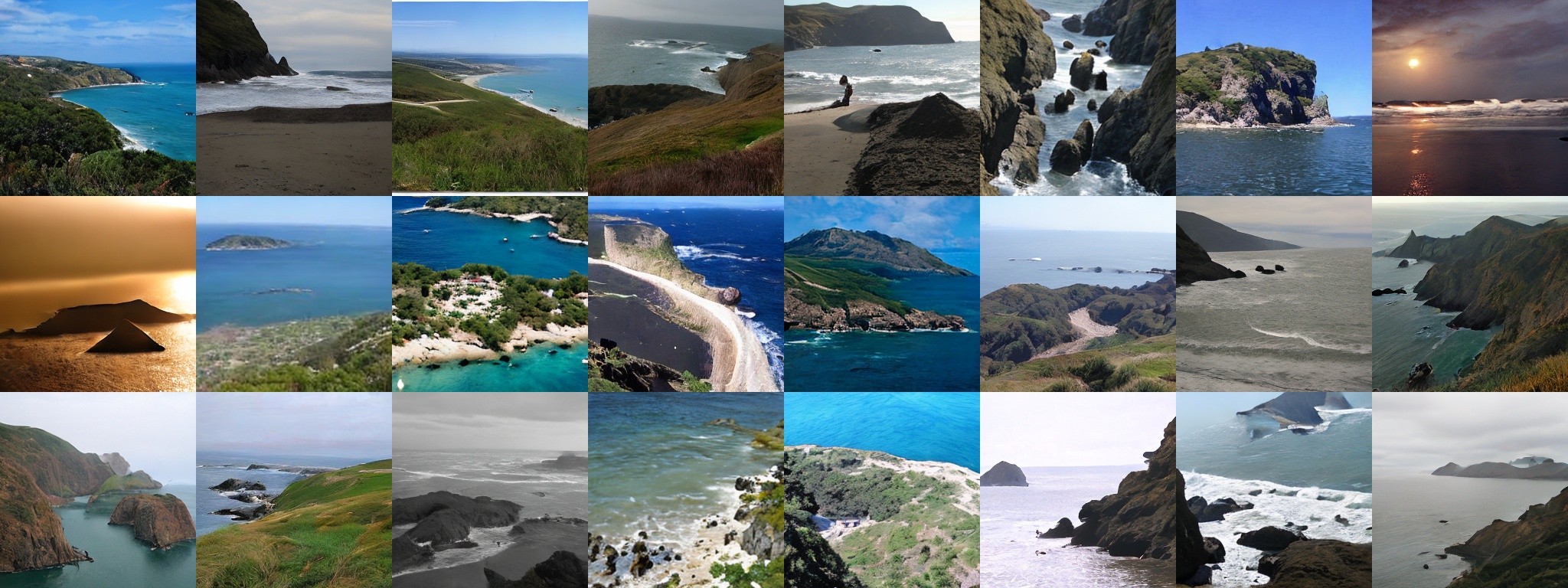}
    \end{minipage}\\[-0.2em]
    \centering\small Class 976: Promontory\\[0.2em]

    \begin{minipage}[b]{0.48\textwidth}
        \includegraphics[width=\linewidth]{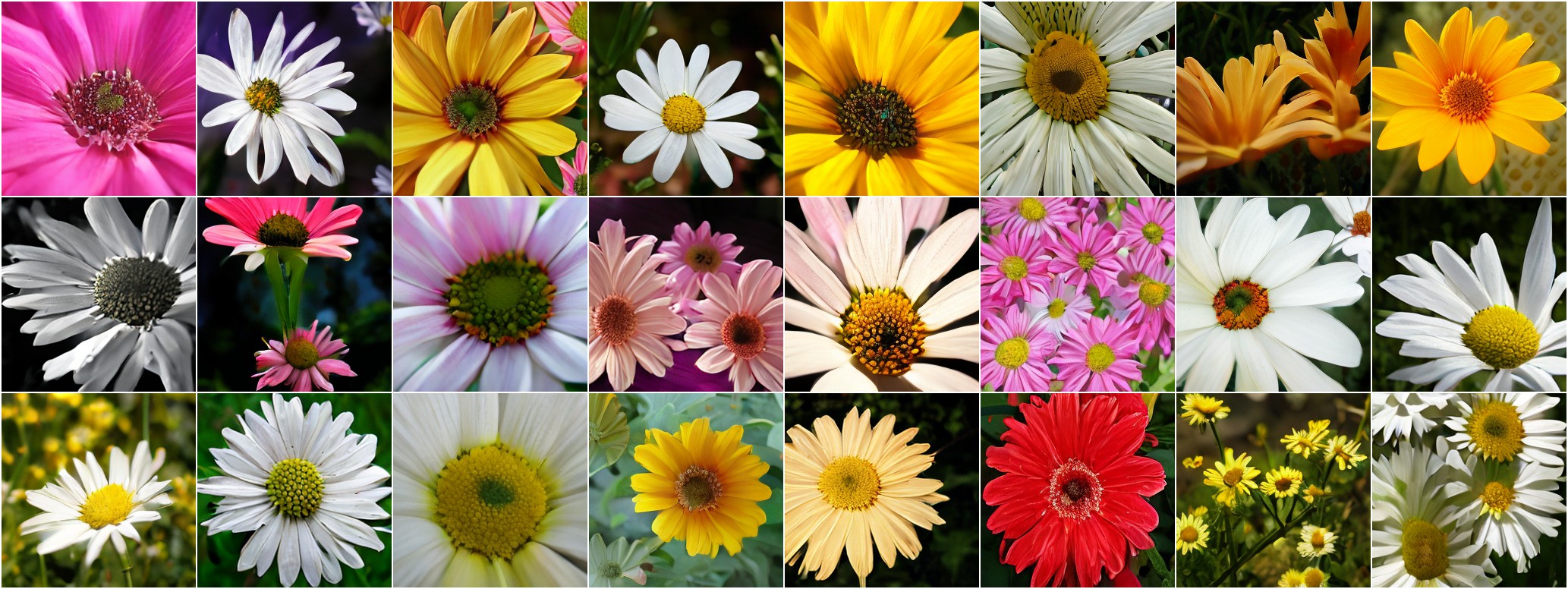}
    \end{minipage}\hfill
    \begin{minipage}[b]{0.48\textwidth}
        \includegraphics[width=\linewidth]{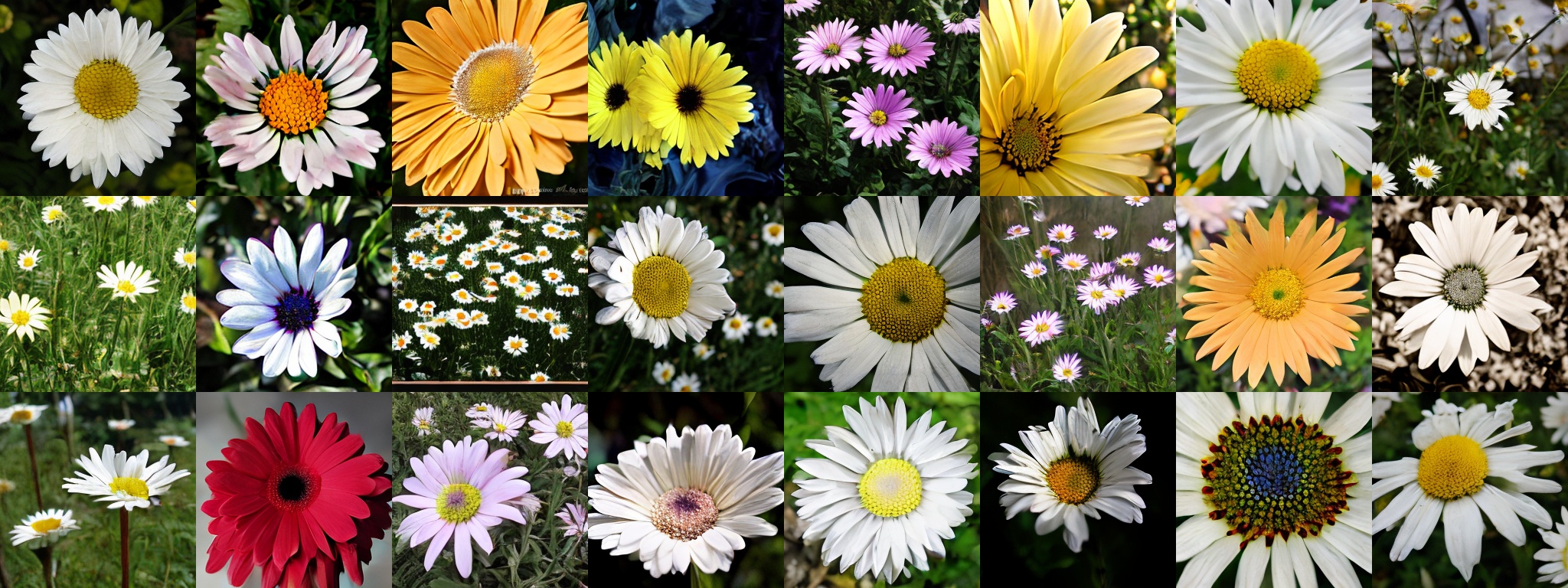}
    \end{minipage}\\[-0.2em]
    \centering\small Class 985: Daisy
    
    \caption{\textbf{Side-by-side comparison with improved MeanFlow (iMF)~\cite{geng2025improved} (page 5/5).} Uncurated samples from our method (left) and iMF (right) on \textbf{all} ImageNet classes visualized in the iMF paper. Both methods generate images with a single neural function evaluation (1-NFE). The iMF visualizations use CFG $\omega{=}6.0$ and interval $[t_\text{min}, t_\text{max}]{=}[0.2, 0.8]$, achieving FID 3.92 and IS 348.2 (DiT-XL/2). For fair comparison, we set the CFG scale to match the IS of iMF visualizations, which leads to FID 3.01 and IS 354.4 (at CFG=1.5) for our method (DiT-L/2).}
    \label{fig:comparison_imf_5}
\end{figure*}

Fig.~\ref{fig:latent_uncurated_samples_1}-\ref{fig:latent_uncurated_samples_4} show uncurated samples from our model.
Fig.~\ref{fig:comparison_imf_1}-\ref{fig:comparison_imf_5} provide side-by-side comparison with improved MeanFlow (iMF) \cite{geng2025improved}, the current state-of-the-art one-step method.

\section{Supplementary Connections to Prior Work}
\label{app:related_connections}

\subsection{Drifting Fields of Existing Generative Models}
\label{sec:theory_implicit_drifting}

For one-step generative methods that minimize a loss between
two distributions $p$ and $q$, our framework implies the existence of an underlying drifting field $\V$. 
We may define $\V{\propto}{-}\frac{\partial\mathcal L }{\partial {\x}}$, which vanishes at the minimum of $\mathcal L$, attained when $p = q$.
Many one-step generative models can be interpreted as inducing per-sample drifting fields:

\begin{itemize}[itemsep=.2em, topsep=1pt, parsep=1pt, leftmargin=1.2em]
    \item \textbf{GANs}~\cite{goodfellow2014generative}, with the optimal discriminator $D^*=p_\text{data}/(p_\text{data}+p_\theta)$, induce
    $\V(\x)=\nabla_{\x}\log p_\text{data}(\x)-\nabla_{\x}\log\bigl(p_\text{data}(\x)+p_\theta(\x)\bigr)$, which vanishes when $p_\theta=p_\text{data}$.
    \item \textbf{Distribution Matching Distillation} (DMD~\cite{yin2024one}), \textbf{Diff-Instruct}~\cite{luo2023diff}, and \textbf{SiD}~\cite{zhou2024score} induce a score-difference field
    $\V(\x)=\mathbb{E}_{t,\x_t}\bigl[v_\text{real}(\x_t,t)-v_\text{gen}(\x_t,t)\bigr]$ on noisy samples $\x_t$.
    \item \textbf{MMD-based moment matching}~\cite{li2015generative,dziugaite2015training} induces a kernel-gradient field $\V_\text{MMD}(\x)=\mathbb{E}_{\y^+\sim p}[\nabla_{\x}\xi(\x,\y^+)]-\mathbb{E}_{\y^-\sim q}[\nabla_{\x}\xi(\x,\y^-)]$; we provide a detailed derivation in Appendix~\ref{sec:theory_mmd_relation}.
    \item More generally, minimizing a divergence $\min_q L(p,q)$ induces a field of the form $\V(\x)=\nabla_{\x}\frac{\delta L}{\delta q}(\x)$, which is zero at the optimum under suitable regularity conditions.
\end{itemize}

Under this view, our contribution is to elevate $\V$ as an explicit designable object, which enables normalizations, aggregation across feature spaces, and modified equilibria. We also show that a non-parametric instantiation works at ImageNet scale without adversarial training or pretrained diffusion teachers.

\paragraph{Relation to DMD and Score-Difference Fields.}
Our method has a surface-level resemblance to DMD: in a special case, both can be written as score-differences. Under a Gaussian-kernel mean-shift instantiation, our field reduces to
\[
\V_{p,q}(\x)=\sigma^2\bigl[\nabla_{\x}\log\hat p_\sigma(\x)-\nabla_{\x}\log\hat q_\sigma(\x)\bigr],
\]
a smoothed score-difference $\nabla\log\hat p_\sigma-\nabla\log\hat q_\sigma$ on clean samples $\x$, while the DMD update is a score-difference on noisy samples $\x_t$ amortized over the diffusion time $t$. The two thus share the score-difference form but are not the same object.

This score-difference interpretation does not extend beyond the Gaussian-kernel mean-shift case. Our practical ImageNet model instead uses a Laplacian kernel, which does not reduce to a score-difference form and which slightly outperforms the Gaussian variant in the kernel-family ablation of Appendix~\ref{sec:appendix_kernel_family} (8.46 vs.\ 9.19 FID).

Beyond this special case, our approach differs from DMD in two main respects.

\textbf{Framework.}
By treating $\V$ as the central designable object rather than as the gradient of a fixed divergence, our framework expresses several extensions as instances of a single zero-drift condition $\V=0$, none of which has a natural counterpart in the DMD framework, where the update is tied to the score difference.
\emph{(i) Aggregation across temperatures and feature spaces.} Multiple drifting fields, computed at different kernel bandwidths and in different feature encoders, can be combined under one shared zero-drift condition.
\emph{(ii) Kernel and $\V$-normalization.} Because $\V$ is a vector field rather than a loss, it admits normalizations (kernel-bandwidth-aware and $\V$-magnitude-aware) that are hard to express purely as divergence minimization.
\emph{(iii) Modified equilibria.} The zero-drift condition $\V=0$ can be deliberately shifted to a non-$p_\theta=p_\text{data}$ fixed point (e.g.\ to encode CFG-style guidance), while keeping the same training principle.

\textbf{Practical instantiation.}
We compute $\V$ on clean samples $\x$, often in feature spaces and with non-Gaussian kernels, whereas DMD uses score differences on noisy samples $\x_t$, amortized over diffusion time $t$, directly in raw space.
Our ImageNet instantiation estimates $\V$ directly from data and model samples, whereas DMD estimates score differences with parametric networks that are typically initialized from pretrained diffusion teachers and trained alternately with the generator. As a consequence, our generator can be trained from scratch, without a pretrained diffusion teacher and without alternating optimization between an auxiliary score network and the generator.
Our realized field additionally supports batch-level distance normalization, $\V$-normalization, and aggregation across multiple feature spaces, making the update meaningfully different from the DMD construction.

\subsection{Relation to Mean Flow/Consistency Methods.}
Consistency models and related flow-map methods learn one-/few-step generators by matching long jumps along diffusion or flow trajectories~\cite{song2023consistency,song2023improved,boffi2025flow,geng2025mean}.
Methods such as Consistency Models \cite{song2023consistency}, Mean Flows \cite{geng2025mean}, Terminal Velocity Matching (TVM)~\cite{zhou2025terminal}, and Self-Evaluation (Self-E)~\cite{yu2025selfevaluation} have a related stop-gradient target structure, which can be abstractly written as regressing to $\operatorname{sg}(\x+\Delta)$.
In these methods, $\Delta$ is tied to an underlying flow-matching trajectory: Consistency Models / Mean Flows / TVM build $\Delta$ by perturbing the flow-matching endpoint, while Self-E builds $\Delta$ using implicit classification scores.
Our method also trains with a stop-gradient target, but the update is conceptually different: $\Delta$ comes from an explicit drifting field computed directly between data and generated samples, and does not rely on a prescribed FM trajectory.

\subsection{Connection to CACR}
\label{sec:cacr_connection}

CACR~\cite{zheng2021contrastive} is part of the broader family of softmax-normalized contrastive representation-learning objectives.
It explicitly decomposes the contrastive objective into an attraction term and a repulsion term.
In CACR, the attraction weights are normalized over the positive set, while the repulsion weights are normalized over the negative set.
This separate positive/negative normalization is similar in spirit to our mean-shift instantiation in Eqs.~(\ref{eq:meanshift})--(\ref{eq:V_pos_neg}), where positives from $p$ and negatives from $q$ each have their own kernel normalizer.

However, the role of the interaction is different.  CACR uses attraction and repulsion as a representation-learning objective: the optimized variables are encoder parameters, and the goal is to shape an embedding space.
In our framework, the interaction defines a vector-valued sample-update field for generated samples.
The optimized variables are generator parameters, and the central condition is not contrastive separability but zero drift of the generated distribution relative to the data distribution.
Thus the attraction--repulsion form is repurposed from representation learning into a generative matching principle.

\subsection{Relation to Kernel-Based Generators}
\label{sec:kernel_generators}

Kernel-based generators, such as GMMN~\cite{li2015generative,dziugaite2015training}, train a generator by minimizing the MMD between the generated and data distributions.
GMMN computes the MMD in pixel space or in autoencoder codes, and was demonstrated on small datasets such as MNIST; MMD-GAN~\cite{li2017mmd} learns the kernel adversarially.
In our ImageNet setting, an MMD generator in the latent space does not produce reasonable results (214.6 FID; Table~\ref{tab:ladder}, row (a)).

Our method is not derived from a discrepancy between two distributions.
It is formulated around the drifting field $\V$, and it only requires $\V=\mathbf{0}$ when $p=q$.
From this view, MMD relates to one special case: a drifting field with an unnormalized kernel (Appendix~\ref{sec:theory_mmd_relation}).
The designs that our results rely on follow from this formulation: computing the field in feature spaces at multiple scales and locations (Sec.~\ref{subsec:feature_space}), normalized kernels, drift normalization, and guidance.
Table~\ref{tab:ladder} shows the effect of each design.
Rows (b)--(d) of that table are ablations of our method, rather than existing baselines.
In particular, to our knowledge, training a generator from scratch with a kernel-based objective, computed in the feature space of a frozen encoder at multiple scales, first appears in this paper.

Pretrained features have also been used to train generators in other ways.
GFMN~\cite{dossantos2019learning} matches the mean and covariance of the features of a pretrained network, rather than a kernel-based objective.
MMD-DDM~\cite{aiello2023fast} fine-tunes a pretrained diffusion model with MMD in a pretrained feature space.

\subsection{Relation to Coulomb GAN}
\label{sec:coulomb_gan}

Coulomb GAN~\cite{unterthiner2017coulomb} is a GAN whose discriminator learns a kernel-based potential field; the generator moves its samples along the gradient of this potential.
Its ideal field is $\mathbb{E}_{\y\sim p}[\nabla_\x k(\x,\y)]-\mathbb{E}_{\y\sim q}[\nabla_\x k(\x,\y)]$ with the Plummer kernel $k$, which resembles the unnormalized-kernel drifting field of MMD (Eq.~(\ref{eq:mmd_Vmmd2})) and vanishes when $p=q$.
Our method differs in three respects.
\emph{(i) Estimation.}
Coulomb GAN estimates the field with a discriminator network trained alongside the generator. This learned estimate has low variance, but it is biased when the discriminator has not been trained to match the current $q$.
Our field is computed in closed form from the current batch at every step, with no learned approximation: it has no such bias, at the cost of a higher variance from the finite batch.
\emph{(ii) Aggregation.}
A learned potential requires one network per feature space, whereas our fields from many features, scales, and temperatures are aggregated at little cost (Appendix~\ref{sec:appendix_feature_extraction} and \ref{sec:appendix_feature_and_loss_normalization}).
\emph{(iii) Form of the field.}
Coulomb GAN requires the field to be the gradient of a potential.
Our formulation only requires $\V=\mathbf{0}$ when $p=q$, and thus admits fields that do not correspond to any potential, such as our mean-shift field with normalized kernels (Eq.~(\ref{eq:VZZ})).

\section{Additional Derivations}

\subsection{On Identifiability of the Zero-Drift Equilibrium}
\label{sec:theory_identifiability}

In Sec.~\ref{sec:methodology}, we showed that anti-symmetry implies $p=q \Rightarrow \V(\x)\equiv \mathbf{0}$.
Here we investigate the converse: under what conditions does $\V(\x)\approx \mathbf{0}$ imply $p\approx q$?
Generally, this is not guaranteed for arbitrary vector fields.
However, we argue that for our specific construction, the zero-drift condition imposes strong constraints on the distributions.

We assume that $p$ and $q$ have full support on $\mathbb{R}^d$ (e.g., via infinitesimal Gaussian smoothing). 
Consequently, ensuring the equilibrium condition $\V(\x) \approx \mathbf{0}$ for generated samples $\x \sim q$ effectively enforces $\V(\x) \approx \mathbf{0}$ for all $\x \in \mathbb{R}^d$.

We first establish identifiability for mean-shift kernels and two concrete kernel cases that admit clean closed-form arguments (Propositions~\ref{prop:gaussian_kernel} and~\ref{prop:laplacian_gaussian}); we then give a general bilinear-expansion argument that applies to broader drifting field.

\paragraph{Gaussian kernel: identifiability via score-difference.}
Consider the (normalized) mean-shift drifting field of Eq.~(\ref{eq:meanshift})--(\ref{eq:V_pos_neg}) with the Gaussian kernel $k_\sigma(\x,\y)=\exp(-\|\x-\y\|^2/2\sigma^2)$. A direct calculation, using $\nabla_\x k_\sigma(\x,\y)=k_\sigma(\x,\y)(\y-\x)/\sigma^2$, shows that
\begin{equation}
\Vp{p}(\x)
=\frac{\mathbb{E}_{\y^+\sim p}[k_\sigma(\x,\y^+)(\y^+-\x)]}{\mathbb{E}_{\y^+\sim p}[k_\sigma(\x,\y^+)]}
=\sigma^2\nabla_\x \log\hat p_\sigma(\x),
\label{eq:gaussian_score}
\end{equation}
where $\hat p_\sigma(\x)=\int p(\y)\,\mathcal{N}(\x;\y,\sigma^2 I)\,d\y$ is the Gaussian-smoothed density of $p$. Therefore
\begin{equation}
\V_{p,q}(\x)
=\sigma^2\bigl[\nabla_\x \log\hat p_\sigma(\x)-\nabla_\x \log\hat q_\sigma(\x)\bigr],
\end{equation}
which is a (smoothed) score-difference field, under the Gaussian kernel special case.

\begin{proposition}[Identifiability under Gaussian kernel]
\label{prop:gaussian_kernel}
For the normalized mean-shift drifting field with Gaussian kernel $k_\sigma$, $\V_{p,q}(\x)=\mathbf{0}$ for all $\x\in\mathbb{R}^d$ implies $p=q$.
\end{proposition}

\textit{Proof.} If $\V_{p,q}\equiv\mathbf{0}$, then $\nabla_\x\log\hat p_\sigma=\nabla_\x\log\hat q_\sigma$, so $\log\hat p_\sigma-\log\hat q_\sigma$ is a constant. Since $\hat p_\sigma$ and $\hat q_\sigma$ both integrate to $1$, the constant must be $0$, giving $\hat p_\sigma=\hat q_\sigma$. Gaussian convolution is injective (its Fourier transform is non-vanishing), so $p=q$. $\hfill\square$

\paragraph{Small-bandwidth intuition for general radial kernels.}
The Gaussian-kernel result above gives an exact identifiability statement through a smoothed score-difference field.
A similar intuition applies more broadly in the small-bandwidth regime.
Consider a radial kernel, meaning that the weight assigned to a pair $(\x,\y)$ depends only on their distance $\|\x-\y\|$.
For example, one may write
\[
k_\sigma(\x,\y)=\kappa\!\left(\frac{\|\x-\y\|^2}{\sigma^2}\right),
\]
where $\kappa$ is a smooth, rapidly decaying one-dimensional function.
Under standard smoothness and full-support assumptions on $p$, the normalized mean-shift vector has the local expansion
\[
\Vp{p}(\x)
=
c_K\sigma^2\nabla_\x \log p(\x)+o(\sigma^2),
\qquad \sigma\to 0,
\]
where $c_K>0$ depends only on the kernel.
Therefore, in the small-bandwidth regime,
\[
\V_{p,q}(\x)
=
c_K\sigma^2
\bigl(\nabla_\x\log p(\x)-\nabla_\x\log q(\x)\bigr)
+o(\sigma^2).
\]
This gives another intuition for identifiability: if zero drift holds for a family of $\sigma$ that approaching $0$, then 
$\nabla\log p\approx \nabla\log q$, hence $\log p-\log q$ is roughly constant, and normalization gives $p=q$.
Though for finite samples, very small bandwidths can make the drift estimate high-variance, especially in high-dimensional spaces.

\paragraph{Laplacian kernel, Gaussian distributions.}
The kernel actually used in our experiments (Eq.~(\ref{eq:kernel})) is the Laplacian $k_\tau(\x,\y)=\exp(-\|\x-\y\|/\tau)$. We establish identifiability for this kernel when $p$ and $q$ are both multivariate Gaussian distributions.

\begin{proposition}[Identifiability for Gaussians under Laplacian kernel]
\label{prop:laplacian_gaussian}
Let $P=\mathcal{N}(\mu_p,\Sigma_p)$ and $Q=\mathcal{N}(\mu_q,\Sigma_q)$ on $\mathbb{R}^d$ with $\Sigma_p,\Sigma_q\succ 0$, and consider the normalized mean-shift drifting field with Laplacian kernel $k_\tau$. If $\V_{p,q}(\x)=\mathbf{0}$ for all $\x$, then $\mu_p=\mu_q$ and $\Sigma_p=\Sigma_q$, i.e.\ $P=Q$.
\end{proposition}

\textit{Proof sketch.} Fix a unit vector $\mathbf{u}\in\mathbb{R}^d$ and a scalar $r>0$, and consider $\x=r\mathbf{u}$. The conditional measure $p(\y)\,k_\tau(\x,\y)/Z_p(\x)$ used in $\Vp{p}(\x)$ has log-density
\begin{equation}
F_p(\y;\x)=-\tfrac{1}{2}(\y-\mu_p)^\top\Sigma_p^{-1}(\y-\mu_p)-\|\x-\y\|/\tau+\text{const}.
\end{equation}
For large $r$, by Laplace's method, this measure concentrates near $\y^*_p\in\arg\max_\y F_p(\y;r\mathbf{u})$. The first-order condition is $-\Sigma_p^{-1}(\y-\mu_p)+(r\mathbf{u}-\y)/(\tau\|r\mathbf{u}-\y\|)=\mathbf{0}$. As $r\to\infty$ with $\y$ in any bounded set, $(r\mathbf{u}-\y)/\|r\mathbf{u}-\y\|\to\mathbf{u}$, so $\y^*_p\to\mu_p+\Sigma_p\mathbf{u}/\tau$. Hence
\begin{equation}
\Vp{p}(r\mathbf{u})=\mathbb{E}_{p}[\y\mid \x{=}r\mathbf{u}]-r\mathbf{u}\;\to\;\mu_p+\Sigma_p\mathbf{u}/\tau-r\mathbf{u},
\end{equation}
and similarly for $\Vn{q}(r\mathbf{u})$. The translational $r\mathbf{u}$ terms cancel in $\V_{p,q}=\Vp{p}-\Vn{q}$, yielding the asymptotic
\begin{equation}
\V_{p,q}(r\mathbf{u})\;\to\;(\mu_p-\mu_q)+(\Sigma_p-\Sigma_q)\mathbf{u}/\tau,\qquad r\to\infty.
\label{eq:laplacian_gaussian_limit}
\end{equation}
The hypothesis $\V_{p,q}\equiv\mathbf{0}$ forces this limit to vanish for every unit $\mathbf{u}$, which gives $\mu_p-\mu_q=\mathbf{0}$ and $(\Sigma_p-\Sigma_q)\mathbf{u}=\mathbf{0}$ for all $\mathbf{u}$, i.e.\ $\mu_p=\mu_q$ and $\Sigma_p=\Sigma_q$. $\hfill\square$

The results above cover two natural special cases for our framework: a clean general statement for the Gaussian kernel (Proposition~\ref{prop:gaussian_kernel}), and identifiability for Gaussian distributions under the Laplacian kernel actually used in practice (Proposition~\ref{prop:laplacian_gaussian}). We next give a more general argument that does not assume a specific kernel or distribution family.

\paragraph{Setup.}
Consider a general interaction kernel $K(\mathbf{x},\mathbf{y}^+,\mathbf{y}^-) \in \mathbb{R}^d$ and the drifting field
\begin{equation}
\mathbf{V}_{p,q}(\mathbf{x}) := \mathbb{E}_{\mathbf{y}^+\sim p,\ \mathbf{y}^-\sim q} \big[ K(\mathbf{x},\mathbf{y}^+,\mathbf{y}^-) \big].
\label{eq:V_general_K}
\end{equation}
We assume that $p$ and $q$ belong to a finite-dimensional model class spanned by a linearly independent basis $\{\varphi_i\}_{i=1}^m$:
\begin{equation}
p(\mathbf{y})=\sum_{i=1}^m a_i\,\varphi_i(\mathbf{y}),\qquad
q(\mathbf{y})=\sum_{i=1}^m b_i\,\varphi_i(\mathbf{y}),
\label{eq:pq_basis}
\end{equation}
where $\mathbf{a},\mathbf{b}\in\mathbb{R}^m$ are coefficient vectors.

\paragraph{Bilinear expansion over test locations.}
Consider a set of test locations (probes) $\mathcal{X} = \{\mathbf{x}_k\}_{k=1}^N$ with sufficiently large $N$ (e.g., $N \gg m^2$).
For each pair of basis indices $(i,j)$, we define the \emph{induced interaction vector} $\mathbf{U}_{ij} \in \mathbb{R}^{d{\times}N}$ by computing its column:
\begin{equation}
\mathbf{U}_{ij}[:, \mathbf{x}] \triangleq \iint K(\mathbf{x},\mathbf{y}^+,\mathbf{y}^-)\, \varphi_i(\mathbf{y}^+)\,\varphi_j(\mathbf{y}^-)\,d\mathbf{y}^+d\mathbf{y}^-
\end{equation}
evaluated at all $\mathbf{x} \in \mathcal{X}$.
Substituting the basis expansion into Eq.~\eqref{eq:V_general_K}, the drifting field evaluated on $\mathcal{X}$ (stored as a matrix $\mathbf{V}_{\mathcal{X}}$) is a bilinear combination:
\begin{equation}
\mathbf{V}_{\mathcal{X}} \triangleq \sum_{i=1}^m \sum_{j=1}^m a_i b_j \mathbf{U}_{ij}.
\end{equation}
Here, $\mathbf{V}_{\mathcal{X}} \in \mathbb{R}^{d{\times}N}$. At the equilibrium, we have $\mathbf{V}_{\mathcal{X}}=\textbf{0}$, which yields $dN$ linear equations.

\paragraph{Linear independence assumption.}
Our anti-symmetry condition implies that switching $p$ and $q$ negates the field. In terms of basis interactions, this means $\mathbf{U}_{ij} = -\mathbf{U}_{ji}$ (and consequently $\mathbf{U}_{ii} = \mathbf{0}$).
We make the \emph{generic non-degeneracy assumption}:
\textit{The set of vectors $\{\mathbf{U}_{ij}\}_{1 \le i < j \le m}$ is linearly independent in $\mathbb{R}^{dN}$.}
This assumption requires the probes $\mathcal{X}$ and kernel $K$ to be non-degenerate; if all $\mathbf{x}$ yield identical constraints, independence would fail. For generic choices of $K$ and sufficiently diverse probes $\mathcal{X}$ with $dN\gg m^2$, such linear independence is a natural non-degeneracy condition.

\paragraph{Uniqueness of the equilibrium.}
The zero-drift condition $\mathbf{V}(\mathbf{x}) \equiv \mathbf{0}$ implies $\mathbf{V}_{\mathcal{X}} = \mathbf{0}$. Grouping terms by the independent basis vectors $\{\mathbf{U}_{ij}\}_{i<j}$, we have:
\begin{equation}
\sum_{1 \le i < j \le m} (a_i b_j - a_j b_i) \mathbf{U}_{ij} = \mathbf{0}.
\end{equation}
By the linear independence assumption, the coefficients must vanish: $a_i b_j - a_j b_i = 0$ for all $i,j$.
This implies that the vector $\mathbf{a}$ is parallel to $\mathbf{b}$ (i.e., $\mathbf{a} \propto \mathbf{b}$).
Since $p$ and $q$ are probability densities (implying $\int p = \int q = 1$), we must have $\mathbf{a} = \mathbf{b}$, and thus $p=q$.

\paragraph{Connection to the mean shift field.}
The mean-shift field fits this framework.
The update vector (before normalization) is
$\mathbb{E}_{p,q}[ k(\mathbf{x},\mathbf{y}^+) k(\mathbf{x},\mathbf{y}^-) (\mathbf{y}^+ - \mathbf{y}^-) ]$.
Assuming the normalization factors $Z_p$ and $Z_q$ are finite, the condition $\mathbf{V}(\mathbf{x}) = \mathbf{0}$ implies the numerator integral vanishes, which corresponds to an interaction kernel of the form:
\begin{equation}
K(\mathbf{x}, \mathbf{y}^+, \mathbf{y}^-) = k(\mathbf{x},\mathbf{y}^+)\, k(\mathbf{x},\mathbf{y}^-)\, (\mathbf{y}^+ - \mathbf{y}^-).
\end{equation}
This kernel generates the bilinear structure analyzed above.
Since we can choose $N$ such that $dN \gg m^2$, the dimension of the test space is much larger than the number of basis pairs. Thus, the linear independence of $\{\mathbf{U}_{ij}\}$ is expected to hold for generic configurations.
Finally, for general distributions $p$ and $q$, we can approximate them using a sufficiently large basis expansion, turning into $\tilde p$ and $\tilde q$. When the basis approximation is sufficiently accurate, $\tilde p \approx p$ and $\tilde q \approx q$, and the drift field $\mathbf{V}_{\tilde p, \tilde q}\approx \mathbf{V}_{p,q}\approx 0$. By the argument above, $\tilde p\approx \tilde q$, and thus $p\approx q$.

The argument above works for general form of drifting field, under mild anti-degeneracy assumptions.

\paragraph{Applicability to our practical objective.}
\phantomsection\label{sec:theory_scope}
The analysis above considers a single drifting field.
Our practical objective sums the drifting losses of multiple fields: one per feature, in addition to the vanilla drifting loss in the raw space (Appendix~\ref{sec:appendix_feature_extraction}).
The analysis applies to this objective as follows.

\emph{Equilibrium.}
Every field vanishes when $p=q$, so $p=q$ is an equilibrium of the full objective.
This also holds for our mini-batch implementation, which is a Monte Carlo estimate of a drifting field that satisfies $\V_{p,p}=\mathbf{0}$ (Appendix~\ref{sec:appendix_compute_V}).

\emph{Identifiability.}
Conversely, suppose that the drift is zero for the full objective.
Then every field vanishes, including the field in the raw space.
If a vanishing raw-space field implies $p=q$, we obtain $p=q$ without any assumption on the feature encoder.
In other words, the feature-space fields provide richer training signals (Sec.~\ref{subsec:feature_space}), while the guarantee comes from the raw-space field and depends only on its kernel.
Our low-dimensional experiments use the raw space only.

\emph{Kernel.}
What remains is whether a vanishing raw-space field implies $p=q$, which depends on the kernel.
For the Gaussian kernel, this holds for arbitrary distributions (Proposition~\ref{prop:gaussian_kernel}).
For the Laplacian kernel used in our experiments, we show it when $p$ and $q$ are both Gaussian distributions (Proposition~\ref{prop:laplacian_gaussian}).
For general $p$ and $q$, it is supported by the bilinear argument above under a non-degeneracy assumption, but a proof remains open.
Our analysis also does not cover fields aggregated over multiple temperatures.

\subsection{The Drifting Field of MMD}
\label{sec:theory_mmd_relation}

In principle, if a method minimizes a discrepancy between two distributions $p$ and $q$ and reaches minimum at $p=q$, then from the perspective of our framework, a drifting field $\V$ exists that governs sample movement: we can let $\V{\propto}{-}\frac{\partial\mathcal L }{\partial {\x}}$, which is zero when $p=q$.
We discuss the formulation of this $\V$ for a loss based on Maximum Mean Discrepancy (MMD) \cite{li2015generative,dziugaite2015training}.

\newcommand{\kkk}{\tilde{k}}
\paragraph{Gradients of Drifting Loss.}
With $\x=f_\theta(\e)$, our drifting loss in Eq.~(\ref{eq:train_loss}) can be written as:
\begin{equation}
\mathcal{L}
=
\mathbb{E}_{\x{\sim}q}
[
\mathcal{L}(\x)
]
=
\mathbb{E}_{\x{\sim}q}
\Big[
\big\|
\x
-
\text{\texttt{sg}}\big(
\x
+
\V(\x)
\big)
\big\|^2
\Big],
\end{equation}
where ``\texttt{sg}'' is short for stop-gradient. The gradient w.r.t. the parameters $\theta$ is computed by:
\begin{equation}
\frac{\partial\mathcal L }{\partial {\theta}} =
\mathbb{E}_{\x{\sim}q}\Big[
\frac{\partial\mathcal L(\x) }{\partial {\x}} \frac{\partial\x}{\partial\theta}
\Big].
\end{equation}
where
$
\frac{\partial\mathcal L(\x) }{\partial {\x}}{=}2
(
\x
{-}
\text{\texttt{sg}}(
\x
{+}
\V(\x)
)
)
{=}{-}2\V(\x)
$. This gives:
\begin{equation}
\label{eq:loss_2V}
\V(\x) = -\frac{1}{2}\frac{\partial\mathcal L(\x) }{\partial {\x}}
\end{equation}
We note that this formulation is general and imposes no constraints on $\V$, except that $\V=\textbf{0}$ when $p=q$.

Our method does not require $\mathcal{L}$ to define a discrepancy between $p$ and $q$. However, for other methods that depend on minimizing a discrepancy $\mathcal{L}$, we can induce a drifting field via (\ref{eq:loss_2V}). This is valid if $\mathcal{L}$ is minimized when $p=q$.

\paragraph{Gradients of MMD Loss.}
\newcommand{\kk}{\xi}
\newcommand{\LMMD}{\mathcal{L}_{\text{MMD}^2}}
In MMD-based methods (\eg, \citep{li2015generative}), the difference between two distributions $p$ and $q$ is measured by squared MMD:
\begin{equation}
\label{eq:mmd2_def}
\begin{aligned}
\LMMD(p,q)
= &
\mathbb{E}_{\x,\x'\sim q}[\kk(\x,\x')]
-2\,\mathbb{E}_{\y\sim p,\;\x\sim q}[\kk(\y,\x)]
\\
+ &~const.
\end{aligned}
\end{equation}
Here, the constant term is $\mathbb{E}_{\y,\y'\sim p}[\kk(\y,\y')]$, which depends only on the target distribution $p$ and remains unchanged. $\kk$ is a kernel function. 

Consider $\x=f_\theta(\e)$ with $\e\sim p_\e$. The gradient estimation performed in \cite{li2015generative} corresponds to:
\begin{equation}
\frac{\partial\LMMD}{\partial {\theta}} =
\mathbb{E}_{\x{\sim}q}\Big[\frac{\partial\LMMD(\x)}{\partial {\x}} \frac{\partial\x}{\partial\theta}
\Big]
\end{equation}
where the gradient w.r.t $\x$ is computed by:
\begin{equation}
\frac{\partial\LMMD(\x)}{\partial {\x}}
=
2\mathbb{E}_{\x'{\sim}q} \Big[\frac{\partial \kk(\x,\x')}{\partial{\x}}\Big]
-
2\mathbb{E}_{\y{\sim}p} \Big[\frac{\partial \kk(\x,\y)}{\partial{\x}}\Big].
\end{equation}
Using our notation of positives and negatives, we rename the variables and rewrite as:
\begin{equation}
\label{eq:mmd_eq4}
\frac{\partial\mathcal L_{\text{MMD}^2}(\x)}{\partial {\x}}
=
2\mathbb{E}_{\yn{\sim}q} \Big[\frac{\partial \kk(\x,\yn)}{\partial{\x}}\Big]
-
2\mathbb{E}_{\yp{\sim}p} \Big[\frac{\partial \kk(\x,\yp)}{\partial{\x}}\Big].
\end{equation}
Comparing with Eq.~(\ref{eq:loss_2V}), we obtain:
\begin{equation}
\label{eq:mmd_Vmmd}
\V_\text{MMD}(\x)
\triangleq
\mathbb{E}_{\yp{\sim}p} \Big[\frac{\partial \kk(\x,\yp)}{\partial{\x}}\Big]
-
\mathbb{E}_{\yn{\sim}q} \Big[\frac{\partial \kk(\x,\yn)}{\partial{\x}}\Big]
\end{equation}
This is the underlying drifting field that corresponds to the MMD loss $\LMMD$.

For a radial kernel $\kk(\x, \y)=\kk(R)$ where  $R=\|\x - \y\|^2$, the gradient of kernel is:
\begin{equation}
\frac{\partial\kk(\x, \y)}{\partial\x} = 
2\kk'(\|\x - \y\|^2)(\x - \y)
\end{equation}
where $\kk'$ is the derivative of the function $\kk(R)$.
Accordingly, Eq.~(\ref{eq:mmd_Vmmd}) becomes:
\begin{equation}
\begin{aligned}
\label{eq:mmd_Vmmd2}
\V_\text{MMD}(\x)
= &
\mathbb{E}_{\yp{\sim}p} \Big[
2\kk'(\|\x - \yp\|^2)(\x - \yp)
\Big].
\\ - &
\mathbb{E}_{\yn{\sim}q} \Big[
2\kk'(\|\x - \yn\|^2)(\x - \yn)
\Big]
\end{aligned}
\end{equation}
In \cite{li2015generative}, the Gaussian kernel is used: $\kk(\x, \y) = \exp (-\frac{1}{2\sigma^2} \|\x{-}\y\|^2)$, leading to $\kk'(\|\x{-}\y\|^2)=-\frac{1}{2\sigma^2}\exp (-\frac{1}{2\sigma^2} \|\x{-}\y\|^2)$.

\paragraph{Relations and Differences.}
When using our definition of $\V = \V^{+} - \V^{-}$ (\ie, Eq.~(\ref{eq:V_pos_neg})), we have:
\begin{equation}
\begin{aligned}
\label{eq:mmd_eq6}
\V(\x)
= & 
\mathbb{E}_{\yp{\sim}p} \Big[
\tilde{k}(\x, \yp)(\yp - \x)
\Big] \\
- &
\mathbb{E}_{\yn{\sim}q} \Big[
\tilde{k}(\x, \yn)(\yn - \x)
\Big]
\end{aligned}
\end{equation}

Comparing (\ref{eq:mmd_Vmmd2}) with (\ref{eq:mmd_eq6}), we show that the underlying kernel used to build the drifting field of MMD is:
\begin{equation}
\label{eq:mmd_eq7}
\tilde{k}_\text{MMD}(\x, \y)=-2\kk'(\|\x - \y\|^2).
\end{equation}
When $\kk$ is a Gaussian function, we have: $\tilde{k}(\x, \y)=\frac{1}{\sigma^2}\exp (-\frac{1}{2\sigma^2} \|\x{-}\y\|^2)$.
Without normalization, the resulting drift no longer satisfies the assumptions underlying Alg.~\ref{alg:drift_code}, and the mean-shift interpretation breaks down.

As a comparison, our general formulation enables to use \textit{normalized} kernels:
\begin{equation}
\tilde{k}(\x, \y)=\frac{1}{Z(\x)}k(\x, \y)=\frac{1}{\mathbb{E}_\y[k(\x, \y)]}k(\x, \y),
\end{equation}
where the expectation is over $p$ or $q$.
Only when we use normalized kernels, we have (see Eq.~(\ref{eq:VZZ})):
\begin{equation}
\V(\x)
= {\mathbb{E}_{p,q}\!\left[\tilde{k}(\x,\yp)\tilde{k}(\x,\yn)(\yp - \yn)\right]},
\end{equation}
on which our Alg.~\ref{alg:drift_code} is based.

Given this relation, we summarize the key differences between our model and the MMD-based methods as follows:
\begin{enumerate}
[itemsep=1pt, topsep=1pt, parsep=1pt, partopsep=1pt, label=(\roman*)]
    \item Our method is formulated around the drifting field $\V$, which is more flexible and general.
    \item Our method supports and leverages \textit{normalized} kernels $\frac{1}{Z}k(\x, \y)$ that cannot be naturally derived from the MMD perspective.
    \item Our $\V$-centric formulation enables a flexible step size for drifting (\ie, $\x \leftarrow \x{+}\eta\V$) and therefore naturally supports $\V$-normalization (see \ref{sec:appendix_feature_and_loss_normalization}).
    \item Our $\V$-centric formulation allows the equilibrium concept to be naturally extended to support CFG (Sec.~\ref{subsec:additive_guidance}), which cannot be naturally derived from the MMD loss.
\end{enumerate}
In summary, although a special case of our method reduces to MMD, our $\V$-centric framework is more general and enables unique possibilities that are important in practice.

\begin{table}[!ht]
\centering
\tablestyle{5pt}{1.05}
\tablefontsize
\begin{tabular}{c l l r}
\toprule
row & method & introduced in & FID \\
\midrule
(a) & MMD generator in the latent space (GMMN+AE~\cite{li2015generative}) & prior work & 214.6 \\
\midrule
(b) & + drifting in multi-scale feature space (Sec.~\ref{subsec:feature_space}) & this paper & 18.61 \\
(c) & + drift normalization (Appendix~\ref{sec:appendix_feature_and_loss_normalization}) and guidance (Sec.~\ref{subsec:additive_guidance}) & this paper & 10.72 \\
(d) & + normalized drifting field (Eq.~(\ref{eq:VZZ})); \textbf{our default} & this paper & \textbf{8.46} \\
\bottomrule
\end{tabular}
\vspace{.5em}
\caption{
\textbf{From a classical kernel generator to Drifting Models.} Row (a) is the only pre-existing method (our reimplementation, on SD-VAE latents); rows (b)--(d) add the designs of this paper one at a time. Rows (a)--(c) use the drifting field with an unnormalized kernel (Eq.~(\ref{eq:mmd_Vmmd2})), the special case that relates to MMD.
(Setting: B/2 model, 100 epochs; all rows share the data, seed, and numbers of positive and negative samples; rows (b)--(d) share the same frozen encoder.)
}
\label{tab:ladder}
\end{table}

\paragraph{Quantitative comparison.}
Table~\ref{tab:ladder} starts from the classical MMD generator and adds the designs of this paper one at a time, under our ablation setting.
Row (a) is the MMD objective on raw SD-VAE latents, our reimplementation of GMMN+AE~\cite{li2015generative}. It does not produce reasonable results on ImageNet (214.6 FID).
Row (b) computes the same field in our multi-scale feature space (Sec.~\ref{subsec:feature_space}). It is our feature-space drifting loss in Eq.~(\ref{eq:feature_space_multi}) with an unnormalized kernel: an ablation of our method, rather than an existing baseline (Appendix~\ref{sec:kernel_generators}). This design alone improves FID from 214.6 to 18.61.
Row (c) adds drift normalization and training-time guidance (differences (iii) and (iv)), reaching 10.72 FID.
Row (d) replaces the unnormalized kernel with our normalized kernel (difference (ii)), with the same Laplacian kernel, reaching 8.46 FID, which is our ablation default.
Every step from (a) to (d) is thus a design introduced in this paper. With the encoder held fixed, the designs enabled by the $\V$-centric formulation alone more than halve the FID (18.61$\to$8.46).

\section{Submission Checklist Materials}
\label{sec:checklist_materials}

\subsection{Limitations}
\label{sec:appendix_encoder_discussion}

\paragraph{Reliance on a feature encoder.}
On ImageNet, our method relies on a feature encoder (Sec.~\ref{subsec:imagenet_exp}).
We attribute this to the variance of the drifting field when it is estimated from a batch of samples: as with other kernel-based estimates, this variance grows rapidly with the dimension of the space.
The feature encoder maps the data to a space in which a batch of practical size gives a reliable estimate.
This may be a drawback for modalities that lack strong pretrained encoders.
There are two potential mitigations.
First, our encoder is pre-trained on the same data with a standard self-supervised method (MAE; classification fine-tuning is optional, Appendix~\ref{sec:appendix_no_cls_ft}), and generation quality improves with a larger encoder and longer pre-training (Table~\ref{tab:ablation_drift_space}).
The same pre-training could therefore be run on the data of a new modality.
Second, in low-dimensional settings (robotics control, 2D), drifting operates without any encoder.

\paragraph{Frozen encoder.}
Our feature encoder is frozen during generator training. This keeps training non-adversarial, but, unlike a discriminator, a frozen encoder cannot adapt to the generator.
We discuss two potential failure modes.
\emph{(i) Exploiting the encoder.}
A generator could differ from the data in ways that the encoder does not capture.
To reduce this risk, we compute the drifting loss on multi-scale, multi-location features, in addition to the raw space (Appendix~\ref{sec:appendix_feature_extraction}).
Our model also remains competitive under DINO-FD (Table~\ref{tab:appendix_dino_fd}), whose backbone is independent of our encoder.
An adaptive encoder could detect such behavior, but would reintroduce adversarial training.
\emph{(ii) Sensitivity to the encoder architecture.}
Our method works with several encoders (SimCLR, MoCo-v2, and MAE in Table~\ref{tab:ablation_drift_space}; ConvNeXt-V2 in Table~\ref{tab:ablation_latent_vs_pixel}), while its quality depends on the encoder.
Other architectures, such as ViT, are left for future work.

\paragraph{Scope of experiments.}
Our image-generation experiments are confined to class-conditional ImageNet 256$\times$256; text-conditional generation and other modalities remain to be explored.
 
\subsection{Compute}
\label{sec:appendix_compute}

We trained on Google Cloud TPU v6e. Compute breakdown:
\begin{itemize}[itemsep=1pt, topsep=2pt, parsep=1pt, leftmargin=1.2em]
\item \emph{Ablation} (B/2 model, 100 epochs): MAE pretraining $\approx$ 2.5h + drifting-model training $\approx$ 2.5h on 64 TPU v6e.
\item \emph{Final L/2 latent-space model} (1280 epochs): MAE pretraining $\approx$ 22h + drifting-model training $\approx$ 37h on 128 TPU v6e.
\item \emph{Robotics control}: following the training and evaluation protocol of Diffusion Policy~\citep{chi2025diffusion}, each task is trained on a single NVIDIA RTX 4090 GPU and takes approximately 5--19 days depending on the task and observation modality.
\end{itemize}
Pixel-space final runs use a comparable budget; per-step compute is similar to the latent counterpart.
Exploratory and failed runs were not fully tracked.

\subsection{Broader Impacts}
\label{sec:broader_impacts}

This work studies one-step generative modeling for images and robotic control.
Potential positive impacts include reducing inference-time sampling cost for high-quality generation, making generative-model evaluation and iteration faster, and enabling lower-latency policy generation in simulated robotics settings.
Because the method targets efficient generation, these benefits may be especially relevant in settings where repeated multi-step sampling is costly.

At the same time, improvements in fast image generation can also amplify existing risks of synthetic media, including misinformation, fake profiles, spam, or other deceptive uses.
In robotics, lower-latency policy generation should not be interpreted as sufficient for safe real-world deployment without task-specific safety validation.
To support reproducibility, we release code and pretrained weights for our class-conditional ImageNet 256$\times$256 models at \url{https://github.com/lambertae/drifting}. These models are comparable in scale and capability to widely available public ImageNet generators. We do not release new datasets or generated-data assets.

\subsection{Existing Assets and Licenses}
\label{sec:asset_licenses}

Table~\ref{tab:asset_licenses} summarizes the existing datasets, codebases, pretrained models, and evaluation assets used in this work, together with the license or terms reported by the original creators.
We use these assets for research purposes and cite the corresponding papers in the main text.

\begin{table}[!ht]
\centering
\caption{\textbf{Existing assets and licenses or terms.} URLs point to the original asset pages or license files used to identify the terms.}
\label{tab:asset_licenses}
\scriptsize
\setlength{\tabcolsep}{3pt}
\renewcommand{\arraystretch}{1.08}
\begin{tabular}{p{0.20\linewidth}p{0.30\linewidth}p{0.42\linewidth}}
\toprule
Asset & Use in this work & License or terms \\
\midrule
ImageNet-1K & Training and evaluation data for 256$\times$256 image generation. & ImageNet terms of access: non-commercial research and educational use; see \url{https://www.image-net.org/download.php}. \\
SD-VAE & Latent tokenizer and decoder for latent-space experiments and feature extraction. & MIT license for \texttt{stabilityai/sd-vae-ft-mse} on the Hugging Face model page; see \url{https://huggingface.co/stabilityai/sd-vae-ft-mse/tree/main}. \\
DiT reference implementation & Architectural reference for DiT-style Transformer generators; we do not use DiT pretrained weights. & CC-BY-NC 4.0 for the official DiT code and weights; see \url{https://github.com/facebookresearch/DiT}. \\
SimCLR public encoder & ImageNet-pretrained ResNet-50 feature encoder baseline. & Apache-2.0; see \url{https://github.com/google-research/simclr}. \\
MoCo-v2 public encoder & ImageNet-pretrained ResNet-50 feature encoder baseline. & The MoCo README states CC-BY-NC 4.0; see \url{https://github.com/facebookresearch/moco}. \\
MAE reference implementation & Reference for the masked-autoencoding objective used by our custom latent-MAE feature encoder. & CC-BY-NC 4.0; see \url{https://github.com/facebookresearch/mae}. \\
ConvNeXt-V2 & Pixel-space pretrained feature encoder. & Code is MIT; ImageNet pretrained and fine-tuned model weights are CC-BY-NC; see \url{https://github.com/facebookresearch/ConvNeXt-V2}. \\
DINO-FD / DINOv2 evaluation assets & Complementary FID-style evaluation in a self-supervised feature space. & \texttt{dgm-eval} is MIT; DINOv2 code and standard model weights are Apache-2.0; see \url{https://github.com/layer6ai-labs/dgm-eval} and \url{https://github.com/facebookresearch/dinov2}. \\
CLIP & Feature encoder for nearest-neighbor memorization analysis. & MIT; see \url{https://github.com/openai/CLIP}. \\
Diffusion Policy & Robotics-control protocol and baseline comparison. & MIT; see \url{https://github.com/real-stanford/diffusion_policy}. \\
\bottomrule
\end{tabular}
\end{table}

\end{document}